\documentclass{article} 
\usepackage{iclr2021_arxiv,times}

\usepackage[utf8]{inputenc} 
\usepackage[T1]{fontenc}    
\usepackage{hyperref}       
\usepackage{url}            
\usepackage{booktabs}       
\usepackage{amsfonts}       
\usepackage{nicefrac}       
\usepackage{microtype}      
\usepackage{nameref}
\usepackage[shortlabels]{enumitem}
\usepackage{textgreek}

\usepackage{float}
\usepackage{graphicx}
\usepackage{tocbasic    }

\usepackage{color}
\usepackage{amsmath}
\usepackage{amssymb}
\usepackage{amsthm}
\usepackage{mathtools}
\usepackage{thmtools}
\usepackage[mathscr]{eucal}
\usepackage{bm}
\usepackage{bbm}
\usepackage{relsize}
\usepackage{graphics}
\usepackage{wrapfig}
\usepackage{multirow}
\usepackage{enumitem} 

\usepackage{array}
\usepackage{threeparttable}

\usepackage{pgf}
\usepackage{xinttools}
\usepackage{tikz}
\usetikzlibrary{arrows.meta}
\usepackage{textcomp}
\usetikzlibrary{calc,shapes,arrows,positioning,automata,trees}
\usepackage{placeins} 
\usepackage{tabularx}
\usepackage{graphicx}
\usepackage{subcaption} 
\usepackage{listings}
\usepackage{xargs,lipsum,caption,changepage,ifthen}

\usepackage{tikz} 

\usepackage{footmisc}

\usepackage{bigdelim}
\usepackage{colortbl} 
\definecolor{mColor1}{rgb}{0.95,0.95,0.95}
\newcolumntype{a}{>{\columncolor{mColor1}}c}
\newcolumntype{b}{>{\columncolor{mColor1}}l}

\definecolor{HubiLightBlue}{RGB}{135,205,222}
\definecolor{HubiBlue}{RGB}{35,88,215}
\definecolor{HubiGreen}{RGB}{123,198,83}
\definecolor{HubiPink}{RGB}{228,88,215}
\definecolor{HubiGray}{RGB}{181,181,181}
\definecolor{HubiYellow}{RGB}{253,189,0}

\usepackage[toc,page]{appendix}

\newcommand{\stoptocwriting}{%
  \addtocontents{toc}{\protect\setcounter{tocdepth}{-5}}}
\newcommand{\resumetocwriting}{%
  \addtocontents{toc}{\protect\setcounter{tocdepth}{\arabic{tocdepth}}}}

\newtheorem{theorem}{Theorem}
\newtheorem{definition}{Definition}
\newtheorem*{theorem*}{Theorem}
\newtheorem*{definition*}{Definition}

\newtheorem{theoremA}{Theorem}
\newtheorem{definitionA}{Definition}
\newtheorem{corollaryA}{Corollary}%
\newtheorem{lemmaA}{Lemma}%
\newcommand\Ba{\bm{a}}
\newcommand\Bb{\bm{b}}
\newcommand\Bc{\bm{c}}
\newcommand\Bd{\bm{d}}
\newcommand\Be{\bm{e}}

\newcommand\Bh{\bm{h}}

\newcommand\Bm{\bm{m}}

\newcommand\Bp{\bm{p}}

\newcommand\Br{\bm{r}}

\newcommand\Bt{\bm{t}}

\newcommand\Bv{\bm{v}}

\newcommand\Bx{\bm{x}}
\newcommand\By{\bm{y}}
\newcommand\Bz{\bm{z}}

\newcommand\BA{\bm{A}}

\newcommand\BI{\bm{I}}

\newcommand\BK{\bm{K}}

\newcommand\BQ{\bm{Q}}
\newcommand\BR{\bm{R}}

\newcommand\BV{\bm{V}}
\newcommand\BW{\bm{W}}
\newcommand\BX{\bm{X}}
\newcommand\BY{\bm{Y}}
\newcommand\BZ{\bm{Z}}

\newcommand\Bxi{\bm{\xi}}

\newcommand\BXi{\bm{\Xi}}

\newcommand\BOn{\bm{1}}
\newcommand\BZe{\bm{0}}

 \newcommand{\dR}{\mathbb{R}}

\newcommand{\rC}{\mathrm{C}}  
\newcommand{\rE}{\mathrm{E}} 
 
 \newcommand{\rJ}{\mathrm{J}} 
 \newcommand{\rL}{\mathrm{L}}

\newcommand{\rS}{\mathrm{S}}

 \newcommand{\cL}{\mathcal{L}}

\newcommand{\cS}{\mathcal{S}}

 \newcommand{\Rd}{\mathrm{d}}

\newcommand\EXP{\mathbf{\mathrm{E}}}
\newcommand\PR{\mathbf{\mathrm{Pr}}}
\newcommand\VAR{\mathbf{\mathrm{Var}}}

\newcommand\TR{\mathbf{\mathrm{Tr}}}

\newcommand\sgn{\mathop{\mathrm{sgn}\,}}

\newcommand\nn{\mathrm{new}}

\newcommand{\ABS}[1]{{{\left| #1 \right|}}} 
\newcommand{\BRA}[1]{{{\left\{#1\right\}}}} 
\newcommand{\NRM}[1]{{{\left\| #1\right\|}}}

\renewcommand{\leq}{\leqslant}

\newcommand{\soft}{\mathrm{softmax}}
\newcommand{\diag}{\mathrm{diag}}

\makeatletter
\newcommand{\dlmf}[1]{%
\citep[%
  \def\nextitem{\def\nextitem{, }}%
  \@for \el:=#1\do{\nextitem\href{http://dlmf.nist.gov/\el}{(\el)}}%
]{Olver:10}%
}
\makeatother

\usepackage{pdflscape} 

\newcolumntype{R}[1]{>{\raggedright\arraybackslash}p{#1}}
\newcolumntype{C}[1]{>{\centering\arraybackslash}p{#1}}
\newcolumntype{L}[1]{>{\raggedleft\arraybackslash}p{#1}}

\iclrfinalcopy

\hypersetup{
   colorlinks=true,
   linkcolor=blue,
   citecolor=green,
   urlcolor=magenta,
   pdfborder=0 0 0,
   pdftitle={Hopfield Networks is All You Need},
   pdfsubject={Modern Hopfield Network, Energy, Attention, 
               Transformer, Convergence, Storage Capacity}, 
   pdfkeywords={Hopfield Network, Modern Hopfield Network, 
                Attention, Transformer, Continuous State,
                Energy, Convergence, Storage Capacity},
   pdfauthor={},%
   pdfstartview=FitH
}

\title{Hopfield Networks is All You Need}

\author{\vspace{0.1cm}
    Hubert Ramsauer\footnotemark[1] \quad
    Bernhard Sch\"{a}fl\footnotemark[1] \quad 
    Johannes Lehner\footnotemark[1] \quad  
    Philipp Seidl\footnotemark[1] \quad \\ \vspace{0.1cm} \bf
    Michael Widrich\footnotemark[1] \quad 
    Thomas Adler\footnotemark[1] \quad 
    Lukas Gruber\footnotemark[1] \quad 
    Markus Holzleitner\footnotemark[1] \quad \\ \vspace{0.1cm} \bf
     Milena Pavlovi{\'c}\footnotemark[3]~$~^{,}$\footnotemark[4] \quad
    Geir Kjetil Sandve\footnotemark[4] \quad
    Victor Greiff\footnotemark[3] \quad 
    David Kreil\footnotemark[2] \quad \\ \vspace{0.1cm} \bf
    Michael Kopp\footnotemark[2] \quad 
    G\"{u}nter Klambauer\footnotemark[1] \quad
    Johannes Brandstetter\footnotemark[1]  \quad
    Sepp Hochreiter\footnotemark[1]~$~^{,}$\footnotemark[2]\\
  \footnotemark[1]~~ELLIS Unit Linz, LIT AI Lab, Institute for Machine Learning,\\
                  ~~Johannes Kepler University Linz, Austria\\
  \footnotemark[2]~~Institute of Advanced Research in 
                    Artificial Intelligence (IARAI) \\
  \footnotemark[3]~~Department of Immunology, University of Oslo, Norway\\ 
  \footnotemark[4]~~Department of Informatics, University of Oslo, Norway                    
}

\DeclareUnicodeCharacter{2212}{-}

\begin{document}

\maketitle

\begin{abstract}
We introduce a modern Hopfield network
with continuous states and a corresponding update rule.
The new Hopfield network can store 
exponentially (with the dimension of the 
associative space) many patterns, 
retrieves the pattern with
one update, and has exponentially small retrieval errors.
It has three types of energy minima (fixed points of the update):
(1) global fixed point averaging over all patterns, 
(2) metastable states averaging over a subset of patterns, and 
(3) fixed points which store a single pattern.
The new update rule 
is equivalent to the attention mechanism used in transformers.
This equivalence enables a 
characterization of the heads of transformer models.
These heads perform in the first layers preferably
global averaging and in higher layers partial averaging 
via metastable states. 
The new modern Hopfield network can be integrated 
into deep learning architectures 
as layers to allow the storage of and access to 
raw input data, intermediate results, or learned prototypes.
These Hopfield layers enable new ways of deep learning, 
beyond
fully-connected, convolutional, or recurrent networks,
and provide pooling, memory, association, and attention mechanisms.
We demonstrate the broad applicability of the Hopfield layers
across various domains.
Hopfield layers improved state-of-the-art on three out of four
considered multiple instance learning problems as well as
on immune repertoire
classification with several hundreds of thousands of instances.
On the UCI benchmark collections of small classification tasks, 
where deep learning methods typically struggle, 
Hopfield layers yielded a new state-of-the-art
when compared to different machine learning methods.
Finally, Hopfield layers achieved state-of-the-art
on two drug design datasets.
The implementation is available at: \url{https://github.com/ml-jku/hopfield-layers}
\end{abstract}

\stoptocwriting

\section{Introduction}
The deep learning community 
has been looking for alternatives to recurrent neural networks (RNNs)
for storing information.
For example, linear memory networks use a linear autoencoder for sequences 
as a memory \citep{Carta:20}.
Additional memories for RNNs like
holographic reduced representations \citep{Danihelka:16},
tensor product representations \citep{Schlag:18,Schlag:19}
and classical associative memories 
(extended to fast weight approaches) \citep{Schmidhuber:92fastmem,Ba:16,Ba:16arxiv,Zhang:17,Schlag:21}
have been suggested.
Most approaches to new memories are based on attention.
The neural Turing machine (NTM) is equipped with an
external memory and an attention process \citep{Graves:14}.  
Memory networks \citep{Weston:14} use an $\arg\max$ attention
by first mapping a query and patterns into a space and
then retrieving the pattern with the largest dot product.
End to end memory networks (EMN) make this attention scheme differentiable
by replacing $\arg\max$ through a $\soft$ \citep{Sukhbaatar:15,Sukhbaatar:15arxiv}.
EMN with dot products became very popular and implement a key-value
attention \citep{Daniluk:17} for self-attention.
An enhancement of EMN is the transformer \citep{Vaswani:17,Vaswani:17arxiv}
and its extensions \citep{Dehghani:18}.
The transformer has had a great impact on the natural language processing
(NLP) community, in particular via the BERT models \citep{Devlin:18,Devlin:19}. 

{\bf Contribution of this work:}
(i) introducing novel deep learning layers that are equipped 
with a memory via modern Hopfield networks,
(ii) introducing a novel energy function and a novel update rule for 
continuous modern Hopfield networks that are differentiable and 
typically retrieve patterns after one update. 
Differentiability is required for gradient descent parameter updates and
retrieval with one update is compatible with activating the layers of deep networks.

We suggest using modern Hopfield networks 
to store information or learned prototypes in different 
layers of neural networks.
Binary Hopfield networks 
were introduced as associative memories
that can store and retrieve patterns \citep{Hopfield:82}.
A query pattern can retrieve the pattern to which it is most similar
or an average over similar patterns.
Hopfield networks seem to be an ancient technique,
however, new energy functions improved
their properties.
The stability of spurious states or metastable states
was sensibly reduced \citep{Barra:18}.
The largest and most impactful successes are reported
on increasing the storage capacity of Hopfield networks.
In a $d$-dimensional space,
the standard Hopfield model can store $d$ uncorrelated patterns
without errors but only
$C d/\log(d)$ random patterns with
$C<1/2$ for a fixed stable pattern or $C<1/4$ if all patterns
are stable \citep{McEliece:87}.
The same bound holds for nonlinear learning rules \citep{Mazza:93}.
Using tricks-of-trade and allowing
small retrieval errors, the storage capacity
is about $0.138 d$ \citep{Crisanti:86,Hertz:91,Torres:02}.
If the learning rule is not related to the Hebb rule, then up to $d$ 
patterns can be stored \citep{Abu-Mostafa:85}.
For Hopfield networks with non-zero diagonal matrices,
the storage can be 
increased to $C d \log(d)$ \citep{Folli:17}.
In contrast to the storage capacity, the number of energy minima 
(spurious states, stable states) of Hopfield networks 
is exponential in $d$ \citep{Tanaka:80,Bruck:90,Wainrib:13}.

The standard binary Hopfield network has
an energy function that can be expressed as
the sum of interaction functions $F$ with $F(x)=x^2$.
Modern Hopfield networks, also called 
``dense associative memory'' (DAM) models,
use an energy function with interaction functions
of the form $F(x)=x^n$ and, thereby, achieve
a storage capacity proportional to $d^{n-1}$
\citep{Krotov:16,Krotov:18}.
The energy function of modern Hopfield networks 
makes them robust against adversarial attacks \citep{Krotov:18}.
Modern binary Hopfield networks with energy functions based on 
interaction functions of the form $F(x)=\exp(x)$ even lead 
to storage capacity of $2^{d/2}$,
where all stored binary patterns are fixed points but the radius of
attraction vanishes \citep{Demircigil:17}.
However, in order to integrate Hopfield networks into deep learning
architectures, it is necessary to make them differentiable, that is,
we require continuous Hopfield networks \citep{Hopfield:84,Koiran:94}.

\begin{figure}[ht]
        \centering
        \includegraphics[width=1.0\textwidth]{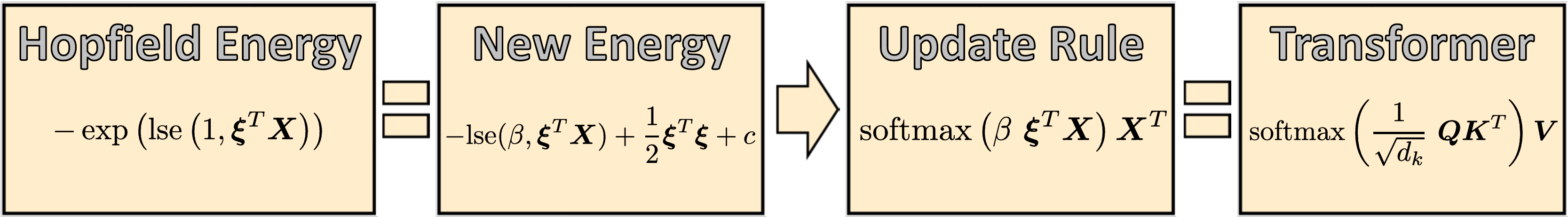}
        \caption[]{We generalize
        the energy
        of binary modern Hopfield networks to continuous states 
        while keeping fast convergence and
        storage capacity properties.
        We also propose a new update
        rule that minimizes the energy. 
        The new update rule is the attention mechanism of the transformer.
        Formulae are modified to express $\soft$ as row vector. 
        ``$=$''-sign means ``keeps the properties''.
        \label{fig:HopfieldToTransformer}}
\end{figure}

Therefore, we generalize the
energy function of \citet{Demircigil:17} that builds on exponential interaction functions
to continuous patterns and states and obtain 
a new modern Hopfield network.
We also propose a new update rule which
ensures global
convergence to stationary points of the energy (local minima or saddle points). 
We prove that our new modern Hopfield network typically
retrieves patterns in one update step ($\epsilon$-close to the fixed point)
with an exponentially low error
and has a storage capacity proportional to $c^{\frac{d-1}{4}}$ (reasonable settings for
$c=1.37$ and $c=3.15$ are given in Theorem~\ref{th:storage}).
The retrieval of patterns with one update is important to integrate 
Hopfield networks in deep learning architectures, 
where layers are activated only once.
Surprisingly, our new update rule is also 
the key-value attention 
as used in transformer and BERT models (see Fig.~\ref{fig:HopfieldToTransformer}).
Our modern Hopfield networks can be integrated as a new layer
in deep learning architectures for pooling, memory, prototype learning, and attention.
We test these new layers on different benchmark datasets and tasks like
immune repertoire classification.

\section{Modern Hopfield Nets with Continuous States}
\label{sec:theory}

\paragraph{New energy function for continuous state Hopfield networks.}
In order to integrate modern Hopfield networks into deep learning
architectures, we have to make them continuous.
To allow for continuous states, we propose a new energy function that is 
a modification of the energy of modern Hopfield networks \citep{Demircigil:17}. 
We also propose a new update rule which 
can be proven to converge to 
stationary points of the energy (local minima or saddle points).

We have $N$ stored (key) patterns $\Bx_i \in \dR^d$ represented by the matrix
$\BX=  \left( \Bx_1,\ldots,\Bx_N \right)$ with
the largest pattern $M = \max_{i} \NRM{\Bx_i}$.
The state (query) pattern is $\Bxi \in \dR^d$.  
For exponential interaction functions, 
we need the {\em log-sum-exp function} ($\mathrm{lse}$) for $0 < \beta$
\begin{align}
  \mathrm{lse}(\beta,\Bx) \ &= \ \beta^{-1} \log \left( \sum_{i=1}^N
    \exp(\beta x_i) \right) \ , 
\end{align}
which is convex (see appendix Eq.~\eqref{th:deflse}, 
and Lemma~\ref{th:LjacobiDefinite}). 
The energy function $\rE$ of the
modern Hopfield networks for binary patterns $\Bx_i$ and a binary state pattern $\Bxi$
is $ \rE= - \sum_{i=1}^N F\left( \Bxi^T \Bx_i \right)$ \citep{Krotov:16}.
Here, $F(x)=x^n$ is the interaction function, where $n=2$ gives the classical 
Hopfield network.
The storage capacity is proportional to $d^{n-1}$ \citep{Krotov:16}.
This model was generalized by \citet{Demircigil:17}
to exponential interaction functions 
$F(x)=\exp(x)$ which gives the energy
$\rE= - \exp(\mathrm{lse}(1, \BX^T \Bxi))$.
This energy leads to an exponential 
storage capacity of $N=2^{d/2}$ for binary patterns.
Furthermore, with a single update, the fixed point 
is recovered with high probability for random patterns.
However, still this modern Hopfield network has binary states.

We generalize this energy function to continuous-valued patterns
while keeping the properties of the modern Hopfield networks like
the exponential storage capacity and the extremely fast convergence
(see Fig.~\ref{fig:HopfieldToTransformer}).
For the new energy we take the logarithm 
of the negative energy of modern Hopfield networks
and add a quadratic term of the current state. 
The quadratic term ensures that the 
norm of the state vector $\Bxi$ remains finite and the energy is bounded.
Classical Hopfield networks do not require to bound the norm of their state vector, 
since it is binary and has fixed length.
We define the novel energy function $\rE$ as
\begin{align}
  \rE \ &= \ - \ \mathrm{lse}(\beta ,\BX^T \Bxi) \ + \
  \frac{1}{2} \Bxi^T \Bxi  \ + \ \beta^{-1} \log N \ + \ 
  \frac{1}{2} M^2 \ .
\end{align}
We have $0 \leq \rE \leq 2 M^2$ (see appendix Lemma~\ref{th:energy}).
Using $\Bp =  \soft ( \beta \BX^T \Bxi)$, we define a novel update rule 
(see Fig.~\ref{fig:HopfieldToTransformer}):
\begin{align}
\label{eq:update}
\Bxi^{\nn} \ &= \ f(\Bxi) \ = \ \BX \Bp \ = \   
\BX \soft ( \beta \BX^T \Bxi) \ .
\end{align}
The next theorem states that the 
update rule Eq.~\eqref{eq:update}   
converges globally.
The proof uses the Concave-Convex Procedure (CCCP) \citep{Yuille:02,Yuille:03},
which is equivalent 
to Legendre minimization \citep{Rangarajan:96,Rangarajan:99}
algorithms \citep{Yuille:03}.
\begin{theorem}
\label{th:globalConv}
The update rule Eq.~\eqref{eq:update} 
converges globally:
For $\Bxi^{t+1} = f(\Bxi^t)$, 
the energy $\rE(\Bxi^t) \to \rE(\Bxi^*)$ for $t \to \infty$
and a fixed point $\Bxi^*$.
\end{theorem}
\begin{proof}
The update rule in Eq.~\eqref{eq:update} 
is the CCCP for minimizing the energy $\rE$, which is the sum of the convex
$1/2 \Bxi^T \Bxi$ and concave $-\mathrm{lse}$ (see details in appendix Theorem~\ref{th:globalConv}).
Theorem~2 in \citet{Yuille:02} yields the global convergence property.
Also, in Theorem~2 in \citet{Sriperumbudur:09} the global convergence of CCCP is proven
via a rigorous analysis using Zangwill's global convergence theory 
of iterative algorithms.
\end{proof}

The global convergence theorem only assures that
for the energy $\rE(\Bxi^t) \to \rE(\Bxi^*)$ for $t \to \infty$ 
but not $\Bxi^t \to \Bxi^*$. 
The next theorem strengthens
Zangwill's global convergence theorem \citep{Meyer:76}
and gives convergence results similar to
those known for expectation maximization \citep{Wu:83}.
\begin{theorem}
\label{th:globalConvergence}
For the iteration Eq.~\eqref{eq:update}
we have $\rE\left(\Bxi^t \right) \to \rE \left(\Bxi^* \right) = \rE^*$
as  $t \to \infty$, for some stationary point $\Bxi^*$. 
Furthermore, $\NRM{\Bxi^{t+1}- \Bxi^t} \to 0$ and
either $\{ \Bxi^t \}_{t=0}^{\infty}$ converges
or, in the other case, the set of limit points of  $\{ \Bxi^t \}_{t=0}^{\infty}$ 
is a connected and compact subset of $\cL \left( \rE^* \right)$, where
$\cL \left( a\right)= \{\Bxi \in \cL \mid \rE\left(\Bxi \right)  = a\} $
and $\cL$ is the set of stationary points of the iteration Eq.~\eqref{eq:update}.
If $\cL \left( \rE^* \right)$ is finite, 
then any sequence $\{ \Bxi^t \}_{t=0}^{\infty}$ 
generated by the iteration Eq.~\eqref{eq:update} 
converges to some $\Bxi^* \in \cL\left( \rE^*\right)$.
\end{theorem}
For a proof, see appendix Theorem~\ref{th:globalConvergence}.
Therefore, all the limit points of any sequence generated by 
the iteration Eq.~\eqref{eq:update} are 
stationary points (local minima or saddle points) of the 
energy function $\rE$. Either the iteration converges or,
otherwise, the set of limit points 
is a connected and compact set.

The next theorem gives the results on the storage capacity
of our new continuous state modern Hopfield network. 
We first define what we mean by storing and retrieving patterns
using a modern Hopfield network with continuous states.
\begin{definition}[Pattern Stored and Retrieved]
We assume that around every pattern $\Bx_i$ a sphere $\rS_i$ is given.
We say $\Bx_i$ {\em is stored} if there is a single fixed point $\Bx_i^* \in \rS_i$ to
which all points $\Bxi \in \rS_i$ converge,
and  $\rS_i \cap \rS_j = \emptyset$ for $i \not= j$.
We say $\Bx_i$ {\em is retrieved} for a given $\epsilon$ if 
iteration (update rule) Eq.~\eqref{eq:update} gives
a point $\tilde{\Bx}_i$ that is at least 
$\epsilon$-close to the single fixed point $\Bx_i^* \in \rS_i$. 
The retrieval error is $\NRM{\tilde{\Bx}_i - \Bx_i}$.
\end{definition}
As with classical Hopfield networks, we consider patterns on the sphere, 
i.e.\ patterns with a fixed norm. 
For randomly chosen patterns, the number of patterns that can be stored
is exponential in the dimension $d$ of the space of the patterns ($\Bx_i \in \dR^d$).
\begin{theorem}
\label{th:storage}
We assume a failure probability $0<p\leq 1$ and randomly chosen patterns 
on the sphere with radius $M:=K \sqrt{d-1}$. 
We define $a := \frac{2}{d-1}  (1 + \ln(2 \beta K^2 p (d-1)))$, 
$b := \frac{2  K^2  \beta}{5}$,
and $c:= \frac{b}{W_0(\exp(a + \ln(b))}$,
where $W_0$ is the upper branch of the Lambert $W$ function \dlmf{4.13},
and ensure $c \geq \left( \frac{2}{ \sqrt{p}}\right)^{\frac{4}{d-1}}$.
Then with probability $1-p$, the number of random patterns 
that can be stored is  
\begin{align} 
 \label{eq:CapacityM}
    N \ &\geq \ \sqrt{p} \ c^{\frac{d-1}{4}}  \ .
\end{align}
Therefore it is proven for $c\geq 3.1546$ with
$\beta=1$, $K=3$, $d= 20$ and $p=0.001$ ($a + \ln(b)>1.27$)
and proven for $c\geq 1.3718$ with $\beta = 1$, $K=1$, $d = 75$, and $p=0.001$
($a + \ln(b)<-0.94$).
\end{theorem}
For a proof, see appendix Theorem~\ref{th:mainStorage}. 

The next theorem states that the update rule typically
retrieves patterns after one update. 
Retrieval of a pattern $\Bx_i$ for fixed point $\Bx_i^*$ and query $\Bxi$
is defined via an $\epsilon$ by $\NRM{f(\Bxi) \ - \ \Bx_i^*} < \epsilon$,
that is, the update is $\epsilon$-close to the fixed point.
Retrieval with one update 
is crucial to integrate modern Hopfield networks into
deep learning architectures, where layers are activated only once.
First we need the concept of separation of a pattern.
For pattern $\Bx_i$ we define its separation $\Delta_i$
to other patterns by:
\begin{align}
 \Delta_i \ &:= \ \min_{j, j \not= i} \left( \Bx_i^T \Bx_i \ - \ \Bx_i^T
    \Bx_j \right) \ = \ \Bx_i^T \Bx_i \ - \ \max_{j, j \not= i} \Bx_i^T
  \Bx_j \ .
\end{align}
The update rule retrieves patterns
with one update for well separated patterns, that is, patterns with large
$\Delta_i$.
\begin{theorem}
With query $\Bxi$, after one update the distance of the new point $f(\Bxi)$
to the fixed point $\Bx_i^*$ is exponentially small in the separation $\Delta_i$.
The precise bounds using the Jacobian $\rJ = \frac{\partial
  f(\Bxi)}{\partial \Bxi}$ and its value $\rJ^m$ in the mean value
theorem are:
\begin{align}
  \NRM{f(\Bxi) \ - \ \Bx_i^*}
  \ &\leq \  \NRM{\rJ^m}_2 \ \NRM{\Bxi \ - \ \Bx_i^*}  \ , \\
   \NRM{\rJ^m}_2  \ &\leq \
  2 \ \beta \ N \ M^2 \ (N-1) \exp(- \ \beta \
  (\Delta_i \ - \ 2 \  \max \{ \NRM{\Bxi  \ - \ \Bx_i} , \NRM{\Bx_i^* \ - \ \Bx_i} \}  \ M) )\ .
\end{align}
For given $\epsilon$ and 
sufficient large $\Delta_i$, we have $\NRM{f(\Bxi) \ - \ \Bx_i^*} < \epsilon$,
that is, retrieval with one update.
\end{theorem}
See proof in appendix Theorem~\ref{th:oneUpdate}.

At the same time,
the retrieval error decreases exponentially with the separation $\Delta_i$.
\begin{theorem}[Exponentially Small Retrieval Error]
The retrieval error $\NRM{f(\Bxi) \ - \ \Bx_i}$ of pattern $\Bx_i$
is bounded by
\begin{align}
   \NRM{f(\Bxi) \ - \ \Bx_i} \ &\leq \ 2 \ (N-1) \ \exp(- \ \beta \ 
   (\Delta_i \ - \ 2 \   \max \{ \NRM{\Bxi  \ - \ \Bx_i} , \NRM{\Bx_i^* \ - \ \Bx_i} \} 
   \ M) )  \ M  
 \end{align}
and for 
$\NRM{\Bx_i - \Bx_i^*} \leq \frac{1}{2 \ \beta \ M } $ 
together with $\NRM{\Bx_i - \Bxi} \leq \frac{1}{2 \ \beta \ M } $
by
\begin{align}
 \NRM{\Bx_i \ - \ \Bx_i^*} \ &\leq \ 2 \ e \ (N-1) \ M \ \exp(- \ \beta \ \Delta_i )  \ .  
\end{align}
\end{theorem}
See proof in appendix Theorem~\ref{th:retrievalError}.

\paragraph{Metastable states and one global fixed point.}
So far, we considered patterns $\Bx_i$ that are well separated
and the iteration converges to a fixed point which is
near a pattern $\Bx_i$. 
If no pattern $\Bx_i$ is well separated from the others,
then the iteration converges to a global fixed point close to
the arithmetic mean of the vectors.
In this case the $\soft$ vector $\Bp$ is close to uniform, that is, $p_i=1/N$.
If some vectors are similar to each other and well separated from all
other vectors, then a metastable state near the similar
vectors exists. Iterations that start near the metastable state converge
to this metastable state, also if initialized by one of the similar patterns.
For convergence proofs to one global fixed point
and to metastable states see appendix 
Lemma~\ref{th:banach} and Lemma~\ref{th:banachM}, respectively.

\paragraph{Hopfield update rule is attention of the transformer.}
The Hopfield network update rule 
is the attention mechanism used
in transformer and BERT models
(see Fig.~\ref{fig:HopfieldToTransformer}).
To see this, we assume $N$ stored (key) patterns $\By_i$ 
and $S$ state (query) patterns $\Br_i$ that are mapped to the
Hopfield space of dimension $d_k$.
We set $\Bx_i = \BW_K^T \By_i$, $\Bxi_i = \BW_Q^T \Br_i$,
and multiply the result of our update rule with $\BW_V$.
The matrices $\BY=(\By_1,\ldots,\By_N)^T$ and $\BR=(\Br_1,\ldots,\Br_S)^T$ combine the $\By_i$ and $\Br_i$ 
as row vectors.
We define the matrices $\BX^T=\BK = \BY \BW_K $, $\BXi^T = \BQ = \BR \BW_Q$,
and $\BV=\BY \BW_K \BW_V=\BX^T \BW_V$, where 
$\BW_K \in \dR^{d_y\times d_k}, \BW_Q \in \dR^{d_r\times d_k}, \BW_V \in \dR^{d_k\times d_v}$. 
If $\beta = 1/\sqrt{d_k}$ and $\soft \in \dR^N$ is changed to a row vector, we obtain
for the update rule Eq.~\eqref{eq:update} multiplied by $\BW_V$:
\begin{align}
 \label{eq:transformer_attention}
 \BZ \ = \ \soft \left( 1/\sqrt{d_k} \ \BQ \ \BK^T \right) \ \BV \ = \ \soft \left( \beta \ \BR \ \BW_{\BQ} \ \BW_{\BK}^T\BY^T \right) \ \BY \ \BW_{\BK}\BW_{\BV} \ .
\end{align}
The left part of Eq.~\eqref{eq:transformer_attention} 
is the transformer attention. In the transformer self-attention $\BR=\BY$, and 
$\BW_{\BK}\BW_{\BV}$ replaced by just $\BW_{\BV}$.
Besides the attention mechanism,
Hopfield networks allow for other functionalities
in deep network architectures,
which we introduce via specific layers 
in the next section. 
The right part of Eq.~\eqref{eq:transformer_attention} 
serves to explain these specific layers.

\section{New Hopfield Layers for Deep Learning}
Modern Hopfield networks with continuous states 
can be integrated into deep learning
architectures, because they are continuous and
differentiable with respect to their parameters.
Furthermore, they typically retrieve patterns with
one update, which is conform to deep learning layers
that are activated only once.
For these two reasons, modern Hopfield networks
can serve as specialized layers in 
deep networks to equip them with memories.
Below, we introduce three types of Hopfield layers: 
{\tt Hopfield}, {\tt HopfieldPooling}, and 
{\tt HopfieldLayer}.
Possible applications of Hopfield layers
in deep network architectures comprise:
\begin{itemize}
\item multiple instance learning (MIL) \citep{Dietterich:97},
\item processing of and learning with point sets \citep{Qi:17ieee,Qi:17,Xu:18spider},
\item set-based and permutation invariant learning \citep{Guttenberg:16,Ravanbakhsh:16,Zaheer:17,Korshunova:18,Ilse:18,Zhai:20},
\item attention-based learning \citep{Vaswani:17},
\item deep learning with associative memories \citep{Graves:14,Weston:14,Ba:16,Ba:16arxiv,Schlag:18,Schlag:19},
\item natural language processing \citep{Devlin:18,Devlin:19},
\item sequence analysis and time series prediction \citep{Hochreiter:91a,Hochreiter:97,Cho:14}, and
\item storing and retrieving reference data, e.g.\ 
the training data, outliers, high error data points, 
prototypes or cluster centers, support vectors \& border cases.
\end{itemize}

Hopfield network layers can substitute
existing layers like 
pooling layers, 
permutation equivariant layers \citep{Guttenberg:16,Ravanbakhsh:16}, 
GRU \citep{Cho:14} \& 
LSTM \citep{Hochreiter:91a,Hochreiter:97} layers, and 
attention layers \citep{Vaswani:17,Vaswani:17arxiv, Bahdanau:14}.


\setlength\intextsep{0pt}
\begin{wrapfigure}[19]{r}{0.35\textwidth}
        \centering
        \includegraphics[width=0.35\textwidth]{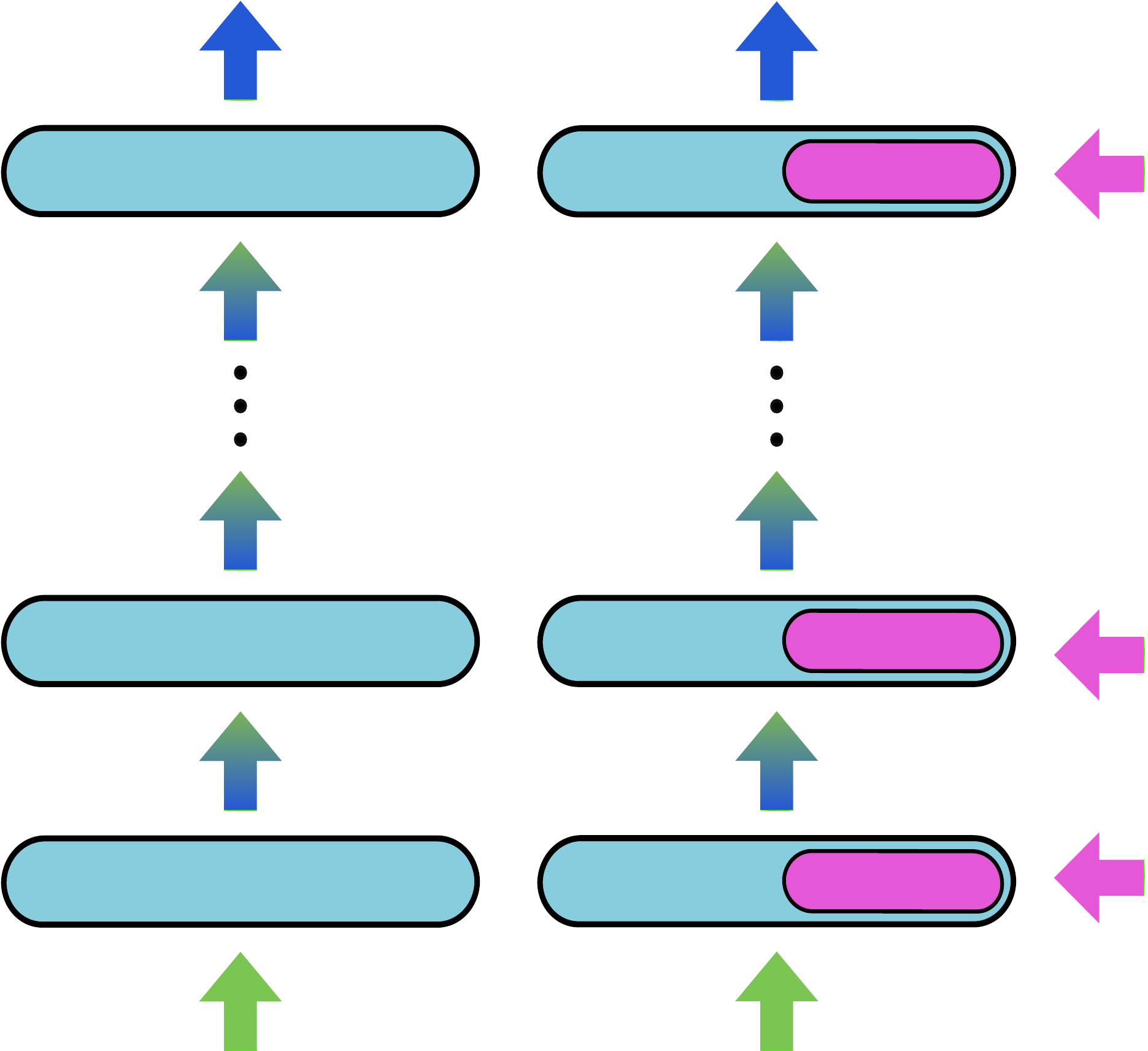}
        \caption[]{Left: A standard deep network with layers 
        ({\color{HubiLightBlue} $\blacksquare$}) propagates 
        either a vector or a set of vectors from the input to the output.
        Right: A deep network, where layers ({\color{HubiLightBlue} $\blacksquare$})
        are equipped with associative memories
        via Hopfield layers ({\color{HubiPink} $\blacksquare$}).
        \label{fig:NNlayers}}
\end{wrapfigure}%

\paragraph{Types of neural networks.}
We consider two types of feed-forward neural networks: 
(I) Neural networks that propagate an activation vector from the input layer 
to the output layer. Examples are fully-connected or convolutional neural networks.
(II) Neural networks that propagate a set of vectors from the input layer to
the output layer, where each layer applies the same operation to each element of the set 
and the output layer may summarize the set via a vector.
An example is the transformer.
Recurrent neural networks are networks of type (I), 
which are iteratively applied to a set or a sequence,
where intermediate results are stored in a memory and can be reused.
Modern Hopfield networks can be integrated into both types of 
neural network architectures and enable to equip each of their layers
with associative memories. See Fig.~\ref{fig:NNlayers}.

\begin{figure}[ht]
        \centering
        \includegraphics[width=1.0\textwidth]{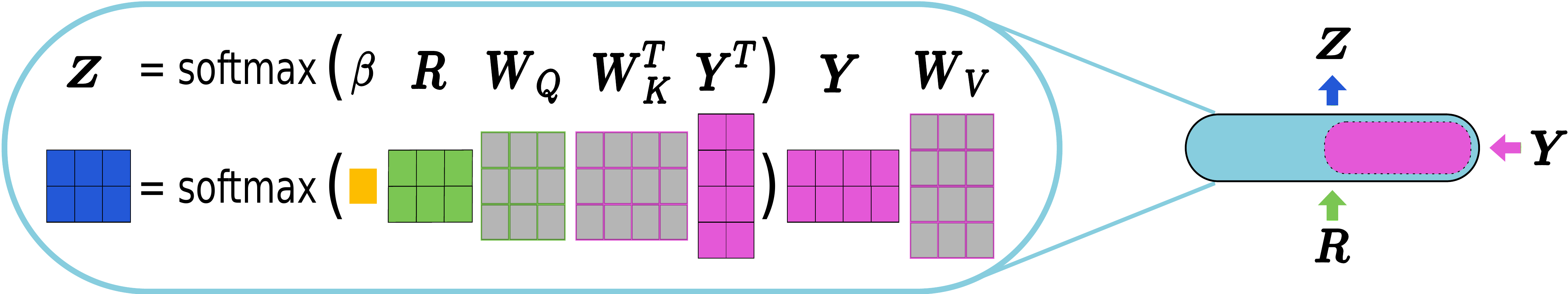}
        \caption[]{The layer {\tt Hopfield} allows the association of two sets $\BR$ ({\color{HubiGreen} $\blacksquare$}) and $\BY$ ({\color{HubiPink} $\blacksquare$}).
        It can be integrated into deep networks that propagate sets of vectors.
        The Hopfield memory 
        is filled
        with a set from either the input or previous layers. 
        The output is a set of vectors $\BZ$ ({\color{HubiBlue} $\blacksquare$}).
        \label{fig:HopfieldLayer}}
\end{figure}

\paragraph{Types of new Hopfield layers.}
We introduce three types of Hopfield layers: 
{\tt Hopfield}, {\tt HopfieldPooling}, and 
{\tt HopfieldLayer}.
The continuous modern Hopfield network
results in a plethora of new deep learning architectures, 
since we can  
(a) propagate sets or single vectors,
(b) propagate queries, stored patterns, or both, 
(c) learn static queries or stored patterns,
(d) fill the memory by training sets, prototypes, or external data.
Next, we provide three useful types of Hopfield layers.
The implementation is available at: \url{https://github.com/ml-jku/hopfield-layers}

{\bf (1)} Layer {\tt Hopfield} 
for networks that {\bf propagate sets of vectors via 
state (query) patterns $\BR$ and stored (key) patterns $\BY$}. 
The layer {\tt Hopfield} is the realization of 
formula~\eqref{eq:transformer_attention}.
The memory of the {\tt Hopfield} layer can be 
{\em filled with sets from the input or previous layers},
see Fig.~\ref{fig:HopfieldLayer}.
The memory may be filled with a reference set, 
which is covered by providing the reference set as additional input.
Thus, the layer {\tt Hopfield} allows the association of two sets.  
A prominent example of a layer that performs such association
is the transformer attention mechanism, 
which associates keys and queries, e.g.\
two point sets that have to be compared. 
This layer allows for different kinds of
sequence-to-sequence learning, 
point set operations, 
and retrieval-based methods.
The layer {\tt Hopfield} with skip connections in a ResNet architecture
is identical to the popular transformer and BERT models.
In the experiments, we analyzed these Hopfield layers in transformer architectures.
In our experiments in which we compare machine learning methods
on small datasets of the UCI benchmark collection
the layer {\tt Hopfield} is also used.

\vspace{3mm}
\vspace{3mm}

\begin{figure}[ht]
        \centering
        \includegraphics[width=1.0\textwidth]{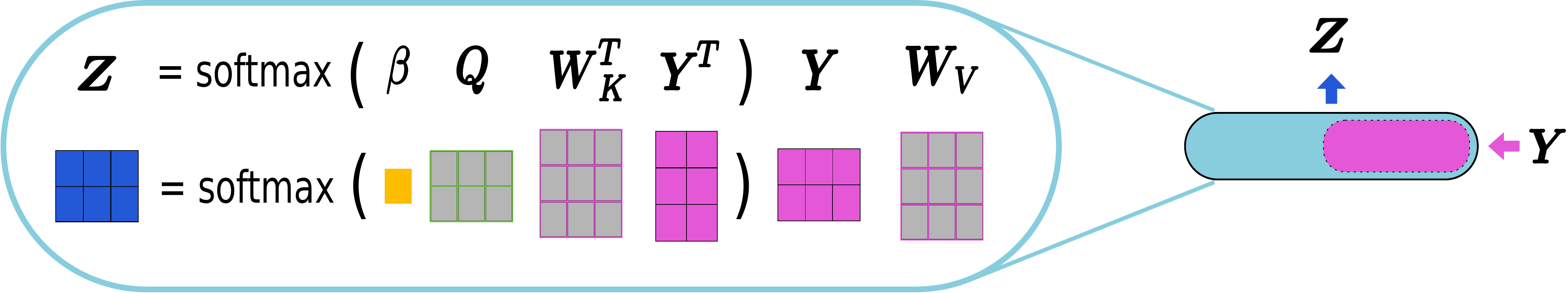}
        \caption[]{The layer {\tt HopfieldPooling} enables 
        pooling or summarization of sets,
        which are obtained from the input or from previous layers. 
        The input $\BY$ ({\color{HubiPink} $\blacksquare$}) 
        can be either a set or a sequence.
        The query patterns of each layer are static and can be learned. 
        The output is a set of vectors $\BZ$ ({\color{HubiBlue} $\blacksquare$}), 
        where the number of vectors equals the number of query patterns.
        The layer {\tt HopfieldPooling} can realize multiple instance learning.        
          \label{fig:HopfieldPooling}}
\end{figure}

{\bf (2)} Layer {\tt HopfieldPooling} 
for networks that {\bf propagate patterns via the stored (key) patterns $\BY$}.
This layer performs 
a pooling or summarization of sets $\BY$ obtained 
from queries in previous layers or the input.
The memory of the {\tt HopfieldPooling} layer is
{\em filled with sets from the input or previous layers}.
The {\tt HopfieldPooling} layer uses the queries to search for 
patterns in the memory, the stored set. 
If more patterns are similar to
a particular search pattern (query), then the result is an average 
over these patterns. 
The state (query) patterns of each layer are static and can be learned.
Multiple queries supply a set to the next layer, where each query
corresponds to one element of the set.
Thus, the layer {\tt HopfieldPooling} enables
fixed pattern search, 
pooling operations, 
and memories like LSTMs or GRUs.
The static pattern functionality is typically needed if particular
patterns must be identified in the data. \\
A single {\tt HopfieldPooling} layer allows for 
multiple instance learning.
Static state (query) patterns together with position encoding in the
keys allows for performing pooling operations. The position encoding can
be two-dimensional, where standard convolutional filters can be
constructed as in convolutional neural networks (CNNs).
The {\tt HopfieldPooling} layer can substitute pooling, averaging, LSTM,  
and permutation equivariant layers. 
See Fig.~\ref{fig:HopfieldPooling}.
The layer {\tt HopfieldPooling} is used for
experiments with multiple instance learning tasks, e.g.\ 
for immune repertoire classification in the experiments.

\vspace{3mm}
\vspace{3mm}

\begin{figure}[ht]
        \centering
        \includegraphics[width=1.0\textwidth]{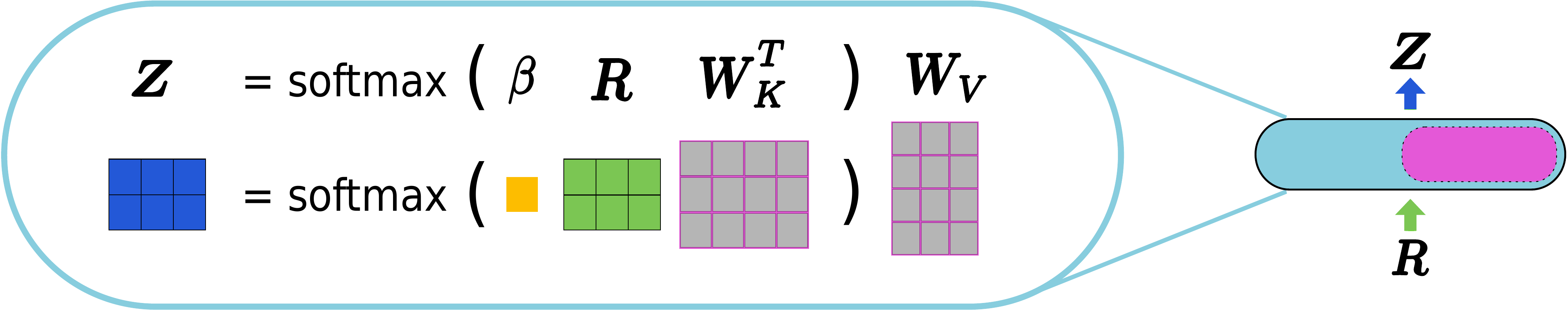}
        \caption[]{The layer {\tt HopfieldLayer} enables
        multiple queries of the training set,
        a reference set, prototype set, or a learned set (a learned matrix).
        The queries for each layer are computed from the results of previous layers.
        The input is a set of vectors $\BR$ ({\color{HubiGreen} $\blacksquare$}). The output is also a set of vectors $\BZ$ ({\color{HubiBlue} $\blacksquare$}),
        where the number of output vectors equals the number of input vectors.
        The layer {\tt HopfieldLayer} can realize 
        SVM models, $k$-nearest neighbor, and LVQ. \label{fig:LayerHopfieldLayer}}
\end{figure}

{\bf (3)} Layer {\tt HopfieldLayer} 
for networks that {\bf propagate a vector or a set of vectors via state (query) patterns $\BR$}.
The queries $\BR$ can be input vectors or queries
that are computed from the output of previous layers.
The memory of the {\tt HopfieldLayer} layer is
{\em filled with a fixed set}, which can be the training set,
a reference set, prototype set, or a learned set (a learned matrix).
The stored (key) patterns are static and can be learned.
If the training set is stored in the memory, 
then each layer constructs a new set of queries based on the
query results of previous layers.
The stored patterns can be initialized by the training set or 
a reference set and then learned, in which case they deviate from the training set. 
The stored patterns can be interpreted as weights from the state (query)
to hidden neurons that have a softmax activation function \citep{Krotov:20}. 
The layer {\tt HopfieldLayer} can substitute a fully connected layer,
see Fig.~\ref{fig:LayerHopfieldLayer}.
A single {\tt HopfieldLayer} layer also allows for 
approaches similar to support vector machines (SVMs), 
approaches similar to $k$-nearest neighbor, 
approaches similar to learning vector quantization,
and pattern search.
For classification, the raw data 
$\By_i=(\Bz_i,\Bt_i)$ can be the concatenation of input $\Bz_i$ and target $\Bt_i$.
In this case, the matrices $\BW_K$ and  $\BW_V$
can be designed such that 
inside the softmax the input $\Bz_i$ is used and outside the softmax
the target $\Bt_i$. 
Thus, the softmax provides a weighted average of the target vectors
based on the similarity between the query and the inputs.
Also SVM models, $k$-nearest neighbor, and learning vector quantization can
be considered as weighted averages of the targets.
The encoder-decoder attention layer of the transformers
are a {\tt HopfieldLayer} layer, where the memory is filled
with the encoder output set.
In our experiments with the drug design benchmark datasets,
the layer {\tt HopfieldLayer} has been applied and compared to 
other machine learning methods.

\vspace{3mm}
\vspace{3mm}

\paragraph{Additional functionality of new Hopfield layers.}
The insights about energy, convergence, and storage properties
provide all new Hopfield layers with additional functionalities:
i) {\em multiple updates} to control how
precise fixed points are found without additional parameters needed.
ii) {\em variable $\beta$} to determine the kind of fixed points
such as the size of metastable states.
The variable $\beta$ controls over how many patterns is averaged.
As observed in the experiments,
the variable is relevant in combination with the learning rate to steer the learning
dynamics.
The parameter $\beta$ governs the fixed point dynamics
and can be learned, too.
iii) {\em controlling the storage capacity} via the dimension of the
associative space. The storage capacity can be relevant 
for tasks with a huge number of instances as in
the immune repertoire classification experiment.
iv) {\em pattern normalization} controls, like the layernorm, 
the fixed point dynamics by the norm and shift of the patterns. 
For more details see appendix, Section~\ref{chap:Hopfield_layer}.

\section{Experiments}
We show that our proposed Hopfield layers can be applied successfully
to a wide range of tasks.
The tasks are from natural language processing,
contain multiple instance learning problems, 
a collection of small classification tasks,
and drug design problems.

\paragraph{Analysis of transformer and BERT models.}
\label{sec:Analysis}
Transformer and BERT models can be implemented by the layer {\tt Hopfield}.
The kind of fixed point of the Hopfield net is determined by
how the pattern $\Bx_i$ is separated from others patterns.
(a) {\em a global fixed point}: no separation of a pattern from the others,
(b) {\em a fixed point close to a single pattern}: pattern is
separated from other patterns,
(c) {\em metastable state}: some patterns are similar 
to each other and well separated from all other vectors.
We observed that the attention heads of transformer and BERT models 
are predominantly in metastable states, which
are categorized into four classes:
(I) averaging over a very large number of patterns (very large metastable state or fixed point (a)), 
(II) averaging over a large number of patterns (large metastable state),
(III) averaging over a medium number of patterns (medium metastable state),
(IV) averaging over a small number of patterns (small metastable state or fixed point (c)).
For analyzing the metastable states, we calculated 
the minimal number $k$ of $\soft$ values required to sum up to $0.90$. 
Hence, $k$ indicates the size of a metastable state.
To determine in which of the four classes a head is mainly operating,
we computed the distribution of $k$ across sequences.
Concretely, for $N$ tokens and for $\bar{k}$ as the median of the distribution, 
a head is classified 
as operating in class (I) if  $1/2 N < \bar{k}$, 
as operating in class (II) if $1/8 N < \bar{k} \leq  1/2 N$,
as operating in class (III) if  $1/32 N < \bar{k} \leq 1/8 N$, and
as operating in class (IV) if $\bar{k} \leq 1/32 N$.
We analyzed pre-trained BERT models from Hugging 
Face Inc.\ \citep{Wolf:19} according to these operating classes. 
In Fig.~\ref{fig:bert_analysis} in the appendix the distribution of the pre-trained 
bert-base-cased
model is depicted (for other models see 
appendix Section~\ref{sec:attention_learning_dynamics}).
Operating classes (II) (large metastable states) and 
(IV) (small metastable states) 
are often observed in the middle layers.
Operating class (I) (averaging over a very large number of patterns)
is abundant in lower layers.
Similar observations have been reported in other studies \citep{Toneva:19,Toneva:19arxiv,Tay:20}. 
Operating class (III) (medium metastable states) is predominant in the last layers.

\paragraph{Multiple Instance Learning Datasets.}
For multiple instance learning (MIL) \citep{Dietterich:97}, 
we integrate our new Hopfield network
via the layer {\tt HopfieldPooling} into deep learning architectures.
Recently, deep learning methods 
have been applied to MIL problems \citep{Ilse:18}, but still
the performance on many datasets lacks improvement. 
Thus, MIL datasets still pose an 
interesting challenge, in which Hopfield layers equipped with memory 
are a promising approach.

{\textbullet \em Immune Repertoire Classification.} 
The first MIL task is immune repertoire classification,
where a deep learning architecture with {\tt HopfieldPooling} 
(DeepRC) was used \citep{Widrich:20,Widrich:20nips}.
Immune repertoire classification \citep{Emerson:17}
typically requires 
to extract few patterns from a large set of sequences, the repertoire, 
that are indicative for the respective immune status.
The datasets contain $\approx$ 
300,000 instances per immune repertoire,
which represents one of the largest multiple instance 
learning experiments ever conducted \citep{Carbonneau:18}.
Most MIL methods fail due the large number of instances. 
This experiment comprises real-world and simulated datasets. 
Simulated datasets are generated by implanting 
sequence motifs \citep{Akbar:19,Weber:19} with low 
frequency into simulated or experimentally-observed immune receptor sequences.
The performance of 
DeepRC was compared with other machine learning methods:
(i) known motif,
(ii) SVM using $k$-mers
and MinMax or Jaccard kernel,
(iii) $K$-Nearest Neighbor (KNN) with $k$-mers,
(iv) logistic regression with $k$-mers,
(v) burden test with $k$-mers, and
(vi) logistic multiple instance learning (lMIL).
On the real-world dataset DeepRC achieved
an AUC of $0.832\pm0.022$, followed by
the SVM with MinMax kernel (AUC $0.825\pm0.022$) and the burden
test with an AUC of $0.699\pm0.041$.
Across datasets, DeepRC outperformed all competing methods 
with respect to average AUC \citep{Widrich:20,Widrich:20nips}. 

{\textbullet \em  MIL benchmark datasets.}
We apply Hopfield layers to further 
MIL datasets \citep{Ilse:18,Kuccukacsci:18, Cheplygina:16}:
Elephant, Fox and Tiger  
for image annotation \citep{Andrews:03}.
These datasets consist of color images from the Corel dataset 
that have been preprocessed and
segmented. An image consists
of a set of segments (or blobs), 
each characterized by color, texture and shape
descriptors. 
The datasets have 100 positive and 100 negative
example images. 
The latter have been randomly drawn from a pool of photos of
other animals. 
Elephant comprises 1,391 instances and 230 features, 
Fox 1,320 instances and 230 features, and
Tiger has 1,220 instances and 230 features.
Furthermore, we use the 
UCSB breast cancer classification \citep{Kandemir:14}
dataset, which consists of 2,002 instances across 58 input objects. An instance 
represents a patch of a histopathological image of cancerous or normal tissue.
The layer \texttt{HopfieldPooling} is used, which allows for 
computing a per-input-object representation by extracting an 
average of instances that are indicative for one of the two classes. 
The input to the layer \texttt{HopfieldPooling} 
is a set of embedded instances $\BY$. 
A trainable but fixed state (query) pattern $\BQ$ is used for averaging 
over class-indicative instances.
This averaging enables a compression of variable-sized bags to a 
fixed-sized representation to discriminate the bags. 
More details in appendix Sec.~\ref{sec:deeprc_details}. 
%
Our approach has set a new state-of-the-art and has 
outperformed other methods \citep{Kuccukacsci:18, Carbonneau:16}
on the datasets Tiger, Elephant and UCSB Breast Cancer 
(see Table~\ref{tab:MIL}).

\begin{table}[]
    \centering
    \resizebox{1.0\textwidth}{!}{
    \begin{threeparttable}
        \begin{tabular}{lcccc} \toprule
            Method &                             tiger & fox & elephant & UCSB   \\ \midrule
            Hopfield (ours)                      & $\mathbf{91.3\pm0.5}$
                                                 & $64.05\pm0.4$
                                                 & $\mathbf{94.9\pm0.3}$
                                                 & $\mathbf{89.5\pm0.8}$ \\ 
            Path encoding \citep{Kuccukacsci:18} & $91.0\pm1.0$\tnote{a}
                                                 & $71.2\pm1.4$\tnote{a}
                                                 & $94.4\pm0.7$\tnote{a}
                                                 & $88.0\pm2.2$\tnote{a} \\ 
            MInD \citep{Cheplygina:16}           & $85.3\pm1.1$\tnote{a}
                                                 & $70.4\pm1.6$\tnote{a}
                                                 & $93.6\pm0.9$\tnote{a}
                                                 & $83.1\pm2.7$\tnote{a} \\
            MILES \citep{Chen:06}                & $87.2\pm1.7$\tnote{b}
                                                 & $\mathbf{73.8\pm1.6}$\tnote{a}
                                                 & $92.7\pm0.7$\tnote{a}
                                                 & $83.3\pm2.6$\tnote{a} \\
            APR \citep{Dietterich:97}            & $77.8\pm0.7$\tnote{b}
                                                 & $54.1\pm0.9$\tnote{b} 
                                                 & $55.0\pm1.0$\tnote{b}
                                                 & --- \\
            Citation-kNN \citep{Wang:00}     	 & $85.5\pm0.9$\tnote{b}
                                                 & $63.5\pm1.5$\tnote{b}
                                                 & $89.6\pm0.9$\tnote{b}
                                                 & $70.6\pm3.2$\tnote{a} \\
            DD \citep{Maron:98}                  & $84.1$\tnote{b}
                                                 & $63.1$\tnote{b}
                                                 & $90.7$\tnote{b}
                                                 & --- \\
            
            \bottomrule
        \end{tabular}
     \caption[]{Results for MIL datasets Tiger, Fox, Elephant, and 
     UCSB Breast Cancer in terms of AUC. 
     Results for all methods except the first are taken from either 
     \tnote{a}\citep{Kuccukacsci:18} or \tnote{b}\citep{Carbonneau:16}, 
     depending on which reports the higher AUC.\label{tab:MIL} }
      \end{threeparttable}
      }
\end{table}

\paragraph{UCI Benchmark Collection.}

So far deep learning struggled with small datasets.
However, Hopfield networks are promising for handling small datasets,
since they can store the training data points or their representations
to perform similarity-based, nearest neighbor, or
learning vector quantization methods.
Therefore, we test the Hopfield layer {\tt Hopfield} on the small datasets of 
the UC Irvine (UCI) Machine Learning Repository that
have been used to benchmark
supervised learning methods \citep{Fernandez:14,Wainberg:16,Khan:18} 
and also feed-forward neural networks \citep{Klambauer:17,Wu:18}, where
our Hopfield networks could exploit their memory. 
The whole 121 datasets in the collection 
vary strongly with respect to their size, number of 
features, and difficulties \citep{Fernandez:14}, such that they have
been divided into 75 ``small datasets'' with less than 1,000 samples and
45 ``large datasets'' with more than or equal 
to 1,000 samples in \citet{Klambauer:17}.
\begin{wraptable}[11]{r}{0.5\textwidth}

\centering

\begin{tabular}{llc}
  \toprule
 Method       & avg. rank diff. & $p$-value    \\ \midrule
Hopfield (ours)& $\mathbf{-3.92}$   &   ---              \\ 
SVM           & $-3.23$   &    $0.15$   \\ 
SNN           & $-2.85$  &    $0.10$   \\ 
RandomForest  & $-2.79$  &    $0.05$   \\ 
\ldots           &   ~~~~\ldots       &   \ldots      \\ 
Stacking      & \phantom{$-$}$8.73$  &  $1.2$e$-11$ \\
   \bottomrule
\end{tabular}

\caption[]{Results on 75 small datasets of the UCI benchmarks 
given as difference to average rank. \label{tab:uci_results}}
\end{wraptable}
On the 75 small datasets, 
Random Forests (RFs) and Support Vector Machines (SVM) are highly accurate, whereas
on the large datasets, deep learning methods and neural networks are in the lead \citep{Klambauer:17,Klambauer:17arxiv,Wu:18}.
We applied a modern Hopfield network via the layer {\tt HopfieldLayer},
where a self-normalizing net (SNN) maps the
input vector to $\BY$ and $\BR$.
The output $\BZ$ of {\tt HopfieldLayer} enters
a softmax output. 
We compared our modern Hopfield networks against deep learning methods
(e.g.\ SNNs, resnet), 
RFs, SVMs, boosting, bagging, and many other machine
learning methods of \citet{Fernandez:14}.
Since for each method, multiple variants and implementations had been included, we 
used method groups and representatives as defined by \citet{Klambauer:17}. 
For each dataset, a ranking of the methods was calculated which is presented in 
Table~\ref{tab:uci_results}. 
We found that Hopfield networks outperform all other methods on the 
small datasets, setting a new state-of-the-art for 10 datasets. 
The difference is significant except for the first three runner-up methods
(Wilcoxon signed rank test).
See appendix Section~\ref{appsec:uci} for details.

\textbf{Drug Design Benchmark Datasets.}
We test the Hopfield layer {\tt HopfieldLayer}, on four drug design datasets.
These datasets represent four main areas of modeling tasks in drug design, 
concretely to develop accurate models for predicting  
a) new anti-virals (HIV) by the Drug Therapeutics Program (DTP) AIDS Antiviral Screen, 
b) new protein inhibitors, concretely human $\beta$-secretase (BACE) inhibitors by \citet{Subramanian:16},
c) metabolic effects as blood-brain barrier permeability (BBBP) \citep{Martins:12} and
d) side effects of a chemical compound from the Side Effect Resource (SIDER) \cite{Kuhn:16}.
We applied the Hopfield layer {\tt HopfieldLayer}, 
where the training data is used
as stored patterns $\BY$, the input vector as state pattern $\BR$, and 
the corresponding training label to project the output of
the Hopfield layer $\BY \BW_V$. 
%
Our architecture with {\tt HopfieldLayer} has reached state-of-the-art 
for predicting side 
effects on SIDER $0.672\pm0.019$ 
as well as for predicting $\beta$-secretase BACE $0.902\pm0.023$. 
For details, see Table~\ref{tab:drug_discovery} in the appendix.

{\bf Conclusion.}
We have introduced a modern Hopfield network with continuous states
and the corresponding new update rule.
This network can store exponentially many patterns, retrieves patterns with 
one update, and has exponentially small retrieval errors.
We analyzed the attention heads of BERT models.
The new modern Hopfield networks have been integrated 
into deep learning architectures 
as layers to allow the storage of and access to 
raw input data, intermediate results, or learned prototypes. 
These Hopfield layers enable new ways of deep learning, beyond 
fully-connected, convolutional, or recurrent networks,
and provide pooling, memory, association, and attention mechanisms.
Hopfield layers that equip neural network layers with memories 
improved state-of-the-art 
in three out of four
considered multiple instance learning problems and
on immune repertoire classification, and
on two drug design dataset. 
They yielded the best results among 
different machine learning methods on 
the UCI benchmark collections of small classification tasks. 

\section*{Acknowledgments}
The ELLIS Unit Linz, the LIT AI Lab and the 
Institute for Machine Learning are supported by
the Land Ober\"{o}sterreich,
LIT grants DeepToxGen (LIT-2017-3-YOU-003),
and AI-SNN (LIT-2018-6-YOU-214),	
the Medical Cognitive Computing Center (MC3),
Janssen Pharmaceutica,
UCB Biopharma,
Merck Group,
Audi.JKU Deep Learning Center, Audi Electronic Venture GmbH,
TGW,
Primal, S3AI (FFG-872172),
Silicon Austria Labs (SAL),
Anyline,
FILL,
EnliteAI,
Google Brain,
ZF Friedrichshafen AG,
Robert Bosch GmbH,
TÜV Austria,
DCS,
and the NVIDIA Corporation. 
IARAI is supported by Here Technologies.

\newpage
\appendix

\section{Appendix}

\resumetocwriting

This appendix consists of six sections (A.1--A.6).
Section~A.1 introduces the new modern Hopfield network with continuous
states and its update rule. Furthermore, Section~A.1 provides 
a thorough and profound theoretical analysis of 
this new Hopfield network. 
Section~A.2 provides the mathematical background for Section~A.1.
Section~A.3 reviews {\em binary} Modern Hopfield Networks of Krotov \&
Hopfield.
Section~A.4 shows that the Hopfield update rule is the attention
mechanism of the transformer.
Section~A.5 gives details on the
experiments.
Section~A.6 describes the PyTorch implementation of
layers based on the new Hopfield networks and how to use them.

\setcounter{tocdepth}{3}
\setcounter{secnumdepth}{4}
\renewcommand\thefigure{\thesection.\arabic{figure}}    
\setcounter{figure}{0}    
\renewcommand\thetable{\thesection.\arabic{table}}    
\setcounter{table}{0}

\renewcommand{\contentsname}{Contents of the appendix}
\tableofcontents

\renewcommand{\listtheoremname}{List of theorems}
\listoftheorems[ignoreall,show={theoremA}]

\renewcommand{\listtheoremname}{List of definitions}
\listoftheorems[ignoreall,show={definitionA}]

\renewcommand{\listfigurename}{List of figures}
\listoffigures

\renewcommand{\listtablename}{List of tables}
\listoftables

\subsection{Continuous State Modern Hopfield Networks (A New Concept)}
\label{sec:newHopfield}

\subsubsection{Introduction}
In Section~\ref{sec:newHopfield}
our new modern Hopfield network is introduced.
In Subsection~\ref{sec:newEnergy} we present the new energy function.
Then in Subsection~\ref{sec:newUpdate},
our new update rule is introduced.
In Subsection~\ref{sec:GlobalConv}, we show that this update rule 
ensures global convergence.
We show that all the limit points of any sequence generated by 
the update rule are the
stationary points (local minima or saddle points) of the 
energy function.
In Section~\ref{sec:LocalConv}, we consider the local convergence 
of the update rule and see that patterns are retrieved with one update.
In Subsection~\ref{sec:Properties}, we consider the 
properties of the fixed points that are associated with the
stored patterns.
In Subsection~\ref{sec:ExpStorage},
we show that exponentially many patterns can be stored.
The main result is given in Theorem~\ref{th:mainStorage}:
For random patterns on a sphere we can store and retrieve 
exponentially (in the dimension of the Hopfield space) many patterns. 
Subsection~\ref{sec:ConvError} reports that 
patterns are typically retrieved with one update step and that the
retrieval error is exponentially small.

In Subsection~\ref{sec:LearningAss}, we consider how
associations for the new Hopfield networks can be learned.
In Subsection~\ref{sec:LearningDirect}, we analyze if the
association is learned directly by a bilinear form.
In Subsection~\ref{sec:LearningMap}, we analyze if 
stored patterns and query patterns are mapped to the space of
the Hopfield network. Therefore, we treat the architecture of
the transformer and BERT. 
In Subsection~\ref{sec:sequential}, we introduce 
a temporal component into the new Hopfield network that
leads to a forgetting behavior. 
The forgetting allows us to treat infinite memory capacity 
in Subsection~\ref{sec:infinite}.
In Subsection~\ref{sec:forgetting}, we consider the controlled 
forgetting behavior.

In Section~\ref{sec:MathProperties}, we provide the mathematical
background that is needed for our proofs.
In particular we give lemmas on
properties of the softmax, 
the log-sum-exponential, the Legendre transform, and the 
Lambert $W$ function.

In Section~\ref{sec:Krotov}, we review the new Hopfield network
as introduced by Krotov and Hopfield in 2016. 
However in contrast to our new Hopfield network,
the Hopfield network of Krotov and Hopfield is binary, that is, a network with binary states.
In Subsection~\ref{sec:KrotovIntro}, we give an introduction to 
neural networks equipped with associative memories and new Hopfield networks.
In Subsection~\ref{sec:KrotovNN}, we discuss neural networks that are 
enhanced by an additional external memory and by attention mechanisms.
In Subsection~\ref{sec:KrotovOverview}, we give an overview over
the modern Hopfield networks.
Finally, in Subsection~\ref{sec:KrotovEnergy}, we present
the energy function and the update rule for the modern, binary 
Hopfield networks.

~~~~

~~~~

~~~~

~~~~

\subsubsection{New Energy Function}
\label{sec:newEnergy}

We have patterns $\Bx_1,\ldots,\Bx_N$
that are represented by the matrix
\begin{align}
  \BX \ &= \ \left( \Bx_1,\ldots,\Bx_N \right) \ .
\end{align}
The largest norm of a pattern is
\begin{align}
  M \ &= \ \max_{i} \NRM{\Bx_i}  \ .
\end{align}
The query or state of the Hopfield network is $\Bxi$.

The energy function $\rE$ 
in the new type of Hopfield models of Krotov and Hopfield 
is $ \rE= - \sum_{i=1}^N F\left( \Bxi^T \Bx_i \right)$ 
for binary patterns $\Bx_i$ and binary state $\Bxi$
with interaction function $F(x)=x^n$, where $n=2$ gives classical 
Hopfield model \citep{Krotov:16}.
The storage capacity is proportional to $d^{n-1}$ \citep{Krotov:16}.
This model was generalized by Demircigil et al.~\citep{Demircigil:17}
to exponential interaction functions
$F(x)=\exp(x)$, which gives the energy
$\rE= - \exp(\mathrm{lse}(1, \BX^T \Bxi))$.
This energy leads to an exponential 
storage capacity of $N=2^{d/2}$ for binary patterns.
Furthermore, with a single update the fixed point 
is recovered with high probability. See more details in Section~\ref{sec:Krotov}.

In contrast to the these binary modern Hopfield networks, 
we focus on modern Hopfield networks
with {\em continuous state}s that can store {\em continuous patterns}.
We generalize the energy of Demircigil et al.~\citep{Demircigil:17}
to continuous states while keeping the $\mathrm{lse}$ properties which
ensure high storage capacity and fast convergence.
Our new energy $\rE$ for a continuous query or state $\Bxi$ is defined as
\begin{align}
 \label{eq:DefEnergy}
  \rE \ &= \ - \ \mathrm{lse}(\beta ,\BX^T \Bxi) \ + \
  \frac{1}{2} \Bxi^T \Bxi  \ + \ \beta^{-1} \ln N \ + \ 
  \frac{1}{2} M^2 \\
  &= \ - \ \beta^{-1} \ln \left( \sum_{i=1}^N
    \exp(\beta \Bx_i^T \Bxi) \right)  \ + \ \beta^{-1} \ln N \ + \
  \frac{1}{2} \Bxi^T \Bxi \ + \ \frac{1}{2} M^2 \\
   &= \ - \ \beta^{-1} \ln  \left( \frac{1}{N} \ \sum_{i=1}^N
   \exp \left( - \ \frac{1}{2} \ \beta \ \left( M^2 \ - \   \NRM{\Bx_i}^2  \right) \right) \ 
    \exp \left(- \ \frac{1}{2} \ \beta \ \NRM{\Bx_i \ -  \ \Bxi}^2 \right)  \right)   \ .
\end{align}

First let us collect and prove some properties of $\rE$.
The next lemma gives bounds on the energy $\rE$.
\begin{lemmaA} \label{th:energy}
The energy $\rE$ is larger than zero:
\begin{align}
  0 \ &\leq \ \rE \ .
\end{align}
For $\Bxi$ in the simplex defined by the patterns,
the energy $\rE$ is upper bounded by:
\begin{align}
  \rE \ &\leq \ \beta^{-1} \ln N \ + \ \frac{1}{2} \ M^2 \ , \\
  \rE \ &\leq \ 2 \ M^2 \ .
\end{align}
\end{lemmaA}

\begin{proof}
We start by deriving the lower bound of zero.
The pattern most similar to query or state $\Bxi$ is
$\Bx_{\Bxi}$:
\begin{align}
  \Bx_{\Bxi} \ &= \ \Bx_k \ , \quad k \ = \ \arg\max_{i} \Bxi^T \Bx_i \ .
\end{align}
We obtain
\begin{align}
  \rE  \ &= \ - \ \beta^{-1} \ln \left( \sum_{i=1}^N
    \exp(\beta \Bx_i^T \Bxi) \right) \ + \ \beta^{-1} \ln N  \ + \
  \frac{1}{2} \Bxi^T \Bxi  \ + \ \frac{1}{2} \ M^2\\ \nonumber
  &= \ - \ \beta^{-1} \ln \left( \frac{1}{N} \ \sum_{i=1}^N
    \exp(\beta \Bx_i^T \Bxi) \right)  \ + \
  \frac{1}{2} \Bxi^T \Bxi  \ + \ \frac{1}{2} \ M^2\\ \nonumber
   &\geq \ - \ \beta^{-1} \ln \left( \frac{1}{N} \ \sum_{i=1}^N
    \exp(\beta \Bx_i^T \Bxi) \right)  \ + \
  \frac{1}{2} \ \Bxi^T \Bxi  \ + \ \frac{1}{2} \ \Bx_{\Bxi}^T \Bx_{\Bxi}\\ \nonumber
  &\geq  \ - \ \beta^{-1} \ln \left(  
    \exp(\beta \Bx_{\Bxi}^T \Bxi) \right)  \ + \
  \frac{1}{2} \Bxi^T \Bxi \ + \ \frac{1}{2} \ \Bx_{\Bxi}^T \Bx_{\Bxi}\\ \nonumber
  &=  \ - \ \Bx_{\Bxi}^T \Bxi  \ + \
  \frac{1}{2} \ \Bxi^T \Bxi \ + \ \frac{1}{2} \ \Bx_{\Bxi}^T \Bx_{\Bxi}\\ \nonumber
  &=  \frac{1}{2} \ \left(\Bxi \ - \  \Bx_{\Bxi} \right)^T 
  \left(\Bxi \ - \  \Bx_{\Bxi} \right)
  \ = \ \frac{1}{2} \ \NRM{\Bxi \ - \  \Bx_{\Bxi}}^2
  \ \geq \ 0 \ .
\end{align}
The energy is zero and, therefore, the bound attained, if all $\Bx_i$ are equal,
that is, $\Bx_i=\Bx$ for all $i$ and 
$\Bxi = \Bx$.

For deriving upper bounds on the energy $\rE$,
we require the the query $\Bxi$
to be in the simplex defined by the patterns, that is,
\begin{align}
 &\Bxi \ = \ \sum_{i=1}^N p_i \ \Bx_i \ ,  \quad 
 \sum_{i=1}^N p_i \ = \ 1 \ , \quad
 \forall_i: \ 0 \ \leq \ p_i \ .  
\end{align}
The first upper bound is.
\begin{align}
  &\rE   \ = \ - \ \beta^{-1} \ln \left( \sum_{i=1}^N
    \exp(\beta \Bx_i^T \Bxi) \right)  \ + \
  \frac{1}{2} \ \Bxi^T \Bxi \ + \ \beta^{-1} \ln N 
  \ + \ \frac{1}{2} \ M^2\\ \nonumber
  &\leq \  -   \sum_{i=1}^N p_i \ (\Bx_i^T \Bxi)   \ + \
  \frac{1}{2} \ \Bxi^T \Bxi \ + \ \beta^{-1} \ln N \ + \ \frac{1}{2} \ M^2\\ \nonumber
  &= \  - \  \frac{1}{2} \ \Bxi^T \Bxi \ + \ \beta^{-1} \ln N \ + \ \frac{1}{2} \ M^2 
  \ \leq \  \beta^{-1} \ln N \ + \ \frac{1}{2} \ M^2  \ .
\end{align}
For the first inequality we applied
Lemma \ref{th:Lentropy} to $-\mathrm{lse}(\beta,\BX^T \Bxi)$ with $\Bz=\Bp$ giving
\begin{align}
  - \ \mathrm{lse}(\beta,\BX^T \Bxi) \ &\leq \ -   \ \sum_{i=1}^N p_i \
  (\Bx_i^T \Bxi ) \ + \ \beta^{-1} \sum_{i=1}^N p_i \ln p_i \
  \leq \  -  \ \sum_{i=1}^N p_i \ (\Bx_i^T \Bxi) \ ,
\end{align}
as the term involving the logarithm is non-positive.

Next we derive the second upper bound, for which
we need the mean $\Bm_{\Bx}$ of the patterns 
\begin{align}
  \Bm_{\Bx} \ &= \  \frac{1}{N} \ \sum_{i=1}^N \Bx_i  \ .
\end{align}
We obtain
\begin{align}
  &\rE   \ = \ - \ \beta^{-1} \ln \left( \sum_{i=1}^N
    \exp(\beta \Bx_i^T \Bxi) \right)  \ + \
  \frac{1}{2} \ \Bxi^T \Bxi \ + \ \beta^{-1} \ln N 
  \ + \ \frac{1}{2} \ M^2\\ \nonumber
  &\leq \  -   \sum_{i=1}^N \frac{1}{N} \ \Bx_i^T \Bxi   \ + \
  \frac{1}{2} \ \Bxi^T \Bxi \ + \ \frac{1}{2} \ M^2\\ \nonumber
 &= \  -  \ \Bm_{\Bx}^T \Bxi   \ + \
  \frac{1}{2} \ \Bxi^T \Bxi \ + \ \frac{1}{2} \ M^2\\ \nonumber
 &\leq \  \NRM{\Bm_{\Bx}} \ \NRM{\Bxi}   \ + \
  \frac{1}{2} \ \NRM{\Bxi}^2 \ + \ \frac{1}{2} \ M^2\\ \nonumber
 &\leq \  2 \ M^2 \ ,
\end{align}
where for the first inequality we again applied
Lemma~\ref{th:Lentropy} with $\Bz=(1/N,\ldots,1/N)$
and $\beta^{-1} \sum_i 1/N \ln(1/N) = -\beta^{-1} \ln(N)$.
This inequality also follows from Jensen's inequality.
The second inequality uses the Cauchy-Schwarz inequality.
The last inequality uses
\begin{align}
  \NRM{\Bxi} \ &= \ \NRM{\sum_i p_i \ \Bx_i} \ \leq \ \sum_i p_i \
  \NRM{\Bx_i} \ \leq \ \sum_i p_i M \ =  \ M
\end{align}
and 
\begin{align}
  \NRM{\Bm_{\Bx}} \ &= \ \NRM{\sum_i (1/N) \ \Bx_i} \ \leq \ 
  \sum_i (1/N) \ \NRM{\Bx_i} \ \leq \ \sum_i (1/N) \ M \ = \ M \ .
\end{align}
  
\end{proof}

\subsubsection{New Update Rule}
\label{sec:newUpdate}

We now introduce an update rule for minimizing the energy function $\rE$.
The new update rule is
\begin{align}
\label{eq:newUpdate}
\Bxi^{\nn} \ &= \ \BX \Bp \ = \   \BX \soft ( \beta \BX^T \Bxi) \ ,
\end{align}
where we used 
\begin{align}
  \Bp \ &= \ \soft (\beta \BX^T \Bxi) \ .
\end{align}
The new state $\Bxi^{\nn}$ is in the simplex defined by the patterns, no matter
what the previous state $\Bxi$ was.
For comparison, the synchronous update rule for the classical Hopfield network with threshold zero is
\begin{align}
\Bxi^{\nn} \ &= \  \sgn(\BX  \BX^T \Bxi) \ . 
\end{align}
Therefore, instead of using the vector $\BX^T \Bxi$ as in the
classical Hopfield network, its softmax version $\soft (\beta \BX^T \Bxi)$ is used.

In the next section (Section~\ref{sec:GlobalConv}) we show 
that the update rule Eq.~\eqref{eq:newUpdate} ensures global convergence.
We show that all the limit points of any sequence generated by 
the update rule are the
stationary points (local minima or saddle points) of the 
energy function $\rE$.
In Section~\ref{sec:LocalConv} we consider the local convergence 
of the update rule Eq.~\eqref{eq:newUpdate} 
and see that patterns are retrieved with one update.

\subsubsection{Global Convergence of the Update Rule}
\label{sec:GlobalConv}

We are interested in the {\em global convergence},
that is, convergence from each initial point,  
of the iteration
\begin{align}
\label{eq:iterateCCCP1}
\Bxi^{\nn} \ &= \ f(\Bxi) \ = \ \BX \Bp \ = \   
\BX \soft ( \beta \BX^T \Bxi) \ ,
\end{align}
where we used
\begin{align}
  \Bp \ &= \   \soft ( \beta \BX^T \Bxi) \ . 
\end{align}
We defined the energy function
\begin{align}
  \rE \ &= \ - \ \mathrm{lse}(\beta ,\BX^T \Bxi) \ + \
  \frac{1}{2} \Bxi^T \Bxi  \ + \ \beta^{-1} \ln N \ + \ 
  \frac{1}{2} M^2 \\
  &= \ - \ \beta^{-1} \ln \left( \sum_{i=1}^N
    \exp(\beta \Bx_i^T \Bxi) \right)  \ + \ \beta^{-1} \ln N \ + \
  \frac{1}{2} \Bxi^T \Bxi \ + \ \frac{1}{2} M^2 \ .
\end{align}
We will show that
the update rule in Eq.~\eqref{eq:iterateCCCP1} 
is the Concave-Convex Procedure (CCCP) for minimizing the energy $\rE$. 
The CCCP is proven to converge globally.

\begin{theoremA}[Global Convergence (Zangwill): Energy]
\label{th:globalConvA}
The update rule Eq.~\eqref{eq:iterateCCCP1} 
converges globally:
For $\Bxi^{t+1} = f(\Bxi^t)$, 
the energy $\rE(\Bxi^t) \to \rE(\Bxi^*)$ for $t \to \infty$
and a fixed point $\Bxi^*$.
\end{theoremA}

\begin{proof}
The Concave-Convex Procedure (CCCP) \citep{Yuille:02,Yuille:03}
minimizes a function that is the sum of a concave function and a convex function.
CCCP is equivalent to Legendre minimization \citep{Rangarajan:96,Rangarajan:99}
algorithms \citep{Yuille:03}.
The Jacobian of the 
softmax is positive semi-definite
according to Lemma~\ref{th:LjacobiDefinite}. 
The Jacobian of the softmax is the 
Hessian of the $\mathrm{lse}$, therefore $\mathrm{lse}$ is a convex and
$-\mathrm{lse}$ a concave function.
Therefore, the energy function $\rE(\Bxi)$ is the sum of the convex function
$\rE_1(\Bxi)=1/2 \Bxi^T \Bxi +C_1$
and the concave function $\rE_2(\Bxi)=-\mathrm{lse}$:
\begin{align}
  \rE(\Bxi) \ &= \ \rE_1(\Bxi) \ + \ \rE_2(\Bxi) \ , \\
  \rE_1(\Bxi) \ &= \ \frac{1}{2} \Bxi^T \Bxi  \ + \ \beta^{-1} \ln N \ + \ 
  \frac{1}{2} M^2 \ = \ \frac{1}{2} \Bxi^T \Bxi  \ + \ C_1 \ , \\
  \rE_2(\Bxi) \ &= \ - \ \mathrm{lse}(\beta ,\BX^T \Bxi) \ ,
\end{align}
where $C_1$ does not depend on $\Bxi$.

The Concave-Convex Procedure (CCCP) \citep{Yuille:02,Yuille:03}
applied to $\rE$ is
\begin{align}
  \nabla_{\xi} \rE_1 \left(\Bxi^{t+1}\right) \ &= \ - \ \nabla_{\xi} \rE_2 \left(\Bxi^{t} \right) \ ,
 \end{align}
which is 
\begin{align}
  \nabla_{\xi} \left(\frac{1}{2} \Bxi^T \Bxi  \ + \ C_1 \right) \left( \Bxi^{t+1} \right)
   \ &= \ \nabla_{\xi} \mathrm{lse}(\beta ,\BX^T \Bxi^t)  \ .
 \end{align}
The resulting update rule is
\begin{align}
   \Bxi^{t+1}  \ &= \  \BX \Bp^t \ = \   \BX \soft ( \beta \BX^T \Bxi^t)  
 \end{align}
using
\begin{align}
  \Bp^t \ &= \   \soft ( \beta \BX^T \Bxi^t) \ . 
\end{align}
This is the update rule in Eq.~\eqref{eq:iterateCCCP1}.

Theorem~2 in \citet{Yuille:02} and Theorem~2 in \citet{Yuille:03} state that
the update rule Eq.~\eqref{eq:iterateCCCP1}
is guaranteed to monotonically decrease the energy $\rE$ 
as a function of time.
See also Theorem~2 in \citet{Sriperumbudur:09}.
\end{proof}
Although the objective converges in all cases, 
it does not necessarily converge to a local minimum \citep{Lipp:16}.

However the convergence proof of CCCP in \citet{Yuille:02,Yuille:03}
was not as rigorous as required.
In \citet{Sriperumbudur:09} a rigorous analysis of the convergence of CCCP
is performed using Zangwill's global convergence theory
of iterative algorithms.

In \citet{Sriperumbudur:09} the minimization problem
\begin{align}
  \min_{\Bxi} &{\mbox{\ ~} } \rE_1 \ + \  \rE_2 \\ \nonumber
  \mbox{ s.t. } &{\mbox{\ ~} } \Bc(\Bxi) \leq \BZe \ , \quad \Bd(\Bxi) \ = \ \BZe 
\end{align}
is considered with $\rE_1$ convex, $-\rE_2$ convex, $\Bc$
component-wise convex function, and $\Bd$ an affine function.
The CCCP algorithm solves this minimization problem by linearization of the 
concave part and is defined in \citet{Sriperumbudur:09} as
\begin{align}
  \Bxi^{t+1} \ \in \ \arg\min_{\Bxi} &{\mbox{\ ~} } \rE_1\left(\Bxi \right) 
  \ + \ \Bxi^T
  \nabla_{\xi} \rE_2\left(\Bxi^t \right) \\ \nonumber
  \mbox{ s.t. } &{\mbox{\ ~} } \Bc(\Bxi) \leq \BZe \ , \quad 
  \Bd(\Bxi) \ = \ \BZe \ .
\end{align}
We define the upper bound $\rE_{\rC}$ on the energy:
\begin{align}
\rE_{\rC}\left(\Bxi,\Bxi^t \right) 
   \ &:= \ \rE_1\left(\Bxi \right) \ + \ 
      \rE_2\left( \Bxi^t \right) \ + \ \left(\Bxi \ - \ \Bxi^t\right)^T
      \nabla_{\xi} \rE_2\left(\Bxi^t \right) \ .
\end{align}
$\rE_{\rC}$ is equal to the energy $\rE\left(\Bxi^t \right)$ for $\Bxi=\Bxi^t$:
\begin{align}
\rE_{\rC}\left(\Bxi^t,\Bxi^t \right) 
   \ &= \ \rE_1\left(\Bxi^t \right) \ + \ 
      \rE_2\left( \Bxi^t \right) \ = \ \rE\left(\Bxi^t \right) \ .
\end{align}
Since $-\rE_2$ is convex, the first order characterization of convexity
holds (Eq.~3.2 in \citet{Boyd:09}):
\begin{align}
   - \ \rE_2\left( \Bxi \right)  \ &\geq \  - \ 
      \rE_2\left( \Bxi^t \right) \ - \ \left(\Bxi \ - \ \Bxi^t\right)^T
      \nabla_{\xi} \rE_2\left(\Bxi^t \right) \ ,
\end{align}
that is
\begin{align}
    \rE_2\left( \Bxi \right)  \ &\leq \   
      \rE_2\left( \Bxi^t \right) \ + \ \left(\Bxi \ - \ \Bxi^t\right)^T
      \nabla_{\xi} \rE_2\left(\Bxi^t \right) \ .
\end{align}

Therefore, for $\Bxi\not=\Bxi^t$ the function $\rE_{\rC}$ is an upper bound on the energy:
\begin{align}
\label{eq:CCCPlin}
   \rE\left(\Bxi \right) \ &\leq \ \rE_{\rC}\left(\Bxi,\Bxi^t \right) 
   \ = \ \rE_1\left(\Bxi \right) \ + \ 
      \rE_2\left( \Bxi^t \right) \ + \ \left(\Bxi \ - \ \Bxi^t\right)^T
      \nabla_{\xi} \rE_2\left(\Bxi^t \right) \\ \nonumber
  &= \ \rE_1\left(\Bxi \right) \ + \ \Bxi^T \nabla_{\xi} \rE_2\left(\Bxi^t \right) \ + \ C_2 \ ,
\end{align}
where $C_2$ does not depend on $\Bxi$.
Since we do not have constraints, $\Bxi^{t+1}$ is defined as
\begin{align}
  \Bxi^{t+1} \ \in \ \arg\min_{\Bxi} &{\mbox{\ ~} } \rE_{\rC}\left(\Bxi,\Bxi^t \right) \ ,
\end{align}
hence
$\rE_{\rC}\left(\Bxi^{t+1},\Bxi^t \right) \leq \ \rE_{\rC}\left(\Bxi^t,\Bxi^t \right) $.
Combining the inequalities gives:
\begin{align}
   \rE\left(\Bxi^{t+1} \right) \ &\leq \ \rE_{\rC}\left(\Bxi^{t+1},\Bxi^t \right) 
   \ \leq \  \rE_{\rC}\left(\Bxi^t,\Bxi^t \right) \ = \  \rE\left(\Bxi^t \right) \ .
\end{align}

Since we do not have constraints, $\Bxi^{t+1}$ is the minimum of
\begin{align}
\label{eq:CCCPlin1}
   \rE_{\rC}\left(\Bxi,\Bxi^t \right) 
   \ &= \ \rE_1\left(\Bxi \right) \ + \ \Bxi^T \nabla_{\xi} \rE_2\left(\Bxi^t \right) \ + \ C_2 
\end{align}
as a function of $\Bxi$.

For a minimum not at the border, the derivative has to be the zero vector
\begin{align}
   \frac{\partial \rE_{\rC}\left(\Bxi,\Bxi^t \right)}{\partial \Bxi} \ &= \ 
  \Bxi \ + \ \nabla_{\xi} \rE_2\left(\Bxi^t \right) \ = \ 
  \Bxi \ - \ \BX \soft ( \beta \BX^T \Bxi^t) \ = \ \BZe
\end{align}
and the Hessian must be positive semi-definite
\begin{align}
\label{eq:Hessian}
   \frac{\partial^2 \rE_{\rC}\left(\Bxi,\Bxi^t \right)}{\partial \Bxi^2} \ &= \ 
  \BI \ .
\end{align}
The Hessian is strict positive definite everywhere, therefore the optimization
problem is strict convex (if the domain is convex) and there exist only one minimum,
which is a global minimum.
$\rE_{\rC}$ can even be written as a quadratic form:
\begin{align}
\label{eq:quadraticForm}
   \rE_{\rC}\left(\Bxi,\Bxi^t \right) 
   \ &= \ \frac{1}{2} \ \left( \Bxi \ + \ \nabla_{\xi} \rE_2\left(\Bxi^t \right) \right)^T 
   \left( \Bxi \ + \ \nabla_{\xi} \rE_2\left(\Bxi^t \right) \right) \ + \ C_3 \ ,
\end{align}
where $C_3$ does not depend on $\Bxi$.

Therefore, the minimum is
\begin{align}
\label{eq:iterCCCP}
  \Bxi^{t+1} \ &= \ - \ \nabla_{\xi} \rE_2\left(\Bxi^t \right) \ = \
  \BX \soft ( \beta \BX^T \Bxi^t)  
\end{align}
if it is in the domain as we assume.

Using  $M = \max_{i} \NRM{\Bx_i}$,
$\Bxi^{t+1}$ is in the sphere $\rS=\{\Bx \mid  \NRM{\Bx} \leq 
M\}$ which is a convex and compact set.  
Hence, if $\Bxi^0 \in \rS$, then the iteration is a mapping from $\rS$ to $\rS$.
Therefore, the point-set-map defined by the iteration Eq.~\eqref{eq:iterCCCP}
is uniformly compact on $\rS$ according to Remark~7 in \citet{Sriperumbudur:09}.
Theorem~2 and Theorem~4 in \citep{Sriperumbudur:09} states that
all the limit points of the iteration Eq.~\eqref{eq:iterCCCP}
are stationary points.
These theorems follow from
Zangwill’s global convergence theorem:
Convergence~Theorem~A, page 91 in \citet{Zangwill:69} and page 3
in \citet{Wu:83}.

The global convergence theorem only assures that for the sequence
$\Bxi^{t+1} = f(\Bxi^t)$ and 
a function $\Phi$ we have $\Phi(\Bxi^t) \to \Phi(\Bxi^*)$ for $t \to \infty$ 
but not $\Bxi^t \to \Bxi^*$. 
However, if $f$ is strictly monotone with respect to $\Phi$, then 
we can strengthen Zangwill's global convergence theorem \citep{Meyer:76}.
We set $\Phi=\rE$ and show $\rE(\Bxi^{t+1})<\rE(\Bxi^t)$ if $\Bxi^t$ is
not a stationary point of $\rE$, that is, $f$ is strictly monotone with respect 
to $\rE$.
The following theorem is similar to the convergence results for the 
expectation maximization (EM) algorithm in \citet{Wu:83} which
are given in theorems 1 to 6 in \citet{Wu:83}.
The following theorem is also very similar to Theorem~8 in \citet{Sriperumbudur:09}.
\begin{theoremA}[Global Convergence: Stationary Points]
\label{th:globalConvergenceA}
For the iteration Eq.~\eqref{eq:iterCCCP}
we have $\rE\left(\Bxi^t \right) 
\to \rE\left(\Bxi^* \right) = \rE^*$
as  $t \to \infty$, for some stationary point $\Bxi^*$. 
Furthermore $\NRM{\Bxi^{t+1}- \Bxi^t} \to 0$ and
either $\{ \Bxi^t \}_{t=0}^{\infty}$ converges
or, in the other case, the set of limit points of  $\{ \Bxi^t \}_{t=0}^{\infty}$ 
is a connected and compact subset of $\cL\left( \rE^* \right)$, where
$\cL\left( a\right)= \{\Bxi \in \cL \mid \rE\left(\Bxi \right) = a\} $
and $\cL$ is the set of stationary points of the iteration Eq.~\eqref{eq:iterCCCP}.
If $\cL\left( \rE^* \right)$ is finite, then any sequence $\{ \Bxi^t \}_{t=0}^{\infty}$ 
generated by the iteration Eq.~\eqref{eq:iterCCCP} 
converges to some $\Bxi^* \in \cL\left( \rE^*\right)$.
\end{theoremA}

\begin{proof}
We have $\rE\left(\Bxi^t \right) = 
\rE_1\left(\Bxi^t \right) + \rE_2\left(\Bxi^t \right)$.
The gradient
$\nabla_{\xi} \rE_2\left(\Bxi^t \right) = - 
\nabla_{\xi} \mathrm{lse}(\beta ,\BX^T \Bxi)$ is
continuous. Therefore, Eq.~\eqref{eq:CCCPlin1} has minimum in the sphere $\rS$, which is
a convex and compact set. 
If $\Bxi^{t+1} \not= \Bxi^t$, 
then $\Bxi^t$ was not the minimum of
Eq.~\eqref{eq:CCCPlin} as the derivative at $\Bxi^t$ is not equal to zero.
Eq.~\eqref{eq:Hessian} shows that the optimization problem Eq.~\eqref{eq:CCCPlin} 
is strict convex, hence it has only one minimum, which is a global minimum.
Eq.~\eqref{eq:quadraticForm} shows that the optimization problem Eq.~\eqref{eq:CCCPlin} 
is even a quadratic form.
Therefore, we have
\begin{align}
\label{eq:strictE}
   \rE\left(\Bxi^{t+1} \right) \ &\leq \ \rE_{\rC}\left(\Bxi^{t+1},\Bxi^t \right) 
   \ < \  \rE_{\rC}\left(\Bxi^t,\Bxi^t \right) \ = \  \rE\left(\Bxi^t \right) \ .
\end{align}

Therefore, the point-set-map defined by the iteration Eq.~\eqref{eq:iterCCCP}
(for definitions see \citep{Sriperumbudur:09})
is strictly monotonic with respect to $\rE$.
Therefore, we can apply Theorem~3 in \citet{Sriperumbudur:09} or
Theorem~3.1 and Corollary~3.2 in \citet{Meyer:76},
which give the statements of the theorem.

\end{proof}

We showed global convergence of the iteration Eq.~\eqref{eq:iterateCCCP1}.
We have shown that all the limit points of any sequence generated by 
the iteration Eq.~\eqref{eq:iterateCCCP1} are the
stationary points (critical points; local minima or saddle points) of the 
energy function $\rE$.
Local maxima as stationary points are only possible 
if the iterations exactly hits a local maximum. 
However, convergence to a local maximum without being there 
is not possible because  
Eq.~\eqref{eq:strictE} ensures a strict decrease of the 
energy $\rE$. Therefore, almost sure local maxima are not 
obtained as stationary points.
Either the iteration converges or,
in the second case, the set of limit points 
is a connected and compact set.
But what happens if $\Bxi^0$ is in an $\epsilon$-neighborhood 
around a local minimum $\Bxi^*$? 
Will the iteration Eq.~\eqref{eq:iterateCCCP1} converge to $\Bxi^*$?
What is the rate of convergence? 
These questions are about 
{\em local convergence} which will be treated in detail in next section.

\subsubsection{Local Convergence of the Update Rule: Fixed Point Iteration}
\label{sec:LocalConv}

For the proof of local convergence to a fixed point 
we will apply Banach fixed point theorem.
For the rate of convergence we will rely on properties of a contraction mapping.

\paragraph{General Bound on the Jacobian of the Iteration.}

We consider the iteration
\begin{align} \label{eq:iter}
\Bxi^{\nn} \ &= \ f(\Bxi) \ = \ \BX \Bp \ = \   \BX \soft ( \beta \BX^T \Bxi) 
\end{align}
using
\begin{align}
  \Bp \ &= \   \soft ( \beta \BX^T \Bxi) \ . 
\end{align}

The Jacobian $\rJ$ is symmetric and has the following form:
\begin{align}
 \label{eq:theJacobian}
  \rJ \ &= \ \frac{\partial f(\Bxi)}{\partial \Bxi} \ = \   \beta \ \BX \left(
  \diag(\Bp) - \Bp \Bp^T \right) \BX^T  \ = \  \BX \rJ_s \BX^T \ ,
\end{align}
where $\rJ_s$ is Jacobian of the softmax. 

\begin{figure}[ht]
    \centering
    \includegraphics[width=\textwidth]{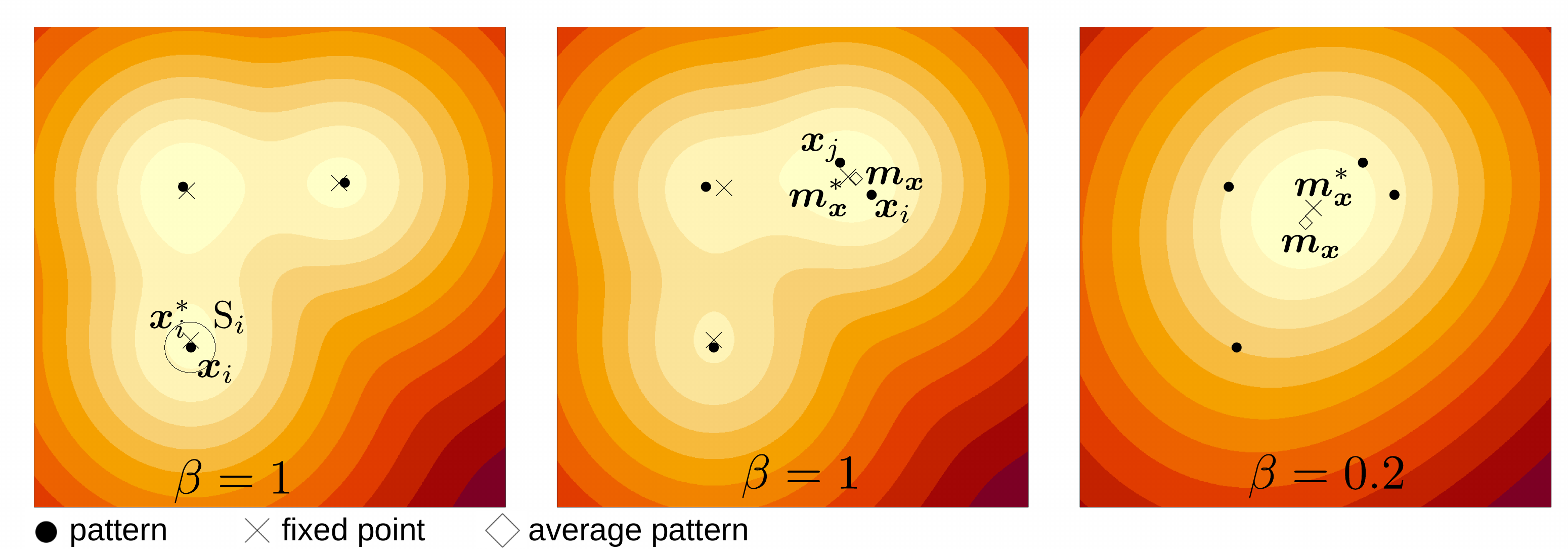}
    \caption[The three cases of fixed points]{The three cases of fixed points. 
    {\bf a) Stored patterns (fixed point is single pattern)}: 
    patterns are stored if they are well separated. 
    Each pattern $\Bx_i$ has a single fixed point $\Bx_i^*$ close to it.
    In the sphere $\rS_i$, pattern $\Bx_i$ is the only pattern and $\Bx_i^*$ the only fixed point.
    {\bf b) Metastable state (fixed point is average of similar patterns)}: 
    $\Bx_i$ and $\Bx_j$ are similar to each other and not
    well separated. The fixed point $\Bm_{\Bx}^*$ is a metastable state 
    that is close to the mean $\Bm_{\Bx}$ of the similar patterns.
    {\bf c) Global fixed point (fixed point is average of all patterns)}: 
    no pattern is well separated from the others. 
    A single global fixed point $\Bm_{\Bx}^* $ exists that is close to 
    the arithmetic mean $\Bm_{\Bx}$ of all patterns.}
    \label{fig:main_figure}
\end{figure}

To analyze the local convergence of the iteration,
we distinguish between the following three cases (see also Fig.~\ref{fig:main_figure}).
Here we only provide an informal discussion to give the reader some
intuition.
A rigorous formulation of the results can be found in the corresponding subsections.
\begin{enumerate}[a)]

\item If the patterns $\Bx_i$ are not well separated,
the iteration goes to a fixed point close to
the arithmetic mean of the vectors.
In this case  $\Bp$ is close to $p_i=1/N$.

\item If the patterns $\Bx_i$ are well separated,
then the iteration goes to the pattern to which the initial
$\Bxi$ is similar.
If the initial $\Bxi$ is similar to a vector $\Bx_i$ then it will
converge to a vector close to $\Bx_i$ and $\Bp$ will converge to a vector
close to $\Be_i$.

\item If some vectors are similar to each other but well separated from all
other vectors, then a so called metastable state between the similar
vectors exists.
Iterations that start near the metastable state converge
to this metastable state.
\end{enumerate}

We begin with a bound on the Jacobian of the iteration, thereby
heavily relying on
the Jacobian of the softmax from Lemma~\ref{th:JacobiB}. 
\begin{lemmaA}
\label{th:boundJ}
For $N$ patterns $\BX=(\Bx_1,\ldots,\Bx_N)$, 
$\Bp =  \soft ( \beta \BX^T \Bxi)$, $M = \max_{i} \NRM{\Bx_i}$, and
$m= \max_i p_i (1-p_i)$, the spectral norm of the Jacobian
$\rJ$ of the fixed point iteration
is bounded:
\begin{align} \label{eq:jacobi2general}
    \NRM{\rJ}_2 \ &\leq \
  2 \ \beta \ \NRM{\BX}_2^2 \ m 
  \ \leq \   2 \ \beta \ N \ M^2 \ m \ .
\end{align}
If $p_{\max}=\max_i p_i \geq 1-\epsilon$, then for the spectral norm of
the Jacobian holds
\begin{align}
\label{eq:boundJpmax}
 \NRM{\rJ}_2 \ &\leq \  2 \ \beta \ N \ M^2 \ \epsilon 
 \ - \  2 \ \epsilon^2 \ \beta \ N \ M^2 
 \ < \  2 \ \beta \ N \ M^2 \ \epsilon \ .
\end{align}

\end{lemmaA}

\begin{proof}
With
\begin{align}
  \Bp \ &= \   \soft ( \beta \BX^T \Bxi) \ , 
\end{align}
the symmetric Jacobian $\rJ$ is
\begin{align}
  \rJ \ &= \ \frac{\partial f(\Bxi)}{\partial \Bxi} \ = \   \beta \ \BX \left(
  \diag(\Bp) - \Bp \Bp^T \right) \BX^T  \ = \  \BX \rJ_s \BX^T \ ,
\end{align}
where $\rJ_s$ is Jacobian of the softmax. 

With $m= \max_i p_i (1-p_i)$, 
Eq.~\eqref{eq:softjacobi2-m} from
Lemma~\ref{th:JacobiB} is
\begin{align}
 \NRM{\rJ_s}_2 \ &= \ \beta \ \NRM{\diag(\Bp) - \Bp \Bp^T}_2
 \ \leq \ 2 \ m \ \beta \ .
\end{align}
Using this bound on $ \NRM{\rJ_s}_2$, we obtain
\begin{align}
 \NRM{\rJ}_2 \ &\leq \ \beta \ \NRM{\BX^T}_2 \   \NRM{\rJ_s}_2  \ \NRM{\BX}_2
 \ \leq \ 2 \ m \ \beta  \ \NRM{\BX}_2^2 \ .
\end{align}

The spectral norm  $\NRM{.}_2$ is bounded by the Frobenius norm
$\NRM{.}_F$ which can be expressed by the norm squared of its
column vectors:
\begin{align}
  \NRM{\BX}_2 \ &\leq \  \NRM{\BX}_F \ = \ \sqrt{\sum_i\NRM{\Bx_i}^2} \ .
\end{align}
Therefore, we obtain the first statement of the lemma:
\begin{align}
    \NRM{\rJ}_2 \ &\leq \
  2 \ \beta \ \NRM{\BX}_2^2 \ m 
  \ \leq \   2 \ \beta \ N \ M^2 \ m \ .
\end{align}

With $p_{\max}=\max_i p_i \geq 1-\epsilon$
Eq.~\eqref{eq:softjacobi2-eps}
in Lemma~\ref{th:JacobiB} is
\begin{align}
 \NRM{\rJ_s}_2 \ &\leq \ 2 \ \beta \ \epsilon \ - \  2 \ \epsilon^2 \ \beta 
 \ < \  2 \ \beta \ \epsilon \ .
\end{align}
Using this inequality, 
we obtain the second statement of the lemma:
\begin{align}
 \NRM{\rJ}_2 \ &\leq \  2 \ \beta \ N \ M^2 \ \epsilon 
 \ - \  2 \ \epsilon^2 \ \beta \ N \ M^2 
 \ < \  2 \ \beta \ N \ M^2 \ \epsilon \ .
\end{align}

\end{proof}

We now define the ``separation'' $\Delta_i$ of a pattern $\Bx_i$ from
data $\BX=(\Bx_1,\ldots,\Bx_N)$ here, since it has an important role
for the convergence properties of the iteration.
\begin{definition}[Separation of Patterns]
We define $\Delta_i$, i.e.\ {\em the separation of pattern $\Bx_i$ from data $\BX=(\Bx_1,\ldots,\Bx_N)$} as:
\begin{align}
  \Delta_i \ &= \ \min_{j,j \not= i} \left( \Bx_i^T \Bx_i \ - \ \Bx_i^T
    \Bx_j \right) \ = \ \Bx_i^T \Bx_i \ - \ \max_{j,j \not= i} \Bx_i^T
  \Bx_j \ .
\end{align}
The pattern is separated from the other data if $0<\Delta_i$.
Using the parallelogram identity,  $\Delta_i$ can also be expressed as
\begin{align} \label{eq:diwithnormD}
  \Delta_i \ &= \ \min_{j,j \not= i} \frac{1}{2} \ \left(
  \NRM{\Bx_i}^2 \ - \  \NRM{\Bx_j}^2 \ + \  \NRM{\Bx_i \ - \ \Bx_j}^2
  \right) \\ \nonumber
  &= \ \frac{1}{2} \NRM{\Bx_i}^2  \ - \
  \frac{1}{2} \ \max_{j,j \not= i}  \left(  \NRM{\Bx_j}^2 \ - \
    \NRM{\Bx_i \ - \ \Bx_j}^2 \right) \ .
\end{align}
For $\NRM{\Bx_i} =  \NRM{\Bx_j}$ we have $\Delta_i= 1/2 \min_{j,j \not= i}
\NRM{\Bx_i \ - \ \Bx_j}^2$.

Analog we say for a query $\Bxi$ and data $\BX=(\Bx_1,\ldots,\Bx_N)$,
that $\Bx_i$ is least separated from $\Bxi$ while 
being separated from other $\Bx_j$ with $j \not= i$ if
\begin{align}
  i \ &= \ \arg\max_k \min_{j,j \not= k} \left( \Bxi^T \Bx_k \ - \ \Bxi^T
    \Bx_j \right) \ = \ \arg\max_k \left( \Bxi^T \Bx_k \ - \ \max_{j,j \not= k} \Bxi^T
  \Bx_j \right) \\
  0 \ &\leq \ c \ = \ \max_k \min_{j,j \not= k} \left( \Bxi^T \Bx_k \ - \ \Bxi^T
    \Bx_j \right) \ = \ \max_k  \left( \Bxi^T \Bx_k \ - \ \max_{j, j \not= k} \Bxi^T
  \Bx_j \right) \ .
\end{align}

\end{definition}

Next we consider the case where the iteration has only
one stable fixed point.

\paragraph{One Stable State: Fixed Point Near the Mean of the Patterns.}
\label{sec:global}

We start with the case where
no pattern is well separated from the others.

{\textbullet \em Global fixed point near the global mean: Analysis using the data center.}

We revisit the bound on the Jacobian of the iteration by
utilizing properties of pattern distributions.
We begin with a probabilistic interpretation where
we consider $p_i$ as the probability of selecting
the vector $\Bx_i$.
Consequently, we define
expectations as $\EXP_{\Bp}[f(\Bx)]=\sum_{i=1}^N p_i f(\Bx_i)$.
In this setting the matrix 
\begin{align}
  \BX \left(
  \diag(\Bp) - \Bp \Bp^T \right) \BX^T 
\end{align}
is the covariance matrix of data $\BX$ when its 
vectors are selected according 
to the probability $\Bp$:
\begin{align}
 &\BX \left(  \diag(\Bp) \ - \ \Bp \Bp^T \right) \BX^T 
 \ = \ \BX \diag(\Bp) \BX^T \ - \ \BX \Bp \Bp^T \BX^T \\
 &= \ \sum_{i=1}^N p_i \ \Bx_i \ \Bx_i^T \ - \ 
  \left( \sum_{i=1}^N p_i \ \Bx_i \right) 
   \left( \sum_{i=1}^N p_i \ \Bx_i \right)^T  \\
 &= \ \EXP_{\Bp}[\Bx \ \Bx^T] \ - \ \EXP_{\Bp}[\Bx] \ \EXP_{\Bp}[\Bx]^T
 \ = \ \VAR_{\Bp}[\Bx] \ ,
\end{align}
therefore we have
\begin{align}
\label{eq:jacVar}
 \rJ \ &= \ \beta \ \VAR_{\Bp}[\Bx]  \ .
\end{align}

The largest eigenvalue of the covariance matrix
(equal to the largest singular value)
is the variance in the direction of the eigenvector
associated with the largest eigenvalue.

We define:
\begin{align}  \label{eq:meanPattern}
  \Bm_{\Bx} \ &= \ \frac{1}{N} \ \sum_{i=1}^N \ \Bx_i \ , \\  \label{eq:meanmaxdist}
  m_{\max} \ &= \  \max_{1\leq i \leq N} \NRM{\Bx_i \ - \ \Bm_{\Bx}}_2  \ .
\end{align}
$\Bm_{\Bx}$ is the arithmetic mean (the center) of the patterns. 
$m_{\max}$ is the maximal distance of the patterns to the center $\Bm_{\Bx}$ .

The variance of the patterns is
\begin{align}
   \VAR_{\Bp}[\Bx] \ &= \  
  \sum_{i=1}^N p_i \ \Bx_i \ \Bx_i^T
  \ - \   \left(\sum_{i=1}^N p_i \ \Bx_i\right) \ 
  \left(\sum_{i=1}^N p_i \ \Bx_i\right)^T \\ \nonumber
  &= \ 
    \sum_{i=1}^N p_i \ 
    \left(\Bx_i \ - \ \sum_{i=1}^N p_i \Bx_i \right) \ 
     \left(\Bx_i \ - \ \sum_{i=1}^N p_i \Bx_i \right)^T   \ .
\end{align}

The maximal distance to the center $m_{\max}$ allows the derivation of a
bound on the norm of the Jacobian.

Next lemma gives a condition for a global fixed point.
\begin{lemmaA}  \label{th:jacobibound}
The following bound on the norm $\NRM{\rJ}_2$ of the
Jacobian of the fixed point iteration $f$ holds independent of $\Bp$ or
the query $\Bxi$.
\begin{align}
   \NRM{\rJ}_2 \ &\leq \ \beta \ m_{\max}^2  \ .
\end{align}
For $\beta \ m_{\max}^2  < 1$ there exists 
a unique fixed point (global fixed point) of iteration $f$ in each compact set. 
\end{lemmaA}

\begin{proof}

In order to bound the variance 
we compute the vector $\Ba$ that minimizes 
\begin{align}
 f(\Ba) \ &= \  \sum_{i=1}^N p_i \NRM{\Bx_i \ - \ \Ba}^2
 \ = \  \sum_{i=1}^N p_i (\Bx_i \ - \ \Ba)^T (\Bx_i \ - \ \Ba)  \ .
\end{align}

The solution to 
\begin{align}
 \frac{\partial f(\Ba)}{\partial \Ba} \ &= \ 
 2 \ \sum_{i=1}^N p_i (\Ba \ - \ \Bx_i ) \ = \ 0 
\end{align}
is
\begin{align}
 \Ba \ &= \  \sum_{i=1}^N p_i \Bx_i \ .
\end{align}
The Hessian of $f$ is positive definite since
\begin{align}
 \frac{\partial^2 f(\Ba)}{\partial \Ba^2} \ &= \ 
 2 \ \sum_{i=1}^N p_i \ \BI \ = \ 2 \ \BI 
\end{align}
and $f$ is a convex function.
Hence, the mean
\begin{align}
\bar{\Bx} \ &:= \ \sum_{i=1}^N p_i \ \Bx_i 
\end{align}
minimizes 
$\sum_{i=1}^N p_i \NRM{\Bx_i  - \Ba}^2$.
Therefore, we have
\begin{align}  \label{eq:estmeandev}
  \sum_{i=1}^N p_i \NRM{\Bx_i \ - \ \bar{\Bx}}^2 \ &\leq \ \sum_{i=1}^N p_i 
  \NRM{\Bx_i \ - \ \Bm_{\Bx}}^2 \ \leq \ m_{\max}^2 \ .
\end{align}

Let us quickly recall that the spectral norm of an outer product of two vectors 
is the product of the Euclidean norms of the vectors:
\begin{align} \label{eq:normouter}
\NRM{\Ba \Bb^T}_2 \ &= \ \sqrt{\lambda_{\max}(\Bb \Ba^T \Ba \Bb^T) }
 \ = \ \NRM{\Ba} \ \sqrt{\lambda_{\max}(\Bb \Bb^T)} \ = \ \NRM{\Ba} \ \NRM{\Bb} \ ,
\end{align}
since $\Bb \Bb^T$ has eigenvector $\Bb/\NRM{\Bb}$ with
eigenvalue $\NRM{\Bb}^2$ and otherwise zero eigenvalues.

We now bound the variance of the patterns:
\begin{align}
 \NRM{\VAR_{\Bp}[\Bx]}_2 
    \ &\leq \ \sum_{i=1}^N p_i \NRM{\left( \Bx_i \ - \ \bar{\Bx} \right) \
   \left( \Bx_i \ - \ \bar{\Bx} \right)^T}_2 \\ \nonumber
   &= \ \sum_{i=1}^N p_i \NRM{\Bx_i \ - \ \bar{\Bx}}^2
   \ \leq \ \sum_{i=1}^N p_i 
  \NRM{\Bx_i \ - \ \Bm_{\Bx}}^2 \ \leq \  \ m_{\max}^2 \ .
\end{align}

The bound of the lemma on $\NRM{\rJ}_2$
follows from Eq.~\eqref{eq:jacVar}.

For $\NRM{\rJ}_2 \leq \beta \ m_{\max}^2  < 1$ we have a 
contraction mapping on each compact set. Banach fixed point theorem
says there is a unique fixed point in the compact set. 

\end{proof}

Now let us further investigate the tightness of the bound on $\NRM{\VAR_{\Bp}[\Bx]}_2$ via
$\NRM{\Bx_i - \bar{\Bx}}^2$: we
consider the trace,  which
is the sum $\sum_{k=1}^d e_k$ of the w.l.o.g.\ ordered nonnegative eigenvalues $e_k$ of $\VAR_{\Bp}[\Bx]$ 
The spectral norm is equal to the 
largest eigenvalue $e_1$, which is equal to the largest singular
value, as we have positive semidefinite matrices.
We obtain:
\begin{align} \label{eq:varandeigenvs}
  \NRM{\VAR_{\Bp}[\Bx]}_2 
    \ &= \ \TR \left(\sum_{i=1}^N p_i \left( \Bx_i \ - \ 
    \bar{\Bx} \right) \
   \left( \Bx_i \ - \ \bar{\Bx} \right)^T \right) \ - \ 
   \sum_{k=2}^d e_k \\ \nonumber
   &= \ \sum_{i=1}^N p_i \TR \left( \left( \Bx_i \ - \ 
    \bar{\Bx} \right) \
   \left( \Bx_i \ - \ \bar{\Bx} \right)^T \right)
   \ - \ 
   \sum_{k=2}^d e_k \\ \nonumber
   &= \ \sum_{i=1}^N p_i \NRM{\Bx_i \ - \ \bar{\Bx}}^2  \ - \ 
   \sum_{k=2}^d e_k \ .
\end{align}
Therefore, the tightness of the bound depends on eigenvalues
which are not the largest. Hence variations which are not
along the largest variation weaken the bound.

\vspace{1cm}

Next we investigate the location of
fixed points which existence is ensured by the global convergence
stated in Theorem~\ref{th:globalConvergenceA}.
For $N$ patterns $\BX=(\Bx_1,\ldots,\Bx_N)$, 
we consider the iteration
\begin{align}
\label{eq:updateN}
\Bxi^{\nn} \ &= \ f(\Bxi) \ = \ \BX \Bp \ = \   \BX \soft ( \beta \BX^T \Bxi) 
\end{align}
using
\begin{align}
  \Bp \ &= \   \soft ( \beta \BX^T \Bxi) \ . 
\end{align}
$\Bxi^{\nn}$ is in the simplex of the patterns, that is,
$\Bxi^{\nn}=\sum_i p_i \Bx_i$ with 
$\sum_i p_i = 1$ and $0 \leq p_i$.
Hence, after one update $\Bxi$ is in the simplex of the pattern and stays there.
If the center $\Bm_{\Bx}$ is the zero vector $\Bm_{\Bx}=\BZe$, that is, the data is centered,
then the mean is a fixed point of the iteration.
For $\Bxi=\Bm_{\Bx}=\BZe$ we have
\begin{align}
  \Bp \ &= \  1/N \ \BOn  
\end{align}
and 
\begin{align}
\Bxi^{\nn} \ &= \  1/N \ \BX \ \BOn \ = \  \Bm_{\Bx} \ = \  \Bxi \ .
\end{align}
In particular normalization methods like batch normalization would 
promote the mean as a fixed point.

We consider the differences of dot products 
for $\Bx_i$: $\Bx_i^T \Bx_i - \Bx_i^T \Bx_j = \Bx_i^T (\Bx_i - \Bx_j)$,
for fixed point $\Bm_{\Bx}^* $:
$(\Bm_{\Bx}^* )^T \Bx_i - (\Bm_{\Bx}^* )^T \Bx_j = (\Bm_{\Bx}^* )^T (\Bx_i - \Bx_j)$, 
and for the center $\Bm_{\Bx}$:
$\Bm_{\Bx}^T \Bx_i - \Bm_{\Bx}^T \Bx_j = \Bm_{\Bx}^T (\Bx_i - \Bx_j)$.
Using the Cauchy-Schwarz inequality, we get
\begin{align}
 \left| \Bxi^T (\Bx_i \ - \ \Bx_j) \right| \ &\leq \  \NRM{\Bxi} \ \NRM{\Bx_i \ - \ \Bx_j} 
 \ \leq \ \NRM{\Bxi} \ ( \NRM{\Bx_i \ - \ \Bm_{\Bx}} \ + \  \NRM{\Bx_j \ - \ \Bm_{\Bx}}) \\ \nonumber
 &\leq \ 2 \  m_{\max} \ \NRM{\Bxi}  \ .
\end{align}
This inequality gives:
\begin{align}
 \left| \Bxi^T (\Bx_i \ - \ \Bx_j) \right| \ &\leq \  2 \  m_{\max} \ (  m_{\max} \ + \ \NRM{\Bm_{\Bx}} ) \ , \\ \nonumber
 \left| \Bxi^T (\Bx_i \ - \ \Bx_j) \right| \ &\leq \  2 \  m_{\max} \ M \ ,
\end{align}
where we used $\NRM{\Bxi - \BZe} \leq \NRM{\Bxi - \Bm_{\Bx}} + \NRM{\Bm_{\Bx}- \BZe}$,
$\NRM{\Bxi - \Bm_{\Bx}}=\NRM{\sum_i p_i \Bx_i - \Bm_{\Bx}}
\leq \sum_i p_i \NRM{\Bx_i - \Bm_{\Bx}} \leq m_{\max}$,
and $M = \max_{i} \NRM{\Bx_i}$.
In particular 
\begin{align}
 \beta \ \left| \Bm_{\Bx}^T (\Bx_i \ - \ \Bx_j) \right| \ &\leq \ 2 \  \beta \ m_{\max} \ \NRM{\Bm_{\Bx}} \ , \\
 \beta \ \left| (\Bm_{\Bx}^* )^T (\Bx_i \ - \ \Bx_j) \right| \ &\leq \ 2 \  \beta \ m_{\max} \ \NRM{\Bm_{\Bx}^* } \ \leq \ 
  2 \  \beta \ m_{\max} \ (  m_{\max} \ + \ \NRM{\Bm_{\Bx}} ) \ , \\
 \beta \ \left| \Bx_i^T (\Bx_i \ - \ \Bx_j) \right| \ &\leq \ 2 \  \beta \ m_{\max} \ \NRM{\Bx_i} \ \leq \ 
  2 \  \beta \ m_{\max} \ (  m_{\max} \ + \ \NRM{\Bm_{\Bx}} ) \ .
\end{align}

Let $i=\arg\max_j \Bxi^T \Bx_j$, therefore the maximal softmax component
is $i$.
For the maximal softmax component $i$ we have:
\begin{align}
  &[ \soft( \beta \ \BX^T \Bxi)]_i \ = \ \frac{1}{1 \ + \ \sum_{j \not= i}
    \exp(- \ \beta \ ( \Bxi^T \Bx_i \ - \ \Bxi^T \Bx_j ) )} \\ \nonumber
  &\leq \
  \frac{1}{1 \ + \ \sum_{j \not= i}  \exp( - \ 2 \  \beta \ m_{\max} \ (  m_{\max} \ + \ \NRM{\Bm_{\Bx}} ) )} \\ \nonumber
  &= \ \frac{1}{1 \ + \ (N-1) \exp(- \ 2 \  \beta \ m_{\max} \ (  m_{\max} \ + \ \NRM{\Bm_{\Bx}} ) )} \\ \nonumber
  &= \ \frac{\exp(2 \  \beta \ m_{\max} \ 
  (  m_{\max} \ + \ \NRM{\Bm_{\Bx}} ) )}{\exp(2 \  \beta \ m_{\max} \
  (  m_{\max} \ + \ \NRM{\Bm_{\Bx}} ) ) \ + \ (N-1) } \\ \nonumber
  &\leq \ 1/N \ \exp(2 \  \beta \ m_{\max} \ (  m_{\max} \ + \ \NRM{\Bm_{\Bx}} ) ) \ .
\end{align}
Analogously we obtain for $i=\arg\max_j \Bm_{\Bx}^T \Bx_j$,
a bound on the maximal softmax component $i$ if the center is put into
the iteration:
\begin{align}
  [ \soft( \beta \ \BX^T \Bm_{\Bx})]_i \ &\leq \ 
  1/N \ \exp(2 \  \beta \ m_{\max} \ \NRM{\Bm_{\Bx}}) \ .
  \label{eq:bound_max_softmax_center}
\end{align}
Analog we obtain a bound for $i=\arg\max_j (\Bm_{\Bx}^* )^T \Bx_j$ 
on the maximal softmax component $i$ of the fixed point:
\begin{align}
\label{eq:bound_max_softmax_fixed}
  [ \soft( \beta \ \BX^T \Bm_{\Bx}^* )]_i \ &\leq \ 
  1/N \ \exp(2 \  \beta \ m_{\max} \ \NRM{\Bm_{\Bx}^* }) \\ \nonumber 
  &\leq \ 
  1/N \ \exp(2 \  \beta \ m_{\max} \ (  m_{\max} \ + \ 
  \NRM{\Bm_{\Bx}} ) ) \ .
\end{align}

The two important terms are 
$m_{\max}$, the variance or spread of 
the data and $\NRM{\Bm_{\Bx}}$, which tells how well
the data is centered.
For a contraction mapping we already required $\beta  m_{\max}^2<1$, therefore
the first term in the exponent is $2\beta  m_{\max}^2<2$. 
The second term $2 \beta  m_{\max} \NRM{\Bm_{\Bx}}$ is small if the data is centered.

{\textbullet \em Global fixed point near the global mean: Analysis using softmax values.}

If $\Bxi^T \Bx_i \approx \Bxi^T \Bx_j$ for all $i$ and $j$,
then $p_i \approx 1/N$ and we have $m= \max_i p_i (1-p_i) < 1/N$.
For $M\leq 1/\sqrt{2 \beta}$ we obtain from
Lemma~\ref{th:boundJ}:
\begin{align}
  \NRM{\rJ}_2 \ &< \ 1 \ .
\end{align}
The local fixed point is $\Bm_{\Bx}^* \approx \Bm_{\Bx} = (1/N) \sum_{i=1}^N \Bx_i$
with $p_i \approx 1/N$.

\vspace{0.5cm}

We now treat this case more formally.
First we discuss conditions that ensure that the iteration is a
contraction mapping. 
We consider the iteration Eq.~\eqref{eq:iter} in the variable $\Bp$:
\begin{align}
  \Bp^{\nn} \ &= \ g(\Bp) \ = \ \soft ( \beta \BX^T \BX \Bp) \ .
\end{align}
The Jacobian is
\begin{align}
  \rJ(\Bp) \ &= \ \frac{\partial g(\Bp)}{\partial \Bp} \ = \ 
   \BX^T \BX \ \rJ_s 
\end{align}
with
\begin{align}
  \rJ_s(\Bp^{\nn}) \ &= \  \beta \left( \diag(\Bp^{\nn}) 
  \ - \ \Bp^{\nn} (\Bp^{\nn})^T  \right) \ .
\end{align}
The version of the mean value theorem in Lemma~\ref{th:MVT} states for 
$\rJ^m = \int_{0}^1 \rJ(\lambda \Bp) \ \Rd \lambda
= \BX^T \BX \rJ_s^m $ with the symmetric matrix
$\rJ_s^m = \int_{0}^1 \rJ_s(\lambda \Bp) \ \Rd \lambda$:

\begin{align}
  \Bp^{\nn} \ &= \ g(\Bp) \ = \  g(\BZe) \ + \  
  (\rJ^m)^T \Bp  \ = \  g(\BZe) \ + \ 
  \rJ_s^m \ \BX^T \BX \ \Bp
  \ = \ 1/N \ \BOn \ + \  \rJ_s^m \ \BX^T \BX \ \Bp \ .
\end{align}

With $m= \max_i p_i (1-p_i)$, 
Eq.~\eqref{eq:softjacobi2-m} from
Lemma~\ref{th:JacobiB} is
\begin{align}
 \NRM{\rJ_s(\Bp)}_2 \ &= \ \beta \ \NRM{\diag(\Bp) - \Bp \Bp^T}_2
 \ \leq \ 2 \ m \ \beta \ .
\end{align}
First observe that
$\lambda p_i (1-\lambda p_i) \leq p_i (1-p_i)$ for $p_i\leq 0.5$ and $\lambda \in [0,1]$, since
$p_i(1-p_i) - \lambda p_i (1-\lambda p_i)=(1-\lambda) p_i (1-(1+\lambda) p_i) \geq 0$.
For $\max_i p_i \leq 0.5$ this observation leads to the following bound for $\rJ_s^m$:
\begin{align}
 \NRM{\rJ_s^m}_2  \ &\leq \ 2 \ m \ \beta \ .
\end{align}
Eq.~\eqref{eq:softjacobi2-beta} in Lemma~\ref{th:JacobiB} states that every $\rJ_s$ is bounded by $1/2 \beta$,
therefore also the mean:
\begin{align}
 \NRM{\rJ_s^m}_2 \ &\leq \ 0.5 \ \beta \ .
\end{align}
Since
$m= \max_i p_i (1-p_i)<\max_i p_i =p_{\max}$, the previous bounds can be combined as follows:
\begin{align}
 \NRM{\rJ_s^m}_2 \ &\leq \  2 \ \min \{0.25,p_{\max} \} \ \beta \ .
\end{align}
Consequently, 
\begin{align}
 \label{eq:boundJM}
 \NRM{\rJ^m}_2 \ &\leq  \ 
  N \ M^2  \ 2 \ \min \{0.25,p_{\max} \} \ \beta \ ,
\end{align}
where we used Eq.~\eqref{eq:normX}.
$\NRM{\BX^T \BX}_2=\NRM{\BX \BX^T}_2$,
therefore $\NRM{\BX^T \BX}_2$ is $N$ times the maximal second moment of the data squared.

Obviously, $g(\Bp) $ is a contraction mapping in compact sets, where
\begin{align}
  & N \ M^2  \ 2 \ \min \{0.25,p_{\max} \} \ \beta \ <  \ 1 \ .
\end{align}

$\rS$ is the sphere around the origin $\BZe$ with radius one.
For
\begin{align}
\label{eq:mvtA}
  \Bp^{\nn} \ &= \ g(\Bp) \ = \  1/N \ \BOn \ + \  \rJ^m \ \Bp \ ,
\end{align}
we have $\NRM{\Bp} \leq \NRM{\Bp}_1 = 1$ and 
$\NRM{\Bp^{\nn}} \leq \NRM{\Bp^{\nn}}_1 = 1$.
Therefore, $g$ maps points from $\rS$ into $\rS$.
$g$ is a contraction mapping for
\begin{align}
 \NRM{\rJ^m}_2 \ &\leq  \  
 N \ M^2  \ 2 \ \min \{0.25,p_{\max} \} \ \beta \ = \ c
  \ < \ 1 \ .
\end{align}
According to Banach fixed point theorem $g$ has
a fixed point in the sphere $\rS$.

H\"{o}lder's inequality gives:
\begin{align}
  \NRM{\Bp}^2\ &=  \  \Bp^T \Bp \ \leq \
 \NRM{\Bp}_1  \NRM{\Bp}_{\infty} \ = \ \NRM{\Bp}_{\infty} \ = \  p_{\max} \ .
\end{align}
Alternatively:
\begin{align}
  \NRM{\Bp}^2\ &=  \  \sum_i p_i^2 \ = \  p_{\max} \sum_i \frac{p_i}{p_{\max}}\ p_i
  \ \leq \  p_{\max} \sum_i p_i \ = \ p_{\max} \ .
\end{align}
Let now $\rS$ be the sphere around the origin $\BZe$ 
with radius $1/\sqrt{N} + \sqrt{p_{\max}}$
and let $\NRM{\rJ^m(\Bp)}_2 \leq c <1$ for $\Bp \in \rS$. 
The old $\Bp$ is in the sphere $\rS$ ($\Bp \in \rS$) since $p_{\max}<\sqrt{p_{\max}}$ for $p_{\max}<1$. 
We have 
\begin{align}
\label{eq:pnew1}
  \NRM{\Bp^{\nn}} \ &\leq \  1/\sqrt{N} \ + \  \NRM{\rJ^m}_2 \ \NRM{\Bp}
  \ \leq \  1/\sqrt{N} \ + \  \sqrt{p_{\max}} \ .
\end{align}
Therefore, $g$ is a mapping from $\rS$ into $\rS$ and a contraction mapping.
According to Banach fixed point theorem, a fixed point exists in $\rS$.

\vspace{1cm}
For the 1-norm, we use Lemma~\ref{th:JacobiB} and $\NRM{\Bp}_1=1$
to obtain from Eq.~\eqref{eq:mvtA}:
\begin{align}
  \NRM{\Bp^{\nn}\ - \ 1/N \ \BOn }_1 \ &\leq \  \NRM{\rJ^m}_1 
  \ \leq \ 2 \ \beta \ m  \ \NRM{\BX}_{\infty} \ M_1 \ , \\
  \NRM{\Bp^{\nn}\ - \ 1/N \ \BOn }_1 \ &\leq \  \NRM{\rJ^m}_1 
  \ \leq \ 2 \ \beta \ m \ N \ M_{\infty} \ M_1 \ , \\
   \NRM{\Bp^{\nn}\ - \ 1/N \ \BOn }_1 \ &\leq \  \NRM{\rJ^m}_1 
  \ \leq \ 2 \ \beta \ m \ N \ M^2 \ ,
\end{align}
where $m= \max_i p_i (1-p_i)$, $M_1= \NRM{\BX}_1 = \max_i \NRM{\Bx_i}_1$,
$M =  \max_i \NRM{\Bx_i}$,
$\NRM{\BX}_{\infty}=\NRM{\BX^T}_1=\max_i \NRM{[X^T]_i}_1$ (maximal absolute row sum norm),
and $M_{\infty}= \max_i \NRM{\Bx_i}_{\infty}$.
Let us quickly mention some auxiliary estimates related to $\BX^T \BX$:
\begin{align}
  \NRM{\BX^T \BX}_1 \ &= \  \max_i \sum_{j=1}^N \left| \Bx_i^T \Bx_j \right|
  \ \leq \ \max_i \sum_{j=1}^N  \NRM{\Bx_i}_{\infty} \ \NRM{\Bx_j}_1 \\ \nonumber
  &\leq \  M_{\infty} \ \sum_{j=1}^N  M_1
  \ = \ N \ M_{\infty} \ M_1 \ , 
 \end{align}
where the first inequaltiy is from H\"{o}lder's inequality.
We used
\begin{align}
  \NRM{\BX^T \BX}_1 \ &= \  \max_i \sum_{j=1}^N \left| \Bx_i^T \Bx_j \right|
  \ \leq \ \max_i \sum_{j=1}^N  \NRM{\Bx_i} \ \NRM{\Bx_j} \\ \nonumber
  &\leq \ M \ \sum_{j=1}^N  M
  \ = \ N \ M^2  \ , 
 \end{align}
where the first inequality is from H\"{o}lder's inequality (here the same as the Cauchy-Schwarz inequality).
See proof of Lemma~\ref{th:JacobiB} for the 1-norm bound on $J_s$. Everything else
follows from the fact that the 1-norm is sub-multiplicative as induced matrix norm.

\vspace{1cm}

We consider the minimal $\NRM{\Bp}$.
\begin{align}
  \min_{\Bp} &{\mbox{\ ~} } \NRM{\Bp}^2 \\ \nonumber
  \mbox{ s.t. } &{\mbox{\ ~} } \sum_i p_i = 1 \\ \nonumber
  &{\mbox{\ ~} } \forall_i: \ \ p_i \ \geq \ 0  \ .
\end{align}
The solution to this minimization problem is
$\Bp = (1/N) \BOn$.
Therefore, we have $1/\sqrt{N} \leq \NRM{\Bp}$ 
and $1/N \leq \NRM{\Bp}^2$ 
Using Eq.~\eqref{eq:pnew1} we obtain
\begin{align}
   1/\sqrt{N} \ &\leq \ \NRM{\Bp^{\nn}} 
  \ \leq \  1/\sqrt{N} \ + \  \sqrt{p_{\max}} \ .
\end{align}

Moreover
\begin{align}
  \NRM{\Bp^{\nn}}^2 \ &= \ (\Bp^{\nn})^T \Bp^{\nn} \ = \  1/N  \ + \ (\Bp^{\nn})^T \rJ^m \ \Bp 
  \ \leq \   1/N  \ + \  \NRM{\rJ^m}_2 \ \NRM{\Bp} \\ \nonumber
  &\leq \   1/N  \ + \  \NRM{\rJ^m}_2 \ ,
\end{align}
since $\Bp^{\nn} \in \rS$ and $\Bp \in \rS$.

For the fixed point, we have
\begin{align}
  \NRM{\Bp^*}^2 \ &= \ (\Bp^*)^T \Bp^* \ = \  1/N  \ + \ (\Bp^*)^T \rJ^m \ \Bp^* 
  \ \leq \   1/N  \ + \  \NRM{\rJ^m}_2 \ \NRM{\Bp^*}^2  \ ,
\end{align}
and hence
\begin{align}
  1/N \ &\leq \ \NRM{\Bp^*}^2 \ \leq \ 1/N  \frac{1}{1 \ - \ \NRM{\rJ^m}_2} \ = \ 
  1/ N  \ (1 \ + \ \frac{\NRM{\rJ^m}_2}{1 \ - \ \NRM{\rJ^m}_2}) \ .
\end{align}
Therefore, for small $\NRM{\rJ^m}_2$ we have $\Bp^* \approx (1/N) \BOn$.

\paragraph{Many Stable States: Fixed Points Near Stored Patterns.}
\label{sec:fixedPoints}

We move on to the next case, where the patterns $\Bx_i$ are well separated.
In this case the iteration goes to the pattern to which the initial
$\Bxi$ is most similar.
If the initial $\Bxi$ is similar to a vector $\Bx_i$ then it will
converge to $\Bx_i$ and $\Bp$ will be $\Be_i$.
The main ingredients are again Banach's Theorem and estimates on the Jacobian norm.

{\textbullet \em Proof of a fixed point by Banach Fixed Point Theorem.}

{$\rightarrow$ \em Mapped Vectors Stay in a Compact Environment.}
We show that if $\Bx_i$ is sufficient dissimilar to
other $\Bx_j$ then there is an compact environment of $\Bx_i$
(a sphere) 
where the fixed point iteration maps this environment into
itself.
The idea of the proof is to define a sphere around $\Bx_i$
for which points from the sphere are mapped by $f$ into the sphere.

We first need following lemma which bounds the distance
$\NRM{\Bx_i \ - \ f(\Bxi)}$, where $\Bx_i$ is the 
pattern that is least separated from $\Bxi$ but 
separated from other patterns.
\begin{lemmaA}
\label{th:similar}
For a query $\Bxi$ and data $\BX=(\Bx_1,\ldots,\Bx_N)$,
there exists a $\Bx_i$ that is least separated from $\Bxi$ while 
being separated from other $\Bx_j$ with $j \not= i$:
\begin{align}
  i \ &= \ \arg\max_k \min_{j,j \not= k} \left( \Bxi^T \Bx_k \ - \ \Bxi^T
    \Bx_j \right) \ = \ \arg\max_k \left( \Bxi^T \Bx_k \ - \ \max_{j,j \not= k} \Bxi^T
  \Bx_j \right) \\
  0 \ &\leq \ c \ = \ \max_k \min_{j,j \not= k} \left( \Bxi^T \Bx_k \ - \ \Bxi^T
    \Bx_j \right) \ = \ \max_k  \left( \Bxi^T \Bx_k \ - \ \max_{j, j \not= k} \Bxi^T
  \Bx_j \right) \ .
\end{align}
For $\Bx_i$, the following holds:
\begin{align}
  \NRM{\Bx_i \ - \ f(\Bxi)} \ &\leq \ 2 \ \epsilon \ M
  \ ,  
\end{align}
where 
\begin{align}
  M \ &= \ \max_{i} \NRM{\Bx_i}  \ , \\
  \epsilon \ &= \ (N-1) \ \exp(- \ \beta \ c ) \ .
\end{align}
\end{lemmaA}

\begin{proof}
For the softmax component $i$ we have:
\begin{align} \label{eq:softlower}
  [ \soft( \beta \ \BX^T \Bxi)]_i \ &= \ \frac{1}{1 \ + \ \sum_{j \not= i}
    \exp(\beta \ ( \Bxi^T \Bx_j \ - \ \Bxi^T \Bx_i ) )} \ \geq \
  \frac{1}{1 \ + \ \sum_{j \not= i}  \exp(- \ \beta \ c )} \\ \nonumber
  &= \ \frac{1}{1 \ + \ (N-1) \exp(- \ \beta \ c )}
  \ = \ 1 \ - \  \frac{(N-1) \exp(- \ \beta \ c )}{1 \ + \ (N-1) \exp(- \ \beta \ c )}\\ \nonumber
  &\geq \ 1 \ - \ (N-1) \exp(- \ \beta \ c ) \ = \ 1 \ - \ \epsilon \
\end{align}
For softmax components $k \not= i$ we have
\begin{align}
  [ \soft( \beta \BX^T \Bxi)]_k \ &= \ \frac{\exp(\beta \ ( \Bxi^T \Bx_k \ - \ \Bxi^T \Bx_i ) )}{1 \ + \ \sum_{j \not= i}
    \exp(\beta \ ( \Bxi^T \Bx_j \ - \ \Bxi^T \Bx_i ) )} \ \leq \
   \exp(- \ \beta \ c ) \ = \  \frac{\epsilon}{N-1} \ .
\end{align}

The iteration $f$ can be written as
\begin{align}
 f(\Bxi) \ &= \   \BX \soft ( \beta \BX^T \Bxi) \ = \ \sum_{j=1}^N
 \Bx_j \  [ \soft(\beta \BX^T \Bxi)]_j \ .
\end{align}

We now can bound  $\NRM{\Bx_i \ - \ f(\Bxi)}$:
\begin{align}
  \NRM{ \Bx_i \ - \ f(\Bxi) } \ &= \  \NRM{ \Bx_i \ - \ \sum_{j=1}^N
     [ \soft(\beta \BX^T \Bxi)]_j \ \Bx_j } \\ \nonumber
  &= \ \NRM{ (1- [ \soft(\beta \BX^T \Bxi)]_i) \ \Bx_i \ - \ \sum_{j=1,j\not= i}^N  [ \soft(\beta \BX^T \Bxi)]_j \ 
  \Bx_j} \\ \nonumber
  &\leq \ 
   \epsilon \ \NRM{ \Bx_i} \ + \ \frac{\epsilon}{N-1} \
    \sum_{j=1,j\not= i}^N   \NRM{\Bx_j} \\ \nonumber
  &\leq \  \epsilon \ M \ + \ \frac{\epsilon}{N-1} \
    \sum_{j=1,j\not= i}^N   M \ = \  2 \ \epsilon \ M
  \ . 
\end{align}
  
\end{proof}

We define $\Delta_i$, i.e.\ the separation of pattern $\Bx_i$ from data
$\BX=(\Bx_1,\ldots,\Bx_N)$ as:
\begin{align}
 \Delta_i \ &= \ \min_{j,j \not= i} \left( \Bx_i^T \Bx_i \ - \ \Bx_i^T
    \Bx_j \right) \ = \ \Bx_i^T \Bx_i \ - \ \max_{j,j \not= i} \Bx_i^T
  \Bx_j \ .
\end{align}
The pattern is separated from the other data if $0<\Delta_i$.
Using the parallelogram identity,  $\Delta_i$ can also be expressed as
\begin{align} \label{eq:diwithnorm}
  \Delta_i \ &= \ \min_{j,j \not= i} \frac{1}{2} \ \left(
  \NRM{\Bx_i}^2 \ - \  \NRM{\Bx_j}^2 \ + \  \NRM{\Bx_i \ - \ \Bx_j}^2
  \right) \\ \nonumber
  &= \ \frac{1}{2} \NRM{\Bx_i}^2  \ - \
  \frac{1}{2} \ \max_{j,j \not= i}  \left(  \NRM{\Bx_j}^2 \ - \
    \NRM{\Bx_i \ - \ \Bx_j}^2 \right) \ .
\end{align}
For $\NRM{\Bx_i} =  \NRM{\Bx_j}$ we have $\Delta_i= 1/2 \min_{j,j \not= i}  \NRM{\Bx_i \ - \ \Bx_j}^2$.

Next we define the sphere where we want to apply 
Banach fixed point theorem.
\begin{definition}[Sphere $\rS_i$]
\label{th:DefSphere}
The sphere $\rS_i$ is defined as
\begin{align}
 \rS_i \ &:= \ \left\{ \Bxi \mid \NRM{\Bxi \ - \ \Bx_i} \ 
 \leq \  \frac{1}{\beta \ N \ M} \right\} \  .
\end{align}
\end{definition}

\begin{lemmaA}
\label{th:stayInside}
With $\Bxi$ given, if the assumptions
\begin{enumerate}[label=A\arabic*:]
\item $\Bxi$ is inside sphere: $\Bxi \in \rS_i$,
\item data point $\Bx_i$ is well separated from the other data:
\begin{align}
 \Delta_i \ &\geq \  \frac{2}{\beta \ N} \ + \  \frac{1}{\beta} \ \ln \left( 2 \ (N-1) \ N \ \beta \
        M^2 \right) 
\end{align}
\end{enumerate}
hold, then $f(\Bxi)$ is inside the sphere: $f(\Bxi) \in \rS_i$.
Therefore, with assumption (A2),
$f$ is a mapping from $\rS_i$ into $\rS_i$.
\end{lemmaA}

\begin{proof}

We need the separation $\tilde{\Delta}_i$ of
$\Bxi$ from the data.
\begin{align}
  \tilde{\Delta}_i \ &= \ \min_{j,j \not= i} \left( \Bxi^T \Bx_i \ - \
    \Bxi^T \Bx_j \right) \ .
\end{align}
Using the Cauchy-Schwarz inequality, we obtain for $1 \leq j \leq N$:
\begin{align}
  \left| \Bxi^T \Bx_j \ - \ \Bx_i^T \Bx_j \right|  
 &\leq \ \NRM{\Bxi \ - \ \Bx_i} \  \NRM{\Bx_j} \ \leq
 \  \NRM{\Bxi \ - \ \Bx_i} \   M \ .
\end{align}
We have the lower bound
\begin{align} \label{eq:di_lower}
  \tilde{\Delta}_i \ &\geq \ \min_{j,j \not= i} \left(
    \left( \Bx_i^T \Bx_i \ - \  \NRM{\Bxi \ - \ \Bx_i }
    \   M \right) \ - \
    \left( \Bx_i^T \Bx_j \ + \  \NRM{\Bxi \ - \ \Bx_i}
    \   M \right) \right) \\ \nonumber
  &= \ - \ 2 \  \NRM{\Bxi \ - \ \Bx_i}  \   M
  \ + \ \min_{j, j \not= i} \left(
    \Bx_i^T \Bx_i \ - \ \Bx_i^T \Bx_j \right) \ = \
  \Delta_i \ - \ 2 \  \NRM{\Bxi \ - \ \Bx_i}  \ M \\ \nonumber
  &\geq \  \Delta_i \ - \ \frac{2}{\beta \ N}   \ ,
\end{align}
where we used the assumption (A1) of the lemma.

From the proof in Lemma~\ref{th:similar} we have
\begin{align}
  p_{\max} \ &= \  [\soft(\beta \BX^T \Bxi)]_i \ \geq \
  1 \ - \ (N-1) \ \exp(- \ \beta \ \tilde{\Delta}_i) \ = \ 1 \ - \
  \tilde{\epsilon} \ .
\end{align}
Lemma~\ref{th:similar} states that 
\begin{align}
  \NRM{\Bx_i \ - \ f(\Bxi)} \ &\leq \ 2 \ \tilde{\epsilon} \
  M \ = \ 2 \ (N-1) \ \exp(- \ \beta \ \tilde{\Delta}_i) \
  M \\ \nonumber
  &\leq \  2 \ (N-1) \ \exp(- \ \beta \  (\Delta_i \ - \ \frac{2}{\beta \ N}) ) \
  M  \ .  
\end{align}

We have
\begin{align}
  &\NRM{\Bx_i \ - \ f(\Bxi)} \\ \nonumber
  &\leq  \
  2 \ (N-1) \ \exp(- \ \beta \  ( \frac{2}{\beta \ N} \ + \  \frac{1}{\beta} \ \ln \left( 2 \ (N-1) \ N \ \beta \
        M^2 \right) \ - \ \frac{2}{\beta \ N}) ) \
      M \\ \nonumber
      &= \  2 \ (N-1) \ \exp(- \ \ln \left( 2 \ (N-1) \ N \ \beta \
        M^2 \right) ) \
      M \\ \nonumber
      &= \ \frac{1}{ N \ \beta \ M }    \ ,  
\end{align}
where we used assumption (A2) of the lemma.
Therefore, $f(\Bxi) $ is a mapping from the 
sphere $\rS_i$ into the sphere $\rS_i$: 
If $\Bxi \in \rS_i$ then $f(\Bxi) \in \rS_i$.
\end{proof}

{\textbullet \em Contraction mapping.}

For applying Banach fixed point theorem we need to show that
$f$ is contraction in the compact environment $\rS_i$.
\begin{lemmaA}
\label{th:contraction}
Assume that
\begin{enumerate}[label=A\arabic*:]
\item
\begin{align} \label{eq:diassumption}
 \Delta_i \ &\geq \  \frac{2}{\beta \ N} \ + \  \frac{1}{\beta} \ \ln \left( 2 \ (N-1) \ N \ \beta \
        M^2 \right) \ ,
\end{align}
\end{enumerate}    
then $f$ is a contraction mapping in $\rS_i$.
\end{lemmaA}

\begin{proof}

The version of the mean value theorem Lemma~\ref{th:MVT} states for 
$\rJ^m = \int_{0}^1 \rJ(\lambda \Bxi + (1-\lambda)\Bx_i) \ \Rd \lambda$:
\begin{align}
  f(\Bxi) \ &= \  f(\Bx_i) \ + \ \rJ^m \
  ( \Bxi \ - \ \Bx_i )  \ . 
\end{align}
Therefore
\begin{align}
  \NRM{f(\Bxi) \ - \ f(\Bx_i)} \ &\leq \   \NRM{\rJ^m}_2 \
  \NRM{ \Bxi \ - \ \Bx_i}  \ . 
\end{align}

We define $\tilde{\Bxi}=\lambda \Bxi + (1-\lambda)\Bx_i$ for some $\lambda \in [0,1]$.
From the proof in Lemma~\ref{th:similar} we have
\begin{align}
  \label{eq:boundDefs}
  p_{\max}(\tilde{\Bxi}) \ &= \  [\soft(\beta \ \BX^T \ \tilde{\Bxi})]_i \ \geq \
  1 \ - \ (N-1) \ \exp(- \ \beta \ \tilde{\Delta}_i) \ = \ 1 \ - \
  \tilde{\epsilon} \ , \\ 
  \tilde{\epsilon} \ &= \  (N-1) \ \exp(- \ \beta \ \tilde{\Delta}_i) \ , \\
  \tilde{\Delta}_i \ &= \ \min_{j,j \not= i} \left( \tilde{\Bxi}^T \Bx_i \ - \
    \tilde{\Bxi}^T \Bx_j \right) \ .
\end{align}

First we compute an upper bound on $\tilde{\epsilon}$.
We need the separation $\tilde{\Delta}_i$ of
$\Bxi$ from the data.
Using the Cauchy-Schwarz inequality, we obtain for $1 \leq j \leq N$:
\begin{align}
  \left| \tilde{\Bxi}^T \Bx_j \ - \ \Bx_i^T \Bx_j \right|  
 &\leq \ \NRM{\tilde{\Bxi} \ - \ \Bx_i} \  \NRM{\Bx_j} \ \leq
 \  \NRM{\tilde{\Bxi} \ - \ \Bx_i} \   M \ .
\end{align}
We have the lower bound on $\tilde{\Delta}_i$:
\begin{align}
\label{eq:boundD}
  \tilde{\Delta}_i \ &\geq \ \min_{j,j \not= i} \left(
    \left( \Bx_i^T \Bx_i \ - \  \NRM{\tilde{\Bxi} \ - \ \Bx_i }
    \   M \right) \ - \
    \left( \Bx_i^T \Bx_j \ + \  \NRM{\tilde{\Bxi} \ - \ \Bx_i}
    \   M \right) \right) \\ \nonumber
  &= \ - \ 2 \  \NRM{\tilde{\Bxi} \ - \ \Bx_i}  \   M
  \ + \ \min_{j, j \not= i} \left(
    \Bx_i^T \Bx_i \ - \ \Bx_i^T \Bx_j \right) \ = \
  \Delta_i \ - \ 2 \  \NRM{\tilde{\Bxi} \ - \ \Bx_i}  \ M \\ \nonumber
  &\geq \  \Delta_i \ - \ 2 \  \NRM{\Bxi \ - \ \Bx_i}  \ M\ ,
\end{align}
where we used $\NRM{\tilde{\Bxi}  -  \Bx_i} = \lambda \NRM{\Bxi - \Bx_i}
\leq \NRM{\Bxi - \Bx_i}$.
From the definition of $\tilde{\epsilon}$ in Eq.~\eqref{eq:boundDefs}
we have
\begin{align}
  \tilde{\epsilon} \ &= \
  (N-1) \ \exp(- \ \beta \ \tilde{\Delta}_i) \\ \nonumber
  &\leq  \  (N-1) \ 
  \exp \left(- \ \beta \ \left(  \Delta_i \ - \ 2 \  \NRM{\Bxi \ - \ \Bx_i} 
  \ M \right) \right)  \\ \nonumber
  &\leq \ (N-1) \ \exp\left(- \ \beta \ \left(\Delta_i \ - \ \frac{2}{\beta \ N} \right) \right)\ ,
\end{align}
where we used $\Bxi \in \rS_i$, therefore $\NRM{\Bxi \ - \ \Bx_i} \ 
 \leq \  \frac{1}{\beta \ N \ M}$.

Next we compute an lower bound on $\tilde{\epsilon}$.
We start with an upper on $\tilde{\Delta}_i$:
\begin{align}
\label{eq:boundDu}
  \tilde{\Delta}_i \ &\leq \ \min_{j,j \not= i} \left(
    \left( \Bx_i^T \Bx_i \ + \  \NRM{\tilde{\Bxi} \ - \ \Bx_i }
    \   M \right) \ - \
    \left( \Bx_i^T \Bx_j \ - \  \NRM{\tilde{\Bxi} \ - \ \Bx_i}
    \   M \right) \right) \\ \nonumber
  &= \ 2 \  \NRM{\tilde{\Bxi} \ - \ \Bx_i}  \   M
  \ + \ \min_{j, j \not= i} \left(
    \Bx_i^T \Bx_i \ - \ \Bx_i^T \Bx_j \right) \ = \
  \Delta_i \ + \ 2 \  \NRM{\tilde{\Bxi} \ - \ \Bx_i}  \ M  \\ \nonumber
  &\leq \ \Delta_i \ + \ 2 \  \NRM{\Bxi \ - \ \Bx_i}  \ M   \ ,
\end{align}
where we used $\NRM{\tilde{\Bxi}  -  \Bx_i} = \lambda \NRM{\Bxi - \Bx_i}
\leq \NRM{\Bxi - \Bx_i}$.
From the definition of $\tilde{\epsilon}$ in Eq.~\eqref{eq:boundDefs}
we have
\begin{align}
  \tilde{\epsilon} \ &= \
  (N-1) \ \exp(- \ \beta \ \tilde{\Delta}_i) \\ \nonumber
  &\geq \ (N-1) \ \exp\left(- \ \beta \ \left(\Delta_i \ + \ 2 \  
  \NRM{\Bxi \ - \ \Bx_i}  \ M \right) \right)
  \\ \nonumber
  &\geq \ (N-1) \ \exp\left(- \ \beta \ \left(\Delta_i \ + \ \frac{2}{\beta \ N} \right) \right)\ ,
\end{align}
where we used $\Bxi \in \rS_i$, therefore $\NRM{\Bxi \ - \ \Bx_i} \ 
 \leq \  \frac{1}{\beta \ N \ M}$.

Now we bound the Jacobian.
We can assume $\tilde{\epsilon} \leq 0.5$ otherwise 
$(1 -  \tilde{\epsilon}) \leq 0.5$ in the following.
From the proof of
Lemma~\ref{th:JacobiB} we know
for $p_{\max}(\tilde{\Bxi}) \geq 1-\tilde{\epsilon}$, 
then $p_i(\tilde{\Bxi}) \leq \tilde{\epsilon}$ 
for $p_i(\tilde{\Bxi})\not= p_{\max}(\tilde{\Bxi})$.
Therefore, $p_i(\tilde{\Bxi}) (1-p_i(\tilde{\Bxi})) \leq m \leq
\tilde{\epsilon} (1 -  \tilde{\epsilon})$ for all $i$.
Next we use the derived upper and lower bound on $\tilde{\epsilon}$ 
in previous Eq.~\eqref{eq:boundJpmax} in Lemma~\ref{th:boundJ}:
\begin{align}
  \label{eq:boundJJ1}
  \NRM{\rJ(\tilde{\Bxi})}_2 
  \ &\leq \  2 \ \beta \ N \ M^2 \ \tilde{\epsilon}
 \ - \  2 \ \tilde{\epsilon}^2 \ \beta \ N \ M^2 \\ \nonumber
 &\leq \ 2 \ \beta \ N \ M^2 \  (N-1) \ 
  \exp \left(- \ \beta \ \left(  \Delta_i \ - \frac{2}{\beta \ N}\right) \right) 
 \ - \\ \nonumber
 &2 \ (N-1)^2 \ \exp\left(- \ 2 \ \beta \ \left(\Delta_i \ + \ \frac{2}{\beta \ N} \right) \right) \ \beta \ N \ M^2   \ .
\end{align}

The bound Eq.~\eqref{eq:boundJJ1} holds for the mean $\rJ^m$, too,
since it averages over $\rJ(\tilde{\Bxi})$:
\begin{align}
  \label{eq:boundJJ}
  \NRM{\rJ^m}_2 \ &\leq \ 2 \ \beta \ N \ M^2 \  (N-1) \ 
  \exp \left(- \ \beta \ \left(  \Delta_i \ - \ \frac{2}{\beta \ N} \right) \right) 
 \ - \\ \nonumber
 &2 \ (N-1)^2 \ \exp\left(- \ 2 \ \beta \ \left(\Delta_i \ + \ \frac{2}{\beta \ N} \right) \right) \ \beta \ N \ M^2   \ .
\end{align}

The assumption of the lemma is
\begin{align}
 \Delta_i \ &\geq \  \frac{2}{\beta \ N} \ + \  \frac{1}{\beta} \ \ln \left( 2 \ (N-1) \ N \ \beta \
        M^2 \right) \ ,
\end{align}
This is 
\begin{align}
 \Delta_i \ - \ \frac{2}{\beta \ N} \ &\geq \ \frac{1}{\beta} \ \ln \left( 2 \ (N-1) \ N \ \beta \
        M^2 \right) \ ,
\end{align}

Therefore, the spectral norm  $\NRM{\rJ}_2$ can be bounded by:    
\begin{align}
\label{eq:boundDJ}
  &\NRM{\rJ^m}_2 \  \leq \
  2 \ \beta \ (N-1) \ \exp\left(- \ \beta \ \frac{1}{\beta} \ \ln \left( 2 \ (N-1) \ N \ \beta \
        M^2 \right)  \right) \ N \
  M^2 \ - \\ \nonumber
 &2 \ (N-1)^2 \ \exp\left(- \ 2 \ \beta \ \left(\Delta_i \ + \ \frac{2}{\beta \ N} \right) \right) \ \beta \ N \ M^2 \\ \nonumber
  &= \  \ 2 \ \beta \ (N-1) \ \frac{1}{2 \ (N-1) \ N \ \beta \
        M^2}  \ N \
  M^2 \ - \\ \nonumber
 &2 \ (N-1)^2 \ \exp\left(- \ 2 \ \beta \ \left(\Delta_i \ + \ \frac{2}{\beta \ N} \right) \right) \ \beta \ N \ M^2 \\ \nonumber 
 &=  \  1 \ -
 2 \ (N-1)^2 \ \exp\left(- \ 2 \ \beta \ \left(\Delta_i \ + \ \frac{2}{\beta \ N} \right) \right) \ \beta \ N \ M^2  \ < 1 \ .
\end{align}

Therefore, $f$ is a contraction mapping in $\rS_i$.
\end{proof}

{\textbullet \em  Banach Fixed Point Theorem.}
Now we have all ingredients to apply Banach fixed point theorem.
\begin{lemmaA}
\label{th:banach}
Assume that
\begin{enumerate}[label=A\arabic*:]
\item
\begin{align}
 \Delta_i \ &\geq \  \frac{2}{\beta \ N} \ + \  \frac{1}{\beta} \ \ln \left( 2 \ (N-1) \ N \ \beta \
        M^2 \right) \ ,
    \end{align}
\end{enumerate}    
then $f$ has a fixed point in $\rS_i$.
\end{lemmaA}

\begin{proof}
We use Banach fixed point theorem: 
Lemma~\ref{th:stayInside} says that $f$ maps from $\rS_i$ into $\rS_i$.
Lemma~\ref{th:contraction} says that $f$ is a contraction mapping in
$\rS_i$.
\end{proof}

{\textbullet \em Contraction mapping with a fixed point.}
\label{sec:Mapfixed}

We have shown that a fixed point exists. We want to know
how fast the iteration converges to the fixed point.
Let $\Bx_i^*$ be the fixed point of the iteration $f$ in the sphere $\rS_i$.
Using the mean value theorem Lemma~\ref{th:MVT}, we have with
$\rJ^m = \int_{0}^1 \rJ(\lambda \Bxi + (1-\lambda)\Bx_i^*) \ \Rd \lambda$:
\begin{align}
  \NRM{f(\Bxi) \ - \ \Bx_i^*} \ &= \ \NRM{f(\Bxi) \ - \ f(\Bx_i^*)} 
  \ \leq \  \NRM{\rJ^m}_2 \ \NRM{\Bxi \ - \ \Bx_i^*}
\end{align}

According to Lemma~\ref{th:JacobiB},
if $p_{\max}=\max_i p_i \geq 1-\epsilon$ for all 
$\tilde{\Bx}=\lambda \Bxi + (1-\lambda)\Bx_i^*$, 
then the spectral norm of
the Jacobian is bounded by 
\begin{align}
 \NRM{\rJ_s (\tilde{\Bx}) }_2 \ &< \ 2 \ \epsilon  \ \beta \ .
\end{align}
The norm of Jacobian at $\tilde{\Bx}$ is bounded
\begin{align}
  \NRM{\rJ (\tilde{\Bx}) }_2  \ &\leq \
  2 \ \beta \ \NRM{\BX}_2^2 \ \epsilon 
  \ \leq \ 2 \ \beta \ N M^2 \ \epsilon \ .
\end{align}
We used that
the spectral norm  $\NRM{.}_2$ is bounded by the Frobenius norm
$\NRM{.}_F$ which can be expressed by the norm squared of its
column vectors:
\begin{align}
  \NRM{\BX}_2 \ &\leq \  \NRM{\BX}_F \ = \ \sqrt{\sum_i\NRM{\Bx_i}^2} \ .
\end{align}
Therefore
\begin{align}
 \label{eq:normX}
  \NRM{\BX}_2^2 \ &\leq \  N \ M^2  \ .
\end{align}
The norm of Jacobian of the fixed point iteration is bounded
\begin{align}
 \label{eq:Jbound1}
  \NRM{\rJ^m}_2  \ &\leq \
  2 \ \beta \ \NRM{\BX}_2^2 \ \epsilon 
  \ \leq \ 2 \ \beta \ N M^2 \ \epsilon \ .
\end{align}

The separation of pattern $\Bx_i$ from data $\BX=(\Bx_1,\ldots,\Bx_N)$
is 
\begin{align}
 \Delta_i \ &= \ \min_{j, j \not= i} \left( \Bx_i^T \Bx_i \ - \ \Bx_i^T
    \Bx_j \right) \ = \ \Bx_i^T \Bx_i \ - \ \max_{j, j \not= i} \Bx_i^T
  \Bx_j \ .
\end{align}
We need the separation $\tilde{\Delta}_i$ of
$\tilde{\Bx}=\lambda \Bxi + (1-\lambda)\Bx_i^*$ from the data:
\begin{align}
  \tilde{\Delta}_i \ &= \ \min_{j, j \not= i} \left( \tilde{\Bx}^T \Bx_i \ - \
\tilde{\Bx}^T \Bx_j \right) \ .
\end{align}
We compute a lower bound on $\tilde{\Delta}_i$.
Using the Cauchy-Schwarz inequality, we obtain for $1 \leq j \leq N$:
\begin{align}
  \left| \tilde{\Bx}^T \Bx_j \ - \ \Bx_i^T \Bx_j \right|  
 &\leq \ \NRM{\tilde{\Bx} \ - \ \Bx_i} \  \NRM{\Bx_j} \ \leq
 \  \NRM{\tilde{\Bx} \ - \ \Bx_i} \   M \ .
\end{align}
We have the lower bound
\begin{align}
  \tilde{\Delta}_i \ &\geq \ \min_{j, j \not= i} \left(
    \left( \Bx_i^T \Bx_i \ - \  \NRM{\tilde{\Bx} \ - \ \Bx_i }
    \   M \right) \ - \
    \left( \Bx_i^T \Bx_j \ + \  \NRM{\tilde{\Bx} \ - \ \Bx_i}
    \   M \right) \right) \\ \nonumber
  &= \ - \ 2 \  \NRM{\tilde{\Bx} \ - \ \Bx_i}  \   M
  \ + \ \min_{j, j \not= i} \left(
    \Bx_i^T \Bx_i \ - \ \Bx_i^T \Bx_j \right) \ = \
  \Delta_i \ - \ 2 \  \NRM{\tilde{\Bx} \ - \ \Bx_i}  \
  M \ .
\end{align}
Since
\begin{align}
  \NRM{\tilde{\Bx} \ - \ \Bx_i} \ &= \ 
  \NRM{\lambda \Bxi + (1-\lambda)\Bx_i^* \ - \ \Bx_i} \\ \nonumber
  &\leq \ \lambda \ \NRM{\Bxi  \ - \ \Bx_i} \ + \ 
  (1-\lambda) \ \NRM{\Bx_i^* \ - \ \Bx_i} \\ \nonumber
  &\leq \ \max \{ \NRM{\Bxi  \ - \ \Bx_i} , \NRM{\Bx_i^* \ - \ \Bx_i} \} \ ,
\end{align}
we have
\begin{align} \label{eq:dilowerfixedp}
  \tilde{\Delta}_i \ &\geq \ 
  \Delta_i \ - \ 2 \  \max \{ \NRM{\Bxi  \ - \ \Bx_i} , \NRM{\Bx_i^* \ - \ \Bx_i} \}  \
  M \ .
\end{align}

For the softmax component $i$ we have:
\begin{align}
  &[ \soft( \beta \ \BX^T \tilde{\Bxi})]_i \ = \ \frac{1}{1 \ + \ \sum_{j \not= i}
    \exp(\beta \ ( \tilde{\Bxi}^T \Bx_j \ - \ \tilde{\Bxi}^T \Bx_i ) )} \\ \nonumber
    &\geq \
  \frac{1}{1 \ + \ \sum_{j \not= i}  \exp(- \ \beta \ 
  (\Delta_i \ - \ 2 \  \max \{ \NRM{\Bxi  \ - \ \Bx_i} , \NRM{\Bx_i^* \ - \ \Bx_i} \}   \ M) )} \\ \nonumber
  &= \ \frac{1}{1 \ + \ (N-1) \exp(- \ \beta \ 
  (\Delta_i \ - \ 2 \  \max \{ \NRM{\Bxi  \ - \ \Bx_i} , \NRM{\Bx_i^* \ - \ \Bx_i} \}  \ M)  )} \\ \nonumber
  &= \ 1 \ - \  \frac{(N-1) \exp(- \ \beta \ 
  (\Delta_i \ - \ 2 \  \max \{ \NRM{\Bxi  \ - \ \Bx_i} , \NRM{\Bx_i^* \ - \ \Bx_i} \}   \ M) )}{1 \ + \ (N-1) \exp(- \ \beta \ 
  (\Delta_i \ - \ 2 \  \max \{ \NRM{\Bxi  \ - \ \Bx_i} , \NRM{\Bx_i^* \ - \ \Bx_i} \}   \ M) )}\\ \nonumber
  &\geq \ 1 \ - \ (N-1) \exp(- \ \beta \ 
  (\Delta_i \ - \ 2 \  \max \{ \NRM{\Bxi  \ - \ \Bx_i} , \NRM{\Bx_i^* \ - \ \Bx_i} \}   \ M) ) \\ \nonumber
  &= \ 1 \ - \ \epsilon \ .
\end{align}
Therefore
\begin{align}
\label{eq:epsA}
 \epsilon \ &= \ (N-1) \exp(- \ \beta \ 
 (\Delta_i \ - \ 2 \  \max \{ \NRM{\Bxi  \ - \ \Bx_i} , \NRM{\Bx_i^* \ - \ \Bx_i} \}  \ M) )  \ .
\end{align}

We can bound the spectral norm of 
the Jacobian, which upper bounds the Lipschitz constant: 
\begin{align}
\label{eq:boundJfix}
  \NRM{\rJ^m}_2  \ &\leq \
  2 \ \beta \ N \ M^2 \ (N-1) \exp(- \ \beta \
  (\Delta_i \ - \ 2 \  \max \{ \NRM{\Bxi  \ - \ \Bx_i} , \NRM{\Bx_i^* \ - \ \Bx_i} \}  \ M) )\ .
\end{align}

For a contraction mapping we require
\begin{align}
  \NRM{\rJ^m}_2  \ < \ 1 \ ,
\end{align}
which can be ensured by
\begin{align}
  &2 \ \beta \ N M^2 \ (N-1) \exp(- \ \beta \ 
  (\Delta_i \ - \ 2 \  \max \{ \NRM{\Bxi  \ - \ \Bx_i} , \NRM{\Bx_i^* \ - \ \Bx_i} \}  \ M) ) \ < \ 1 \ .
\end{align}
Solving this inequality for $\Delta_i$ gives
\begin{align}
\label{eq:boundD1}
 \Delta_i \ &> \  2 \  \max \{ \NRM{\Bxi  \ - \ \Bx_i} , \NRM{\Bx_i^* \ - \ \Bx_i} \}  \ M \ + \ 
 \frac{1}{\beta} \ \ln \left( 2 \ (N-1) \ N \ \beta \  M^2 \right) \ .
\end{align}
In an environment around $\Bx_i^*$ in
which Eq.~\eqref{eq:boundD1} holds, $f$ is a contraction mapping and every
point converges under the iteration $f$ to $\Bx_i^*$ when the iteration stays in
the environment.
After every iteration the mapped point $f(\Bxi)$ is closer to the fixed point $\Bx_i^*$
than the original point $\Bx_i$:
\begin{align}
  \NRM{f(\Bxi) \ - \ \Bx_i^*}
  \ &\leq \  \NRM{\rJ^m}_2 \ \NRM{\Bxi \ - \ \Bx_i^*} \ < \ \NRM{\Bxi \ - \ \Bx_i^*} \ .
\end{align}

Using
\begin{align}
  \NRM{f(\Bxi) \ - \ \Bx_i^*}
  \ \leq \  \NRM{\rJ^m}_2 \ \NRM{\Bxi \ - \ \Bx_i^*} \ \leq \ 
  \NRM{\rJ^m}_2 \ \NRM{\Bxi \ - \ f(\Bxi)} \ + \  
  \NRM{\rJ^m}_2 \ \NRM{f(\Bxi)\ - \ \Bx_i^*}  \ ,
\end{align}
we obtain
\begin{align}
 \NRM{f(\Bxi) \ - \ \Bx_i^*}
  \ \leq \  \frac{\NRM{\rJ^m}_2}{1 \ - \ \NRM{\rJ^m}_2} \  \NRM{\Bxi \ - \ f(\Bxi)}  \ .
\end{align}
For large $\Delta_i$ the iteration is close to the fixed point even after
one update.
This has been confirmed in several experiments.

\paragraph{Metastable States: Fixed Points Near Mean of Similar Patterns.}

The proof concept is the same as for a single pattern
but now for the arithmetic mean of similar patterns.

{\textbullet \em Bound on the Jacobian.}

The Jacobian of the fixed point iteration is
\begin{align}
 \rJ \ &= \   \beta \ \BX \left(
  \diag(\Bp) - \Bp \Bp^T \right) \BX^T \ = \ \BX \rJ_s \BX^T \ .
\end{align}

If we consider $p_i$ as the probability of selecting
the vector $\Bx_i$, then we can define
expectations as $\EXP_{\Bp}[f(\Bx)]=\sum_{i=1}^N p_i f(\Bx_i)$.
In this setting the matrix 
\begin{align}
  \BX \left(
  \diag(\Bp) - \Bp \Bp^T \right) \BX^T 
\end{align}
is the covariance matrix of data $\BX$ when its 
vectors are selected according 
to the probability $\Bp$:
\begin{align}
 &\BX \left(  \diag(\Bp) \ - \ \Bp \Bp^T \right) \BX^T 
 \ = \ \BX \diag(\Bp) \BX^T \ - \ \BX \Bp \Bp^T \BX^T \\
 &= \ \sum_{i=1}^N p_i \ \Bx_i \ \Bx_i^T \ - \ 
  \left( \sum_{i=1}^N p_i \ \Bx_i \right) 
   \left( \sum_{i=1}^N p_i \ \Bx_i \right)^T  \\
 &= \ \EXP_{\Bp}[\Bx \ \Bx^T] \ - \ \EXP_{\Bp}[\Bx] \ \EXP_{\Bp}[\Bx]^T
 \ = \ \VAR_{\Bp}[\Bx] \ ,
\end{align}
therefore we have
\begin{align}
 \rJ \ &= \ \beta \ \VAR_{\Bp}[\Bx]  \ .
\end{align}
We now elaborate more on this interpretation as variance.
Specifically the singular values of $\rJ$ (or in other words: the
covariance) should be reasonably small.
The singular values are the key to ensure convergence of the iteration
Eq.~\eqref{eq:iter}.
Next we present some thoughts.
\begin{enumerate}
\item It's clear that the largest eigenvalue of the covariance matrix
(equal to the largest singular value)
is the variance in the direction of the eigenvector
associated with the largest eigenvalue.

\item Furthermore the variance goes to zero as one $p_i$ goes to
one, since only one pattern is chosen and there is
no variance.

\item The variance is reasonable small if all patterns 
are chosen with equal probability.

\item The variance is small if few similar patterns are
chosen with high probability.
If the patterns are sufficient similar, then
the spectral norm of the covariance matrix is
smaller than one.
\end{enumerate}

The first three issues have already been adressed.
Now we focus on the last one in greater detail.
We assume that the first $l$ patterns 
are much more probable (and similar
to one another) than the other patterns.
Therefore, we define:
\begin{align}
 \label{eq:defA1}
 M \ &:= \ \max_i \NRM{\Bx_i} \ , \\
  \gamma \ &= \ \sum_{i=l+1}^N p_i \ \leq \ \epsilon  \ , \\
  1- \gamma \ &= \ \sum_{i=1}^l p_i \ \geq \ 1 \ - \ \epsilon \ , \\
  \tilde{p}_i \ &:= \ \frac{p_i}{1- \gamma } \ \leq \ p_i/(1-\epsilon) \ , \\
  \sum_{i=1}^l \tilde{p}_i \ &= \ 1  \ , \\
  \Bm_{\Bx} \ &= \ \frac{1}{l} \ \sum_{i=1}^l \ \Bx_i \ , \\
  m_{\max} \ &= \  \max_{1\leq i \leq l} \NRM{\Bx_i \ - \ \Bm_{\Bx}}  \ .
\end{align}
$M$ is an upper bound on the Euclidean norm of the patterns, which are vectors.
$\epsilon$ is an upper bound on the probability $\gamma$ of not choosing one of the first $l$ patterns, while 
$1-\epsilon$ is a lower bound the probability $(1-\gamma)$ 
of choosing one of the first $l$ patterns. 
$\Bm_{\Bx}$ is the arithmetic mean (the center) of the first $l$ 
patterns. 
$m_{\max}$ is the maximal distance of the patterns to the center $\Bm_{\Bx}$ .
$\tilde{\Bp}$ is the probability $\Bp$
normalized for the first $l$ patterns.

The variance of the first $l$ patterns is
\begin{align}
\label{eq:defA2}
   \VAR_{\tilde{p}}[\Bx_{1:l}] \ &= \  
  \sum_{i=1}^l \tilde{p}_i \ \Bx_i \ \Bx_i^T
  \ - \   \left(\sum_{i=1}^l \tilde{p}_i \ \Bx_i\right) \ 
  \left(\sum_{i=1}^l \tilde{p}_i \ \Bx_i\right)^T \\ \nonumber
  &= \ 
    \sum_{i=1}^l \tilde{p}_i \ 
    \left(\Bx_i \ - \ \sum_{i=1}^l \tilde{p}_i \Bx_i \right) \ 
     \left(\Bx_i \ - \ \sum_{i=1}^l \tilde{p}_i \Bx_i \right)^T   \ .
\end{align}

\begin{lemmaA}
\label{th:jacobiBoundM}
With the definitions in Eq.~\eqref{eq:defA1}
to Eq.~\eqref{eq:defA2}, 
the following bounds on the norm $\NRM{\rJ}_2$ of the
Jacobian of the fixed point iteration hold.
The $\gamma$-bound for $\NRM{\rJ}_2$ is
\begin{align} \label{eq:jacobiestlpatterns}
   \NRM{\rJ}_2 \ &\leq \ \beta \left(
 (1 - \gamma) \ m_{\max}^2 \ + \ 
   \gamma \ 2 \ (2 \ - \ \gamma ) \ M^2 \right) 
\end{align}
and the $\epsilon$-bound for $\NRM{\rJ}_2$ is:
\begin{align}
   \NRM{\rJ}_2 \ &\leq \ \beta \left(
  \ m_{\max}^2 \ + \ 
   \epsilon \ 2 \ (2 \ - \ \epsilon ) \ M^2 \right) \ .
\end{align}

\end{lemmaA}

\begin{proof}

The variance $\VAR_{\tilde{p}}[\Bx_{1:l}]$ can be expressed as:
\begin{align}
   &(1-\gamma) \ \VAR_{\tilde{p}}[\Bx_{1:l}] \ = \  
     \sum_{i=1}^l p_i \ 
    \left(\Bx_i \ - \ \frac{1}{1-\gamma} \ 
    \sum_{i=1}^l p_i \ \Bx_i \right) \ 
     \left(\Bx_i \ - \ \frac{1}{1-\gamma} \ 
     \sum_{i=1}^l p_i \ \Bx_i \right)^T 
     \\ \nonumber 
     &= \ 
     \sum_{i=1}^l p_i \ \Bx_i \ \Bx_i^T \ - \ 
     \left(\sum_{i=1}^l p_i \ \Bx_i \right) \
     \frac{1}{1-\gamma} \
     \left(\sum_{i=1}^l p_i \ \Bx_i \right)^T \
     - \ 
     \frac{1}{1-\gamma} \ \left(\sum_{i=1}^l p_i \ \Bx_i \right) \ 
     \left(\sum_{i=1}^l p_i \ \Bx_i \right)^T \\  \nonumber
     &+ \ 
     \frac{\sum_{i=1}^l p_i}{(1-\gamma)^2} \ \left(\sum_{i=1}^l p_i \ \Bx_i \right) \ 
     \left(\sum_{i=1}^l p_i \ \Bx_i \right)^T  \ = \ 
     \sum_{i=1}^l p_i \ \Bx_i \ \Bx_i^T \ - \ 
     \frac{1}{1-\gamma} \ \left(\sum_{i=1}^l p_i \ \Bx_i \right) \ 
     \left(\sum_{i=1}^l p_i \ \Bx_i \right)^T \\ \nonumber
     &= \ 
     \sum_{i=1}^l p_i \ \Bx_i \ \Bx_i^T \ - \ \left(\sum_{i=1}^l p_i \ \Bx_i \right) \ 
     \left(\sum_{i=1}^l p_i \ \Bx_i \right)^T \ + \
     \left(1 \ - \ \frac{1}{1-\gamma} \right)  \
     \left(\sum_{i=1}^l p_i \ \Bx_i \right) \ 
     \left(\sum_{i=1}^l p_i \ \Bx_i \right)^T \\ \nonumber
     &= \ 
     \sum_{i=1}^l p_i \ \Bx_i \ \Bx_i^T \ - \ \left(\sum_{i=1}^l p_i \ \Bx_i \right) \ 
     \left(\sum_{i=1}^l p_i \ \Bx_i \right)^T \ - \
      \frac{\gamma}{1-\gamma}  \
     \left(\sum_{i=1}^l p_i \ \Bx_i \right) \ 
     \left(\sum_{i=1}^l p_i \ \Bx_i \right)^T  \ .
\end{align}

Therefore, we have
\begin{align}
  &\sum_{i=1}^l p_i \ \Bx_i \ \Bx_i^T \ - \ \left(\sum_{i=1}^l p_i \ \Bx_i \right) \ 
     \left(\sum_{i=1}^l p_i \ \Bx_i \right)^T \\ \nonumber
     &= \ 
     (1-\gamma) \ \VAR_{\tilde{p}}[\Bx_{1:l}]
     \ + \ \frac{\gamma}{1-\gamma}  \
     \left(\sum_{i=1}^l p_i \ \Bx_i \right) \ 
     \left(\sum_{i=1}^l p_i \ \Bx_i \right)^T  \ .
 \end{align}

We now can reformulate the Jacobian $\rJ$:
\begin{align}
  \rJ \ &= \ \beta \ \left( \sum_{i=1}^l p_i \ \Bx_i \ \Bx_i^T 
  \ + \ \sum_{i=l+1}^N p_i \ \Bx_i \ \Bx_i^T \right. \\ \nonumber 
  &- \ \left. \left( \sum_{i=1}^l p_i \ \Bx_i 
  \ + \ \sum_{i=l+1}^N p_i \ \Bx_i \right) 
  \left( \sum_{i=1}^l p_i \ \Bx_i 
  \ + \ \sum_{i=l+1}^N p_i \ \Bx_i \right)^T \right) \\ \nonumber
  &= \ \beta \ \left( \sum_{i=1}^l p_i \ \Bx_i \ \Bx_i^T 
  \ - \ \left( \sum_{i=1}^l p_i \ \Bx_i \right) \ 
  \left( \sum_{i=1}^l p_i \ \Bx_i \right)^T  \right. \\ \nonumber
   &+ \ \left. \sum_{i=l+1}^N p_i \ \Bx_i \ \Bx_i^T  
  \ - \  \left( \sum_{i=l+1}^N p_i \ \Bx_i \right) \
  \left( \sum_{i=l+1}^N p_i \ \Bx_i \right)^T 
  \right. \\ \nonumber 
  &- \ \left. 
  \left( \sum_{i=1}^l p_i \ \Bx_i \right) \
  \left(\sum_{i=l+1}^N p_i \ \Bx_i \right)^T
  \ - \ 
   \left( \sum_{i=l+1}^N p_i \ \Bx_i \right) 
  \left( \sum_{i=1}^l p_i \ \Bx_i \right)^T 
  \right) \\ \nonumber
  &= \ \beta \ \left(  (1-\gamma) \ \VAR_{\tilde{p}}[\Bx_{1:l}]
  \ + \ \frac{\gamma}{1-\gamma}  \
     \left(\sum_{i=1}^l p_i \ \Bx_i \right) \ 
     \left(\sum_{i=1}^l p_i \ \Bx_i \right)^T   \right. \\ \nonumber
   &+ \ \left. \sum_{i=l+1}^N p_i \ \Bx_i \ \Bx_i^T  
  \ - \  \left( \sum_{i=l+1}^N p_i \ \Bx_i \right) \
  \left( \sum_{i=l+1}^N p_i \ \Bx_i \right)^T 
  \right. \\ \nonumber 
  &- \ \left. 
  \left( \sum_{i=1}^l p_i \ \Bx_i \right) \
  \left(\sum_{i=l+1}^N p_i \ \Bx_i \right)^T
  \ - \ 
   \left( \sum_{i=l+1}^N p_i \ \Bx_i \right) 
  \left( \sum_{i=1}^l p_i \ \Bx_i \right)^T 
  \right) \ .
\end{align}

The spectral norm of an outer product of two vectors 
is the product of the Euclidean norms of the vectors:
\begin{align}
\NRM{\Ba \Bb^T}_2 \ &= \ \sqrt{\lambda_{\max}(\Bb \Ba^T \Ba \Bb^T) }
 \ = \ \NRM{\Ba} \ \sqrt{\lambda_{\max}(\Bb \Bb^T)} \ = \ \NRM{\Ba} \ \NRM{\Bb} \ ,
\end{align}
since $\Bb \Bb^T$ has eigenvector $\Bb/\NRM{\Bb}$ with
eigenvalue $\NRM{\Bb}^2$ and otherwise zero eigenvalues.

We now bound the norms of some matrices and vectors:
\begin{align}
  \NRM{\sum_{i=1}^l p_i \ \Bx_i}
  \ &\leq \ \sum_{i=1}^l p_i \ \NRM{\Bx_i} \ \leq \ (1- \gamma)
  \ M \ , \\ 
  \NRM{\sum_{i=l+1}^N p_i \ \Bx_i}
  \ &\leq \ \sum_{i=l+1}^N p_i  \ \NRM{\Bx_i} \ \leq \ \gamma \ M \ , \\ 
  \NRM{\sum_{i=l+1}^N p_i \ \Bx_i \ \Bx_i^T }_2  
  \ &\leq \ \sum_{i=l+1}^N p_i \ \NRM{\Bx_i \ \Bx_i^T }_2 \ =  \ 
  \sum_{i=l+1}^N p_i \ \NRM{\Bx_i}^2 \ \leq \ \sum_{i=l+1}^N p_i  \ M^2 \ = \ \gamma
  \ M^2  \ .
\end{align}

In order to bound the variance of the
first $l$ patterns,
we compute the vector $\Ba$ that minimizes 
\begin{align}
 f(\Ba) \ &= \  \sum_{i=1}^l p_i \NRM{\Bx_i \ - \ \Ba}^2
 \ = \  \sum_{i=1}^l p_i (\Bx_i \ - \ \Ba)^T (\Bx_i \ - \ \Ba)  \ .
\end{align}

The solution to 
\begin{align}
 \frac{\partial f(\Ba)}{\partial \Ba} \ &= \ 
 2 \ \sum_{i=1}^N p_i (\Ba \ - \ \Bx_i ) \ = \ 0 
\end{align}
is
\begin{align}
 \Ba \ &= \  \sum_{i=1}^N p_i \Bx_i \ .
\end{align}
The Hessian of $f$ is positive definite since
\begin{align}
 \frac{\partial^2 f(\Ba)}{\partial \Ba^2} \ &= \ 
 2 \ \sum_{i=1}^N p_i \ \BI \ = \ 2 \ \BI  
\end{align}
and $f$ is a convex function.
Hence, the mean
\begin{align}
\bar{\Bx} \ &:= \ \sum_{i=1}^N p_i \ \Bx_i 
\end{align}
minimizes 
$\sum_{i=1}^N p_i \NRM{\Bx_i  - \Ba}^2$.
Therefore, we have
\begin{align}
  \sum_{i=1}^l p_i \NRM{\Bx_i \ - \ \bar{\Bx}}^2 \ &\leq \ \sum_{i=1}^l p_i 
  \NRM{\Bx_i \ - \ \Bm_{\Bx}}^2 \ \leq \ (1 \ - \ \gamma) \ m_{\max}^2 \ .
\end{align}

We now bound the variance on the first $l$ patterns:
\begin{align} \label{eq:varest}
  (1-\gamma) \ \NRM{\VAR_{\tilde{p}}[\Bx_{1:l}]}_2 
    \ &\leq \ \sum_{i=1}^l p_i \NRM{\left( \Bx_i \ - \ \bar{\Bx} \right) \
   \left( \Bx_i \ - \ \bar{\Bx} \right)^T}_2 \\ \nonumber
   &= \ \sum_{i=1}^l p_i \NRM{\Bx_i \ - \ \bar{\Bx}}^2
   \ \leq \ \sum_{i=1}^l p_i 
  \NRM{\Bx_i \ - \ \Bm_{\Bx}}^2 \ \leq \ (1 \ - \ \gamma) \ m_{\max}^2 \ .
\end{align}

We obtain for the spectral norm of $\rJ$:
\begin{align}
  \NRM{\rJ}_2 \ &\leq \ \beta \left(
   (1-\gamma) \ \NRM{\VAR_{\tilde{p}}[\Bx_{1:l}]}_2  \right. \\ \nonumber
  &+ \ \left. \frac{\gamma}{1-\gamma} 
  \ \NRM{\left( \sum_{i=1}^l p_i \ \Bx_i \right) \ 
  \left( \sum_{i=1}^l p_i \ \Bx_i \right)^T}_2  \right. \\ \nonumber
   &+ \ \left. \NRM{\sum_{i=l+1}^N p_i \ \Bx_i \ \Bx_i^T}_2  
  \ + \  \NRM{\left( \sum_{i=l+1}^N p_i \ \Bx_i \right) \
  \left( \sum_{i=l+1}^N p_i \ \Bx_i \right)^T}_2 
  \right. \\ \nonumber 
  &+ \ \left. 
  \NRM{\left( \sum_{i=1}^l p_i \ \Bx_i \right) \
  \left(\sum_{i=l+1}^N p_i \ \Bx_i \right)^T}_2
  \ + \ 
   \NRM{\left( \sum_{i=l+1}^N p_i \ \Bx_i \right) 
  \left( \sum_{i=1}^l p_i \ \Bx_i \right)^T}_2 \right) \\ \nonumber
  &\leq \ \beta \left(
   (1 - \gamma) \ \NRM{\VAR_{\tilde{p}}[\Bx_{1:l}]}_2  \ + \  
   \gamma \ (1- \gamma) \ M^2  
   \ + \  \gamma \ M^2  
  \ + \  \gamma^2 \ M^2\ + \right. \\ \nonumber  
  &\left. \gamma \ (1- \gamma) \ M^2
  \ + \ 
   \gamma \ (1- \gamma) \ M^2  \right) \\ \nonumber
   &= \ \beta \left(
   (1 - \gamma) \ \ \NRM{\VAR_{\tilde{p}}[\Bx_{1:l}]}_2  \ + \ 
   \gamma \ 2 \ (2 \ - \ \gamma ) \ M^2 \right) \ . 
\end{align}

Combining the previous two estimates immediately
leads to Eq.~\eqref{eq:jacobiestlpatterns}.

The function $h(x)=x  2  (2  -  x ) $ has the derivative $h'(x)=4(1-x)$.
Therefore, $h(x)$ is monotone increasing for $x<1$.
For $0 \leq \gamma \leq \epsilon < 1$,
we can immediately deduce that
$ \gamma  2  (2  -  \gamma ) \leq \epsilon  2  (2  -  \epsilon )$.
Since $\epsilon$ is larger than $\gamma$,
we obtain the following $\epsilon$-bound for $\NRM{\rJ}_2$:
\begin{align}
   \NRM{\rJ}_2 \ &\leq \ \beta \left(
  \ m_{\max}^2 \ + \ 
   \epsilon \ 2 \ (2 \ - \ \epsilon ) \ M^2 \right) \ .
\end{align}

\end{proof}

We revisit the bound on $(1-\gamma) \ \VAR_{\tilde{p}}[\Bx_{1:l}]$.
The trace $\sum_{k=1}^d e_k$ is the sum of the eigenvalues
$e_k$. The spectral norm is equal to the 
largest eigenvalue $e_1$, that is, the largest singular value.
We obtain:
\begin{align}
  \NRM{\VAR_{\tilde{p}}[\Bx_{1:l}]}_2 
    \ &= \ \TR \left(\sum_{i=1}^l p_i \left( \Bx_i \ - \ 
    \bar{\Bx} \right) \
   \left( \Bx_i \ - \ \bar{\Bx} \right)^T \right) \ - \ 
   \sum_{k=2}^d e_k \\ \nonumber
   &= \ \sum_{i=1}^l p_i \TR \left( \left( \Bx_i \ - \ 
    \bar{\Bx} \right) \
   \left( \Bx_i \ - \ \bar{\Bx} \right)^T \right)
   \ - \ 
   \sum_{k=2}^d e_k \\ \nonumber
   &= \ \sum_{i=1}^l p_i \NRM{\Bx_i \ - \ \bar{\Bx}}^2  \ - \ 
   \sum_{k=2}^d e_k \ .
\end{align}
Therefore, the tightness of the bound depends on eigenvalues
which are not the largest. That is variations which are not
along the strongest variation weaken the bound.

{\textbullet \em Proof of a fixed point by Banach Fixed Point Theorem.}

Without restricting the generality,
we assume that the first $l$ patterns are 
much more probable (and similar
to one another) than the other patterns.
Therefore, we define:
\begin{align}
 M \ &:= \ \max_i \NRM{\Bx_i} \ , \\
  \gamma \ &= \ \sum_{i=l+1}^N p_i \ \leq \ \epsilon  \ , \\
  1- \gamma \ &= \ \sum_{i=1}^l p_i \ \geq \ 1 \ - \ \epsilon \ , \\
  \tilde{p}_i \ &:= \ \frac{p_i}{1- \gamma } \ \leq \ p_i/(1-\epsilon) \ , \\
  \sum_{i=1}^l \tilde{p}_i \ &= \ 1  \ , \\
  \Bm_{\Bx} \ &= \ \frac{1}{l} \ \sum_{i=1}^l \ \Bx_i \ , \\
  m_{\max} \ &= \  \max_{1\leq i \leq l} \NRM{\Bx_i \ - \ \Bm_{\Bx}}  \ .
\end{align}
$M$ is an upper bound on the Euclidean norm of the patterns, which are vectors.
$\epsilon$ is an upper bound on the probability $\gamma$ of not choosing one of the first $l$ patterns, while 
$1-\epsilon$ is a lower bound the probability $(1-\gamma)$ 
of choosing one of the first $l$ patterns. 
$\Bm_{\Bx}$ is the arithmetic mean (the center) of the first $l$ 
patterns. 
$m_{\max}$ is the maximal distance of the patterns to the center $\Bm_{\Bx}$ .
$\tilde{\Bp}$ is the probability $\Bp$
normalized for the first $l$ patterns.

{\textbullet \em Mapped vectors stay in a compact environment.}
We show that if $\Bm_{\Bx}$ is sufficient dissimilar to
other $\Bx_j$ with $l <j$ 
then there is an compact environment of $\Bm_{\Bx}$
(a sphere) 
where the fixed point iteration maps this environment into
itself.
The idea of the proof is to define a sphere around $\Bm_{\Bx}$
for which the points from the sphere are mapped by $f$ into the sphere.

We first need following lemma which bounds the distance
$\NRM{\Bm_{\Bx} \ - \ f(\Bxi)}$ of a $\Bxi$ which is close
to $\Bm_{\Bx}$.
\begin{lemmaA}
\label{th:similarM}
For a query $\Bxi$ and data $\BX=(\Bx_1,\ldots,\Bx_N)$,
we define 
\begin{align}
  0 \ &\leq \ c \ = \ \min_{j, l<j} \left( \Bxi^T \Bm_{\Bx} \ - \ \Bxi^T
    \Bx_j \right) \ = \ \Bxi^T \Bm_{\Bx} \ - \ \max_{j, l<j} \Bxi^T
  \Bx_j \ .
\end{align}
The following holds:
\begin{align}
  \NRM{\Bm_{\Bx} \ - \ f(\Bxi)} \ &\leq \
  m_{\max} \ + \ 2 \ \gamma \ M \ \leq \
  m_{\max} \ + \ 2 \ \epsilon \ M \ ,  
\end{align}
where 
\begin{align}
  M \ &= \ \max_{i} \NRM{\Bx_i}  \ , \\
  \epsilon \ &= \ (N-l) \ \exp(- \ \beta \ c ) \ .
\end{align}
\end{lemmaA}

\begin{proof}
Let $s=\arg\max_{j,j\leq l} \Bxi^T \Bx_j$,
therefore $\Bxi^T \Bm_{\Bx}= \frac{1}{l} \ \sum_{i=1}^l \ \Bxi^T \Bx_i
\leq \frac{1}{l} \ \sum_{i=1}^l \ \Bxi^T \Bx_s = \Bxi^T \Bx_s$.
For softmax components $j$ with $l<j$ we have
\begin{align}
  [ \soft( \beta \BX^T \Bxi)]_j \ &= \ \frac{\exp(\beta \ ( \Bxi^T \Bx_j
  \ - \ \Bxi^T \Bx_s ) )}{1 \ + \ \sum_{k,k\not=s}
    \exp(\beta \ ( \Bxi^T \Bx_k \ - \ \Bxi^T \Bx_s  ) )} \ \leq \
   \exp(- \ \beta \ c ) \ = \  \frac{\epsilon}{N-l} \ ,
\end{align}
since $\Bxi^T \Bx_s - \Bxi^T \Bx_j \geq \Bxi^T \Bm_{\Bx} -  \Bxi^T \Bx_j$ for each $j$ with $l<j$,
therefore $\Bxi^T \Bx_s - \Bxi^T \Bx_j \geq c$

The iteration $f$ can be written as
\begin{align}
 f(\Bxi) \ &= \   \BX \soft ( \beta \BX^T \Bxi) \ = \ \sum_{j=1}^N
 \Bx_j \  [ \soft(\beta \BX^T \Bxi)]_j \ .
\end{align}

We set $p_i=[ \soft(\beta \BX^T \Bxi)]_i$,
therefore $\sum_{i=1}^l p_i = 1 - \gamma \geq 1 - \epsilon$ 
and $\sum_{i=l+1}^N p_i = \gamma \leq \epsilon$.
Therefore
\begin{align} \label{eq:mxbound}
 &\NRM{ \Bm_{\Bx} \ - \ \sum_{j=1}^l  \frac{p_j}{1-\gamma} \ 
  \Bx_j}^2 \ = \ \NRM{ \sum_{j=1}^l  \frac{p_j}{1-\gamma} 
  \left(\Bm_{\Bx} \ - \ \Bx_j \right) }^2 \\ \nonumber
  &= \ \sum_{j=1,k=1}^l  \frac{p_j}{1-\gamma} \frac{p_k}{1-\gamma} 
  \left(\Bm_{\Bx} \ - \ \Bx_j \right)^T \left(\Bm_{\Bx} \ - \ \Bx_k \right) \\ \nonumber
  &= \ \frac{1}{2} \ \sum_{j=1,k=1}^l  \frac{p_j}{1-\gamma} \frac{p_k}{1-\gamma} 
  \left( \NRM{ \Bm_{\Bx} \ - \ \Bx_j}^2 \ + \ \NRM{ \Bm_{\Bx} \ - \ \Bx_k}^2  
  \ - \ \NRM{ \Bx_j \ - \ \Bx_k}^2 \right)  \\ \nonumber
  &= \ \sum_{j=1}^l \frac{p_j}{1-\gamma} \ \NRM{ \Bm_{\Bx} \ - \ \Bx_j}^2 \ - \
  \frac{1}{2} \ \sum_{j=1,k=1}^l  \frac{p_j}{1-\gamma} \frac{p_k}{1-\gamma} 
  \NRM{ \Bx_j \ - \ \Bx_k}^2 \\ \nonumber
  &\leq \ \sum_{j=1}^l \frac{p_j}{1-\gamma} \ \NRM{ \Bm_{\Bx} \ - \ \Bx_j}^2 \ \leq \ 
   m_{\max}^2 \ .
 \end{align}
It follows that
\begin{align}
 \NRM{ \Bm_{\Bx} \ - \ \sum_{j=1}^l  \frac{p_j}{1-\gamma} \ 
  \Bx_j} \ &\leq \   m_{\max}
\end{align}

We now can bound  $\NRM{\Bm_{\Bx} \ - \ f(\Bxi)}$:
\begin{align}
  \NRM{ \Bm_{\Bx} \ - \ f(\Bxi) } \ &= \  \NRM{ \Bm_{\Bx} \ - \ \sum_{j=1}^N
    p_j \ \Bx_j } \\ \nonumber
  &= \ \NRM{ \Bm_{\Bx} \ - \ \sum_{j=1}^l  p_j \ 
  \Bx_j \ - \ \sum_{j=l+1}^N  p_j \ 
  \Bx_j} \\ \nonumber
  &= \ \NRM{ \Bm_{\Bx} \ - \ \sum_{j=1}^l  \frac{p_j}{1-\gamma} \ 
  \Bx_j \ + \ \frac{\gamma}{1-\gamma} \ \sum_{j=1}^l  p_j \ 
  \Bx_j \ - \ \sum_{j=l+1}^N  p_j \ 
  \Bx_j} \\ \nonumber
  &\leq \ \NRM{ \Bm_{\Bx} \ - \ \sum_{j=1}^l  \frac{p_j}{1-\gamma} \ 
  \Bx_j} \ + \ \frac{\gamma}{1-\gamma} \ \NRM{\sum_{j=1}^l  p_j \ 
  \Bx_j} \ + \ \NRM{\sum_{j=l+1}^N  p_j \ 
  \Bx_j} \\ \nonumber
  &\leq \ \NRM{ \Bm_{\Bx} \ - \ \sum_{j=1}^l  \frac{p_j}{1-\gamma} \ 
  \Bx_j} \ + \ \frac{\gamma}{1-\gamma} \ \sum_{j=1}^l  p_j \ 
  M \ + \ \sum_{j=l+1}^N  p_j \ 
  M \\ \nonumber
  &\leq \ \NRM{ \Bm_{\Bx} \ - \ \sum_{j=1}^l  \frac{p_j}{1-\gamma} \ 
  \Bx_j} \ + \ 2 \ \gamma \ M \\ \nonumber
  &\leq \  m_{\max} \ + \ 2 \ \gamma \ M \ \leq \
  m_{\max} \ + \ 2 \ \epsilon \ M  \ ,
\end{align}
where we applied Eq.~\eqref{eq:mxbound} in the penultimate inequality.
This is the statement of the lemma.
\end{proof}

The separation of the center (the arithmetic mean) 
$\Bm_{\Bx}$ of the first $l$ 
from data $\BX=(\Bx_{l+1},\ldots,\Bx_N)$
is $\Delta_m$, defined as 
\begin{align}
 \Delta_m \ &= \ \min_{j,l<j} \left( \Bm_{\Bx}^T \Bm_{\Bx} \ - \ 
 \Bm_{\Bx}^T  \Bx_j \right) \ = \ 
 \Bm_{\Bx}^T \Bm_{\Bx} \ - \ \max_{j,l<j} \Bm_{\Bx}^T \Bx_j \ .
\end{align}
The center is separated from the other data $\Bx_j$
with $l<j$ if $0<\Delta_m$.
By the same arguments as in Eq.~\eqref{eq:diwithnorm}, $\Delta_m$ can also be expressed as
\begin{align}
  \Delta_m \ &= \ \min_{j,l<j} \frac{1}{2} \ \left(
  \NRM{\Bm_{\Bx}}^2 \ - \  \NRM{\Bx_j}^2 \ + \  \NRM{\Bm_{\Bx} \ - \ \Bx_j}^2
  \right) \\ \nonumber
  &= \ \frac{1}{2} \NRM{\Bm_{\Bx}}^2  \ - \
  \frac{1}{2} \ \max_{j,l<j}  \left(  \NRM{\Bx_j}^2 \ - \
    \NRM{\Bm_{\Bx} \ - \ \Bx_j}^2 \right) \ .
\end{align}
For $\NRM{\Bm_{\Bx}} =  \NRM{\Bx_j}$ we have $\Delta_m= 1/2 \min_{j,l<j}  \NRM{\Bm_{\Bx} \ - \ \Bx_j}^2$.

Next we define the sphere where we want to apply 
Banach fixed point theorem.
\begin{definition}[Sphere $\rS_m$]
\label{th:DefSphereM}
The sphere $\rS_m$ is defined as
\begin{align}
 \rS_m \ &:= \ \left\{ \Bxi \mid \NRM{\Bxi \ - \ \Bm_{\Bx}} \ \leq \ 
 \frac{1}{\beta \ m_{\max} }  \right\} \  .
\end{align}
\end{definition}

\begin{lemmaA}
\label{th:stayInsideM}
With $\Bxi$ given, if the assumptions
\begin{enumerate}[label=A\arabic*:]
\item $\Bxi$ is inside sphere: $\Bxi \in \rS_m$,
\item the center $\Bm_{\Bx}$ is well separated from other data $\Bx_j$
with $l<j$:
\begin{align}
 \Delta_m \ &\geq \  \frac{2 \ M}{\beta \ m_{\max}} \ - \  
 \frac{1}{\beta} \ \ln \left(  \frac{ 1 \ - \  \beta \ m_{\max}^2    }{
    2 \ \beta \ (N-l) \ M \ \max \{ m_{\max} \ , \ 2 \ M \} } \right)  \ ,
\end{align}
\item the distance $m_{\max}$ of similar patterns to the center is sufficient small:
\begin{align}
  \beta \ m_{\max}^2 \ &\leq \ 1 
\end{align}
\end{enumerate}
hold, then $f(\Bxi) \in \rS_m$.
Therefore, under conditions (A2) and (A3), $f$ is a mapping from $\rS_m$ into $\rS_m$.
\end{lemmaA}

\begin{proof}
We need the separation $\tilde{\Delta}_m$ of
$\Bxi$ from the rest of the data, which is the last $N-l$ 
data points  $\BX=(\Bx_{l+1},\ldots,\Bx_N)$.
\begin{align}
  \tilde{\Delta}_m \ &= \ \min_{j,l<j} \left( \Bxi^T \Bm_{\Bx} \ - \
    \Bxi^T \Bx_j \right) \ .
\end{align}
Using the Cauchy-Schwarz inequality, we obtain for $l+1 \leq j \leq N$:
\begin{align}
  \left| \Bxi^T \Bx_j \ - \ \Bm_{\Bx}^T \Bx_j \right|  
 &\leq \ \NRM{\Bxi \ - \ \Bm_{\Bx}} \  \NRM{\Bx_j} \ \leq
 \  \NRM{\Bxi \ - \ \Bm_{\Bx}} \   M \ .
\end{align}
We have the lower bound
\begin{align}
  \tilde{\Delta}_m \ &\geq \ \min_{j,l<j} \left(
    \left( \Bm_{\Bx}^T \Bm_{\Bx} \ - \  \NRM{\Bxi \ - \ \Bm_{\Bx} }
    \   M \right) \ - \
    \left( \Bm_{\Bx}^T \Bx_j \ + \  \NRM{\Bxi \ - \ \Bm_{\Bx}}
    \   M \right) \right) \\ \nonumber
  &= \ - \ 2 \  \NRM{\Bxi \ - \ \Bm_{\Bx}}  \   M
  \ + \ \min_{j,l<j} \left(
    \Bm_{\Bx}^T \Bm_{\Bx} \ - \ \Bm_{\Bx}^T \Bx_j \right) \ = \
  \Delta_m \ - \ 2 \  \NRM{\Bxi \ - \ \Bm_{\Bx}}  \ M \\ \nonumber
  &\geq \  \Delta_m \ - \ 2 \ \frac{M}{\beta \ m_{\max} }  \ ,
\end{align}
where we used the assumption (A1) of the lemma.

From the proof in Lemma~\ref{th:similarM} we have
\begin{align}
  \sum_{i=1}^l p_i \ &\geq \
  1 \ - \ (N-l) \ \exp(- \ \beta \ \tilde{\Delta}_m) \ = \ 1 \ - \
  \tilde{\epsilon}  \ , \\
  \sum_{i=l+1}^N p_i \ &\leq \
  (N-l) \ \exp(- \ \beta \ \tilde{\Delta}_m) \ = \  \tilde{\epsilon}  \ .
\end{align}
Lemma~\ref{th:similarM} states that 
\begin{align}
  \NRM{\Bm_{\Bx} \ - \ f(\Bxi)} \ &\leq \
  m_{\max} \ + \ 2 \ \tilde{\epsilon}  \ M \\ \nonumber
  &\leq \ m_{\max} \ + \ 2 \ (N-l) \ 
  \exp(- \ \beta \  \tilde{\Delta}_m ) \ M  \ .  \\ \nonumber
  &\leq \ m_{\max} \ + \ 2 \ (N-l) \ 
  \exp(- \ \beta \  ( \Delta_m \ - \ 2 \ \frac{M}{\beta \ m_{\max} } ) ) \ M  \ .  
\end{align}

Therefore, we have
\begin{align}
  &\NRM{\Bm_{\Bx} \ - \ f(\Bxi)} \ \leq \
   m_{\max} \ + \ 2 \ (N-l) \ 
  \exp \left(- \ \beta \  ( \Delta_m \ - \ 2 \ \frac{M}{\beta \ m_{\max} } ) \right) \ 
  M \\ \nonumber
  &\leq \
   m_{\max} \ + \ 2 \ (N-l) \ 
  \exp \left( - \ \beta \  \left( \frac{2 \ M}{\beta \ m_{\max}} \ - \right. \right.\\ \nonumber
  & \left. \left. \frac{1}{\beta} \ \ln \left(  \frac{ 1 \ - \  \beta \ m_{\max}^2    }{
     2 \ \beta \ (N-l) \ M \ \max \{ m_{\max} \ , \ 2 \ M \}   }  \right)  \ - \ 2 \ \frac{M}{\beta \ m_{\max} } \right) \right) \ 
  M \\ \nonumber
  &= \
   m_{\max} \ + \ 2 \ (N-l) \ 
    \frac{ 1 \ - \  \beta \ m_{\max}^2    }{
     2 \ \beta \ (N-l) \ M \ \max \{ m_{\max} \ , \ 2 \ M \}   }  \ 
  M \\ \nonumber
  &\leq \
   m_{\max} \ + \ 
    \frac{ 1 \ - \  \beta \ m_{\max}^2    }{
    \beta \  m_{\max} }  \ = \ \frac{1}{ \beta \  m_{\max} }   \ ,
\end{align}
where we used assumption (A2) of the lemma.
Therefore, $f(\Bxi) $ is a mapping from the 
sphere $\rS_m$ into the sphere $\rS_m$. 

\begin{align}
&m_{\max} = \max_{1 \leq i \leq l}  \NRM{\Bx_i -\Bm_{\Bx}} \\
&= \max_{1 \leq i \leq l}  \NRM{\Bx_i - 1/l \sum_{j=1}^l \Bx_j} \\ 
&= \max_{1 \leq i \leq l}  \NRM{ 1/l \sum_{j=1}^l (\Bx_i -\Bx_j)} \\
&\leq \max_{1 \leq i,j  \leq l}  \NRM{\Bx_i -\Bx_j} \\
&\leq \max_{1 \leq i  \leq l}  \NRM{\Bx_i} + \max_{1 \leq j  \leq l}  \NRM{\Bx_i} \\
&\leq 2 M 
\end{align}
\end{proof}

{\textbullet \em Contraction mapping.}

For applying Banach fixed point theorem we need to show that
$f$ is contraction in the compact environment $\rS_m$.
\begin{lemmaA}
\label{th:contractionM}
Assume that
\begin{enumerate}[label=A\arabic*:]
\item
\begin{align} \label{eq:dmcondition}
 \Delta_m \ &\geq \  \frac{2 \ M}{\beta \ m_{\max}} \ - \  
 \frac{1}{\beta} \ \ln \left(  \frac{ 1 \ - \  \beta \ m_{\max}^2    }{
    2 \ \beta \ (N-l) \ M \ \max \{ m_{\max} \ , \ 2 \ M \} } \right)  \ ,
\end{align}
and 
\item
\begin{align}
  \beta \ m_{\max}^2 \ &\leq \ 1 \ ,
\end{align}
\end{enumerate}    
then $f$ is a contraction mapping in $\rS_m$.
\end{lemmaA}

\begin{proof}

The version of the mean value theorem Lemma~\ref{th:MVT} states for the symmetric
$\rJ^m = \int_{0}^1 \rJ(\lambda \Bxi + (1-\lambda)\Bm_{\Bx}) \ \Rd \lambda$:
\begin{align}
  f(\Bxi) \ &= \  f(\Bm_{\Bx}) \ + \ \rJ^m \
  ( \Bxi \ - \ \Bm_{\Bx} )  \ . 
\end{align}
In complete analogy to Lemma~\ref{th:contraction}, we get: 
\begin{align}
  \NRM{f(\Bxi) \ - \ f(\Bm_{\Bx})} \ &\leq \   \NRM{\rJ^m}_2 \
  \NRM{ \Bxi \ - \ \Bm_{\Bx}}  \ . 
\end{align}

We define $\tilde{\Bxi}=\lambda \Bxi + (1-\lambda)\Bm_{\Bx}$ 
for some $\lambda \in [0,1]$.
We need the separation $\tilde{\Delta}_m$ of
$\tilde{\Bxi}$ from the rest of the data, which is the last $N-l$ 
data points  $\BX=(\Bx_{l+1},\ldots,\Bx_N)$.
\begin{align}
  \tilde{\Delta}_m \ &= \ \min_{j,l<j} \left( \tilde{\Bxi}^T \Bm_{\Bx} \ - \
    \tilde{\Bxi}^T \Bx_j \right) \ .
\end{align}
From the proof in Lemma~\ref{th:similarM} we have
\begin{align}
  \tilde{\epsilon} \ &= \ (N-l) \ \exp(- \ \beta \ \tilde{\Delta}_m) \ , \\
  \sum_{i=1}^l p_i(\tilde{\Bxi}) \ &\geq \
  1 \ - \ (N-l) \ \exp(- \ \beta \ \tilde{\Delta}_m) \ = \ 1 \ - \
  \tilde{\epsilon}  \ , \\
  \sum_{i=l+1}^N p_i(\tilde{\Bxi}) \ &\leq \
  (N-l) \ \exp(- \ \beta \ \tilde{\Delta}_m) \ = \  \tilde{\epsilon}  \ .
\end{align}

We first compute an upper bound on $\tilde{\epsilon}$.
Using the Cauchy-Schwarz inequality, we obtain for $l+1 \leq j \leq N$:
\begin{align}
  \left| \tilde{\Bxi}^T \Bx_j \ - \ \Bm_{\Bx}^T \Bx_j \right|  
 &\leq \ \NRM{\tilde{\Bxi} \ - \ \Bm_{\Bx}} \  \NRM{\Bx_j} \ \leq
 \  \NRM{\tilde{\Bxi} \ - \ \Bm_{\Bx}} \   M \ .
\end{align}
We have the lower bound on $\tilde{\Delta}_m$:
\begin{align}
  \tilde{\Delta}_m \ &\geq \ \min_{j,l<j} \left(
    \left( \Bm_{\Bx}^T \Bm_{\Bx} \ - \  \NRM{\tilde{\Bxi} \ - \ \Bm_{\Bx} }
    \   M \right) \ - \
    \left( \Bm_{\Bx}^T \Bx_j \ + \  \NRM{\tilde{\Bxi} \ - \ \Bm_{\Bx}}
    \   M \right) \right) \\ \nonumber
  &= \ - \ 2 \  \NRM{\tilde{\Bxi} \ - \ \Bm_{\Bx}}  \   M
  \ + \ \min_{j,l<j} \left(
    \Bm_{\Bx}^T \Bm_{\Bx} \ - \ \Bm_{\Bx}^T \Bx_j \right) \ = \
  \Delta_m \ - \ 2 \  \NRM{\tilde{\Bxi} \ - \ \Bm_{\Bx}}  \ M \\ \nonumber
  &\geq \ \Delta_m \ - \ 2 \  \NRM{\Bxi \ - \ \Bm_{\Bx}}  \ M\ .
\end{align}
where we used $\NRM{\tilde{\Bxi}  -  \Bm_{\Bx}} = \lambda \NRM{\Bxi - \Bm_{\Bx}}
\leq \NRM{\Bxi - \Bm_{\Bx}}$.
We obtain the upper bound on $\tilde{\epsilon}$:
\begin{align}
  \tilde{\epsilon} \ &\leq \ (N-l) \ \exp \left(- \ \beta \ \left( \Delta_m \ - \ 2 \  \NRM{\Bxi \ - \ \Bm_{\Bx}}  \ M\right) \right) \\ \nonumber
  &\leq \  (N-l) \ \exp \left(- \ \beta \ \left( \Delta_m \ - \ 
  \frac{2 \ M}{\beta \ m_{\max}} \right) \right) 
  \ .
\end{align}
where we used that in the sphere $\rS_i$ holds:
\begin{align}
 \NRM{\Bxi \ - \ \Bm_{\Bx}} \ \leq \ 
 \frac{1}{\beta \ m_{\max} } \ ,
\end{align}
therefore
\begin{align} \label{eq:sphereest}
 2 \ \NRM{\Bxi \ - \ \Bm_{\Bx}} \ M \ \leq \ 
 \frac{2 \ M}{\beta \ m_{\max}}  \ .
 \end{align}

 Next we compute a lower bound on $\tilde{\epsilon}$
 and to this end start with the upper bound on $\tilde{\Delta}_m$
 using the same arguments as in Eq.~\eqref{eq:boundDu} in combination with Eq.~\eqref{eq:sphereest}.
\begin{align}
  \tilde{\Delta}_m \ &\geq \ \min_{j,l<j} \left(
    \left( \Bm_{\Bx}^T \Bm_{\Bx} \ + \  \NRM{\tilde{\Bxi} \ - \ \Bm_{\Bx} }
    \   M \right) \ - \
    \left( \Bm_{\Bx}^T \Bx_j \ - \  \NRM{\tilde{\Bxi} \ - \ \Bm_{\Bx}}
    \   M \right) \right) \\ \nonumber
  &= \  2 \  \NRM{\tilde{\Bxi} \ - \ \Bm_{\Bx}}  \   M
  \ + \ \min_{j,l<j} \left(
    \Bm_{\Bx}^T \Bm_{\Bx} \ - \ \Bm_{\Bx}^T \Bx_j \right) \ = \
  \Delta_m \ + \ 2 \  \NRM{\tilde{\Bxi} \ - \ \Bm_{\Bx}}  \ M \\ \nonumber
  &\geq \ \Delta_m \ + \ 2 \  \NRM{\Bxi \ - \ \Bm_{\Bx}}  \ M\ .
\end{align}
where we used $\NRM{\tilde{\Bxi}  -  \Bm_{\Bx}} = \lambda \NRM{\Bxi - \Bm_{\Bx}}
\leq \NRM{\Bxi - \Bm_{\Bx}}$.
We obtain the lower bound on $\tilde{\epsilon}$:
\begin{align}
  \tilde{\epsilon} \ &\geq \ (N-l) \ \exp \left(- \ \beta \ \left( \Delta_m \ + \ \frac{2 \ M}{\beta \ m_{\max}} \right) \right)  \ ,
\end{align}
where we used that in the sphere $\rS_i$ holds:
\begin{align}
 \NRM{\Bxi \ - \ \Bm_{\Bx}} \ \leq \ 
 \frac{1}{\beta \ m_{\max} } \ ,
\end{align}
therefore
\begin{align}
 2 \ \NRM{\Bxi \ - \ \Bm_{\Bx}} \ M \ \leq \ \frac{2 \ M}{\beta \ m_{\max}}  \ .
 \end{align}

From Lemma~\ref{th:jacobiBoundM} we have
\begin{align}
\label{eq:boundJJ1M}
   \NRM{\rJ(\tilde{\Bxi})}_2 \ &\leq \ \beta \left(
  \ m_{\max}^2 \ + \ 
   \tilde{\epsilon} \ 2 \ (2 \ - \ \tilde{\epsilon} ) \ M^2 \right) \\ \nonumber
  &= \ \beta \left( m_{\max}^2 \ + \ 
   \tilde{\epsilon} 4 \ M^2 \ - \ 2 \ \tilde{\epsilon}^2 \ M^2 \right) \\ \nonumber
  &\leq \ \beta \left( m_{\max}^2 \ + \ 
    (N-l) \ \exp \left(- \ \beta \ \left( \Delta_m \ - \ 
    \frac{2 \ M}{\beta \ m_{\max}} \right) \right)  
    4 \ M^2  \ - \right. \\ \nonumber
    &\left. 2 \ (N-l)^2 \ \exp \left(- \ 2 \ \beta \ \left( \Delta_m \ + \ \frac{2 \ M}{\beta \ m_{\max}} \right) \right)  \ M^2 \right)  \ .
\end{align}

The bound Eq.~\eqref{eq:boundJJ1M} holds for the mean $\rJ^m$, too,
since it averages over $\rJ(\tilde{\Bxi})$:
\begin{align}
  \label{eq:boundJJM}
  \NRM{\rJ^m}_2 \ &\leq \ 
  \beta \left( m_{\max}^2 \ + \ 
    (N-l) \ \exp \left(- \ \beta \ \left( \Delta_m \ - \ 
    \frac{2 \ M}{\beta \ m_{\max}} \right) \right)  
    4 \ M^2  \ - \right. \\ \nonumber
    &\left. 2 \ (N-l)^2 \ \exp \left(- \ 2 \ \beta \ \left( \Delta_m \ + \ \frac{2 \ M}{\beta \ m_{\max}} \right) \right)  \ M^2 \right)   \ .
\end{align}

The assumption of the lemma is
\begin{align}
 \Delta_m \ &\geq \  \frac{2 \ M}{\beta \ m_{\max}} \ - \  
 \frac{1}{\beta} \ \ln \left(  \frac{ 1 \ - \  \beta \ m_{\max}^2    }{
    2 \ \beta \ (N-l) \ M \ \max \{ m_{\max} \ , \ 2 \ M \} } \right) 
    \ ,
\end{align}
Therefore, we have
\begin{align}
 \Delta_m \ - \ \frac{2 \ M}{\beta \ m_{\max}} \ &\geq \  - \  
 \frac{1}{\beta} \ \ln \left(  \frac{ 1 \ - \  \beta \ m_{\max}^2    }{
    2 \ \beta \ (N-l) \ M \ \max \{ m_{\max} \ , \ 2 \ M \} } \right) 
    \ .
\end{align}

Therefore, the spectral norm  $\NRM{\rJ^m}_2$ can be bounded by:    
\begin{align}
  &\NRM{\rJ^m}_2 \ \leq \\ \nonumber 
  &\beta \left( m_{\max}^2 \ + \ 
    (N-l) \ \exp \left(- \ \beta \ \left( - \  
 \frac{1}{\beta} \ \ln \left(  \frac{ 1 \ - \  \beta \ m_{\max}^2    }{
    2 \ \beta \ (N-l) \ M \ \max \{ m_{\max} \ , \ 2 \ M \} } \right)  \right) \right) \right. \\ \nonumber 
    &\left.4 \ M^2  \ - \
     2 \ (N-l)^2 \ \exp \left(- \ 2 \ \beta \ \left( \Delta_m \ + \ \frac{2 \ M}{\beta \ m_{\max}} \right) \right)  \ M^2 \right)   \\ \nonumber
   &= \ \beta \left( m_{\max}^2 \ + \ 
    (N-l) \ \exp \left( 
  \ln \left(  \frac{ 1 \ - \  \beta \ m_{\max}^2    }{
    2 \ \beta \ (N-l) \ M \ \max \{ m_{\max} \ , \ 2 \ M \} } \right)  \right) \right. \\ \nonumber
    &\left. 4 \ M^2  \ - \
     2 \ (N-l)^2 \ \exp \left(- \ 2 \ \beta \ \left( \Delta_m \ + \ \frac{2 \ M}{\beta \ m_{\max}} \right) \right)  \ M^2 \right)  \\ \nonumber
   &= \ \beta \left( m_{\max}^2 \ + \ 
   (N-l) \ \frac{ 1 \ - \  \beta \ m_{\max}^2    }{
    2 \ \beta \ (N-l) \ M \ \max \{ m_{\max} \ , \ 2 \ M \} } 
    \  4 \ M^2 \ - \right. \\ \nonumber
   &\left. 2 \ (N-l)^2 \ \exp \left(- \ 2 \ \beta \ \left( \Delta_m \ + \ \frac{2 \ M}{\beta \ m_{\max}} \right) \right)  \ M^2\right) \\ \nonumber
   &= \  \beta  m_{\max}^2 \ + \ 
    \frac{ 1 \ - \  \beta \ m_{\max}^2    }{
      \ \max \{ m_{\max} \ , \ 2 \ M \} } 
    \  2 \ M \ -  \\ \nonumber
   &\beta \ 2 \ (N-l)^2 \ \exp \left(- \ 2 \ \beta \ \left( \Delta_m \ + \ \frac{2 \ M}{\beta \ m_{\max}} \right) \right)  \ M^2 \\ \nonumber
   &\leq \  \beta  m_{\max}^2 \ + \ 
    1 \ - \ \beta \ m_{\max}^2  \ - \ \beta \ 2 \ (N-l)^2 \ \exp \left(- \ 2 \ \beta \ \left( \Delta_m \ + \ \frac{2 \ M}{\beta \ m_{\max}} \right) \right)  \ M^2\\ \nonumber
    &= \ 1 \ - \ \beta \ 2 \ (N-l)^2 \ \exp \left(- \ 2 \ \beta \ \left( \Delta_m \ + \ \frac{2 \ M}{\beta \ m_{\max}} \right) \right)  \ M^2 \ < 1 \ .
\end{align}
For the last but one inequality we used
$2 M \leq \max \{ m_{\max}  ,  2  M \}$.

Therefore, $f$ is a contraction mapping in $\rS_m$.
\end{proof}

{\textbullet \em Banach Fixed Point Theorem.}
Now we have all ingredients to apply Banach fixed point theorem.
\begin{lemmaA}
\label{th:banachM}
Assume that
\begin{enumerate}[label=A\arabic*:]
\item
\begin{align}
 \Delta_m \ &\geq \  \frac{2 \ M}{\beta \ m_{\max}} \ - \  
 \frac{1}{\beta} \ \ln \left(  \frac{ 1 \ - \  \beta \ m_{\max}^2    }{
    2 \ \beta \ (N-l) \ M \ \max \{ m_{\max} \ , \ 2 \ M \} } \right)  \ ,
\end{align}
and 
\item
\begin{align}
  \beta \ m_{\max}^2 \ &\leq \ 1 \ ,
\end{align}
\end{enumerate}    
then $f$ has a fixed point in $\rS_m$.
\end{lemmaA}

\begin{proof}
We use Banach fixed point theorem: 
Lemma~\ref{th:stayInsideM} says that $f$ maps from the compact set $\rS_m$ into 
the same compact set $\rS_m$.
Lemma~\ref{th:contractionM} says that $f$ is a contraction mapping in
$\rS_m$.
\end{proof}

{\textbullet \em Contraction mapping with a fixed point.}

We assume that the first $l$ patterns 
are much more probable (and similar
to one another) than the other patterns.
Therefore, we define:
\begin{align}
 \label{eq:defA1M}
 M \ &:= \ \max_i \NRM{\Bx_i} \ , \\
  \gamma \ &= \ \sum_{i=l+1}^N p_i \ \leq \ \epsilon  \ , \\
  1- \gamma \ &= \ \sum_{i=1}^l p_i \ \geq \ 1 \ - \ \epsilon \ , \\
  \tilde{p}_i \ &:= \ \frac{p_i}{1- \gamma } \ \leq \ p_i/(1-\epsilon) \ , \\
  \sum_{i=1}^l \tilde{p}_i \ &= \ 1  \ , \\
  \Bm_{\Bx} \ &= \ \frac{1}{l} \ \sum_{i=1}^l \ \Bx_i \ , \\
  m_{\max} \ &= \  \max_{1\leq i \leq l} \NRM{\Bx_i \ - \ \Bm_{\Bx}}  \ .
\end{align}
$M$ is an upper bound on the Euclidean norm of the patterns, which are vectors.
$\epsilon$ is an upper bound on the probability $\gamma$ of not choosing one of the first $l$ patterns, while 
$1-\epsilon$ is a lower bound the probability $(1-\gamma)$ 
of choosing one of the first $l$ patterns. 
$\Bm_{\Bx}$ is the arithmetic mean (the center) of the first $l$ 
patterns. 
$m_{\max}$ is the maximal distance of the patterns to the center $\Bm_{\Bx}$ .
$\tilde{\Bp}$ is the probability $\Bp$
normalized for the first $l$ patterns.

The variance of the first $l$ patterns is
\begin{align}
\label{eq:defA2M}
   \VAR_{\tilde{p}}[\Bx_{1:l}] \ &= \  
  \sum_{i=1}^l \tilde{p}_i \ \Bx_i \ \Bx_i^T
  \ - \   \left(\sum_{i=1}^l \tilde{p}_i \ \Bx_i\right) \ 
  \left(\sum_{i=1}^l \tilde{p}_i \ \Bx_i\right)^T \\ \nonumber
  &= \ 
    \sum_{i=1}^l \tilde{p}_i \ 
    \left(\Bx_i \ - \ \sum_{i=1}^l \tilde{p}_i \Bx_i \right) \ 
     \left(\Bx_i \ - \ \sum_{i=1}^l \tilde{p}_i \Bx_i \right)^T   \ .
\end{align}

We have shown that a fixed point exists. We want to know
how fast the iteration converges to the fixed point.
Let $\Bm_{\Bx}^*$ be the fixed point of the iteration $f$ in the sphere $\rS_m$.
Using the mean value theorem Lemma~\ref{th:MVT}, we have with
$\rJ^m = \int_{0}^1 \rJ(\lambda \Bxi + (1-\lambda)\Bm_{\Bx}^*) \ \Rd \lambda$:
\begin{align}
  \NRM{f(\Bxi) \ - \ \Bm_{\Bx}^*} \ &= \ \NRM{f(\Bxi) \ - \ f(\Bm_{\Bx}^*)} 
  \ \leq \  \NRM{\rJ^m}_2 \ \NRM{\Bxi \ - \ \Bm_{\Bx}^*}
\end{align}

According to Lemma~\ref{th:jacobiBoundM}
the following bounds on the norm $\NRM{\rJ}_2$ of the
Jacobian of the fixed point iteration hold.
The $\gamma$-bound for $\NRM{\rJ}_2$ is
\begin{align}
   \NRM{\rJ}_2 \ &\leq \ \beta \left(
 (1 - \gamma) \ m_{\max}^2 \ + \ 
   \gamma \ 2 \ (2 \ - \ \gamma ) \ M^2 \right) \ ,
\end{align}
while the $\epsilon$-bound for $\NRM{\rJ}_2$ is:
\begin{align}
   \NRM{\rJ}_2 \ &\leq \ \beta \left(
  \ m_{\max}^2 \ + \ 
   \epsilon \ 2 \ (2 \ - \ \epsilon ) \ M^2 \right) \ .
\end{align}

From the last condition we require
for a contraction mapping:
\begin{align}
 \beta \ \ m_{\max}^2 \ &< \ 1 \ .
\end{align}

We want to see how large $\epsilon$ is.
The separation of center 
$\Bm_{\Bx}$ from data $\BX=(\Bx_{l+1},\ldots,\Bx_N)$
is 
\begin{align}
 \Delta_m \ &= \ \min_{j, l<j} \left( \Bm_{\Bx}^T \Bm_{\Bx} \ - \ \Bm_{\Bx}^T
    \Bx_j \right) \ = \ \Bm_{\Bx}^T \Bm_{\Bx} \ - \ \max_{j, l<j} \Bm_{\Bx}^T
  \Bx_j \ .
\end{align}
We need the separation $\tilde{\Delta}_m$ of
$\tilde{\Bx}=\lambda \Bxi + (1-\lambda)\Bm_{\Bx}^*$ from the data.
\begin{align}
  \tilde{\Delta}_m \ &= \ \min_{j, l<j} \left( \tilde{\Bx}^T \Bm_{\Bx} \ - \
\tilde{\Bx}^T \Bx_j \right) \ .
\end{align}
We compute a lower bound on $\tilde{\Delta}_m$.
Using the Cauchy-Schwarz inequality, we obtain for $1 \leq j \leq N$:
\begin{align}
  \left| \tilde{\Bx}^T \Bx_j \ - \ \Bm_{\Bx}^T \Bx_j \right|  
 &\leq \ \NRM{\tilde{\Bx} \ - \ \Bm_{\Bx}} \  \NRM{\Bx_j} \ \leq
 \  \NRM{\tilde{\Bx} \ - \ \Bm_{\Bx}} \   M \ .
\end{align}
We have the lower bound
\begin{align}
  \tilde{\Delta}_m \ &\geq \ \min_{j, l<j} \left(
    \left( \Bm_{\Bx}^T \Bm_{\Bx} \ - \  \NRM{\tilde{\Bx} \ - \ \Bm_{\Bx} }
    \   M \right) \ - \
    \left( \Bm_{\Bx}^T \Bx_j \ + \  \NRM{\tilde{\Bx} \ - \ \Bm_{\Bx}}
    \   M \right) \right) \\ \nonumber
  &= \ - \ 2 \  \NRM{\tilde{\Bx} \ - \ \Bm_{\Bx}}  \   M
  \ + \ \min_{j, l<j} \left(
    \Bm_{\Bx}^T \Bm_{\Bx} \ - \ \Bm_{\Bx}^T \Bx_j \right) \ = \
  \Delta_m \ - \ 2 \  \NRM{\tilde{\Bx} \ - \ \Bm_{\Bx}}  \
  M \ .
\end{align}
Since
\begin{align}
  \NRM{\tilde{\Bx} \ - \ \Bm_{\Bx}} \ &= \ 
  \NRM{\lambda \Bxi + (1-\lambda)\Bm_{\Bx}^* \ - \ \Bm_{\Bx}} \\ \nonumber
  &\leq \ \lambda \ \NRM{\Bxi  \ - \ \Bm_{\Bx}} \ + \ 
  (1-\lambda) \ \NRM{\Bm_{\Bx}^* \ - \ \Bm_{\Bx}} \\ \nonumber
  &\leq \ \max \{ \NRM{\Bxi  \ - \ \Bm_{\Bx}} , \NRM{\Bm_{\Bx}^* \ - \ \Bm_{\Bx}} \} \ ,
\end{align}
we have
\begin{align}
  \tilde{\Delta}_m \ &\geq \ 
  \Delta_m \ - \ 2 \  \max \{ \NRM{\Bxi  \ - \ \Bm_{\Bx}} , \NRM{\Bm_{\Bx}^* \ - \ \Bm_{\Bx}} \}  \
  M \ .
\end{align}

\begin{align}
 \epsilon \ &= \ (N-l) \exp(- \ \beta \ (\Delta_m \ - \ 2 \  \max \{ \NRM{\Bxi  \ - \ \Bm_{\Bx}} , \NRM{\Bm_{\Bx}^* \ - \ \Bm_{\Bx}} \}  \ M) )  \ .
\end{align}

\subsubsection{Properties of Fixed Points Near Stored Pattern}
\label{sec:Properties}

In Subsection~\ref{sec:fixedPoints} 
many stable states that are fixed points near the
stored patterns are considered.
We now consider this case. 
In the fist subsection
we investigate the storage capacity if all patterns are sufficiently 
separated so that metastable states do not appear.
In the next subsection
we look into the updates required and error when
retrieving the stored patterns.
For metastable states we can do the same analyses if each
metastable state is treated as one state like one pattern. 

We see a trade-off that is known from classical Hopfield networks
and for modern Hopfield networks.
Small separation $\Delta_i$ of the pattern $\Bx_i$ from the other 
patterns gives high storage capacity. However the convergence 
speed is lower and the retrieval error higher.
In contrast, large separation $\Delta_i$ of the pattern $\Bx_i$ from the other 
pattern allows the retrieval of patterns with one update step
and exponentially low error.

\paragraph{Exponentially Many Patterns can be Stored.}
\label{sec:ExpStorage}

From Subsection~\ref{sec:fixedPoints} need some definitions.
We assume to have $N$ patterns,
the separation of pattern $\Bx_i$ from the other patterns $\BRA{\Bx_1,\ldots,
\Bx_{i-1}, \Bx_{i+1},\ldots,\Bx_N}$
is $\Delta_i$, defined as 
\begin{align}
 \Delta_i \ &= \ \min_{j,j \not= i} \left( \Bx_i^T \Bx_i \ - \ \Bx_i^T
    \Bx_j \right) \ = \ \Bx_i^T \Bx_i \ - \ \max_{j,j \not= i} \Bx_i^T
  \Bx_j \ .
\end{align}
The pattern is separated from the other data if $0<\Delta_i$.
The separation $\Delta_i$ can also be expressed as
\begin{align}
  \Delta_i \ &= \ \min_{j,j \not= i} \frac{1}{2} \ \left(
  \NRM{\Bx_i}^2 \ - \  \NRM{\Bx_j}^2 \ + \  \NRM{\Bx_i \ - \ \Bx_j}^2
  \right) \\ \nonumber
  &= \ \frac{1}{2} \NRM{\Bx_i}^2  \ - \
  \frac{1}{2} \ \max_{j,j \not= i}  \left(  \NRM{\Bx_j}^2 \ - \
    \NRM{\Bx_i \ - \ \Bx_j}^2 \right) \ .
\end{align}
For $\NRM{\Bx_i} =  \NRM{\Bx_j}$ we have $\Delta_i= 1/2 \min_{j,j \not= i}  \NRM{\Bx_i \ - \ \Bx_j}^2$.
The sphere $\rS_i$ with center $\Bx_i$ is defined as
\begin{align}
 \rS_i \ &= \ \left\{ \Bxi \mid \NRM{\Bxi \ - \ \Bx_i} \ 
 \leq \  \frac{1}{\beta \ N \ M} \right\} \  .
\end{align}
The maximal length of a pattern is $M= \max_{i} \NRM{\Bx_i}$.

We next define what we mean with storing and retrieving a pattern.
\begin{definition}[Pattern Stored and Retrieved]
\label{def:stored}
We assume that around every pattern $\Bx_i$ a sphere $\rS_i$ is given.
We say $\Bx_i$ {\em is stored} if there is a single fixed point $\Bx_i^* \in \rS_i$ to
which all points $\Bxi \in \rS_i$ converge,
and  $\rS_i \cap \rS_j = \emptyset$ for $i \not= j$.
We say $\Bx_i$ {\em is retrieved} for a given $\epsilon$ if 
iteration (update rule) Eq.~\eqref{eq:updateN} gives
a point $\tilde{\Bx}_i$ that is at least 
$\epsilon$-close to the single fixed point $\Bx_i^* \in \rS_i$. 
The retrieval error is $\NRM{\tilde{\Bx}_i - \Bx_i}$.
\end{definition}
The sphere $\rS_i$ around pattern $\Bx_i$ can be any a sphere
and do not have the specific sphere defined in Def.~\ref{th:DefSphere}.

For a query $\Bxi \in \rS_i$ to converge to a fixed point
$\Bx_i^* \in \rS_i$ we required for the application 
of Banach fixed point theorem
and for ensuring a contraction mapping
the following inequality:
\begin{align}
 \Delta_i \ &\geq \  \frac{2}{\beta \ N} \ + \  
 \frac{1}{\beta} \ \ln \left( 2 \ (N-1) \ N \ \beta \ M^2 \right) \ .
\end{align}
This is the assumption in Lemma~\ref{th:banach}
to ensure a fixed point in sphere $\rS_i$.
Since replacing $(N-1)  N $ by $N^2$ gives
\begin{align}
  &\frac{2}{\beta \ N} \ + \  \frac{1}{\beta} \ \ln \left( 2 \  N^2 \ \beta \
        M^2 \right) 
  \ > \  \frac{2}{\beta \ N} \ + \  \frac{1}{\beta} \ \ln \left( 2 \ (N-1) \ N \ \beta \
        M^2 \right) \ ,
\end{align}
the inequality follows from following master inequality 
\begin{align}
\label{eq:masterAi}
 \Delta_i \ &\geq \  \frac{2}{\beta \ N} \ + \  
 \frac{1}{\beta} \ \ln \left( 2 \ N^2 \ \beta \ M^2 \right) \ ,
\end{align}

If we assume that $\rS_i \cap \rS_j \neq \emptyset$ with $i \neq j$, then
the triangle inequality with a point from the intersection gives
\begin{align}
 \NRM{\Bx_i \ - \ \Bx_j} \ &\leq  \ \frac{2}{\beta \ N \ M}   \ .
\end{align}
Therefore, we have using the Cauchy-Schwarz inequality:
\begin{align}
 \Delta_i \ &\leq  \  \Bx_i^T \left( \Bx_i \ - \ \Bx_j \right) 
 \ \leq \ \NRM{\Bx_i} \ \NRM{\Bx_i \ - \ \Bx_j } \ \leq M \ \frac{2}{\beta \ N \ M} 
 \ = \ \frac{2}{\beta \ N } \ .
\end{align}
The last inequality is a contraction to Eq.~\eqref{eq:masterAi}
if we assume that
\begin{align}
\label{eq:assumeA}
  1 \ &< \  2 \ (N-1) \ N \ \beta \ M^2  \ .
\end{align}
With this assumption, the spheres $\rS_i$ and $\rS_j$ do
not intersect. Therefore, each $\Bx_i$ has its separate fixed point in
$\rS_i$.
We define 
\begin{align}
  \Delta_{\min} \ &= \ \min_{1 \leq i \leq N} \Delta_i  
\end{align}
to obtain the master inequality
\begin{align}
\label{eq:masterA}
 \Delta_{\min}  \ &\geq \  \frac{2}{\beta \ N} \ + \  
 \frac{1}{\beta} \ \ln \left( 2 \ N^2 \ \beta \ M^2 \right) \ .
\end{align}

{\textbullet \em Patterns on a sphere.}

For simplicity and in accordance with the results of the classical Hopfield
network, we assume all {\em patterns being on a sphere} with radius $M$:
\begin{align}
 \forall_i: \ \NRM{\Bx_i} \ &= \ M \ .
\end{align}
Under assumption Eq.~\eqref{eq:assumeA} we have only
to show that the master inequality Eq.~\eqref{eq:masterA} is fulfilled 
for each $\Bx_i$ to have a separate fixed point near each $\Bx_i$.

We defined $\alpha_{ij}$ as the angle between $\Bx_i$ and $\Bx_j$.
The minimal angle $\alpha_{\min}$ between two
data points is
\begin{align}
  \alpha_{\min} \ &= \ \min_{1 \leq i < j \leq N} \alpha_{ij} \ .
\end{align}
On the sphere with radius $M$ we have
\begin{align}
 \Delta_{\min} \ &= \ \min_{1 \leq i < j \leq N}  M^2  ( 1 \ - \ \cos(\alpha_{ij}) )
  \ = \  M^2  ( 1 \ - \ \cos(\alpha_{\min}) ) \ ,
\end{align}
therefore it is sufficient to show the master inequality on the sphere:
\begin{align}
 \label{eq:masterAsphere}
 M^2  ( 1 \ - \ \cos(\alpha_{\min}) )  \ &\geq \  \frac{2}{\beta \ N} \ + \  
 \frac{1}{\beta} \ \ln \left( 2 \ N^2 \ \beta \ M^2 \right) \ .
\end{align}

Under assumption Eq.~\eqref{eq:assumeA} we have only
to show that the master inequality Eq.~\eqref{eq:masterA} is fulfilled 
for $\Delta_{\min}$. We consider patterns on the sphere, therefore the master
inequality Eq.~\eqref{eq:masterA} becomes Eq.~\eqref{eq:masterAsphere}.
First we show results when pattern positions 
on the sphere are constructed and $\Delta_{\min}$ is ensured. 
Then we move on to 
random patterns on a sphere, where $\Delta_{\min}$ becomes a random variable.

{\textbullet \em Storage capacity for patterns placed on the sphere.}

Next theorem says how many patterns we can stored
(fixed point with attraction basin near pattern)
if we are allowed to place them on the sphere.
\begin{theoremA}[Storage Capacity (M=2): Placed Patterns]
We assume $\beta=1$ and patterns on the sphere with radius $M$.
If $M= 2 \sqrt{d-1}$ and the dimension $d$ 
of the space is $d\geq 4$ or if 
$M= 1.7 \sqrt{d-1}$ and the dimension $d$ 
of the space is $d\geq 50$,
then
the number of patterns $N$ that can be stored 
(fixed point with attraction basin near pattern) is
at least
\begin{align}
N \ &= \ 2^{2(d-1)} \ .
\end{align}

\end{theoremA}

\begin{proof}
For random patterns on the sphere,
we have to show that the master inequality Eq.~\eqref{eq:masterAsphere}
holds:
\begin{align}
 M^2  ( 1 \ - \ \cos(\alpha_{\min}) )  \ &\geq \  \frac{2}{\beta \ N} \ + \  
 \frac{1}{\beta} \ \ln \left( 2 \ N^2 \ \beta \ M^2 \right) \ .
\end{align}

We now place the patterns equidistant on the sphere where the pattern
are separated by an angle $\alpha_{\min}$:
\begin{align}
 \forall_i: \ \min_{j,j \not= i} \alpha_{ij} \ = \ \alpha_{\min} \ ,
\end{align}
In a $d$-dimensional space we can place
\begin{align}
 N \ &= \ \left( \frac{2 \pi}{\alpha_{\min}} \right)^{d-1}
\end{align}
points on the sphere. 
In a spherical coordinate system a pattern differs from its most
closest patterns by an angle $\alpha_{\min}$ and there are $d-1$ angles.
Solving for $\alpha_{\min}$ gives
\begin{align}
 \alpha_{\min} \ &= \ \frac{2 \pi}{N^{1/(d-1)}} \ .
\end{align}
The number of patterns that can be stored is determined
by the largest $N$ that fulfils
\begin{align}
\label{eq:mainN}
  & M^2  \left( 1 \ - \ \cos\left(  \frac{2 \pi}{N^{1/(d-1)}} \right) \right) 
  \ \geq \  \frac{2}{\beta \ N} \ + \  
  \frac{1}{\beta} \ \ln \left( 2 \  N^2 \ \beta \ M^2 \right) \ .
\end{align}

We set $N=2^{2(d-1)}$ and obtain for Eq.~\eqref{eq:mainN}:
\begin{align}
  & M^2  \left( 1 \ - \ \cos\left(  \frac{\pi}{2} \right) \right) 
  \ \geq \  \frac{2}{\beta \ 2^{3(d-1)}} \ + \  \frac{1}{\beta} \ \ln \left( 2 \ \beta \
        M^2 \right) \ + \  \frac{1}{\beta} \ 4 \ (d-1) \ln 2  \ .
\end{align}
This inequality is equivalent to
\begin{align}
  & \beta \ M^2  
  \ \geq \  \frac{1}{2^{2(d-1)-1}} \ + \   \ln \left( 2 \ \beta \
        M^2 \right) \ + \ 4 \ (d-1) \ln 2  \ .
\end{align}
The last inequality can be fulfilled with $M= K \sqrt{d-1}$ and proper $K$.
For $\beta=1$, $d=4$ and $K=2$ the inequality is fulfilled.
The left hand side minus the right hand side is
$4 (d-1) - 1/2^{2(d-1)-1}  -    \ln ( 8 (d-1)  ) -  4  (d-1) \ln 2 $.
Its derivative with respect to $d$ is strict positive. 
Therefore, the inequality holds for 
$d \geq 4$.

For $\beta=1$, $d=50$ and $K=1.7$ the inequality is fulfilled.
The left hand side minus the right hand side is
$2.89 (d-1) - 1/2^{2(d-1)-1}  -    \ln (5.78 (d-1)  ) -  4  (d-1) \ln 2 $.
Its derivative with respect to $d$ is strict positive. 
Therefore, the inequality holds for 
$d \geq 50$.

\end{proof}

If we want to store considerably more patterns, then
we have to increase the length of the vectors or the dimension
of the space where the vectors live.
The next theorem shows results for the number of patterns $N$ with $N =  2^{3(d-1)}$.
\begin{theoremA}[Storage Capacity (M=5): Placed Patterns]
We assume $\beta=1$ and patterns on the sphere with radius $M$.
If $M= 5 \sqrt{d-1}$ and the dimension $d$ 
of the space is $d\geq 3$ or if 
$M= 4 \sqrt{d-1}$ and the dimension $d$ 
of the space is $d\geq 13$,
then
the number of patterns $N$ that can be stored 
(fixed point with attraction basin near pattern) is
at least
\begin{align}
N \ &= \ 2^{3(d-1)} \ .
\end{align}

\end{theoremA}

\begin{proof}

We set $N=2^{3(d-1)}$ and obtain for Eq.~\eqref{eq:mainN}:
\begin{align}
& M^2  \left( 1 \ - \ \cos\left(  \frac{\pi}{4} \right) \right) 
\ \geq \  \frac{2}{\beta \ 2^{3(d-1)}} \ + \  \frac{1}{\beta} \ \ln \left( 2 \ \beta \
        M^2 \right) \ + \  \frac{1}{\beta} \ 6 \ (d-1) \ln 2  \ .
\end{align}
This inequality is equivalent to
\begin{align}
& \beta \ M^2   \left( 1 \ - \ \frac{\sqrt{2}}{2} \right)
\ \geq \  \frac{1}{2^{3(d-1)-1}} \ + \   \ln \left( 2 \ \beta \
        M^2 \right) \ + \ 6 \ (d-1) \ln 2  \ .
\end{align}
The last inequality can be fulfilled with $M= K \sqrt{d-1}$ and proper $K$.
For $\beta=1$, $d=13$ and $K=4$ the inequality is fulfilled.
The left hand side minus the right hand side is
$4.686292 (d-1) - 1/2^{3(d-1)-1}  -    \ln (32 (d-1)  ) -  6  (d-1) \ln 2 $.
Its derivative with respect to $d$ is strict positive. 
Therefore, the inequality holds for 
$d \geq 13$.

For $\beta=1$, $d=3$ and $K=5$ the inequality is fulfilled.
The left hand side minus the right hand side is
$7.32233 (d-1) - 1/2^{3(d-1)-1}  -    \ln (50 (d-1)  ) -  6  (d-1) \ln 2 $.
Its derivative with respect to $d$ is strict positive. 
Therefore, the inequality holds for 
$d \geq 3$.

\end{proof}

\vspace{1cm}

{\textbullet \em Storage capacity for random patterns on the sphere.}

Next we investigate random points on the sphere.
Under assumption Eq.~\eqref{eq:assumeA} we have
to show that the master inequality Eq.~\eqref{eq:masterAsphere} is fulfilled 
for $\alpha_{\min}$, where now $\alpha_{\min}$ is now a random variable.
We use results on the
distribution of the minimal angles between random patterns on a sphere
according to \citet{Cai:13} and \citet{Brauchart:18}.
Theorem~2 in \citet{Cai:13} gives the distribution of the
minimal angle for random patterns on the unit sphere. 
Proposition~3.5 in \citet{Brauchart:18} gives a lower bound
on the probability of the minimal angle being larger than a given constant.
We require this proposition to derive the probability of pattern having a minimal 
angle $\alpha_{\min}$.
Proposition~3.6 in \citet{Brauchart:18} gives the expectation
of the minimal angle.

We will prove high probability bounds 
for the expected storage capacity. 
We need the following tail-bound on $\alpha_{\min}$ (the minimal angle of
random patterns on a sphere):
\begin{lemmaA}[\citep{Brauchart:18}]
Let $d$ be the dimension of the pattern space,
\begin{align} \label{eq:kappa}
 \kappa_{d} \ &:= \ \frac{1}{d \ \sqrt{\pi}} \ \frac{\Gamma((d+1)/2)}{\Gamma(d/2)} \ .
\end{align}
and $\delta>0$ such that $\frac{\kappa_{d-1}}{2}\delta^{(d-1)} \leq 1$.
Then
\begin{align} \label{eq:alpha_min_tail}
  \PR(N^{\frac{2}{d-1}} \alpha_{\min} \ \geq \ \delta) \ &\geq \ 
  1 \ - \ \frac{\kappa_{d-1}}{2} \ \delta^{d-1} \ .
\end{align}
\end{lemmaA}

\begin{proof}
 The statement of the lemma is Eq.~(3-6) from Proposition 3.5 in \citet{Brauchart:18}.
\end{proof}
Next we derive upper and lower bounds on the constant $\kappa_d$ since we require them
later for proving storage capacity bounds.

\begin{lemmaA} 
For $\kappa_{d}$ defined in Eq.~\eqref{eq:kappa} 
we have the following bounds for every $d \geq 1$:
\begin{align} \label{eq:kappabounds}
   \frac{1}{\exp(1/6) \ \sqrt{e \ \pi \ d}  } \ &\leq \ \kappa_d \ \leq \ \frac{\exp(1/12)}{ \sqrt{2 \ \pi \ d }} \ < \ 1 \ .
\end{align}
\end{lemmaA}

\begin{proof}
We use for $x>0$ the following bound related to Stirling's approximation formula 
for the gamma function, c.f.\ \dlmf{5.6.1}:
\begin{align} \label{eq:stirling}
    1 \ &< \  \Gamma(x) \ (2 \ \pi)^{- \ \frac{1}{2}} x^{\frac{1}{2} \ - \ x}
    \exp(x) \ < \ \exp \left(\frac{1}{12 \ x} \right) \ .
\end{align}

Using Stirling's formula Eq.~\eqref{eq:stirling}, we upper bound $\kappa_{d}$:
\begin{align}
  &\kappa_{d} \ = \ \frac{1}{d \ \sqrt{\pi}} \ \frac{\Gamma((d+1)/2)}{\Gamma(d/2)}
  \ < \ \frac{1}{d \ \sqrt{\pi}} \ \frac{\exp\left(\frac{1}{6(d+1)}\right) \ 
  \exp\left(- \ \frac{d+1}{2}\right) \ \left(\frac{d+1}{2}\right)^{\frac{d}{2}}}
  {\exp\left(- \ \frac{d}{2}\right) \ \left(\frac{d}{2}\right)^{\frac{d}{2} \ - \ \frac{1}{2}}}\\ \nonumber
  &= \ \frac{1}{d \ \sqrt{\pi \ e}} \ \exp\left(\frac{1}{6(d+1)}\right) \ 
  \left(1\ + \ \frac{1}{d} \right)^{\frac{d}{2}}\sqrt{\frac{d}{2}}  \ \leq \ 
  \frac{\exp\left(\frac{1}{12}\right)}{ \sqrt{2 \ \pi} \ \sqrt{d}} \ .
\end{align}
For the first inequality, we applied Eq.~\eqref{eq:stirling}, 
while for the second we used $(1+\frac{1}{d})^{d} < e$ 
for $d \geq 1$. 

Next, we lower bound $\kappa_{d}$ by again applying Stirling's formula Eq.~\eqref{eq:stirling}:
\begin{align}
    &\kappa_{d} \ = \ \frac{1}{d \ \sqrt{\pi}} \ \frac{\Gamma((d+1)/2)}{\Gamma(d/2)}
    \ > \ \frac{1}{d \ \sqrt{\pi}} \ \frac{ \exp \left(- \ \frac{d+1}{2} \right) \
    \left(\frac{d+1}{2}\right)^{\frac{d}{2}}}{\exp\left(\frac{1}{6 \ d}\right) \ 
    \exp\left(-\frac{d}{2}\right) \ \left(\frac{d}{2}\right)^{\frac{d}{2} -  \frac{1}{2}}}\\ \nonumber &= \ \frac{1}{d \ \sqrt{\pi \ e} \ \exp\left(\frac{1}{6\ d}\right)} \ \left(1+\frac{1}{d}\right)^{\frac{d}{2}}\sqrt{\frac{d}{2}}
    \ \geq \ \frac{1}{\exp \left( \frac{1}{6} \right) \ 
    \sqrt{e \ \pi \ d}} \ ,
\end{align}
where the last inequality holds because of monotonicity of $(1+\frac{1}{d})^d$ and using the fact that for $d=1$ it takes on the value 2. 
\end{proof}

We require a bound on $\cos$ to bound the master inequality
Eq.~\eqref{eq:masterAsphere}.
\begin{lemmaA}
 For $0 \leq x \leq \pi$ the function $\cos$ can be upper bounded by:
\begin{align} \label{eq:cosest}
    \cos(x) \ &\leq \  1 \ - \ \frac{x^2}{5} \ .
\end{align}
\end{lemmaA}

\begin{proof}
We use the infinite product representation of $\cos$, c.f.\ \dlmf{4.22.2}:

\begin{align} \label{eq:cos_prod}
    \cos(x) \ &= \ \prod_{n=1}^{\infty}
    \left(1-\frac{4 \ x^2}{(2n-1)^2 \ \pi^2}\right) \ .
\end{align}
Since it holds that
\begin{align}
 1 \ - \ \frac{4\ x^2}{(2n-1)^2 \ \pi^2} \ \leq \ 1  
\end{align}
for $|x| \leq \pi$ and $n \geq 2$, we can get the following upper bound on
Eq.~\eqref{eq:cos_prod}:
\begin{align}
   \cos(x) \ &\leq \  \prod_{n=1}^{2}\left(1-\frac{4 \ x^2}{(2n-1)^2 \pi^2}\right) 
   \ = \ \left( 1\ - \ \frac{4 \ x^2}{\pi^2} \right) \ \left( 1\ - \ \frac{4 \ x^2}{9 \ \pi^2} \right)\\ \nonumber 
   &= \ 
   1 \ -\ \frac{40 \ x^2}{9 \ \pi^2} \ + \ \frac{16 \ x^4}{9 \ \pi^4}
   \ \leq \  1 \ - \ \frac{40 \ x^2}{9 \ \pi^2} \ + \ \frac{16 \ x^2}{9 \ \pi^2} 
   \\ \nonumber 
   &= \  1 \ - \ \frac{24 \ x^2}{9 \ \pi^2} \ \leq \ 1 \ - \ \frac{x^2}{5}  \ .
\end{align}
The last but one inequality uses $x\leq \pi$,
which implies $x/\pi \leq 1$.
Thus Eq.~\eqref{eq:cosest} is proven.

\end{proof}

{\textbullet \em Exponential storage capacity: the base $c$ as a function of 
the parameter $\beta$, the radius of the sphere $M$, the probability $p$, 
and the dimension $d$ of the space.}

We express the number $N$ of stored patterns by an exponential 
function with base $c>1$ and an exponent linear in $d$.
We derive constraints on he base $c$ 
as a function of $\beta$, the radius of the sphere $M$, 
the probability $p$ that all patterns can be stored, 
and the dimension $d$ of the space.
With $\beta>0$, $K > 0$, and $d \geq 2$ (to ensure a sphere), 
the following theorem gives our main result.

\begin{theoremA}[Storage Capacity (Main): Random Patterns]
\label{th:mainStorage}
We assume a failure probability $0<p\leq 1$ and randomly chosen patterns 
on the sphere with radius $M:=K \sqrt{d-1}$. 
We define
\begin{align} \nonumber
  a \ &:= \  \frac{2}{d-1} \ (1 \ + \ \ln(2 \ \beta \ K^2 \  p \  (d-1))) \ , \quad
  b \ := \ \frac{2 \ K^2 \ \beta}{5} \ , \\
  c \ &:= \ \frac{b}{W_0(\exp(a \ + \ \ln(b))} \ ,
\end{align}
where $W_0$ is the upper branch of the Lambert $W$ function  \dlmf{4.13}
and ensure
\begin{align} 
    \label{eq:Cbound}
    c \ &\geq \ \left( \frac{2}{ \sqrt{p}}\right)^{\frac{4}{d-1}} \ .
\end{align}
Then with probability $1-p$, the number of random patterns 
that can be stored is  
\begin{align} 
 \label{eq:CapacityMA}
    N \ &\geq \ \sqrt{p} \ c^{\frac{d-1}{4}}  \ .
\end{align}

Therefore it is proven for $c\geq 3.1546$ with
$\beta=1$, $K=3$, $d= 20$ and $p=0.001$ ($a + \ln(b)>1.27$)
and proven for $c\geq 1.3718$ with $\beta = 1$, $K=1$, $d = 75$, and $p=0.001$
($a + \ln(b)<-0.94$).
\end{theoremA}

\begin{proof}
We consider the probability that the master inequality Eq.~\eqref{eq:masterAsphere}
is fulfilled:
\begin{align} 
  & \PR \left( M^2  ( 1 \ - \cos(\alpha_{\min}) ))
  \ \geq \  \frac{2}{\beta \ N} \ + \  
  \frac{1}{\beta} \ \ln \left( 2 \ N^2 \ \beta \ M^2 \right) \right) \ \geq \ 1\ - \ p \ .
\end{align}
Using Eq.~\eqref{eq:cosest}, we have:
\begin{align}
    1 \ -  \ \cos(\alpha_{\min}) \ &\geq \ \frac{1}{5} \ \alpha_{\min}^2 \ .
\end{align}
Therefore, with probability $1-p$ the storage capacity is largest $N$ that fulfills 
\begin{align}
    &\PR \left( M^2  \frac{\alpha_{min}^2}{5}
  \ \geq \  \frac{2}{\beta \ N} \ + \  
  \frac{1}{\beta} \ \ln \left( 2 \ N^2 \ \beta \ M^2 \right) \right) 
  \ \geq \ 1 \ - \ p \ .
\end{align}
This inequality is equivalent to 
\begin{align} \label{eq:tail_lower_1:c}
       & \PR \left( N^{\frac{2}{d-1}} \ \alpha_{min}
  \ \geq \ \frac{\sqrt{5} \ N^{\frac{2}{d-1}}}{M} \ \left( \frac{2}{\beta \ N} \ + \  
  \frac{1}{\beta} \ \ln \left( 2 \ N^2 \ \beta \ M^2 \right) \right)^{\frac{1}{2}} 
  \right) \ \geq \ 1\ - \ p \ .
\end{align}
We use Eq.~\eqref{eq:alpha_min_tail} to obtain:
\begin{align}
    &\PR \left( N^{\frac{2}{d-1}} \ \alpha_{min}
  \ \geq \ \frac{\sqrt{5} \ N^{\frac{2}{d-1}}}{M} \ \left( \frac{2}{\beta \ N} \ + \  
  \frac{1}{\beta} \ \ln \left( 2 \ N^2 \ \beta \ M^2 \right) \right)^{\frac{1}{2}}  \right) \\ \nonumber
  &\geq \ 1 \ - \ \frac{\kappa_{d-1}}{2} \ 5^{\frac{d-1}{2}} \ N^{2} \ M^{- (d-1)} 
  \left( \frac{2}{\beta \ N} \ + \  
  \frac{1}{\beta} \ \ln \left( 2 \ N^2 \ \beta \ M^2 \right) \right)^{\frac{d-1}{2}} \ .
\end{align}
For Eq.~\eqref{eq:tail_lower_1:c} to be fulfilled, it is sufficient that
\begin{align} \label{eq:mainclaim:c}
    &\frac{\kappa_{d-1}}{2} \ 5^{\frac{d-1}{2}} \ N^{2} \ M^{- (d-1)}
    \left( \frac{2}{\beta \ N} \ + \  
  \frac{1}{\beta} \ \ln \left( 2 \ N^2 \ \beta M^2 \right) \right)^{\frac{d-1}{2}} 
  \ - \ p \ \leq \ 0 \ .
\end{align}

If we insert the assumption Eq.~\eqref{eq:Cbound} of the theorem
into Eq.~\eqref{eq:CapacityMA}, then we obtain $N \geq 2$.
We now apply the upper bound $\kappa_{d-1}/2 < \kappa_{d-1} <1 $ from Eq.~\eqref{eq:kappabounds} 
and the upper bound $\frac{2}{\beta  N} \leq \frac{1}{\beta}$ from $N \geq 2$ to inequality Eq.~\eqref{eq:mainclaim:c}.
In the resulting inequality we insert $N=\sqrt{p} c^{\frac{d-1}{4}}$ to check whether it is fulfilled
with this special value of $N$ and obtain:
\begin{align} \label{eq:mainclaim:c1}
    &5^{\frac{d-1}{2}} \ p \ c^{\frac{d-1}{2}} \ M^{- (d-1)}
    \left( \frac{1}{\beta} \ + \  
  \frac{1}{\beta} \ \ln \left( 2 \  p \ c^{\frac{d-1}{2}}  \ \beta M^2 \right) \right)^{\frac{d-1}{2}} 
   \leq \ p \ .
\end{align}
Dividing by $p$, inserting $M= K \sqrt{d-1}$,
and exponentiation of the left and right side by $\frac{2}{d-1}$ 
gives:
 \begin{align} \label{eq:inequ1:c}
 \frac{5 \ c}{K^2 \  (d-1)}\left( \frac{1}{\beta} \ + \  
   \frac{1}{\beta} \ \ln \left( 2 \ \beta \ c^{\frac{d-1}{2}} \ p \ K^2 \ 
   (d-1) \right) \right) \ - \ 1 \ \leq \ 0 \ .
\end{align}
After some algebraic manipulation, this inequality can be written as
 \begin{align} \label{eq:inequ2:c}
   a \ c \ + \ c \ \ln(c) \ - \ b \  \leq \ 0 \ ,
\end{align}
where we used
\begin{align} \nonumber
  a \ &:= \  \frac{2}{d-1} \ (1 \ + \ \ln(2 \ \beta \ K^2 \  p \  (d-1))) \ , \quad
  b \ := \ \frac{2 \ K^2 \ \beta}{5} \ .
\end{align}
We determine the value $\hat{c}$ of $c$ 
which makes the inequality Eq.~\eqref{eq:inequ2:c} equal to zero.
We solve
\begin{align} 
   a \ \hat{c} \ + \ \hat{c} \ \ln(\hat{c}) \ - \ b \  = \ 0 
\end{align}
for $\hat{c}$:
\begin{align} 
  &a \ \hat{c} \ + \ \hat{c} \ \ln(\hat{c}) \ - \ b \ = \ 0 \\ \nonumber
  \Leftrightarrow  \ \ &a \ + \ \ln(\hat{c}) \ = \ b/\hat{c} \\ \nonumber
  \Leftrightarrow  \ \ &a \ + \ \ln(b) \ + \ \ln(\hat{c}/b) \ = \ b/\hat{c} \\ \nonumber
  \Leftrightarrow \ \ &b/\hat{c} \ + \ \ln(b/\hat{c}) \ = \ a \ + \ \ln(b) \\ \nonumber
  \Leftrightarrow \ \ &b/\hat{c} \ \exp(b/\hat{c}) \  = \ \exp(a \ + \ \ln(b)) \\ \nonumber
  \Leftrightarrow \ \ &b/\hat{c} \ = \ W_0(\exp(a \ + \ \ln(b))) \\ \nonumber
  \Leftrightarrow \ \ &\hat{c} \ = \ \frac{b}{W_0(\exp(a \ + \ \ln(b))} \ ,
\end{align}
where $W_0$ is the upper branch of the Lambert $W$ function (see Def.~\ref{th:lambert}).
Hence, the solution is
\begin{align} 
 \label{eq:hatc}
  \hat{c} \ &= \ \frac{b}{W_0(\exp(a \ + \ \ln(b))} \ .
\end{align}
The solution exist, since the Lambert function $W_0(x)$  \dlmf{4.13} is defined 
for $- 1/e <x$ and we have $0 < \exp(a  + \ln(b)$.

Since $\hat{c}$ fulfills inequality Eq.~\eqref{eq:inequ2:c} 
and therefore also Eq.~\eqref{eq:mainclaim:c1},
we have a lower bound on the storage capacity $N$:
\begin{align} 
    N \ &\geq \ \sqrt{p} \ \hat{c}^{\frac{d-1}{4}}  \ .
\end{align}

\end{proof}

Next we aim at a lower bound on $c$
which does not use the Lambert $W$ function \dlmf{4.13}.
Therefore, we upper bound $W_0(\exp(a  +  \ln(b))$ to obtain a lower
bound on $c$, therefore, also a lower bound on the storage
capacity $N$. The lower bound is given in the next corollary.
\begin{corollaryA}
We assume a failure probability $0<p\leq 1$ and randomly chosen patterns 
on the sphere with radius $M=K \sqrt{d-1}$. 
We define
\begin{align} \nonumber
  a \ &:= \  \frac{2}{d-1} \ (1 \ + \ \ln(2 \ \beta \ K^2 \  p \  (d-1))) \ , \quad
  b \ := \ \frac{2 \ K^2 \ \beta}{5} \ .
\end{align}
Using the omega constant $\Omega \approx 0.56714329$ we set
\begin{align} 
    c \ &= \ 
    \begin{cases}
    b \  \ln \left( \frac{\Omega \ \exp(a \ + \ \ln(b)) \ + \ 1 }{\Omega \ (1 \ + \ \Omega)} \right)^{-1} 
        & \text{for } \ a \ + \ \ln(b) \ \leq \ 0 \ , \\
     b  \ (a \ +\ \ln(b))^{-\frac{a\ + \ \ln(b)}{a\ + \ \ln(b)\ + \ 1}}    & \text{for } \ a \ + \ \ln(b) \ > \ 0 
    \end{cases}
\end{align}
and ensure
\begin{align} 
    \label{eq:CboundC}
    c \ &\geq \ \left( \frac{2}{ \sqrt{p}}\right)^{\frac{4}{d-1}} \ .
\end{align}

Then with probability $1-p$, the number of random patterns 
that can be stored
is  
\begin{align} 
 \label{eq:Capacity1MA}
    N \ &\geq \ \sqrt{p} \ c^{\frac{d-1}{4}}  \ .
\end{align}

\vspace{0.5cm}

Examples are $c\geq 3.1444$ for
$\beta=1$, $K=3$, $d= 20$ and $p=0.001$ ($a + \ln(b)>1.27$)
and  $c\geq 1.2585$ for $\beta = 1$ $K=1$, $d = 75$, and $p=0.001$
($a + \ln(b)<-0.94$).

\end{corollaryA}

\begin{proof}

We lower bound the $c$ defined in Theorem~\ref{th:mainStorage}.
According to \citep[Theorem 2.3]{Hoorfar:08} we have for any real $u$ and $y>\frac{1}{e}$:
\begin{align}
    W_0(\exp(u)) \ &\leq \ \ln \left( \frac{\exp(u) \ + \ y}{1 \ + \ \ln(y)} \right) \ .
\end{align}
To upper bound $W_0(x)$ for $x \in [0,1]$, we set
\begin{align}
 y \ &= \ 1/W_0(1)  \ = \  1/\Omega  \ = \ \exp{\Omega} \ = \ - \ 1/ \ln \Omega  \ \approx \ 1.76322\ ,
\end{align}
where the Omega constant $\Omega$ is
\begin{align}
  \Omega \ &= \ \left(\int_{-\infty}^{\infty} \frac{\Rd t}{\left(e^t \ - \ t\right)^2 \ + \ \pi^2}\right)^{-1}
   \  - \ 1 \ \approx \ 0.56714329 \ .
\end{align}
See for these equations the special values of the Lambert $W$ function in Lemma~\ref{th:lambertVal}.
We have the upper bound on $W_0$:
\begin{align}
 \label{eq:w0general2}
    W_0(\exp(u)) \ &\leq \ \ln \left( \frac{\exp(u) \ + \ 1/\Omega}{1 \ + \ \ln(1/\Omega)} \right)
    \ = \  \ln \left( \frac{\Omega \ \exp(u) \ + \ 1}{\Omega  (1 \ + \ \Omega)} \right) \ .
\end{align}
At the right hand side of interval $[0,1]$,
we have $u=0$ and $\exp(u)=1$ and get:
\begin{align}    
   \ln \left( \frac{\Omega \ 1 \ + \ 1}{\Omega  (1 \ + \ \Omega)} \right)
   \ = \   \ln \left( \frac{1}{\Omega} \right) \ = \  - \  \ln \left( \Omega \right) \ = \
   \Omega \ = \ W_0(1) \ .
\end{align}
Therefore, the bound is tight at the right hand side of of interval $[0,1]$,
that is for $\exp(u)=1$, i.e.\ $u=0$.
We have derived an bound for  $W_0(\exp(u))$ with $\exp(u) \in [0,1]$ 
or, equivalently, $u \in [-\infty , 0]$.
We obtain from \citet[Corollary 2.6]{Hoorfar:08} the following bound 
on $W_0(\exp(u))$ for $1<\exp(u)$, or, equivalently $0<u$: 
\begin{align} \label{eq:w0general}
      W_0(\exp(u)) \ &\leq \ u^{\frac{u}{1 \ + \ u}} \ .
\end{align}

A lower bound on $\hat{c}$ is obtained via the upper bounds Eq.~\eqref{eq:w0general}
and Eq.~\eqref{eq:w0general2} on $W_0$ as $W_0 >0$. 
We set $u=a +\ln(b)$ and obtain 
 \begin{align} 
   W_0(\exp(a \ +\ \ln(b))) \ &\leq \ 
    \begin{cases}
     \ln \left( \frac{\Omega \ \exp(a \ + \ \ln(b)) \ + \ 1 }{\Omega \ (1 \ + \ \Omega)} \right)^{-1} 
        & \text{for } \ a \ + \ \ln(b) \ \leq \ 0 \ , \\
     (a \ +\ \ln(b))^{-\frac{a\ + \ \ln(b)}{a\ + \ \ln(b)\ + \ 1}}    & \text{for } \ a \ + \ \ln(b) \ > \ 0 
    \end{cases}
\end{align}
 
We insert this bound into Eq.~\eqref{eq:hatc}, the solution for $\hat{c}$,
to obtain the statement of the theorem.

\end{proof}

{\textbullet \em Exponential storage capacity: the dimension $d$ of the space 
as a function of the parameter $\beta$, the radius of the sphere $M$, 
and the probability $p$.}

We express the number $N$ of stored patterns by an exponential 
function with base $c>1$ and an exponent linear in $d$.
We derive constraints on the dimension $d$ of the space
as a function of $\beta$, the radius of the sphere $M$, 
the probability $p$ that all patterns can be stored, 
and the base of the exponential storage capacity.
The following theorem gives this result.

\begin{theoremA}[Storage Capacity (d computed): Random Patterns]
\label{th:capacityD}
We assume a failure probability $0<p\leq 1$ and randomly chosen patterns 
on the sphere with radius $M=K \sqrt{d-1}$. 
We define
\begin{align} \nonumber
  a \ &:= \  \frac{\ln(c)}{2} \ - \ \frac{K^2 \ \beta}{5 \ c} \ , \quad
  b \ := \ 1 \ + \ \ln \left( 2 \  p \ \beta \ K^2  \right) \ , \\ \label{eq:capacityDdA}
  d  \ &= \
  \begin{cases}
  1 \ + \ \frac{1}{a} \ W(a \ \exp(- b) ) &\text{for } a\not=0 \ , \\
  1 \ + \ \exp(-b) &\text{for } a=0 \ ,
  \end{cases}
\end{align}
where $W$ is the Lambert $W$ function  \dlmf{4.13}.
For $0<a$ the function $W$ is the upper branch $W_0$ and for $a<0$ we use 
the lower branch $W_{-1}$.
If we ensure that
\begin{align} 
    \label{eq:Cbound2}
    c \ &\geq \ \left( \frac{2}{ \sqrt{p}}\right)^{\frac{4}{d-1}} \ , \quad \ 
     - \ \frac{1}{e} \ \leq \ a  \ \exp(- b)  \ ,
\end{align}
then with probability $1-p$, the number of random patterns 
that can be stored is  
\begin{align} 
 \label{eq:CapacityDA}
    N \ &\geq \ \sqrt{p} \ c^{\frac{d-1}{4}}  \ .
\end{align}
\end{theoremA}

\begin{proof}
We consider the probability that the master inequality Eq.~\eqref{eq:masterAsphere}
is fulfilled:
\begin{align} 
  & \PR \left( M^2  ( 1 \ - \cos(\alpha_{\min}) ))
  \ \geq \  \frac{2}{\beta \ N} \ + \  
  \frac{1}{\beta} \ \ln \left( 2 \ N^2 \ \beta \ M^2 \right) \right) \ \geq \ 1\ - \ p \ .
\end{align}
Using Eq.~\eqref{eq:cosest}, we have:
\begin{align}
    1 \ -  \ \cos(\alpha_{\min}) \ &\geq \ \frac{1}{5} \ \alpha_{\min}^2 \ .
\end{align}
Therefore, with probability $1-p$ the storage capacity is largest $N$ that fulfills 
\begin{align}
    &\PR \left( M^2  \frac{\alpha_{min}^2}{5}
  \ \geq \  \frac{2}{\beta \ N} \ + \  
  \frac{1}{\beta} \ \ln \left( 2 \ N^2 \ \beta \ M^2 \right) \right) 
  \ \geq \ 1 \ - \ p \ .
\end{align}
This inequality is equivalent to 
\begin{align} \label{eq:tail_lower_1:ca}
       & \PR \left( N^{\frac{2}{d-1}} \ \alpha_{min}
  \ \geq \ \frac{\sqrt{5} \ N^{\frac{2}{d-1}}}{M} \ \left( \frac{2}{\beta \ N} \ + \  
  \frac{1}{\beta} \ \ln \left( 2 \ N^2 \ \beta \ M^2 \right) \right)^{\frac{1}{2}} 
  \right) \ \geq \ 1\ - \ p \ .
\end{align}
We use Eq.~\eqref{eq:alpha_min_tail} to obtain:
\begin{align}
    &\PR \left( N^{\frac{2}{d-1}} \ \alpha_{min}
  \ \geq \ \frac{\sqrt{5} \ N^{\frac{2}{d-1}}}{M} \ \left( \frac{2}{\beta \ N} \ + \  
  \frac{1}{\beta} \ \ln \left( 2 \ N^2 \ \beta \ M^2 \right) \right)^{\frac{1}{2}}  \right) \\ \nonumber
  &\geq \ 1 \ - \ \frac{\kappa_{d-1}}{2} \ 5^{\frac{d-1}{2}} \ N^{2} \ M^{- (d-1)} 
  \left( \frac{2}{\beta \ N} \ + \  
  \frac{1}{\beta} \ \ln \left( 2 \ N^2 \ \beta \ M^2 \right) \right)^{\frac{d-1}{2}} \ .
\end{align}
For Eq.~\eqref{eq:tail_lower_1:ca} to be fulfilled, it is sufficient that
\begin{align} \label{eq:mainclaim:cd}
    &\frac{\kappa_{d-1}}{2} \ 5^{\frac{d-1}{2}} \ N^{2} \ M^{- (d-1)}
    \left( \frac{2}{\beta \ N} \ + \  
  \frac{1}{\beta} \ \ln \left( 2 \ N^2 \ \beta M^2 \right) \right)^{\frac{d-1}{2}} 
  \ - \ p \ \leq \ 0 \ .
\end{align}

If we insert the assumption Eq.~\eqref{eq:Cbound2} of the theorem
into Eq.~\eqref{eq:CapacityDA}, then we obtain $N \geq 2$.
We now apply the upper bound $\kappa_{d-1}/2 < \kappa_{d-1} <1 $ from Eq.~\eqref{eq:kappabounds} 
and the upper bound $\frac{2}{\beta  N} \leq \frac{1}{\beta}$ from $N \geq 2$ to inequality Eq.~\eqref{eq:mainclaim:cd}.
In the resulting inequality we insert $N=\sqrt{p} c^{\frac{d-1}{4}}$ to check whether it is fulfilled
with this special value of $N$ and obtain:
\begin{align} \label{eq:mainclaim:c1d}
    &5^{\frac{d-1}{2}} \ p \ c^{\frac{d-1}{2}} \ M^{- (d-1)}
    \left( \frac{1}{\beta} \ + \  
  \frac{1}{\beta} \ \ln \left( 2 \  p \ c^{\frac{d-1}{2}}  \ \beta M^2 \right) \right)^{\frac{d-1}{2}} 
   \leq \ p \ .
\end{align}
Dividing by $p$, inserting $M= K \sqrt{d-1}$,
and exponentiation of the left and right side by $\frac{2}{d-1}$ 
gives:
 \begin{align} \label{eq:inequ1:d}
 \frac{5 \ c}{K^2 \  (d-1)}\left( \frac{1}{\beta} \ + \  
   \frac{1}{\beta} \ \ln \left( 2 \ \beta \ c^{\frac{d-1}{2}} \ p \ K^2 \ 
   (d-1) \right) \right) \ - \ 1 \ \leq \ 0 \ .
\end{align}
This inequality Eq.~\eqref{eq:inequ1:d} can be reformulated as:
 \begin{align} \label{eq:inequ2}
 & 1 \ + \  
   \ln \left( 2 \ p \ \beta \ c^{\frac{d-1}{2}} \ K^2 \ (d-1) \right) \ 
   - \ \frac{(d-1) \ K^2 \ \beta}{5\ c} \ \leq \  0 \ .
\end{align}
Using
\begin{align} \nonumber
  a \ &:= \  \frac{\ln(c)}{2} \ - \ \frac{K^2 \ \beta}{5 \ c} \ , \quad
  b \ := \ 1 \ + \ \ln \left( 2\  p \ \beta \ K^2  \right) \ , \\
\end{align}
we write inequality Eq.~\eqref{eq:inequ2} as
\begin{align} \label{eq:inequ2ab}
  \ln(d-1) \ + \ a \ (d-1) \ + \ b \ \leq \  0 \ .
\end{align}

We determine the value $\hat{d}$ of $d$ 
which makes the inequality Eq.~\eqref{eq:inequ2ab} equal to zero.
We solve
\begin{align} 
   \ln(\hat{d}-1) \ + \ a \ (\hat{d}-1) \ + \ b \ = \  0 \ .
\end{align}
for $\hat{d}$

For $a\not=0$ we have
\begin{align} 
  &\ln(\hat{d}-1) \ + \ a \ (\hat{d}-1) \ + \ b \ = \  0  \\ \nonumber
  \Leftrightarrow  \ \ &a \ (\hat{d}-1) \ + \ \ln(\hat{d}-1)  \ = \ - \ b \\ \nonumber
  \Leftrightarrow  \ \ &(\hat{d}-1) \exp(a \ (\hat{d}-1))  \ = \  \exp(-  b)  \\ \nonumber
  \Leftrightarrow \ \ &a \ (\hat{d}-1) \exp(a \ (\hat{d}-1))  \ = \  a \ \exp(-  b)  \\ \nonumber
  \Leftrightarrow \ \ &a \ (\hat{d}-1) \ = \ W(a \ \exp(-  b) ) \\ \nonumber
  \Leftrightarrow \ \ &\hat{d} \ - \ 1 \ = \ \frac{1}{a} \ W(a \ \exp(- b) )  \\ \nonumber
  \Leftrightarrow \ \ &\hat{d}  \ = \ 1 \ + \ \frac{1}{a} \  W(a \ \exp(- b) ) \ ,
\end{align}
where $W$ is the Lambert $W$ function (see Def.~\ref{th:lambert}).
For $a>0$ we have to use the upper branch $W_0$ of the Lambert $W$ function
and for $a<0$ we use the lower branch $W_{-1}$ of the Lambert $W$ function  \dlmf{4.13}.
We have to ensure that 
$-1/e \leq a  \exp(- b)$ for a solution to exist.
For $a=0$ we have $\hat{d}=1+\exp(-b)$.

Hence, the solution is
\begin{align} 
 \label{eq:hatd}
  \hat{d}  \ &= \ 1 \ + \ \frac{1}{a} \ W(a \exp( - b) ) \ .
\end{align}

Since $\hat{d}$ fulfills inequality Eq.~\eqref{eq:inequ1:d} 
and therefore also Eq.~\eqref{eq:mainclaim:c1d},
we have a lower bound on the storage capacity $N$:
\begin{align} 
    N \ &\geq \ \sqrt{p} \ \hat{c}^{\frac{d-1}{4}}  \ .
\end{align}

\end{proof}

\begin{corollaryA}
We assume a failure probability $0<p\leq 1$ and randomly chosen patterns 
on the sphere with radius $M=K \sqrt{d-1}$. 
We define
\begin{align} \nonumber
  a \ &:= \  \frac{\ln(c)}{2} \ - \ \frac{K^2 \ \beta}{5 \ c} \ , \quad
  b \ := \ 1 \ + \ \ln \left( 2\  p \ \beta \ K^2  \right) \ , \\
  d  \ &= \ 1 \ + \ \frac{1}{a} \ \left(- \ \ln(-a)\ + \ b\right) \ , 
\end{align}
and ensure
\begin{align} 
    \label{eq:Cbound21}
    c \ &\geq \ \left( \frac{2}{ \sqrt{p}}\right)^{\frac{4}{d-1}} \ , \quad \ 
     - \ \frac{1}{e} \ \leq \ a  \ \exp(- b)  \ ,  \quad \ a \ < \ 0 \ ,
\end{align} 
then with probability $1-p$, the number of random patterns 
that can be stored is  
\begin{align} 
 \label{eq:Capacity1dA}
    N \ &\geq \ \sqrt{p} \ c^{\frac{d-1}{4}}  \ .
\end{align}

Setting $\beta=1$, $K=3$, $c=2$ and $p=0.001$ yields $d<24$.
\end{corollaryA}

\begin{proof}
For $a<0$ the
Eq.~\eqref{eq:capacityDdA} from Theorem~\eqref{th:capacityD} 
can be written as 
\begin{align} \label{eq:hatdN}
   d \ = \ 1 \ + \ \frac{W_{-1}(a \exp(-b))}{a} \ = \ 1 \ + \ \frac{W_{-1}(- \exp\left(-(-\ln(-a)+b-1)-1\right))}{a}    
\end{align} 
From \citet[Theorem 3.1]{Alzahrani:18} we get the following bound on $W_{-1}$:
\begin{align} \label{eq:wminus1general}
  - \ \frac{e}{e-1} \ (u+1)\ &< \ W_{-1}(- \ \exp(-u-1)) \ < \ - \ (u +1) \ .
\end{align}
for $u>0$. We apply Eq.~\eqref{eq:wminus1general} to Eq.~\eqref{eq:hatdN} 
with $u=-\ln(-a)+b-1$.

Since $a<0$ we get
\begin{align} 
   d \ > \ 1 \ + \ \frac{-\ln(-a)+b}{a} \ . 
\end{align}

\end{proof}

{\textbullet \em Storage capacity for the expected minimal separation instead of the probability that all patterns
can be stored.}
In contrast to the previous paragraph, we want to argue about the storage capacity
for the expected minimal separation. 
Therefore, we will use the following bound on the expectation of $\alpha_{\min}$ (minimal angle), 
which gives also a bound on the expected of $\Delta_{\min}$ (minimal separation):

\begin{lemmaA}[Proposition 3.6 in \citet{Brauchart:18}]
We have the following lower bound on the expectation of $\alpha_{\min}$:
\begin{align} \label{eq:alpha_min_lower}
   \EXP \left[ N^{\frac{2}{d-1} } \ \alpha_{\min} \right] \ &\geq \ 
   \left( \frac{\Gamma(\frac{d}{2})}{2(d-1) \ \sqrt{\pi} \ \Gamma(\frac{d-1}{2})}\right)^{-\frac{1}{d-1}} \Gamma(1+\frac{1}{d-1}) \ \frac{d^{-\frac{1}{d-1}}}{\Gamma(2+\frac{1}{d-1})} \ := \ C_{d-1}.
\end{align}
The bound is valid for all $N \geq 2$ and $d\geq2$.
\end{lemmaA}
Let us start with some preliminary estimates.
First of all we need some asymptotics for the constant $C_{d-1}$ in Eq.~\eqref{eq:alpha_min_lower}:
\begin{lemmaA}
The following estimate holds for $d\geq 2$:
\begin{align} \label{eq:cd_est}
      C_{d} \ & \geq \ 1 \ - \ \frac{\ln(d+1)}{d} \ .
\end{align}
\end{lemmaA}

\begin{proof}
The recursion formula for the Gamma function is \dlmf{5.5.1}: 
\begin{align}
    \Gamma(x+1) \ &= \ x \ \Gamma(x) \ .
\end{align}
We use Eq.~\eqref{eq:kappabounds} and 
the fact that $d^{\frac1d} \geq 1$ for $d\geq 1$ to obtain:
\begin{align}
    C_d \ &\geq \ (2 \ \sqrt{d})^{\frac{1}{d}} \Gamma(1+\frac{1}{d}) \ 
    \frac{(d+1)^{- \ \frac{1}{d}}}{\Gamma(2+\frac{1}{d})} \ = \ 
    (2 \ \sqrt{d})^{\frac{1}{d}} \frac{(d+1)^{- \ \frac{1}{d}}}{1-\frac{1}{d}}
     \ > \ (d+1)^{\frac{1}{d}}\\ \nonumber
    &= \ \exp(-\frac{1}{d} \ \ln(d+1)) \ \geq \ 1\ - \ \frac{1}{d} \ \ln(d+1) \ ,
\end{align}
where in the last step we used the elementary inequality $\exp(x) \geq 1+x$,
which follows from the mean value theorem.
\end{proof}

The next theorem states the number of stored patterns 
for the expected minimal separation.
\begin{theoremA}[Storage Capacity (expected separation): Random Patterns]
We assume patterns on the sphere with radius $M=K \sqrt{d-1}$
that are randomly chosen. 
Then for all values $c\geq 1$ for which 
\begin{align} \label{eq:storage_main_2}
   \frac{1}{5} \ (d-1) \ K^2 \ c^{-1} (1 \ - \ \frac{\ln(d-1)}{(d-1)})^2 
   \ &\geq \ \frac{2}{\beta \ c^{\frac{d-1}{4}}} \ + \  
  \frac{1}{\beta} \ \ln \left( 2 \ c^{\frac{d-1}{2}} \ \beta \ (d-1) \ K^2 \right)
 \end{align} 
 holds, 
the number of stored patterns for the expected minimal separation is at least
\begin{align} 
 N \ &= \ c^{\frac{d-1}{4}} \ .
 \end{align} 

The inequality Eq.~\eqref{eq:storage_main_2} is e.g.\ fulfilled with $\beta=1$, $K=3$, $c=2$ and $d\geq 17$.
\end{theoremA}

\begin{proof}
Instead of considering the probability that the master inequality Eq.~\eqref{eq:masterAsphere}
is fulfilled we now consider whether this inequality is fulfilled for the expected minimal distance.
We consider the expectation of the minimal distance $\Delta_{\min}$:
\begin{align} 
  \EXP[\Delta_{\min}] \ &= \ \EXP[  M^2  ( 1 \ - \ \cos(\alpha_{\min}) )) ] \ = \
   M^2  ( 1 \ - \ \EXP[ \cos(\alpha_{\min}) )]) \ .
\end{align}
For this expectation, the master inequality Eq.~\eqref{eq:masterAsphere}
becomes
\begin{align} \label{eq:storage_main_1}
  &  M^2  ( 1 \ - \ \EXP[ \cos(\alpha_{\min}) )])
  \ \geq \  \frac{2}{\beta \ N} \ + \  
  \frac{1}{\beta} \ \ln \left( 2 \ N^2 \ \beta \ M^2 \right) \ .
\end{align}
We want to find the largest $N$ that fulfills this inequality.

We apply Eq.~\eqref{eq:cosest} and Jensen's inequality to deduce the following lower bound:
\begin{align} \label{eq:expstoragelower1}
     1 \ - \ \EXP[ \cos(\alpha_{\min})] \ &\geq \ \frac{1}{5} \ \EXP  \left[ \alpha_{\min}^2\right]
  \ \geq \ \frac{1}{5} \ \EXP[ \alpha_{\min}]^2 \ .
\end{align}
Now we use Eq.~\eqref{eq:alpha_min_lower} and Eq.~\eqref{eq:cd_est} to arrive at
\begin{align}
   \EXP[ \alpha_{\min}]^2 \ &\geq \  N^{-\frac{4}{d-1}} \ \EXP[ N^{\frac{2}{d-1}} \ \alpha_{\min}]^2 
   \ \geq \ N^{-\frac{4}{d-1}} \ C_{d-1}^2 \ \geq \ N^{-\frac{4}{d-1}} \ (1-\frac{\ln(d-1)}{(d-1)})^2 \ ,
\end{align}
for sufficiently large $d$. 
Thus in order to fulfill Eq.~\eqref{eq:storage_main_1}, it is 
enough to find values that satisfy Eq.~\eqref{eq:storage_main_2}.

\end{proof}

\paragraph{Retrieval of Patterns with One Update and Small Retrieval Error.}
\label{sec:ConvError}

Retrieval of a pattern $\Bx_i$ for fixed point $\Bx_i^*$ and query $\Bxi$
is defined via an $\epsilon$ by $\NRM{f(\Bxi) \ - \ \Bx_i^*} < \epsilon$,
that is, the update is $\epsilon$-close to the fixed point.
The update rule retrieves a pattern with one update for well separated patterns, that is,
$\Delta_i$ is large.
\begin{theoremA}[Pattern Retrieval with One Update]
\label{th:oneUpdate}
With query $\Bxi$, after one update the distance of the new point $f(\Bxi)$
to the fixed point $\Bx_i^*$ is exponentially small in the separation $\Delta_i$.
The precise bounds using the Jacobian $\rJ = \frac{\partial
  f(\Bxi)}{\partial \Bxi}$ and its value $\rJ^m$ in the mean value
theorem are:
\begin{align}
  \NRM{f(\Bxi) \ - \ \Bx_i^*}
  \ &\leq \  \NRM{\rJ^m}_2 \ \NRM{\Bxi \ - \ \Bx_i^*}  \ , \\
   \NRM{\rJ^m}_2  \ &\leq \
  2 \ \beta \ N \ M^2 \ (N-1) \exp(- \ \beta \
  (\Delta_i \ - \ 2 \  \max \{ \NRM{\Bxi  \ - \ \Bx_i} , \NRM{\Bx_i^* \ - \ \Bx_i} \}  \ M) )\ .
\end{align}
For given $\epsilon$ and 
sufficient large $\Delta_i$, we have $\NRM{f(\Bxi) \ - \ \Bx_i^*} < \epsilon$,
that is, retrieval with one update.
\end{theoremA}

\begin{proof}

From Eq.~\eqref{eq:boundJfix} we have
\begin{align}
  \NRM{\rJ^m}_2  \ &\leq \
  2 \ \beta \ N \ M^2 \ (N-1) \exp(- \ \beta \
  (\Delta_i \ - \ 2 \  \max \{ \NRM{\Bxi  \ - \ \Bx_i} , \NRM{\Bx_i^* \ - \ \Bx_i} \}  \ M) )\ .
\end{align}
After every iteration the mapped point $f(\Bxi)$ is closer to the fixed point $\Bx_i^*$
than the original point $\Bx_i$:
\begin{align}
  \NRM{f(\Bxi) \ - \ \Bx_i^*}
  \ &\leq \  \NRM{\rJ^m}_2 \ \NRM{\Bxi \ - \ \Bx_i^*}  \ .
\end{align}

For given $\epsilon$ and 
sufficient large $\Delta_i$, we have $\NRM{f(\Bxi) \ - \ \Bx_i^*} < \epsilon$,
since $\NRM{\rJ^m}_2$  foes exponentially fast to zero with increasing $\Delta_i$.
\end{proof}

We want to estimate how large $\Delta_i$ is.
For $\Bx_i$ we have:
\begin{align}
 \Delta_i \ &= \ \min_{j, j \not= i} \left( \Bx_i^T \Bx_i \ - \ \Bx_i^T
    \Bx_j \right) \ = \ \Bx_i^T \Bx_i \ - \ \max_{j, j \not= i} \Bx_i^T
  \Bx_j \ .
\end{align}
To estimate how large $\Delta_i$ is,
assume vectors $\Bx \in \dR^d$ and $\By \in \dR^d$ that have as components 
standard normally distributed values.
The expected value of the separation of two points
with normally distributed components is
\begin{align}
 \EXP \left[ \Bx^T \Bx \ - \ \Bx^T\By \right] \ &= \ 
 \sum_{j=1}^d  \EXP \left[ x_j^2 \right] \ + \ \sum_{j=1}^d  \EXP \left[ x_j \right]
 \sum_{j=1}^d \EXP \left[ y_j \right] \ = \ d \ .
\end{align}
The variance of the separation of two points 
with normally distributed components is
\begin{align}
 &\VAR \left[ \Bx^T \Bx \ - \ \Bx^T\By \right] \ = \ 
 \EXP \left[ \left(\Bx^T \Bx \ - \ \Bx^T\By \right)^2 \right] \ - \ d^2 \\ \nonumber
 &= \ \sum_{j=1}^d  \EXP \left[ x_j^4 \right] \ + \ 
 \sum_{j=1,k=1,k\not=j}^d \EXP \left[ x_j^2 \right] \ \EXP \left[ x_k^2 \right] \ 
  \ - \ 2 \ \sum_{j=1}^d  \EXP \left[ x_j^3 \right] \EXP \left[ y_j \right] \ - \\ \nonumber
  & 2 \ \sum_{j=1,k=1,k\not=j}^d  \EXP \left[ x_j^2 \right] \EXP \left[ x_k \right] \EXP \left[ y_k \right] \ + \ 
  \sum_{j=1}^d \EXP \left[ x_j^2 \right] \ \EXP \left[ y_j^2 \right] \ + \\ \nonumber
  &\sum_{j=1,k=1,k\not=j}^d  \EXP \left[ x_j \right] \EXP \left[ y_j \right] 
  \EXP \left[ x_k \right] \EXP \left[ y_k \right] \ - \ d^2 \\ \nonumber
  &= 3 \ d \ + \ d \ (d-1) \ + \ d \ - \ d^2 \ = \ 3 \ d \ .
\end{align}

The expected value for the separation of two random vectors gives:
\begin{align}
  \NRM{\rJ^m}_2  \ &\leq \
  2 \ \beta \ N \ M^2 \ (N-1) \exp(- \ \beta \
  (d \ - \ 2 \  \max \{ \NRM{\Bxi  \ - \ \Bx_i} , \NRM{\Bx_i^* \ - \ \Bx_i} \}  \ M) )\ .
\end{align}
For the exponential storage we set $M= 2 \sqrt{d-1}$.
We see the Lipschitz constant $\NRM{\rJ^m}_2 $ decreases exponentially with the dimension.
Therefore, $\NRM{f(\Bxi) \ - \ \Bx_i^*}$ is exponentially small after just one update.
Therefore, the fixed point is well retrieved after one update.

\vspace{1cm}

The retrieval error decreases exponentially with the separation $\Delta_i$.
\begin{theoremA}[Exponentially Small Retrieval Error]
\label{th:retrievalError}
The retrieval error $\NRM{f(\Bxi) \ - \ \Bx_i}$ of pattern $\Bx_i$
is bounded by
\begin{align}
   \NRM{f(\Bxi) \ - \ \Bx_i} \ &\leq \ 2 \ (N-1) \ \exp(- \ \beta \ 
   (\Delta_i \ - \ 2 \   \max \{ \NRM{\Bxi  \ - \ \Bx_i} , \NRM{\Bx_i^* \ - \ \Bx_i} \} 
   \ M) )  \ M  
 \end{align}
and for 
$\NRM{\Bx_i - \Bx_i^*} \leq \frac{1}{2 \ \beta \ M } $ 
together with $\NRM{\Bx_i - \Bxi} \leq \frac{1}{2 \ \beta \ M } $
by
\begin{align}
 \NRM{\Bx_i \ - \ \Bx_i^*} \ &\leq \ 2 \ e \ (N-1) \ M \ \exp(- \ \beta \ \Delta_i )  \ .  
\end{align}
\end{theoremA}

\begin{proof}
We compute the retrieval error which is just 
$\NRM{f(\Bxi) \ - \ \Bx_i}$.
From Lemma~\ref{th:similar} we have
\begin{align}
  \NRM{\Bx_i \ - \ f(\Bxi)} \ &\leq \ 2 \ \epsilon \ M \ ,  
\end{align}
From Eq.~\eqref{eq:epsA} we have
\begin{align}
\label{eq:inA}
\epsilon \ &= \ (N-1) \exp(- \ \beta \ 
 (\Delta_i \ - \ 2 \  \max \{ \NRM{\Bxi  \ - \ \Bx_i} , \NRM{\Bx_i^* \ - \ \Bx_i} \}  \ M) )  \ .
\end{align}

For 
$\NRM{\Bx_i - \Bx_i^*} \leq \frac{1}{2 \ \beta \ M } $ 
and $\NRM{\Bx_i - \Bxi} \leq \frac{1}{2 \ \beta \ M } $
Eq.~\eqref{eq:inA}
gives
\begin{align}
\epsilon  \ &\leq \  e \ (N-1) \ M \ \exp(- \ \beta \ \Delta_i  )   \ . 
\end{align}

\end{proof}

\subsubsection{Learning Associations}
\label{sec:LearningAss}

We consider three cases of learning associations,
i.e.\ three cases of how sets are associated.
(i) Non of the sets is mapped in an associative space.
The raw state pattern $\Br_n$ is the state (query) pattern $\Bxi_n$, 
i.e.\ $\Bxi_n=\Br_n$, and the raw stored pattern $\By_s$ is
the stored pattern (key), i.e.\  $\Bx_s=\By_s$.
(ii) Either one of the sets is mapped to the space of the
other set or an association matrix is learned.
(iia) The state patterns are equal to the raw patterns, i.e.\ $\Bxi_n=\Br_n$, and 
raw stored patterns are mapped via $\BW$ to the space of the state patterns,
i.e.\ $\Bx_s=\BW \By_s$. 
(iib) The stored patterns are equal to the raw patterns, i.e.\ $\Bx_s=\By_s$, and raw
state patterns are mapped via $\BW$ to the space of the stored patterns, i.e.\ $\Bxi_n= \BW^T \Br_n$.
(iic) The matrix $\BW$ is an association matrix.
We will compute the derivative of the new state pattern 
with respect to $\BW$, which is valid for all sub-cases (iib)--(iic).
(iii) Both set of patterns are mapped in a common associative space.
A raw state pattern $\Br_n$ is mapped by $\BW_Q$ to a state pattern (query) $\Bxi_n$,
that is $\Bxi_n= \BW_Q \Br_n$. 
A raw stored pattern $\By_s$ is mapped via
$\BW_K$  to stored pattern (key) $\Bx_s$, that is $\Bx_s=\BW_K \By_s$.
We will compute the derivative of the new state pattern with respect to both $\BW_Q$ and $\BW_K$.

\paragraph{Association of Raw Patterns -- No Mapping in an Associative Space.}
The sets are associated via their raw patterns,
i.e.\ the raw state pattern $\Br_n$ is the state (query) pattern $\Bxi_n$,
i.e.\ $\Bxi_n=\Br_n$, and raw stored pattern $\By_s$ is
the stored pattern (key), i.e.\  $\Bx_s=\By_s$.
There is no mapping in an associative space.

The update rule is
\begin{align}
 \Bxi^{\nn} \ &= \ \BX \ \Bp \ ,
\end{align}
where we used 
\begin{align}
  \Bp \ &= \ \soft (\beta \ \BX^T \Bxi) \ .
\end{align}

The derivative with respect to $\Bxi$ is
\begin{align}
 \frac{\partial \Bxi^{\nn}}{\partial \Bxi} \ &= \ 
  \beta \ \BX \ \left(
  \diag(\Bp) - \Bp \Bp^T \right) \ \BX^T 
\end{align}

The derivative with respect to $\BX$ is
\begin{align}
 \frac{\partial \Ba^T \Bxi^{\nn}}{\partial \BX} \ &= \ 
 \Ba \ \Bp^T \ + \ 
   \beta \  \BX \ \left(
  \diag(\Bp) - \Bp \Bp^T \right) \  ( \Bxi^T \Ba) \ .
\end{align}

These derivatives allow to apply the chain rule if a 
Hopfield layer is integrated into a deep neural network.

\paragraph{Learning an Association Matrix -- Only One Set is Mapped in an Associative Space.}
\label{sec:LearningDirect}
Only one of the sets $\BR$ or $\BY$ is mapped in the space of
the patterns of the other set. 
Case (a): the state patterns are equal to the raw patterns $\Bxi_n=\Br_n$ and 
raw stored patterns are mapped via $\BW$ to the space of the state patterns, i.e.\
$\Bx_s=\BW \By_s$.
Case (b): the stored patterns are equal to the raw patterns $\Bx_s=\By_s$ 
and raw state patterns are mapped via $\BW$ 
to the space of the stored patterns, i.e.\  $\Bxi_n= \BW^T \Br_n$.
Case (c): the matrix $\BW$ associates the sets $\BR$ and $\BY$. 
This case also includes that $\BW^T = \BW_K^T \BW_Q$, which is treated 
in next subsection. 
The next subsection focuses on a low rank approximation of $\BW$ by defining  
the dimension $d_k$ of associative space and use the matrices
$\BW_K^T$ and $\BW_Q$ to define $\BW$, or equivalently to map $\BR$ and $\BY$
into the associative space.

From a mathematical point of view all these case are equal as they 
lead to the same update rule.
Therefore, we consider in the following Case (a) with
$\Bx_s=\BW \By_s$ and $\Bxi_n=\Br_n$.
Still, the following formula are valid for 
all three cases (a)--(c).

The update rule is
\begin{align}
 \Bxi^{\nn} \ &= \ \BW \ \BY \ \Bp \ ,
\end{align}
where we used 
\begin{align}
  \Bp \ &= \ \soft (\beta \ \BY^T \BW^T \Bxi) \ .
\end{align}

We consider the state (query) pattern $\Bxi$ with result $\Bxi^{\nn}$:
\begin{align}
\label{eq:FullUpdate}
 \Bxi^{\nn} \ &= \ \BW \ \BY \ \Bp \ = \   \BW \ \BY \ \soft ( \beta \ \BY^T \BW^T \Bxi) 
\end{align}
For multiple updates this update rule has to be used.
However for a single update, or the last update we consider a
simplified update rule.

Since new state vector $\Bxi^{\nn}$ is projected 
by a weight matrix $\BW_V$
to another vector, we consider the simplified update rule:
\begin{align}
 \Bxi^{\nn} \ &= \ \BY \ \Bp \ = \  \BY \ \soft ( \beta \ \BY^T \BW^T \Bxi) 
\end{align}

The derivative with respect to $\BW$ is
\begin{align}
 \frac{\partial \Ba^T \Bxi^{\nn}}{\partial \BW} \ &= \   
 \frac{\partial \Bxi^{\nn}}{\partial \BW}  \
 \frac{\partial \Ba^T \Bxi^{\nn}}{\partial \Bxi^{\nn}} \ = \ 
 \frac{\partial \Bxi^{\nn}}{\partial (\BW^T \Bxi)} \
 \frac{\partial (\BW^T \Bxi)}{\partial \BW}  \
 \frac{\partial \Ba^T \Bxi^{\nn}}{\partial \Bxi^{\nn}} \ .
\end{align}

\begin{align}
 \frac{\partial \Bxi^{\nn}}{\partial (\BW^T \Bxi)} \ &= \ 
  \beta \ \BY \ \left(
  \diag(\Bp) - \Bp \Bp^T \right) \ \BY^T 
 \end{align}

\begin{align}
 \frac{\partial \Ba^T \Bxi^{\nn}}{\partial \Bxi^{\nn}} \ &= \  \Ba \  .
\end{align}

We have the product of the 3-dimensional tensor
$\frac{\partial (\BW^T \Bxi)}{\partial \BW}$ with the 
vector $\Ba$ which gives a 2-dimensional tensor, i.e.\ a
matrix:
\begin{align}
 \frac{\partial (\BW^T \Bxi)}{\partial \BW}  \
 \frac{\partial \Ba^T \Bxi^{\nn}}{\partial \Bxi^{\nn}} \ &= \ 
 \frac{\partial (\BW^T \Bxi)}{\partial \BW} \ \Ba \ = \ 
 \Bxi^T \Ba \BI  \ .
\end{align}

\begin{align}
\label{eq:gradW}
 \frac{\partial \Ba^T \Bxi^{\nn}}{\partial \BW} \ &= \   \beta \  \BY \ \left(
  \diag(\Bp) - \Bp \Bp^T \right) \  \BY^T ( \Bxi^T \Ba ) 
  \ = \ \rJ \ ( \Bxi^T \Ba ) \ ,
\end{align}
where $\rJ$ is the Jacobian of the update rule defined in Eq.~\eqref{eq:theJacobian}.

To obtain the derivative of the full update rule Eq.~\eqref{eq:FullUpdate}
we have to add the term
\begin{align}
  \Ba \ \Bp^T \BY^T 
\end{align}
and include the factor $\BW$ to get
\begin{align}
\label{eq:gradWFull}
 \frac{\partial \Ba^T \Bxi^{\nn}}{\partial \BW} \ &= \  
 \Ba \ \Bp^T \BY^T  \ + \ \beta \  \BW \ \BY \ \left(
  \diag(\Bp) - \Bp \Bp^T \right) \ \BY^T  ( \Bxi^T \Ba ) 
  \\ \nonumber
  &= \  \Ba \ \Bp^T \BY^T  \ + \ \BW \ \rJ \ ( \Bxi^T \Ba ) \ .
\end{align}

\paragraph{Learning Two Association Mappings -- Both Sets are Mapped in an Associative Space.}
\label{sec:LearningMap}

Both sets $\BR$ and $\BY$ are mapped in an associative space.
Every raw state pattern $\Br_n$ is mapped via
$\BW_Q$ to a state pattern (query) $\Bxi_n= \BW_Q \Br_n$. 
Every raw stored pattern $\By_s$ is mapped via
$\BW_K$ to a stored pattern (key) $\Bx_s=\BW_K \By_s$.
In the last subsection we considered a single matrix $\BW$.
For $\BW^T = \BW_K^T \BW_Q$ we have the case of the last subsection.
However in this subsection we are looking for a
low rank approximation of $\BW$. 
Toward this end we define the dimension $d_k$ 
of associative space and use the matrices
$\BW_K^T$ and $\BW_Q$ to map to the associative space.

The update rule is
\begin{align}
 \Bxi^{\nn} \ &= \ \BX \ \Bp  \ ,
\end{align}
where we used 
\begin{align}
  \Bp \ &= \ \soft (\beta \ \BX^T \Bxi) \ .
\end{align}
We consider raw state patterns $\Br_n$ that are mapped to 
state patterns $\Bxi_n = \BW_Q \Br_n$ with $\BQ^T=\BXi=\BW_Q \BR$ 
and raw stored pattern $\By_s$ that are mapped to 
stored patterns $\Bx_s = \BW_K \By_s$ with $\BK^T=\BX=\BW_K \BY$. 
The update rule is
\begin{align}
 \label{eq:FullUpdate2set}
 \Bxi^{\nn}  \ &= \  \BW_K  \ \BY \ \Bp \ = \  
  \BW_K \ \BY \ \soft ( \beta \ \BY^T \BW_K^T \BW_Q \ \Br) \ . 
\end{align}

Since new state vector $\Bxi^{\nn}$ is projected 
by a weight matrix $\BW_V$
to another vector, we consider the simplified update rule:
\begin{align}
 \label{eq:SimpleUpdate2set}
 \Bxi^{\nn} \ &= \ \BY \ \Bp \ = \  
 \BY \ \soft ( \beta \ \BY^T \BW_K^T \BW_Q \ \Br) \ .  
\end{align}
For the simplified update rule,
the vector $\Bxi^{\nn}$ does not live in the associative space but
in the space of raw stored pattern $\By$. 
However $\BW_K$ would map it to the associative space.

{\textbullet \em Derivative with respect to $\BW_Q$.}
The derivative with respect to $\BW_Q$ is
\begin{align}
 \frac{\partial \Ba^T \Bxi^{\nn}}{\partial \BW_Q} \ &= \   
 \frac{\partial \Bxi^{\nn}}{\partial \BW_Q}  \
 \frac{\partial \Ba^T \Bxi^{\nn}}{\partial \Bxi^{\nn}} \ = \ 
 \frac{\partial \Bxi^{\nn}}{\partial (\BW_Q \ \Br)} \
 \frac{\partial (\BW_Q \ \Br)}{\partial \BW_Q} \ 
 \frac{\partial \Ba^T \Bxi^{\nn}}{\partial \Bxi^{\nn}} \ .
\end{align}

\begin{align}
 \frac{\partial \Bxi^{\nn}}{\partial (\BW_Q \ \Br)} \ &= \ 
  \beta \ \BY \ \left(
  \diag(\Bp) - \Bp \Bp^T \right) \ \BY^T \BW_K^T 
 \end{align}

\begin{align}
 \frac{\partial \Ba^T \Bxi^{\nn}}{\partial \Bxi^{\nn}} \ &= \  \Ba \  .
\end{align}

We have the product of the 3-dimensional tensor
$\frac{\partial (\BW_Q \Br)}{\partial \BW_Q}$ with the 
vector $\Ba$ which gives a 2-dimensional tensor, i.e.\ a
matrix:
\begin{align}
 \frac{\partial (\BW_Q \ \Br)}{\partial \BW_Q}  \
 \frac{\partial \Ba^T \Bxi^{\nn}}{\partial \Bxi^{\nn}} \ &= \ 
 \frac{\partial (\BW_Q \ \Br)}{\partial \BW_Q} \ \Ba \ = \ 
 \Br^T \Ba \ \BI  \ .
\end{align}

\begin{align}
 \label{eq:gradWQ}
 \frac{\partial \Ba^T \Bxi^{\nn}}{\partial \BW_Q} \ &= \   \beta \  \BY \ \left(
  \diag(\Bp) - \Bp \Bp^T \right) \ \BY^T \ \BW_K^T ( \Br^T \Ba) 
  \ = \ \rJ \ \BW_K^T ( \Br^T \Ba) \ ,
\end{align}
where $\rJ$ is the Jacobian of the update rule defined in Eq.~\eqref{eq:theJacobian}.
 
To obtain the derivative of the full update rule Eq.~\eqref{eq:FullUpdate2set}
we have to include the factor $\BW_K$, then get
\begin{align}
 \label{eq:gradWQFull}
 \frac{\partial \Ba^T \Bxi^{\nn}}{\partial \BW_Q} \ &= \   
 \beta \  \BW_K \ \BY \left(
  \diag(\Bp) - \Bp \Bp^T \right) \ \BY^T \ \BW_K^T ( \Br^T \Ba) 
  \ = \ \BW_K \ \rJ \ \BW_K^T ( \Br^T \Ba) \ .
\end{align}

{\textbullet \em Derivative with respect to $\BW_K$.}
The derivative with respect to $\BW_K$ is
\begin{align}
 \frac{\partial \Ba^T \Bxi^{\nn}}{\partial \BW_K} \ &= \   
 \frac{\partial \Bxi^{\nn}}{\partial \BW_K}  \
 \frac{\partial \Ba^T \Bxi^{\nn}}{\partial \Bxi^{\nn}} \ = \ 
 \frac{\partial \Bxi^{\nn}}{\partial (\BW_K^T \BW_Q \ \Br)} \
 \frac{\partial (\BW_K^T \BW_Q \ \Br)}{\partial \BW_K}  \
 \frac{\partial \Ba^T \Bxi^{\nn}}{\partial \Bxi^{\nn}} \ .
\end{align}

\begin{align}
 \frac{\partial \Bxi^{\nn}}{\partial (\BW_K^T \BW_Q \ \Br)} \ &= \ 
  \beta \ \BY \ \left(
  \diag(\Bp) - \Bp \Bp^T \right) \ \BY^T 
 \end{align}

\begin{align}
 \frac{\partial \Ba^T \Bxi^{\nn}}{\partial \Bxi^{\nn}} \ &= \  \Ba \  .
\end{align}

We have the product of the 3-dimensional tensor
$\frac{\partial (\BW \Br)}{\partial \BW_K}$ with the 
vector $\Ba$ which gives a 2-dimensional tensor, i.e.\ a
matrix:
\begin{align}
 \frac{\partial (\BW_K^T\BW_Q \ \Br)}{\partial \BW_K}  \
 \frac{\partial \Ba^T \Bxi^{\nn}}{\partial \Bxi^{\nn}} \ &= \ 
 \frac{\partial (\BW_K^T \BW_Q \ \Br)}{\partial \BW_K} \ \Ba \ = \ 
 \BW_Q^T \Br^T \Ba \ \BI  \ .
\end{align}

\begin{align}
 \label{eq:gradWK}
 \frac{\partial \Ba^T \Bxi^{\nn}}{\partial \BW_K} \ &= \   \beta \  \BY \ \left(
  \diag(\Bp) - \Bp \Bp^T \right) \ \BY^T \ ( \BW_Q^T \Br^T \Ba ) 
  \ = \ \rJ \ ( \BW_Q^T \Br^T \Ba ) \ ,
\end{align}
where $\rJ$ is the Jacobian of the update rule defined in Eq.~\eqref{eq:theJacobian}.

To obtain the derivative of the full update rule Eq.~\eqref{eq:FullUpdate2set}
we have to add the term
\begin{align}
  \Ba \ \Bp^T \BY^T 
\end{align}
and to include the factor $\BW_K$, then get
\begin{align}
\label{eq:gradWKFull}
 \frac{\partial \Ba^T \Bxi^{\nn}}{\partial \BW_K} \ &= \  
 \Ba \ \Bp^T \BY^T  \ + \ \beta \  \BW_K \ \BY \ \left(
  \diag(\Bp) - \Bp \Bp^T \right) \ \BY^T  ( \BW_Q^T \Br^T \Ba ) 
  \\ \nonumber 
  &= \  \Ba \ \Bp^T \BY^T  \ + \ \BW_K \ \rJ \  ( \BW_Q^T \Br^T \Ba )  \ .
\end{align}

\subsubsection{Infinite Many Patterns and Forgetting Patterns}
\label{sec:sequential}

In the next subsection we show how
the new Hopfield networks can be used for auto-regressive tasks
by causal masking.
In the following subsection, 
we introduce forgetting to the new Hopfield networks 
by adding a negative value to the softmax which is larger
if the pattern was observed more in the past.

\paragraph{Infinite Many Patterns.}
\label{sec:infinite}

The new Hopfield networks can be used for auto-regressive tasks, 
that is time series prediction and similar.
Causal masking  masks out the future by a large negative
value in the softmax.

We assume to have infinite many stored patterns (keys) $\Bx_1,\Bx_2,\ldots$
that are represented by the infinite matrix
\begin{align}
  \BX \ &= \ \left( \Bx_1,\Bx_2,\ldots,\right) \ .
\end{align}
The pattern index is now a time index, that is,
we observe $\Bx_t$ at time $t$.

The pattern matrix at time $t$ is
\begin{align}
  \BX_t \ &= \ \left( \Bx_1,\Bx_2,\ldots,\Bx_t\right) \ .
\end{align}
The query at time $t$ is $\Bxi_t$.

For $M_t=\max_{1\leq i \leq t} \NRM{\Bx_t}$,
the energy function at time $t$ is $\rE_t$
\begin{align}
  \rE_t \ &= \ - \ \mathrm{lse}(\beta, \BX_t^T \Bxi_t) \ + \
  \frac{1}{2} \Bxi_t^T \Bxi_t \ + \ \beta^{-1} \ln t 
  \ + \ \frac{1}{2} M_t^2\\
  &= \ - \ \beta^{-1} \ln \left( \sum_{i=1}^t
    \exp(\beta \Bx_i^T \Bxi_t) \right)  \ + \
  \frac{1}{2} \Bxi_t^T \Bxi_t \ + \ \beta^{-1} \ln t 
  \ + \ \frac{1}{2} M_t^2\ .
\end{align}

The update rule is
\begin{align}
  \Bxi_t^{\nn} \ &= \ \BX_t \ \Bp_t \ = \ \BX_t \ \soft ( \beta \ \BX_t^T \Bxi_t) \ ,
\end{align}
where we used 
\begin{align}
  \Bp_t \ &= \ \soft (\beta \ \BX_t^T \Bxi_t) \ .
\end{align}

We can use an infinite pattern matrix with 
an infinite softmax when using causal masking.
The pattern matrix at time $t$ is
\begin{align}
  \BX_t \ &= \ \left( \Bx_1,\Bx_2,\ldots,\Bx_t,
  -\alpha \Bxi_t,-\alpha \Bxi_t,\ldots\right) \ ,
\end{align}
with the query $\Bxi_t$ and $\alpha \to \infty$.
The energy function at time $t$ is $\rE_t$
\begin{align}
  \rE_t \ &= \ - \ \mathrm{lse}(\beta, \BX_t^T \Bxi_t) \ + \
  \frac{1}{2} \Bxi_t^T \Bxi_t \ + \ \beta^{-1} \ln t 
  \ + \ \frac{1}{2} M_t^2\\
  &= \ - \ \beta^{-1} \ln \left( \sum_{i=1}^{t}
    \exp(\beta \Bx_i^T \Bxi_t) \ + \ 
    \sum_{i=t+1}^{\lfloor \alpha \rfloor}  
    \exp(- \beta \alpha \NRM{\Bxi_t}^2) 
    \right)  \ + \
  \frac{1}{2} \Bxi_t^T \Bxi_t \ + \\ \nonumber
  &~~~~~~~\beta^{-1} \ln t  \ + \ \frac{1}{2} M_t^2 \ .
\end{align}
For $\alpha \to \infty$ and $\NRM{\Bxi_t} > 0$ this becomes
\begin{align}
  \rE_t \ &= \ - \ \mathrm{lse}(\beta , \BX_t^T \Bxi_t) \ + \
  \frac{1}{2} \Bxi_t^T \Bxi_t \ + \ \beta^{-1} \ln t 
  \ + \ \frac{1}{2} M_t^2\\
  &= \ - \ \beta^{-1} \ln \left( \sum_{i=1}^{t}
    \exp(\beta \Bx_i^T \Bxi_t) \right)  \ + \
  \frac{1}{2} \Bxi_t^T \Bxi_t \ +  \
  \beta^{-1} \ln t 
  \ + \ \frac{1}{2} M_t^2\ .
\end{align}

\paragraph{Forgetting Patterns.}
\label{sec:forgetting}

We introduce forgetting to the new Hopfield networks 
by adding a negative value in the softmax which increases 
with patterns that are more in the past.

We assume to have infinite many patterns $\Bx_1,\Bx_2,\ldots$
that are represented by the infinite matrix
\begin{align}
  \BX \ &= \ \left( \Bx_1,\Bx_2,\ldots,\right) \ .
\end{align}
The pattern index is now a time index, that is,
we observe $\Bx_t$ at time $t$.

The pattern matrix at time $t$ is
\begin{align}
  \BX_t \ &= \ \left( \Bx_1,\Bx_2,\ldots,\Bx_t\right) \ .
\end{align}
The query at time $t$ is $\Bxi_t$.

The energy function with forgetting parameter
$\gamma$ at time $t$ is $\rE_t$
\begin{align}
  \rE_t \ &= \ - \ \mathrm{lse}(\beta ,\BX_t^T \Bxi_t \ - \ 
  \gamma (t-1,t-2,\ldots,0)^T ) \ + \
  \frac{1}{2} \Bxi_t^T \Bxi_t \ + \ \beta^{-1} \ln t 
  \ + \ \frac{1}{2} M_t^2\\
  &= \ - \ \beta^{-1} \ln \left( \sum_{i=1}^T
    \exp(\beta \Bx_i^T \Bxi_t \ - \ \gamma (t-i) ) \right)  \ + \
  \frac{1}{2} \Bxi_t^T \Bxi_t \ + \ \beta^{-1} \ln t 
  \ + \ \frac{1}{2} M_t^2\ .
\end{align}

The update rule is
\begin{align}
  \Bxi_t^{\nn} \ &= \ \BX_t \ \Bp_t  \ = \   
 \BX_t \ \soft ( \beta \BX_t^T \Bxi_t) \ ,
\end{align}
where we used 
\begin{align}
  \Bp_t \ &= \ \soft (\beta \BX_t^T \Bxi_t) \ .
\end{align}

\subsubsection{Number of Spurious States}

The energy $\rE$ is defined as
\begin{align}
  \rE \ &= \ - \ \mathrm{lse}(\beta ,\BX^T \Bxi) \ + \
  \frac{1}{2} \Bxi^T \Bxi  \ + \ \beta^{-1} \ln N \ + \ 
  \frac{1}{2} M^2 \\
  &= \ - \ \beta^{-1} \ln \left( \sum_{i=1}^N
    \exp(\beta \Bx_i^T \Bxi) \right)  \ + \ \beta^{-1} \ln N \ + \
  \frac{1}{2} \Bxi^T \Bxi \ + \ \frac{1}{2} M^2 \ .
\end{align}

Since the negative exponential function is strict monotonic decreasing,
$\exp(-\rE)$ has minima, where $\rE$ has maxima, and has maxima, 
where as has minima $\rE$.
\begin{align}
  \exp(-\rE) \ &= \ \exp ( \mathrm{lse}(\beta ,\BX^T \Bxi)) \
  \exp( - \ \frac{1}{2} \Bxi^T \Bxi) \ C  \\ \nonumber 
  &=  \left( \sum_{i=1}^N
    \exp(\beta \Bx_i^T \Bxi) \right)^{\beta^{-1}} \  
    \exp( - \ \frac{1}{2} \Bxi^T \Bxi) \ C \\ \nonumber 
  &=  \left( \sum_{i=1}^N
    \exp(\beta \Bx_i^T \Bxi) \right)^{\beta^{-1}} \  
    \left(\exp( - \ \beta \ \frac{1}{2} \Bxi^T \Bxi)  \right)^{\beta^{-1}} \ C \\ \nonumber 
  &=  \left( \sum_{i=1}^N
    \exp(\beta \ (\Bx_i^T \Bxi \ - \ \frac{1}{2} \Bxi^T \Bxi)) \right)^{\beta^{-1}}   
    \ C \\ \nonumber 
  &=  \left( \sum_{i=1}^N
    \exp(\frac{1}{2} \ \beta \ \Bx_i^T \Bx_i \ - \ \frac{1}{2} \ \beta \ (\Bxi \ - \ \Bx_i)^T  (\Bxi \ - \ \Bx_i) ) \right)^{\beta^{-1}} \  C   \\ \nonumber 
  &=  \left( \sum_{i=1}^N \lambda(\Bx_i,\beta) \ G(\Bxi; \Bx_i, \beta^{-1} \ \BI)
     \right)^{\beta^{-1}} \  C \ ,
\end{align}
where $C$ is a positive constant, 
$\lambda(\Bx_i,\beta )=\exp(\frac{1}{2}  \beta  \Bx_i^T \Bx_i)$ and
$G(\Bxi; \Bx_i, \beta^{-1} \BI)$ is the Gaussian with mean $\Bx_i$ and covariance matrix $ \beta^{-1} \BI$.

Since $C$ is a positive constant and $x^{\beta^{-1}} = \exp(\beta^{-1} \ln x)$ is
strict monotonic for positive $x$, the minima of $\rE$ are the maxima of
\begin{align}
\label{eq:MOG}
  \sum_{i=1}^N \lambda(\Bx_i,\beta) \ G(\Bxi; \Bx_i, \beta^{-1} \ \BI) \ .
\end{align}

In \citet{Carreira-Perpinan:03tr} it was shown
that Eq.~\eqref{eq:MOG} can have more than $N$ modes, that is, more
than $N$ maxima.

\subsection{Properties of Softmax, Log-Sum-Exponential, Legendre Transform, Lambert W Function}
\label{sec:MathProperties}

For $\beta > 0$, the {\em softmax} is defined as
\begin{definitionA}[Softmax]
\begin{align} \label{th:defsoftmax}
  \Bp \ &= \ \soft (\beta \Bx) \\
  p_i \ &= \ [ \soft (\beta \Bx) ]_i \ = \ \frac{\exp(\beta
    x_i)}{\sum_k \exp(\beta x_k)} \ .
\end{align}
\end{definitionA}

We also need the {\em log-sum-exp function} ($\mathrm{lse}$), defined as 
\begin{definitionA}[Log-Sum-Exp Function]
\begin{align} \label{th:deflse}
  \mathrm{lse}(\beta,\Bx) \ &= \ \beta^{-1} \ln \left( \sum_{i=1}^N
    \exp(\beta x_i) \right) \ . 
\end{align}
\end{definitionA}

We can formulate the $\mathrm{lse}$
in another base:
\begin{align}
  \beta_a \ &= \ \frac{\beta}{\ln a} \ , \\
  \mathrm{lse}(\beta,\Bx) \ &= \
  \beta^{-1} \ln \left( \sum_{i=1}^N \exp(\beta \ x_i) \right) \\ \nonumber
  &= \ 
  \left( \beta_a \ \ln a \right)^{-1} \ln \left( \sum_{i=1}^N
    \exp(\beta_a \ \ln a \ x_i) \right) \\ \nonumber
  &= \
  \left( \beta_a \right)^{-1} \log_a \left( \sum_{i=1}^N
    a^{\beta_a \ x_i} \right) \ .
\end{align}
In particular, the base $a=2$ can be used to speed up computations.

Next, we give the relation between the softmax and the $\mathrm{lse}$ function.
\begin{lemmaA}
\label{th:Llsa}
The softmax is the gradient of the $\mathrm{lse}$:
\begin{align} \label{th:softgrad}
   \soft (\beta \Bx)  \ &= \ \nabla_{\Bx} \mathrm{lse}(\beta, \Bx) \ . 
\end{align}
\end{lemmaA}
In the next lemma we report some important properties of the $\mathrm{lse}$ function.
\begin{lemmaA}
\label{th:Lentropy}
We define
\begin{align}
  \rL \ &:= \ \Bz^T \Bx \ - \  
  \beta^{-1} \sum_{i=1}^N z_i \ln z_i  
\end{align}
with $\rL \geq  \Bz^T \Bx$.
The $\mathrm{lse}$ is the maximum of $\rL$ on
the $N$-dimensional 
simplex $D$ with $D=\{\Bz \mid \sum_i z_i=1, 0 \leq z_i\}$:
\begin{align} \label{eq:maxlse}
  \mathrm{lse}(\beta, \Bx) \ &= \ \max_{\Bz \in D} \Bz^T \Bx \ - \  
  \beta^{-1} \sum_{i=1}^N z_i \ln z_i  \ . 
\end{align}
The softmax  $\Bp = \soft (\beta \Bx)$ is the argument 
of the maximum of $\rL$ on
the $N$-dimensional
simplex $D$ with $D=\{\Bz \mid \sum_i z_i=1, 0 \leq z_i\}$:
\begin{align}  \label{eq:argmaxlse}
  \Bp \ &= \ \soft (\beta \Bx) \ = \ \arg\max_{\Bz \in D} \Bz^T \Bx \ - \  
  \beta^{-1} \sum_{i=1}^N z_i \ln z_i  \ . 
\end{align}
\end{lemmaA}
\begin{proof}
  Eq.~\eqref{eq:maxlse} is obtained from Equation~(8) in \citet{Gao:17} and
  Eq.~\eqref{eq:argmaxlse} from Equation~(11) in \citet{Gao:17}.
\end{proof}
From a physical point of view, the $\mathrm{lse}$ function represents the ``free energy''
in statistical thermodynamics \citep{Gao:17}.

Next we consider the Jacobian of the softmax and its properties.
\begin{lemmaA}
\label{th:Ljacobi}
The Jacobian $\rJ_s$ of the softmax  $\Bp = \soft (\beta \Bx)$ is
\begin{align}
  \rJ_s \ &= \ \frac{\partial \soft(\beta \Bx) }{\partial \Bx}
  \ = \  \beta \left( \diag(\Bp) - \Bp \Bp^T  \right) \ ,
\end{align}
which gives the elements
\begin{align}
  [\rJ_s]_{ij} \ &= \
  \begin{cases}
   \beta p_i(1-p_i) & \text{for} \ i = j \\
   - \beta p_i p_j  & \text{for} \ i \not = j 
  \end{cases}
  \ . 
\end{align}
\end{lemmaA}

Next we show that $\rJ_s$ has eigenvalue $0$.
\begin{lemmaA}
\label{th:LjacobiZero}
The Jacobian $\rJ_s$ of the softmax function $\Bp = \soft (\beta \Bx)$
has a zero eigenvalue with eigenvector $\BOn$.
\end{lemmaA}
\begin{proof}
\begin{align}
  [\rJ_s \BOn]_i \ &= \ \beta \left(  p_i(1-p_i) \ - \ \sum_{j,j\not=i} 
  p_i p_j \right)
   \ = \ \beta \ p_i (1 \ - \ \sum_j p_j) \ = 0 \ .
\end{align}
\end{proof}

Next we show that $0$ is the smallest eigenvalue of $\rJ_s$, therefore
$\rJ_s$ is positive semi-definite but not (strict) positive definite. 
\begin{lemmaA}
\label{th:LjacobiDefinite}
The Jacobian  $\rJ_s$ of the softmax  $\Bp = \soft (\beta \Bxi)$
is symmetric and positive semi-definite.
\end{lemmaA}

\begin{proof}
For an arbitrary $\Bz$, we have
\begin{align}
  \Bz^T \left( \diag(\Bp) - \Bp \Bp^T \right) \Bz
  \ &= \ \sum_i p_i z_i^2 - \left( \sum_i p_i z_i \right)^2 \\\nonumber
  &= \  \left( \sum_i p_i z_i^2\right) \ \left( \sum_i p_i \right)
    - \left( \sum_i p_i z_i \right)^2  \ \geq \ 0 \ .
\end{align}
The last inequality hold true because the Cauchy-Schwarz inequality
says $(\Ba^T \Ba)(\Bb^T \Bb) \geq (\Ba^T \Bb)^2$, which is the last
inequality with $a_i = z_i \sqrt{p_i}$ and $b_i= \sqrt{p_i}$.
Consequently $\left( \diag(\Bp) - \Bp \Bp^T \right)$ is positive semi-definite.

Alternatively $\sum_i p_i z_i^2 - \left( \sum_i p_i z_i \right)^2$ can
be viewed as the expected second moment minus the mean squared which
gives the variance that is larger equal to zero.

The Jacobian is $0<\beta$ times a positive semi-definite matrix, which is
a positive semi-definite matrix.
\end{proof}

Moreover, the softmax is a monotonic map, as described in the next lemma. 
\begin{lemmaA}
\label{th:LsoftMonotone}
  The softmax $\soft (\beta \Bx)$ is monotone for $\beta >0$, that is,
\begin{align}
  \left( \soft(\beta \Bx) \ - \ \soft(\beta \Bx') \right)^T \left( \Bx \ - \
    \Bx'\right)\ &\geq \ 0 \ .
\end{align}
\end{lemmaA}
\begin{proof}
We use the version of mean value theorem Lemma~\ref{th:MVT} with the symmetric matrix
$\rJ_s^m = \int_{0}^1 \rJ_s(\lambda \Bx \ + \ (1-\lambda) \Bx') \ \Rd \lambda$:
\begin{align}
  \soft(\Bx) \ - \ \soft(\Bx') \ &= \ 
  \rJ_s^m \ \left( \Bx \ - \ \Bx' \right) \ .
\end{align}
Therefore
\begin{align}
  &\left( \soft(\Bx) \ - \ \soft(\Bx') \right)^T \left( \Bx \ - \
    \Bx'\right) \ = \
    \left( \Bx \ - \ \Bx' \right)^T \rJ_s^m \ \left( \Bx \ - \ \Bx' \right) 
    \ \geq \ 0 \ ,
\end{align}
since $\rJ_s^m$ is positive semi-definite.
For all $\lambda$ the Jacobians 
$\rJ_s(\lambda \Bx \ + \ (1-\lambda) \Bx')$
are positive semi-definite 
according to Lemma~\ref{th:LjacobiDefinite}. 
Since 
\begin{align}
 \Bx^T \rJ_s^m \Bx \ &= \ \int_{0}^1 \Bx^T \rJ_s(\lambda \Bx \ + \ (1-\lambda) \Bx') \ \Bx \ \Rd \lambda
 \ \geq \ 0  
\end{align}
is an integral over positive values for every $\Bx$, 
$\rJ_s^m$ is positive semi-definite, too. 
\end{proof}

Next we give upper bounds on the norm of $\rJ_s$.
\begin{lemmaA}
\label{th:JacobiB}
For a softmax $\Bp =  \soft ( \beta \Bx)$ with
$m= \max_i p_i (1-p_i)$, the spectral norm of
the Jacobian $\rJ_s$ of the softmax is bounded:
\begin{align} \label{eq:softjacobi2-m}
 \NRM{\rJ_s}_2 \ &\leq \ 2 \ m \ \beta \ ,  \\ \label{eq:softjacobi1-m}
 \NRM{\rJ_s}_1 \ &\leq \ 2 \ m \ \beta \ , \\ \label{eq:softjacobiInf-m}
 \NRM{\rJ_s}_{\infty}  \ &\leq \ 2 \ m  \ \beta \ .
\end{align}
In particular everywhere holds
\begin{align} \label{eq:softjacobi2-beta}
 \NRM{\rJ_s}_2 \ &\leq \ \frac{1}{2} \ \beta \ .
\end{align}
If $p_{\max}=\max_i p_i \geq 1-\epsilon \geq 0.5$, then for the spectral norm of
the Jacobian holds
\begin{align}  \label{eq:softjacobi2-eps}
 \NRM{\rJ_s}_2 \ &\leq \  2 \ \epsilon \  \beta \ 
 \ - \  2 \ \epsilon^2 \ \beta \ 
 \ < \ 2 \ \epsilon  \ \beta \ .
\end{align}
\end{lemmaA}

\begin{proof}
We consider the maximum absolute column sum norm
\begin{align}
 \NRM{\BA}_1 \ &= \ \max_j \sum_i \ABS{a_{ij}}  
\end{align}
and the maximum absolute row sum norm
\begin{align}
 \NRM{\BA}_{\infty} \ &= \ \max_i \sum_j \ABS{a_{ij}}  \ .
\end{align}

We have for $\BA =  \rJ_s = \beta \left( \diag(\Bp) - \Bp \Bp^T \right)$
\begin{align}
 \sum_j \ABS{a_{ij}}  \ &= \ \beta \ \left( p_i (1-p_i) \ + \ \sum_{j,j\not=i} p_i
 p_j \right) \ = \ \beta \ p_i  \ ( 1 \ - \ 2 p_i \ + \ \sum_{j} p_j ) \\ \nonumber 
 &= \ 2
 \ \beta \ p_i \ (1-p_i) \ \leq \ 2 \ m \ \beta \ ,\\
 \sum_i \ABS{a_{ij}}  \ &= \ \beta \  \left( p_j \ (1-p_j) \ + \ \sum_{i,i\not=j} p_j
 p_i \right) \ = \ \beta \ p_j \ ( 1 \ - \ 2 p_j \ + \ \sum_{i} p_i ) \\ \nonumber 
 &= \ 2
 \ \beta \ p_j \ (1-p_j) \ \leq \ 2 \ m \ \beta \ .
\end{align}
Therefore, we have
\begin{align}
 \NRM{\rJ_s}_1 \ &\leq \ 2 \ m \ \beta \ , \\
 \NRM{\rJ_s}_{\infty}  \ &\leq \ 2 \ m  \ \beta\ ,\\
 \NRM{\rJ_s}_2 \ &\leq \ \sqrt{\NRM{\rJ_s}_1 \NRM{\rJ_s}_{\infty} } \ \leq \
 2 \ m \ \beta \ .
\end{align}
The last inequality is a direct consequence of  H\"{o}lder's inequality. 

For $0 \leq p_i \leq 1$, we have $p_i(1-p_i) \leq 0.25$. 
Therefore, $m \leq 0.25$ for all values of $p_i$.

If $p_{\max} \geq 1-\epsilon \geq 0.5$ ($\epsilon \leq 0.5$), then
$1-p_{\max} \leq \epsilon$ and for $p_i\not= p_{\max}$
$p_i \leq \epsilon$.
The derivative $\partial x (1-x) / \partial x = 1 - 2 x > 0$ for $x < 0.5$, therefore
$x (1-x)$ increases with $x$ for $x < 0.5$.
Using $x=1-p_{\max}$ and for $p_i\not= p_{\max}$
$x= p_i$, we obtain
$p_i (1-p_i) \leq \epsilon (1-\epsilon)$ for all $i$.
Consequently, we have $m \leq \epsilon (1-\epsilon)$.
\end{proof}

Using the bounds on the norm of the Jacobian,
we give some Lipschitz properties of the softmax function. 
\begin{lemmaA}
\label{th:LsoftLipschitz}
The softmax function $\Bp =  \soft ( \beta \Bx)$
is $(\beta/2)$-Lipschitz.
The softmax function $\Bp =  \soft ( \beta \Bx)$
is $(2 \beta m)$-Lipschitz 
in a convex environment $U$ for which 
$m= \max_{\Bx \in U} \max_i p_i (1-p_i)$.
For $p_{\max}= \min_{\Bx \in U} \max_i p_i= 1-\epsilon$,
the softmax function $\Bp =  \soft ( \beta \Bx)$
is $(2 \beta \epsilon)$-Lipschitz.
For $\beta<2 m$, the softmax $\Bp =  \soft ( \beta \Bx)$
is contractive in $U$ on which $m$ is defined.
\end{lemmaA}
\begin{proof}
The version of mean value theorem Lemma~\ref{th:MVT} states for the symmetric matrix
$\rJ_s^m = \int_{0}^1 \rJ(\lambda \Bx + (1-\lambda)\Bx') \ \Rd \lambda$:
\begin{align}
  \soft(\Bx) \ - \ \soft(\Bx') \ &= \ 
  \rJ_s^m \ \left( \Bx \ - \ \Bx' \right) \ .
\end{align}
According to Lemma~\ref{th:JacobiB} 
for all $\tilde{\Bx}= \lambda \Bx + (1-\lambda)\Bx') $
\begin{align}
  \NRM{\rJ_s(\tilde{\Bx})}_2 \ \leq \  2 \ \tilde{m} \ \beta  \ ,
\end{align}
where $\tilde{m}=\max_i \tilde{p}_i (1-\tilde{p}_i)$.
Since $\Bx \in U$ and $\Bx' \in U$ we have $\tilde{\Bx} \in U$,
since $U$ is convex.
For $m= \max_{\Bx \in U} \max_i p_i (1-p_i)$ we have
$\tilde{m} \leq m$ for all $\tilde{m}$. Therefore, we have
\begin{align}
  \NRM{\rJ_s(\tilde{\Bx})}_2 \ \leq \  2 \ m \ \beta  
\end{align}
which also holds for the mean:
\begin{align}
  \NRM{\rJ_s^m}_2 \ \leq \  2 \ m \ \beta  \ .
\end{align}
Therefore, 
\begin{align}
  \NRM{\soft(\Bx) \ - \ \soft(\Bx')} \ &\leq \ 
  \NRM{\rJ_s^m}_2 \ \NRM{ \Bx \ - \ \Bx'}
  \ \leq \ 2 \ m \ \beta \ \NRM{ \Bx \ - \ \Bx'} \ .
\end{align}
From Lemma~\ref{th:JacobiB} we know $m \leq 1/4$ globally. 
For  $p_{\max}=\min_{\Bx \in U} \max_i p_i= 1-\epsilon$ we have
according to Lemma~\ref{th:JacobiB}: $m \leq \epsilon$.
\end{proof}

For completeness we present a result about cocoercivity of the softmax:
\begin{lemmaA}
\label{th:LsoftCocoercive}
 For $m= \max_{\Bx \in U} \max_i p_i (1-p_i)$, 
 softmax function $\Bp =  \soft ( \beta \Bx)$
 is $1/(2 m \beta)$-cocoercive in $U$, that is,
\begin{align}
  \left( \soft(\Bx) \ - \ \soft(\Bx') \right)^T \left( \Bx \ - \
    \Bx'\right)\ &\geq \ \frac{1}{2 \ m \ \beta} \NRM{\soft(\Bx) \ - \
    \soft(\Bx')}  .
\end{align}
In particular
the softmax function $\Bp =  \soft ( \beta \Bx)$
is $(2/\beta)$-cocoercive everywhere.
With $p_{\max}=\min_{\Bx \in U} \max_i p_i= 1-\epsilon$,
the softmax function $\Bp =  \soft ( \beta \Bx)$
is $1/(2 \beta \epsilon)$-cocoercive in $U$.
\end{lemmaA}
\begin{proof}
   We apply the Baillon-Haddad theorem (e.g.\ Theorem~1 in \citet{Gao:17})
   together with Lemma~\ref{th:LsoftLipschitz}. 
\end{proof}

Finally, we introduce the Legendre transform and use it
to describe further properties of the $\mathrm{lse}$.
We start with the definition of the convex conjugate.
\begin{definitionA}[Convex Conjugate]
  The {\em Convex Conjugate (Legendre-Fenchel transform)}
  of a function $f$ from a Hilbert Space $X$ to $[-\infty, \infty]$ is $f^*$ which is defined as
\begin{align}
  f^*(\Bx^*) \ &= \  \sup_{\Bx \in X} ( \Bx^T \Bx^* \ - \  f(\Bx)) \ , \quad
  \Bx^* \in X
\end{align}
\end{definitionA}
See page 219 Def.~13.1 in \citet{Bauschke:17} and page 134 in \citet{Garling:17}.
Next we define the Legendre transform, which is a more restrictive version of the convex
conjugate.
\begin{definitionA}[Legendre Transform]
The {\em Legendre transform} of a convex function $f$ from a convex set
$X \subset \dR^n$ to $\dR$ ($f: X \rightarrow \dR$) is $f^*$, which is
defined as
\begin{align}
  f^*(\Bx^*) \ &= \  \sup_{\Bx \in X} ( \Bx^T \Bx^* \ - \  f(\Bx)) \ , \quad
  \Bx^* \in X^* \ , \\
  X^* \ &= \ \left\{\Bx^* \in \dR^n \mid 
  \sup_{\Bx \in X} ( \Bx^T \Bx^* \ - \  f(\Bx)) < \infty \right\} \ .
\end{align}
\end{definitionA}
See page 91 in \citet{Boyd:09}.

\begin{definitionA}[Epi-Sum]
Let $f$ and $g$ be two functions from $X$ to $(-\infty, \infty]$, then the infimal convolution (or epi-sum) of $f$ and $g$ is
\begin{align}
    f \Box g: X \rightarrow [-\infty, \infty]\ ,\ \Bx \mapsto \inf_{\By \in X} \left( f(\By) + g(\Bx-\By) \right)
\end{align}
\end{definitionA}
See Def.~12.1 in \citet{Bauschke:17}.

\begin{lemmaA}
Let $f$ and $g$ be functions from $X$ to $(-\infty,\infty]$. Then
the following hold:
\begin{enumerate}
\item Convex Conjugate of norm squared
\begin{align}
  \left( \frac{1}{2} \NRM{.}^2 \right)^* \ &= \ \frac{1}{2} \NRM{.}^2 \ .
\end{align}
\item Convex Conjugate of a function multiplied by scalar $0<\alpha \in \dR$
\begin{align}
  \left( \alpha \ f \right)^* \ &= \ \alpha \ f^*(./ \alpha )\ .
\end{align}
\item Convex Conjugate of the sum of a function and a scalar $\beta \in \dR$
\begin{align}
  \left( f \ + \ \beta \right)^* \ &=  \ f^* \ - \ \beta \ .
\end{align}
\item Convex Conjugate of affine transformation of the arguments. Let $\BA$ be 
a non-singular matrix and $\Bb$ a vector
\begin{align}
 \left( f\left( \BA \Bx \ + \ \Bb \right)\right)^* \ &= \  
  f^* \left(\BA^{-T} \Bx^* \right)  \ - \ \Bb^T \BA^{-T} \Bx^* \ .
\end{align}
\item Convex Conjugate of epi-sums
\begin{align}
  \left(f \Box g\right)^* \ &= \ f^* + g^* \ .
\end{align}
\end{enumerate}
\end{lemmaA}

\begin{proof}
\begin{enumerate}
    \item Since $h(t):= \frac{t^2}{2}$ is a non-negative convex function and $h(t) = 0 \iff t= 0$ we have because of Proposition 11.3.3 in \citet{Garling:17} that $h\left( \NRM{x} \right)^* = h^* \left( \NRM{x^*} \right) $. Additionally, by example (a) on page 137 we get for $1 < p < \infty$ and $\frac{1}{p} + \frac{1}{q} = 1$ that $\left( \frac{|t|^p}{p} \right)^* = \frac{|t^*|^q}{q}$. Putting all together we get the desired result. The same result can also be deduced from page 222 Example~13.6 in \citet{Bauschke:17}.
    \item Follows immediately from the definition since
    \begin{align*}
        \alpha f^* \left( \frac{\Bx^*}{\alpha} \right) = \alpha \sup_{\Bx \in X} \left( \Bx^T \frac{\Bx^*}{\alpha} \ - \  f(\Bx) \right) = \sup_{\Bx \in X} (\Bx^T \Bx^* - \alpha f(\Bx)) = (\alpha f)^*(\Bx^*)
    \end{align*}
    \item $(f + \beta)^* := \sup_{\Bx \in X} \left( \Bx^T \Bx^* - f(\Bx) - \beta  \right) 
    =: f^* - \beta$
    \item
    \begin{align*}
        \left( f\left( \BA \Bx + \Bb \right) \right)^*(\Bx^*) &= \sup_{\Bx \in X} \left( \Bx^T \Bx^* - f\left( \BA \Bx + \Bb \right) \right) \\
        &= \sup_{\Bx \in X} \left( \left( \BA \Bx + \Bb \right)^T \BA^{-T} \Bx^* - f\left(\BA \Bx + \Bb\right)  \right)- \Bb^T \BA^{-T} \Bx^* \\
        &= \sup_{\By \in X} \left( \By^T \BA^{-T} \Bx^* - f\left( \By \right)  \right)- \Bb^T \BA^{-T} \Bx^* \\
        &= f^* \left(\BA^{-T} \Bx^* \right)   -  \Bb^T \BA^{-T} \Bx^*
    \end{align*}
    \item From Proposition~13.24 (i) in \citet{Bauschke:17} and Proposition~11.4.2 in \citet{Garling:17} we get
    \begin{align*}
        \left( f \Box g \right)^*(\Bx^*) &=
        \sup_{\Bx \in X} \left( \Bx^T \Bx^* - \inf_{\By \in X} \left( f(\By) - g(\Bx-\By) \right) \right)\\
        &= \sup_{\Bx,\By \in X} \left( \Bx^T \Bx^* - f(\By) - g(\Bx-\By)  \right)\\
        &= \sup_{\Bx,\By \in X} \left( \left(\By^T \Bx^* - f(\By) \right) + \left( \left( \Bx - \By \right)^T \Bx^* - g(\Bx-\By) \right) \right)\\
        &= f^*(\Bx^*) + g^*(\Bx^*)
    \end{align*}
\end{enumerate}
\end{proof}

\begin{lemmaA}
\label{th:legendreLSE}
The Legendre transform of the $\mathrm{lse}$ is the 
negative entropy function, restricted to the probability simplex
and vice versa.
For the log-sum exponential
\begin{align}
  f(\Bx) \ &= \ \ln \left( \sum_{i=1}^n \exp(x_i) \right) \ ,
\end{align}
the Legendre transform is the 
negative entropy function, restricted to the probability simplex:
\begin{align}
  f^*(\Bx^*) \ &= \ 
  \begin{cases}
  \sum_{i=1}^n x_i^* \ln(x^*_i) & \text{ for } \ 0 \leq x^*_i \ \text{ and } \ \sum_{i=1}^n x^*_i = 1 \\
  \infty & \text{ otherwise }
  \end{cases} \ .
\end{align}
For the negative entropy function, restricted to the probability simplex:
\begin{align}
  f(\Bx) \ &= \ 
  \begin{cases}
  \sum_{i=1}^n x_i \ln(x_i) & \text{ for } \ 0 \leq x_i \ \text{ and } \ \sum_{i=1}^n x_i = 1 \\
  \infty & \text{ otherwise }
  \end{cases} \ .
\end{align}
the Legendre transform is the 
log-sum exponential
\begin{align}
  f^*(\Bx^*) \ &= \ \ln \left( \sum_{i=1}^n \exp(x_i^*) \right) \ ,
\end{align}
\end{lemmaA}

\begin{proof}
See page 93 Example~3.25 in \citet{Boyd:09}
and \citep{Gao:17}.
If $f$ is a regular convex function (lower semi-continuous convex function), 
then $f^{**}=f$ according to
page 135 Exercise~11.2.3 in \citet{Garling:17}.
If $f$ is lower semi-continuous and convex, then $f^{**}=f$ according to
Theorem~13.37 (Fenchel-Moreau) in \citet{Bauschke:17}.
The log-sum-exponential is continuous and convex.
\end{proof}
 
\begin{lemmaA}
Let $\BX \BX^T$ be non-singular and $X$ a Hilbert space. 
We define 
\begin{align}
 X^* \ = \  \left\{ \Ba \mid 
 0 \ \leq \ \BX^T \left(\BX \BX^T \right)^{-1} \Ba \ , \ \
 \BOn^T \BX^T \left(\BX \BX^T \right)^{-1} \Ba \ = \ 1 \right\} \ .
\end{align}
and
\begin{align}
 X^v \ = \  \left\{ \Ba \mid \Ba =\BX^T \Bxi \ ,  \  \ \Bxi \in X
  \right\} \ .
\end{align}

The Legendre transform of $\mathrm{lse}(\beta, \BX^T \Bxi)$
with $\Bxi \in X$ is
\begin{align}
 \left(\mathrm{lse}(\beta, \BX^T \Bxi) \right)^* (\Bxi^*)
 \ &= \ 
 \left(\mathrm{lse}(\beta, \Bv) \right)^* \left( \BX^T \left(\BX \BX^T \right)^{-1} \Bxi^* \right) \ ,
\end{align}
with $\Bxi^* \in X^*$ and $\Bv \in X^v$.
The domain of $\left(\mathrm{lse}(\beta, \BX^T \Bxi) \right)^*$ is $X^*$.

Furthermore we have
\begin{align}
 \left(\mathrm{lse}(\beta, \BX^T \Bxi) \right)^{**}
 \ &= \ \mathrm{lse}(\beta, \BX^T \Bxi) \ .
\end{align}

\end{lemmaA}

\begin{proof}
We use the definition of the Legendre transform:
\begin{align}
 &\left(\mathrm{lse}(\beta, \BX^T \Bxi) \right)^* (\Bxi^*)
 \ = \ \sup_{\Bxi \in X} \Bxi^T \Bxi^* \ - \  \mathrm{lse}(\beta, \BX^T \Bxi)  \\ \nonumber
 &= \ 
 \sup_{\Bxi \in X} \left(\BX^T \Bxi \right)^T \BX^T \left(\BX \BX^T \right)^{-1} \Bxi^* \ - \  \mathrm{lse}(\beta, \BX^T \Bxi) \\ \nonumber
 &= \ \sup_{\Bv \in X^v} \Bv^T \BX^T \left(\BX \BX^T \right)^{-1} \Bxi^* \ - \  \mathrm{lse}(\beta, \Bv) \\ \nonumber
 &= \ \sup_{\Bv \in X^v} \Bv^T \Bv^* \ - \  \mathrm{lse}(\beta, \Bv) \\ \nonumber
 &= \ \left(\mathrm{lse}(\beta, \Bv) \right)^* ( \Bv^*) \ = \  
 \left(\mathrm{lse}(\beta, \Bv) \right)^* \left( \BX^T \left(\BX \BX^T \right)^{-1} \Bxi^* \right) \ ,
\end{align}
where we used
$\Bv^*=\BX^T \left(\BX \BX^T \right)^{-1} \Bxi^*$.

According to page 93 Example~3.25 in \citet{Boyd:09}, 
the equations for the maximum $\max_{\Bv \in X^v} \Bv^T \Bv^* \ - \  \mathrm{lse}(\beta, \Bv)$
are solvable if and only if $0 < \Bv^*=\BX^T \left(\BX \BX^T \right)^{-1} \Bxi^*$ and
$\BOn^T \Bv^* = \BOn^T \BX^T \left(\BX \BX^T \right)^{-1} \Bxi^* = 1$.
Therefore, we assumed $\Bxi^* \in X^*$.

The domain of $\left(\mathrm{lse}(\beta, \BX^T \Bxi) \right)^*$ is $X^*$, since
on page 93 Example~3.25 in \citet{Boyd:09} it was shown that outside $X^*$
the $\sup_{\Bv \in X^v} \Bv^T \Bv^* \ - \  \mathrm{lse}(\beta, \Bv)$ is not bounded.

Using
\begin{align}
  \Bp \ &= \   \soft ( \beta \BX^T \Bxi) \ ,
\end{align}
the Hessian of $\mathrm{lse}(\beta, \BX^T \Bxi)$
\begin{align}
  \frac{\partial^2 \mathrm{lse}(\beta, \BX^T \Bxi)}{\partial \Bxi^2}
  \ &= \   \beta \ \BX \left( \diag(\Bp) - \Bp \Bp^T \right) \BX^T
\end{align}
is positive semi-definite since $\diag(\Bp) - \Bp \Bp^T$ is positive semi-definite
according to Lemma~\ref{th:LjacobiDefinite}.
Therefore, $\mathrm{lse}(\beta, \BX^T \Bxi)$ is convex and continuous.

If $f$ is a regular convex function (lower semi-continuous convex function), 
then $f^{**}=f$ according to
page 135 Exercise~11.2.3 in \citet{Garling:17}.
If $f$ is lower semi-continuous and convex, then $f^{**}=f$ according to
Theorem~13.37 (Fenchel-Moreau) in \citet{Bauschke:17}.
Consequently we have
\begin{align}
 \left(\mathrm{lse}(\beta, \BX^T \Bxi) \right)^{**}
 \ &= \ \mathrm{lse}(\beta, \BX^T \Bxi) \ .
\end{align}

\end{proof}

We introduce the Lambert $W$ function and some of its properties, since it is
needed to derive bounds on the storage capacity of our new Hopfield networks.
\begin{definitionA}[Lambert Function]
\label{th:lambert}
The {\em Lambert $W$ function}  \dlmf{4.13} is the inverse function of
\begin{align}
 f(y) \ &= \ y e^{y} \ .
\end{align}
The Lambert $W$ function has an upper branch $W_0$ for $-1 \leq y$ and a lower
branch $W_{-1}$ for $y \leq -1$.
We use $W$ if a formula holds for both branches.
We have
\begin{align}
 W(x) \ &= \ y \ \Rightarrow  y e^{y} \ = \ x \ .
\end{align}
\end{definitionA}

We present some identities for the Lambert $W$ function \dlmf{4.13}:
\begin{lemmaA}
\label{th:lambertId}
Identities for the Lambert $W$ function are
\begin{align}
  W(x) \ e^{W(x)} \ &= \ x \ ,\\
  W(x e^x) \ &= \ x \ , \\
  e^{W(x)} \ &= \ \frac{x}{W(x)} \ , \\
  e^{-W(x)} \ &= \ \frac{W(x)}{x} \ , \\
  e^{n W(x)} \ &= \ \left(\frac{x}{W(x)}\right)^n \ , \\
  W_0\left(x \ \ln x\right) \ &= \ \ln x \quad \text{for } \ 
     x \ \geq \ \frac{1}{e} \ , \\ 
  W_{-1}\left(x \ \ln x\right) \ &= \ \ln x \quad \text{for } \ 
     x \ \leq \ \frac{1}{e} \ , \\
  W(x) \ &= \ \ln \frac{x}{W(x)} \quad \text{for } \ 
     x \ \geq \ - \ \frac{1}{e} \ , \\
  W\left( \frac{n \ x^n}{W\left(x\right)^{n-1}} \right) \ &= \ n \ W(x) \quad 
  \text{for } \ n, x \ > \ 0 \ , \\
  W(x) \ + \ W(y) \ &= \ W\left(x \ y \ \left(\frac{1}{W(x)} \ + \ \frac{1}{W(y)}\right)\right) 
     \quad \text{for } \ x, y \ > \ 0  \ , \\
  W_0\left(- \ \frac{\ln x}{x}\right) \ &= \ - \ \ln x  \quad \text{for }\ 
   0 \ < \ x \ \leq \ e  \ , \\
  W_{-1}\left(- \ \frac{\ln x}{x}\right) \ &= \ - \ \ln x \quad \text{for } \ 
    x \ > \ e \ , \\
  e^{-\ W(- \ \ln x)} \ &= \ \frac{W(- \ \ln x)}{- \ \ln x} \quad \text{for } \ 
    x \ \neq \ 1 \ .
\end{align}
\end{lemmaA}

We also present some special values for the Lambert $W$ function \dlmf{4.13}:
\begin{lemmaA}
\label{th:lambertVal}
\begin{align}
W(0) \ &= \  0\ , \\
W(e)  \ &= \  1\ , \\
W\left(-\frac{1}{e}\right)  \ &= \  -1\ , \\
W\left(e^{1+e}\right)  \ &= \  e\ , \\
W\left(2 \ln 2 \right)  \ &= \  \ln 2\ , \\
W(1)  \ &= \  \Omega \ , \\
W(1)  \ &= \ e^{-W(1)} \ = \ \ln\left(\frac{1}{W(1)}\right) \ = \ - \ \ln W(1)\ , \\
W\left(-\frac{\pi}{2}\right)  \ &= \  \frac{i\pi}{2}\ , \\
W(-1) \ &\approx \ -0.31813+1.33723i \ ,
\end{align}
where the Omega constant $\Omega$ is
\begin{align}
  \Omega \ &= \ \left(\int_{-\infty}^{\infty} \frac{\Rd t}{\left(e^t \ - \ t\right)^2 \ + \ \pi^2}\right)^{-1}
   \  - \ 1 \ \approx \ 0.56714329 \ .
\end{align}
\end{lemmaA}

We need in some proofs a version of the mean value theorem as
given in the next lemma.
\begin{lemmaA}[Mean Value Theorem]
\label{th:MVT}
Let $U \subset \dR^n$ be open, $f: U \to \dR^m$  continuously differentiable, and
$\Bx \in U$ as well as $\Bh \in \dR^n$ vectors such that the line segment 
$\Bx+ t \Bh$ for $0 \leq t \leq 1$ is in $U$. Then the following holds:
\begin{align}
   f(\Bx \ + \ \Bh) \ - \ f(\Bx) \ &= \  \left( \int_{0}^{1} J(\Bx \ + \ t \ \Bh) \ \Rd t \right) \ \Bh \ ,
\end{align}
where $J$ is the Jacobian of $f$ and the integral of the matrix is component-wise.
\end{lemmaA}

\begin{proof}
Let $f_1,\ldots,f_m$ denote the components of $f$ and define
$g_i: [0,1] \to \dR$ by
\begin{align}
  g_i(t) \ &= \ f_i(\Bx \ + \ t \ \Bh ) \ ,
\end{align}
then we obtain
\begin{align}
  &f_i(\Bx \ + \  \Bh ) \ - \ f_i(\Bx) \ = \ g_i(1) \ - \ g_i(0) \ = \ \int_{0}^{1} g'(t) \ \Rd t \\ \nonumber
 &\int_{0}^{1} \left( \sum_{j=1}^n \frac{\partial f_i}{\partial x_j}(\Bx \ + \ t  \ \Bh) \ h_j \right)  \ \Rd t
 \ = \ \sum_{j=1}^n \left( \int_{0}^{1}  \frac{\partial f_i}{\partial x_j}(\Bx \ + \ t  \ \Bh)  
 \ \Rd t \right) \  h_j   \ .
\end{align}
The statement follows since the Jacobian $J$ has as entries $\frac{\partial f_i}{\partial x_j}$.
\end{proof}

\subsection{Modern Hopfield Networks: Binary States (Krotov and Hopfield)}
\label{sec:Krotov}

\subsubsection{Modern Hopfield Networks: Introduction}
\label{sec:KrotovIntro}

\paragraph{Additional Memory and Attention for Neural Networks.}
\label{sec:KrotovNN}

Modern Hopfield networks may serve as additional memory for neural networks.
Different approaches have been suggested to equip neural networks with
an additional memory beyond recurrent connections.
The neural Turing machine (NTM) is a neural network equipped with an
external memory and an attention process \citep{Graves:14}.  
The NTM can write to the memory and can read from it.
A memory network \citep{Weston:14} consists of a memory together
with the components: 
(1) input feature map (converts the incoming input to the internal feature representation)
(2) generalization (updates old memories given the new input),
(3) output feature map (produces a new output),
(4) response (converts the output into the response format).
Memory networks are generalized to an end-to-end trained model, where
the $\arg\max$ memory call is
replaced by a differentiable $\soft$ \citep{Sukhbaatar:15,Sukhbaatar:15arxiv}.
Linear Memory Network use a linear autoencoder for sequences 
as a memory \citep{Carta:20}.

To enhance RNNs with additional associative memory like Hopfield networks 
have been proposed \citep{Ba:16,Ba:16arxiv}.
The associative memory stores hidden states of the RNN, retrieves 
stored states if they are similar to actual ones, and has a forgetting parameter.
The forgetting and storing parameters of the RNN associative memory 
have been 
generalized to learned matrices \citep{Zhang:17}. 
LSTMs with associative memory via Holographic Reduced Representations have
been proposed \citep{Danihelka:16}.

Recently most approaches to new memories are based on attention.
The neural Turing machine (NTM) is equipped with an
external memory and an attention process \citep{Graves:14}.  
End to end memory networks (EMN) make the attention scheme 
of memory networks \citep{Weston:14} differentiable
by replacing $\arg\max$ through a $\soft$ \citep{Sukhbaatar:15,Sukhbaatar:15arxiv}.
EMN with dot products became very popular and implement a key-value
attention \citep{Daniluk:17} for self-attention.
An enhancement of EMN is the transformer \citep{Vaswani:17,Vaswani:17arxiv}
and its extensions \citep{Dehghani:18}.
The transformer had great impact on the natural language processing
(NLP) community as new records in NLP benchmarks have been achieved
\citep{Vaswani:17,Vaswani:17arxiv}.
MEMO uses the transformer
attention mechanism for reasoning over longer distances \citep{Banino:20}.
Current  state-of-the-art for language processing is
a transformer architecture called
``the Bidirectional Encoder Representations from Transformers''
(BERT) \citep{Devlin:18,Devlin:19}.

\paragraph{Modern Hopfield networks: Overview.}
\label{sec:KrotovOverview}

The storage capacity of classical binary Hopfield networks \citep{Hopfield:82}
has been shown to be very limited.
In a $d$-dimensional space,
the standard Hopfield model can store $d$ uncorrelated patterns
without errors but only
$C d/\ln(d)$ random patterns with
$C<1/2$ for a fixed stable pattern or $C<1/4$ if all patterns
are stable \citep{McEliece:87}.
The same bound holds for nonlinear learning rules \citep{Mazza:93}.
Using tricks-of-trade and allowing
small retrieval errors, the storage capacity
is about $0.138 d$ \citep{Crisanti:86,Hertz:91,Torres:02}.
If the learning rule is not related to the Hebb rule then up to $d$ 
patterns can be stored \citep{Abu-Mostafa:85}.
Using Hopfield networks with non-zero diagonal matrices,
the storage can be 
increased to $C d \ln(d)$ \citep{Folli:17}.
In contrast to the storage capacity, the number of energy minima 
(spurious states, stable states) of Hopfield networks 
is exponentially in $d$ \citep{Tanaka:80,Bruck:90,Wainrib:13}.

Recent advances 
in the field of binary Hopfield networks \citep{Hopfield:82}
led to new properties of Hopfield networks.
The stability of spurious states or metastable states
was sensibly reduced by a Hamiltonian treatment for 
the new relativistic Hopfield model \citep{Barra:18}.
Recently the storage capacity of Hopfield networks
could be increased by new energy functions.
Interaction functions of the form $F(x)=x^n$ lead to storage capacity of
$\alpha_n d^{n-1}$, where $\alpha_n$ depends on the allowed error
probability \citep{Krotov:16,Krotov:18,Demircigil:17} 
(see \citep{Krotov:18} for the non-binary case).
Interaction functions of the form $F(x)=x^n$ lead to storage capacity of
$\alpha_n \frac{d^{n-1}}{c_n \ln d}$ for $c_n>2(2n-3)!!$ \citep{Demircigil:17}.

Interaction functions of the form $F(x)=\exp(x)$ 
lead to {\em exponential} storage capacity of
$2^{d/2}$ where all stored patterns are fixed points but the radius of
attraction vanishes \citep{Demircigil:17}.
It has been shown that the network converges with high probability 
after one update \citep{Demircigil:17}.

\subsubsection{Energy and Update Rule for Binary Modern Hopfield Networks}
\label{sec:KrotovEnergy}

We follow \citep{Demircigil:17} where the goal is to store a set of input data
$\Bx_1,\ldots,\Bx_N$
that are represented by the matrix
\begin{align}
  \BX \ &= \ \left( \Bx_1,\ldots,\Bx_N \right) \ .
\end{align}
The 
$\Bx_i$ is pattern with binary components
$x_{ij} \in\{-1, +1\}$ for all $i$ and $j$.
$\Bxi$ is the actual state of the units of the Hopfield model.
Krotov and Hopfield \citep{Krotov:16} defined the energy function $\rE$
with the interaction function $F$ that evaluates 
the dot product between patterns $\Bx_i$
and the actual state $\Bxi$:
\begin{align}
\label{eq:krotovEnergy}
  \rE \ &= \ - \ \sum_{i=1}^N F\left( \Bxi^T \Bx_i \right) 
\end{align}
with $F(a)=a^n$, where $n=2$ gives the energy function of the classical 
Hopfield network. This allows to store $\alpha_n d^{n-1}$ patterns \citep{Krotov:16}.
Krotov and Hopfield \citep{Krotov:16} suggested for minimizing this energy
an asynchronous updating dynamics $T=(T_j)$ for component $\xi_j$:
\begin{align}
\label{eq:KrotovUpdate}
  T_j(\Bxi) \ &:= \
  \sgn\Bigl[
  \sum\limits_{i=1}^N
  \bigl( 
  F\bigl( x_{ij} \ + \ \sum\limits_{l\neq j} x_{il} \ \xi_l \bigr)
  \ - \ 
  F\bigl( - \ x_{ij} \ + \ \sum\limits_{l \neq j} x_{il} \  \xi_l \bigr) 
  \bigr) 
  \Bigr]
\end{align}

While Krotov and Hopfield used $F(a)=a^n$,
Demircigil et al.\ \citep{Demircigil:17} went a step further and analyzed
the model with the energy function $F(a)=\exp(a)$, which
leads to an exponential 
storage capacity of $N=2^{d/2}$. 
Furthermore with a single update the final pattern 
is recovered with high probability.
These statements are given in next theorem.
\begin{theoremA}[Storage Capacity for Binary Modern Hopfield Nets (Demircigil et al.\ 2017)]
\label{th:Demircigil}
Consider the generalized Hopfield model with the dynamics 
described in Eq.~\eqref{eq:KrotovUpdate} and 
interaction function $F$ given by $F(x)= e^{x}$.
For a fixed $0 < \alpha < \ln(2)/2$ 
let $N=\exp\left(\alpha d  \right)+1$ and let $\Bx_1, \ldots, \Bx_N$ 
be $N$ patterns chosen uniformly at random from $\{-1,+1\}^d$. 
Moreover fix $\varrho  \in [0, 1/2)$. 
For any $i$ and any $\widetilde{\Bx}_i$ taken uniformly at random 
from the Hamming sphere with radius $\varrho  d$ centered in $\Bx_i$, 
$\cS(\Bx_i,\varrho  d)$, where $\varrho  d$ is assumed to be an integer, 
it holds that
\begin{align}
\nonumber
  \PR \left( \exists i \; \exists j  : \  T_j\left(\widetilde{\Bx}_i \right) \ \neq \ 
  x_{ij}\right) \ \rightarrow \ 0 \ ,
\end{align}
if $\alpha$ is chosen in dependence of $\varrho$ such that
\begin{align}
\nonumber
  \alpha \ < \ \frac{I(1-2\varrho )}{2}
\end{align}
with 
\begin{align}
\nonumber
  I: \ a \ \mapsto \ \frac{1}{2}\left((1+a) \ln (1+a) \ + \ (1-a) \ln(1-a)\right) \ .
\end{align}
\end{theoremA}

\begin{proof}
The proof can be found in \citet{Demircigil:17}.
\end{proof}

The number of patterns $N=\exp\left(\alpha d  \right)+1$ is
exponential in the number $d$ of components.
The result
\begin{align}
\nonumber
  \PR \left( \exists i \; \exists j  : \  T_j\left(\widetilde{\Bx}_i \right) \ \neq \ 
  x_{ij}\right) \ \rightarrow \ 0 
\end{align}
means that one update for each component 
is sufficient to recover the pattern with high probability.
The constraint $\alpha < \frac{I(1-2\varrho )}{2}$ on $\alpha$ gives the 
trade-off between the radius of attraction $\varrho d$ and the number 
$N=\exp\left(\alpha d  \right)+1$ of
pattern that can be stored.

Theorem~\ref{th:Demircigil} in particular implies that
\begin{align}
\nonumber
  \PR \left( \exists i \; \exists j  : \
  T_j\left(\Bx_i \right) \ \neq \ x_{ij} \right) \ \rightarrow \ 0
\end{align}
as $d \rightarrow \infty$, i.e.\ 
with a probability converging to $1$, 
all the patterns are fixed points 
of the dynamics. 
In this case we can have
$\alpha \to \frac{I(1)}{2}=\ln(2)/2$.

Krotov and Hopfield define the update dynamics  $T_j(\Bxi)$
in Eq.~\eqref{eq:KrotovUpdate} via 
energy differences of the energy in Eq.~\eqref{eq:krotovEnergy}.
First we express the energy in Eq.~\eqref{eq:krotovEnergy} with
$F(a)=\exp(a)$ \citep{Demircigil:17}
by the $\mathrm{lse}$ function.
Then we use the mean value theorem to express
the update dynamics  $T_j(\Bxi)$
in Eq.~\eqref{eq:KrotovUpdate} by the softmax function.
For simplicity, we set $\beta=1$ in the following.
There exists a $v\in [-1,1]$ with
\begin{align}
 \nonumber
 T_j(\Bxi) \ &= \ \sgn\Bigl[
 - \ \rE(\xi_j=1) \ + \ \rE(\xi_j=-1) \Bigr]
 \ = \ 
 \sgn\Bigl[ \exp(\mathrm{lse}(\xi_j=1)) \ - \ \exp(\mathrm{lse}(\xi_j=-1)) \Bigr]\\
 \label{eq:KrotovSoftmax}
 &= \ 
 \sgn\Bigl[-\ (2 \Be_j)^T \nabla_{\Bxi} \rE(\xi_j=v)  \Bigr]\ = \ 
 \sgn\Bigl[\exp(\mathrm{lse}(\xi_j=v)) \ (2 \Be_j)^T\frac{\mathrm{lse}(\xi_j=v)}{\partial \Bxi}\Bigr] \\ \nonumber
 &= \ \sgn\Bigl[ \exp(\mathrm{lse}(\xi_j=1)) \ (2 \Be_j)^T \BX \soft (\BX^T \Bxi(\xi_j=v))  \Bigr]
 \\ \nonumber
 &= \ \sgn\Bigl[ [\BX \soft (\BX^T \Bxi(\xi_j=v))]_j  \Bigr] 
 \ = \ \sgn\Bigl[ [\BX \Bp(\xi_j=v) ]_j  \Bigr] \ ,
\end{align}
where $\Be_j$ is the Cartesian unit vector with a one at position $j$ and zeros elsewhere,
$[.]_j$ is the projection to the $j$-th component, and 
\begin{align}
  \Bp \ &= \ \soft (\BX^T \Bxi) \ .
\end{align}

\vspace{2cm} 

\subsection{Hopfield Update Rule is Attention of The Transformer}
\label{sec:Transformer}
\begin{figure}[ht]
        \centering
        \includegraphics[width=1.0\textwidth]{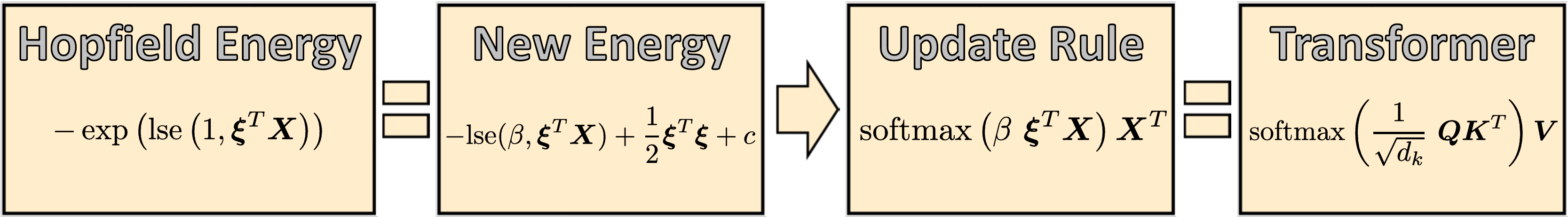}
        \caption[From binary Hopfield network to transformer]{We generalized
        the energy
        of binary modern Hopfield networks for allowing continuous states 
        while keeping fast convergence and
        storage capacity properties.
        We defined for the new energy also a new update
        rule that minimizes the energy. 
        The new update rule is the attention mechanism of the transformer.
        Formulae are modified to express $\soft$ as row vector as for transformers. 
        "$=$"-sign means "keeps the properties".
        \label{fig:AHopfieldToTransformer}}
\end{figure}

The Hopfield network update rule 
is the attention mechanism used
in transformer and BERT models
(see Fig.~\ref{fig:AHopfieldToTransformer}).
To see this, we assume $N$ stored (key) patterns $\By_i$ 
and $S$ state (query) patterns $\Br_i$ that are mapped to the
Hopfield space of dimension $d_k$.
We set $\Bx_i = \BW_K^T \By_i$, $\Bxi_i = \BW_Q^T \Br_i$,
and multiply the result of our update rule with $\BW_V$.
The matrices $\BY=(\By_1,\ldots,\By_N)^T$ and $\BR=(\Br_1,\ldots,\Br_S)^T$ combine the $\By_i$ and $\Br_i$ 
as row vectors.
We define the matrices $\BX^T=\BK = \BY \BW_K $, $\BXi^T = \BQ = \BR \BW_Q$,
and $\BV=\BY \BW_K \BW_V=\BX^T \BW_V$, where 
$\BW_K \in \dR^{d_y\times d_k}, \BW_Q \in \dR^{d_r\times d_k}, \BW_V \in \dR^{d_k\times d_v}$. 
If $\beta = 1/\sqrt{d_k}$ and $\soft \in \dR^N$ is changed to a row vector, we obtain
for the update rule Eq.~\eqref{eq:update} multiplied by $\BW_V$:
\begin{align}
 \label{eq:transformer_attention_detailed_appendix}
 \soft \left( 1/\sqrt{d_k} \ \BQ \ \BK^T \right) \ \BV \ = \soft \left( \beta \ \BR\BW_{\BQ} \ \BW_{\BK}^T\BY^T \right) \ \BY\BW_{\BK}\BW_{\BV} \ .
\end{align}
The left part of Eq.~\eqref{eq:transformer_attention_detailed_appendix} is the transformer attention.
Besides the attention mechanism,
Hopfield networks allow for other functionalities
in deep network architectures,
which we introduce via specific layers 
in the next section. The right part of Eq.~\eqref{eq:transformer_attention_detailed_appendix} serves as starting point for these specific layers.

\subsection{Experiments}

\subsubsection{Experiment 1: Attention in Transformers described by Hopfield dynamics} 
\paragraph{Analysis of operating modes of the heads of a pre-trained BERT model.}
We analyzed pre-trained BERT models from Hugging 
Face Inc.\ \citep{Wolf:19} according to these operating classes. 
In Fig.~\ref{fig:bert_analysis} in the appendix the distribution of the pre-trained 
bert-base-cased
model is depicted (for other models see 
appendix Section~\ref{sec:attention_learning_dynamics}).
Operating classes (II) (large metastable states) and 
(IV) (small metastable states) 
are often observed in the middle layers.
Operating class (I) (averaging over a very large number of patterns)
is abundant in lower layers.
Similar observations have been reported in other studies \citep{Toneva:19,Toneva:19arxiv,Tay:20}. 
Operating class (III) (medium metastable states) is predominant in the last layers.

\begin{figure}[htp]
    \centering
    \includegraphics[width=\textwidth]{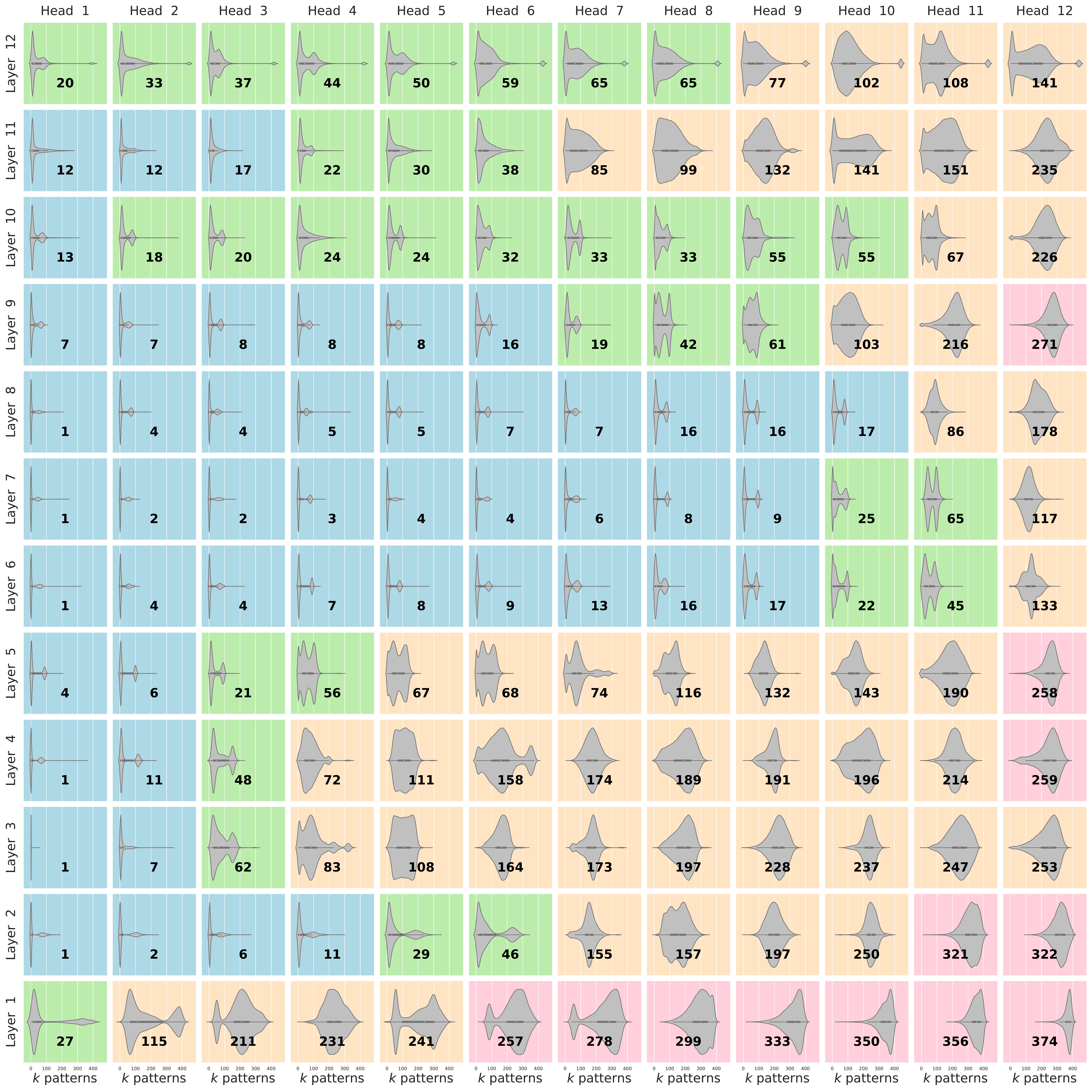}
    \caption[]{Analysis of operating modes of the heads of a pre-trained BERT model. 
     For each head in each layer, the distribution of the minimal number 
    $k$ of patterns required to sum up the $\soft$ values to $0.90$ is displayed as a violin plot in a panel. 
    $k$ indicates the size of a metastable state.
    The bold number in the center of each panel gives the median $\bar{k}$ of the distribution.
    The heads in each layer are sorted according to $\bar{k}$.
    Attention heads belong to the class they mainly operate in. 
    \textbf{Class (IV) in blue:} 
     Small metastable state or fixed point close to a single pattern, which 
     is abundant in the middle layers (6, 7, and 8).
    \textbf{Class (II) in orange:} Large metastable state, which is 
     prominent in middle layers (3, 4, and 5).
    \textbf{Class (I) in red:} Very large metastable state or global fixed point,
    which is predominant in the first layer. 
    These heads can potentially be replaced by averaging  operations. 
    \textbf{Class (III) in green:} Medium metastable state,
    which is frequently observed in higher layers.
    We hypothesize that these heads are used to collect information required to 
    perform the respective task. 
    These heads should be the main target to improve transformer and BERT models.
    }
    \label{fig:bert_analysis}
\end{figure}

\paragraph{Experimental Setup.}
\label{sec:experimental_setup}
Transformer architectures are known for their high computational demands. 
To investigate the learning dynamics of such a model and at the same time keeping training time manageable, we adopted the BERT-small setting from ELECTRA \citep{Clark:20}. It has $12$ layers, $4$ heads and 
a reduced hidden size, the sequence length is shortened from $512$ to $128$ tokens and the batch size is reduced from
$256$ to $128$. Additionally, the hidden dimension is reduced from $768$ to $256$ and the embedding dimension is
reduced from $768$ to $128$ \citep{Clark:20}. The training of such a BERT-small model for $1.45$ million update steps takes roughly four days on a single NVIDIA V100 GPU.

As the code base we use the \textit{transformers} repository from Hugging Face, Inc \citep{Wolf:19}.
We aim to reproduce the dataset of \citet{Devlin:19} as close as possible, which consists of
the English Wikipedia dataset and the Toronto BookCorpus dataset \citep{Zhu:15}. Due to recent copyright claims the later is not publicly available anymore. Therefore, the pre-training experiments use an uncased snapshot of the original BookCorpus dataset.

\paragraph{Hopfield Operating Classes of Transformer and BERT Models.}
\label{sec:operating_classes}

To better understand how operation modes in attention heads develop, we tracked the distribution of counts $k$ (see main paper) over time in a BERT-small model.
At the end of training we visualized the count distribution, grouped into four classes (see Figure~\ref{fig:count_small}). 
The thresholds for the classes were chosen according to the thresholds of Figure~2 in the main paper.
However, they are divided by a factor of $4$ to adapt to the shorter sequence length of $128$ compared to $512$.
From this plot it is clear, that the attention in heads of \textbf{Class IV} commit very early to the operating class of small metastable states. 

\paragraph{Learning Dynamics of Transformer and BERT Models.}
\label{sec:attention_learning_dynamics}

To observe this behavior in the early phase of training, we created a ridge plot of the distributions of counts $k$ for the 
first $20,000$ steps (see Figure~\ref{fig:fine_grained} (a)). This plot shows that the attention in heads of middle layers often 
change the operation mode to \textbf{Class IV} around $9,000$ to $10,000$ steps. At the same time the second big drop in the loss occurs.
The question arises whether this is functionally important or whether it is an artefact which could be even 
harmful. To check if the attention mechanism is still able to learn after the change in the operation mode 
we analyzed the gradient flow through the $\soft$ function.
For every token we calculate the Frobenius norm of the Jacobian of the $\soft$ over multiple samples.
Then, for every head we plot the distribution of the norm (see Figure~\ref{fig:fine_grained}(b)).
The gradients with respect to the weights are determined by the Jacobian $\rJ$ defined in Eq.~\eqref{eq:theJacobian}
as can be seen in Eq.~\eqref{eq:gradW}, Eq.~\eqref{eq:gradWQ}, and Eq.~\eqref{eq:gradWK}.
We can see that 
the attention
in heads of \textbf{Class IV} remain almost unchanged during the rest of the training.

 \begin{figure}[htp]
    \centering
    \includegraphics[width=0.9\textwidth]{figures/ridgeplots_full_and_violine.pdf}
    \caption[Ridge plots of the distribution of counts]{\textbf{Left}: Ridge plots of the distribution of counts $k$ over time for BERT-small \textbf{Right}: Violin plot of counts $k$ after $1,450000$ steps, divided into the four classes from the main paper. The thresholds were adapted to the shorter sequence length.}
    \label{fig:count_small}
\end{figure}

\begin{figure}
    \centering
    \begin{minipage}[t]{0.49\linewidth}
    \centering
    \includegraphics[width=.9\textwidth]{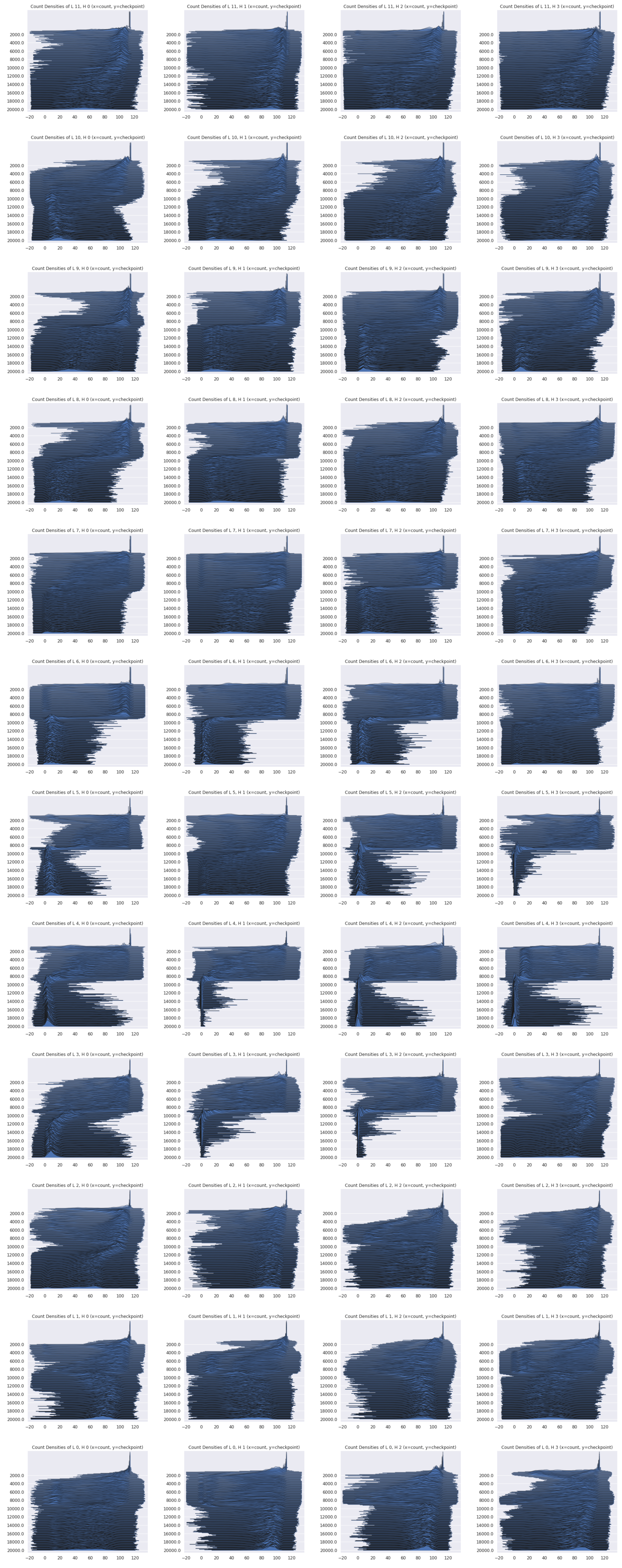} \\
    (a) Densities
    \end{minipage}
    \begin{minipage}[t]{0.49\linewidth}
    \centering
    \includegraphics[width=.9\textwidth]{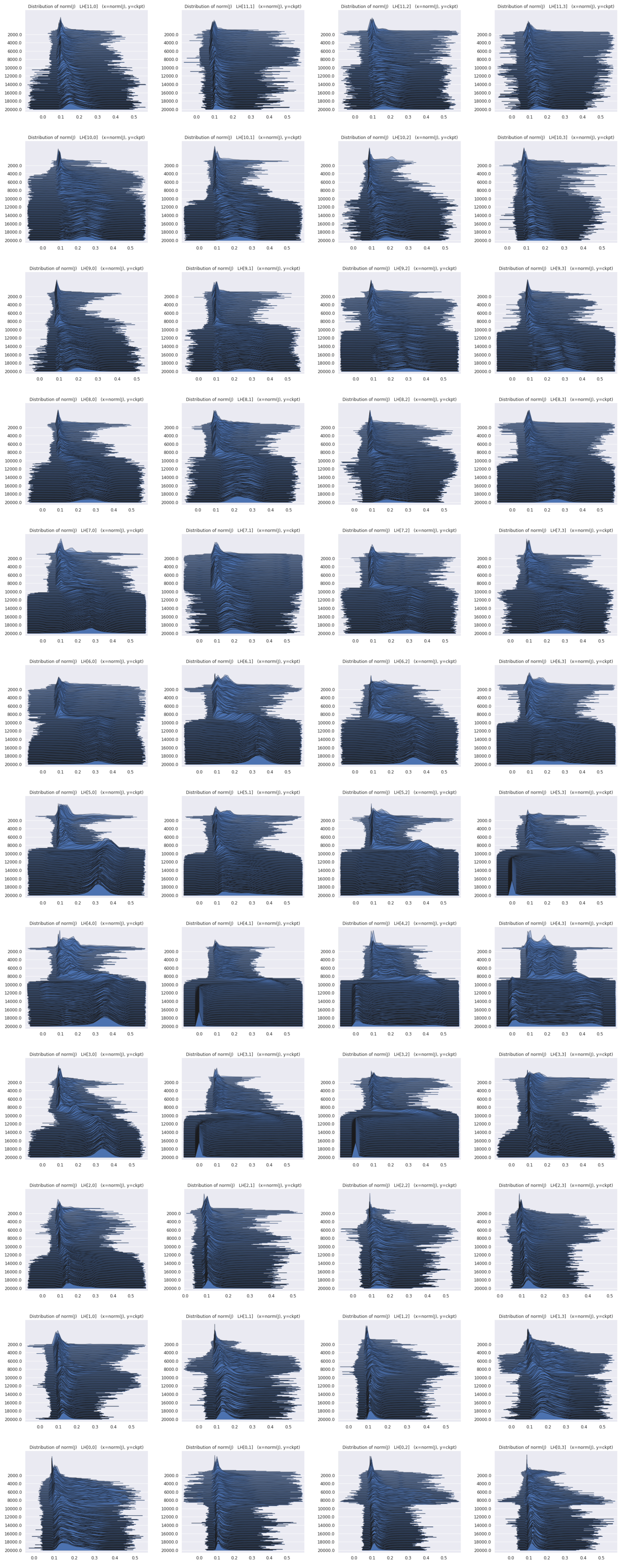} \\ 
    (b) Norm of Jacobian
    \end{minipage}
    \caption[Change of count density during training]{\textbf{(a)}: change of count density during training is depicted for the first $20,000$ steps.
    \textbf{(b)}: the corresponding distribution of the Frobenius norm of the Jacobian of the $\soft$ function
    is depicted. 
    The gradients with respect to the weights are determined by the Jacobian $\rJ$ defined in Eq.~\eqref{eq:theJacobian}
    as can be seen in Eq.~\eqref{eq:gradW}, Eq.~\eqref{eq:gradWQ}, and Eq.~\eqref{eq:gradWK}.
    }
    \label{fig:fine_grained}
\end{figure}

\paragraph{Attention Heads Replaced by Gaussian Averaging Layers.}
\label{sec:gaussian_layers}

\begin{figure}[htp]
    \centering
    \includegraphics[width=0.7\textwidth]{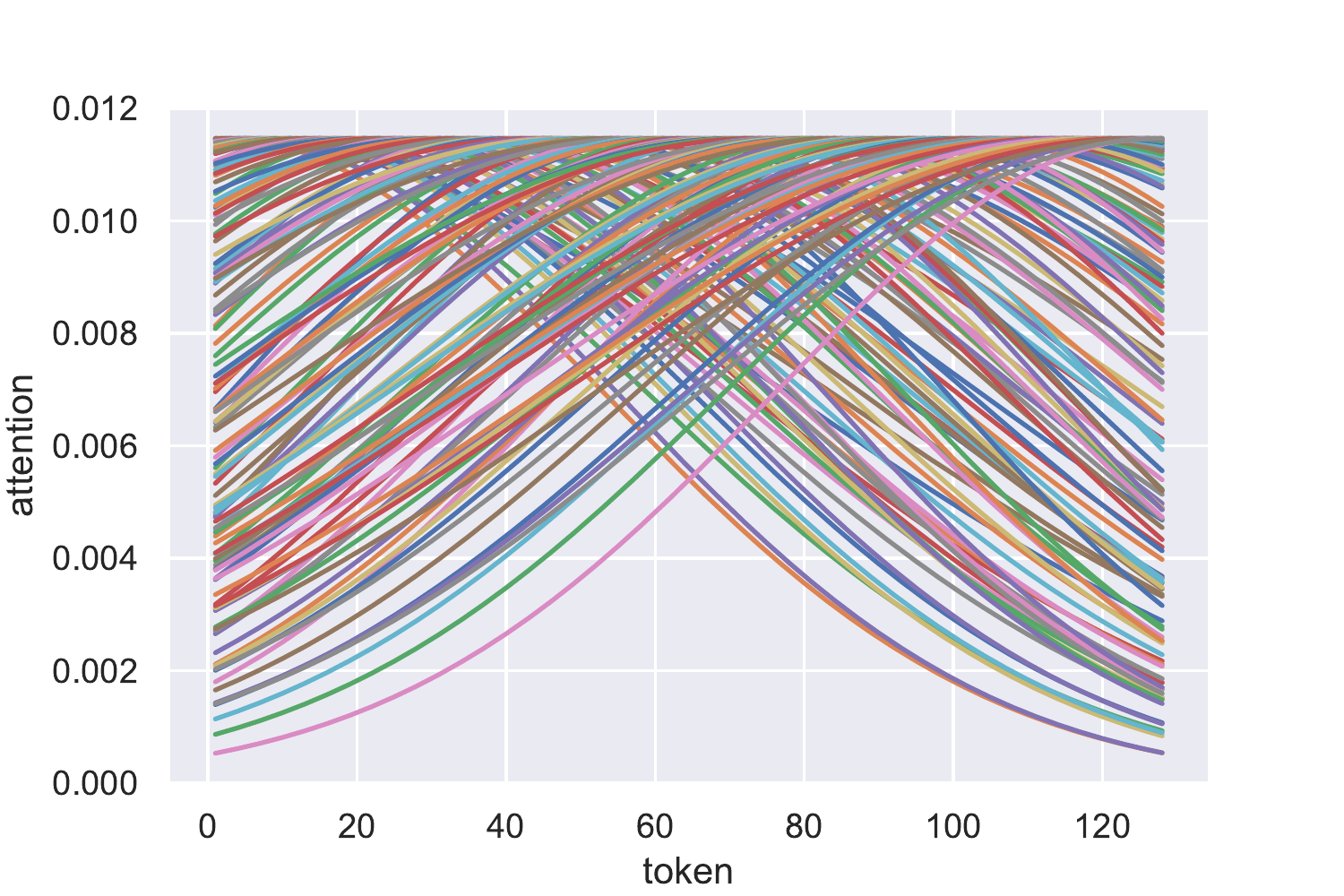}
    \caption[Attentions of a Gaussian averaging heads]{Attentions of a Gaussian averaging head at initialization for sequence length $N = 128$. Every line depicts one Gaussian kernel. Here, the location parameters are initialized with the value of the supporting points $\mu_i = s_i$.
   \label{fig:avg_init}}
\end{figure}

The self-attention mechanism proposed in \citet{Vaswani:17} utilizes the $\soft$ function to compute
the coefficients of a convex combination over the embedded tokens, where the $\soft$ is conditioned on the input. 
However, our analysis showed that especially in lower layers many heads perform averaging over a very 
large number of patterns. This suggests that at this level neither the dependency on the input 
nor a fine grained attention to individual positions is necessary. As an alternative to the original 
mechanism we propose Gaussian averaging heads which 
are computationally more efficient. Here, the $\soft$ function is replaced by a discrete Gaussian kernel,
where the location $\mu$ and the scale $\sigma$ are learned. In detail, for a sequence length of 
$N$ tokens we are given a vector of location parameters $\boldsymbol{\mu} = (\mu_1, \ldots, \mu_N)^T$ and a vector of
corresponding scale parameters $\boldsymbol{\sigma} = (\sigma_1, \ldots, \sigma_N)^T$.
We subdivide the interval $[-1,1]$ into $N$ equidistant supporting points
$\{s_j\}_{j=1}^N$, where 
\begin{align*}
    s_j = \frac{(j-1) - 0.5~(N-1)}{0.5~(N-1)}.
\end{align*}
The attention $[A]_{i,j}$ from the $i$-th token to the $j$-th position is calculated as
\begin{align*}
    [A]_{i,j} = \frac{1}{z_i} \exp \left\{ -\frac{1}{2} \big(\frac{ s_j - \mu_i}{\sigma_i}\big)^2  \right\},
\end{align*}
where $z_i$ normalizes the $i$-th row of the attention matrix $A$ to sum up to one:
\begin{align*}
    z_i = \sum_{j=1}^N \exp \left\{ -\frac{1}{2} \big(\frac{ s_j - \mu_i}{\sigma_i}\big)^2 \right\}.
\end{align*}
For initialization we uniformly sample a location vector $\boldsymbol{\mu} \in [-1,1]^N$ and 
a scale vector $\boldsymbol{\sigma} \in [0.75, 1.25]^N$ per head. A simple way to consider the individual 
position of each token at initialization is to use the supporting points $\mu_i = s_i$
(see Figure~\ref{fig:avg_init}). In practice no difference to the random initialization was observed.

{\textbullet \em Number of parameters.}
Gaussian averaging heads can reduce the number of parameters significantly.
For an input size of $N$ tokens, there are $2 \cdot N$ parameters per head. 
In contrast, a standard self-attention head 
with word embedding dimension $d_y$ and projection dimension $d_k$ has two weight matrices
$W_Q, W_K \in \mathbb{R}^{d_k \times d_y}$, which together amount to $2 \cdot d_k\cdot d_y$ parameters.
As a concrete example, the BERT-base model from \citet{Devlin:19} has an embedding dimension $d_y = 768$, 
 a projection dimension $d_k = 64$ and a sequence length of $N = 512$. 
Compared to the Gaussian head, in this case $ (2 \cdot 768 \cdot 64) / (2 \cdot 512) ~ = 95.5$ times more 
parameters are trained for the attention mechanism itself. 
Only for very long sequences (and given
that the word embedding dimension stays the same) the dependence on $N$ may become a disadvantage.
But of course, due to the independence from the input the Gaussian averaging head is less expressive 
in comparison to the original attention mechanism.
A recently proposed input independent replacement for self-attention is the so called Random Synthesizer 
\citep{Tay:20}. Here the $\soft$-attention is directly parametrized with an $N \times N$ 
matrix. This amounts to $0.5 \cdot N$ more parameters than Gaussian averaging. 

\subsubsection{Experiment 2: Multiple Instance Learning Datasets.}
\label{sec:deeprc_details}
\label{sec:deeprc_implementation_details}

\paragraph{Immune Repertoire Classification.}

An architecture called DeepRC,
is  based on our modern Hopfield networks, for 
immune repertoire classification and compared to
other machine learning approaches. 
For DeepRC, we consider immune repertoires as input objects,
which are represented as bags of instances. In a bag, each instance is
an immune receptor sequence and each bag can contain a large number of sequences. 
At its core, DeepRC consists of a modern 
Hopfield network that
extracts information from each repertoire.
The stored patterns (keys) are representations of the immune amino acid
sequences (instances) that are obtained by an 1D convolutional network
with position encoding.
Each state pattern (query) is static and learned via backpropagation.
For details see \citet{Widrich:20,Widrich:20nips}.

Our new Hopfield network has
been integrated into a deep learning architecture
for immune repertoire classification, a massive 
multiple instance learning task \citep{Widrich:20,Widrich:20nips}.
Theorem~\ref{th:storage} states that modern Hopfield networks possess an exponential 
storage capacity which enables to tackle massive multiple instance learning (MIL) 
problems \citep{Dietterich:97}.
Immune repertoire classification \citep{Emerson:17}  
typically requires 
to extract few patterns from a large set of sequences, the repertoire, 
that are indicative for the respective immune status.
Most MIL methods fail due the large number of instances.

Data is obtained by
experimentally observed immune receptors as well as simulated sequences
sequence motifs \citep{Akbar:19,Weber:19} with low yet varying degrees 
of frequency are implanted.
Four different categories of 
datasets are constructed: 
(a) Simulated immunosequencing data with implanted motifs, 
(b) immunosequencing data generated by long short-term memory (LSTM) with 
implanted motifs, (c) real-world immunosequencing data with implanted motifs, and (d) 
real-world immunosequencing data with known immune 
status \citep{Emerson:17}. Categories (a), (b), and (d) contain approx. 
300,000 instances per immune repertoire. With over 30 billion sequences in total, this 
represents one of the largest multiple instance 
learning experiments ever conducted \citep{Carbonneau:18}.
Despite the massive number of instances as well as the low frequency of sequences 
indicative of the respective immune status, deep learning
architectures with modern Hopfield networks outperform 
all competing methods with respect to average area under the ROC curve in all four 
categories, (a), (b), (c) and (d) (for details see \citet{Widrich:20}).

We evaluate and compare the performance of 
DeepRC to a set of machine learning methods that
serve as baseline, were suggested, or 
can readily be adapted to immune repertoire 
classification.
The methods comprise
(i) known motif,
which counts how often the known implanted motifs occur,
(ii) Support Vector Machine (SVM) approach
that uses a fixed mapping from a bag 
of sequences to the corresponding $k$-mer counts
and used the MinMax and Jaccard kernel,
(iii) $k$-Nearest Neighbor (KNN) with $k$-mer representation,
transforming MinMax and Jaccard kernel to distances,
(iv) logistic regression on the $k$-mer representation,
(v)  burden test that first identifies sequences or $k$-mers and
then computes a burden 
score per individual, and
(vi) logistic multiple instance learning (lMIL).
On the real-world dataset DeepRC achieved
an AUC of $0.832\pm0.022$, followed by
the SVM with MinMax kernel (AUC $0.825\pm0.022$) and the burden
test with an AUC of $0.699\pm0.041$.
Overall on all datasets, DeepRC outperformed all competing methods 
with respect to average AUC
(see \citet{Widrich:20,Widrich:20nips}). 

Table~\ref{tab:results_full} reports the average
performance in the simulated immunosequencing datasets 
(last column) and the performance on datasets of
the remaining three categories. 
DeepRC outperforms all competing methods 
with respect to average AUC.
Across categories, the runner-up methods are either the SVM for 
MIL problems with MinMax kernel or the burden test.

\begingroup
\setlength{\tabcolsep}{2pt} 
\renewcommand{\arraystretch}{1.5}

\begin{table}[htp]%
    \begin{center}%
        \resizebox{\textwidth}{!}{%
        \begin{tabular}{lcacacacacac}
         \toprule
        & Real-world & \multicolumn{4}{c}{ Real-world data with  implanted signals} & \multicolumn{5}{c}{ LSTM-generated data} &  Simulated\\
    & CMV & OM 1\% & OM 0.1\% & MM 1\% & MM 0.1\% & 10\% & 1\% & 0.5\% & 0.1\% & 0.05\%  & avg.\\
        \cmidrule(r){2-2}
        \cmidrule(r){3-6}
        \cmidrule(r){7-11}
        \cmidrule(r){12-12}
        DeepRC & {\bf 0.832} \footnotesize{$\pm$ 0.022} & {\bf 1.00} \footnotesize{$\pm$ 0.00} & {\bf 0.98}\footnotesize{$\pm$ 0.01} & {\bf 1.00}\footnotesize{$\pm$ 0.00} & {\bf 0.94}\footnotesize{$\pm$0.01} & {\bf 1.00}\footnotesize{$\pm$ 0.00} & {\bf 1.00}\footnotesize{$\pm$ 0.00} & {\bf 1.00}\footnotesize{$\pm$ 0.00} & {\bf 1.00}\footnotesize{$\pm$ 0.00} & {\bf 1.00}\footnotesize{$\pm$ 0.00} & {\bf 0.846}\footnotesize{$\pm$ 0.223} \\
        SVM (MM) & 0.825 \footnotesize{$\pm$ 0.022} & {\bf 1.00} \footnotesize{$\pm$ 0.00} & 0.58\footnotesize{$\pm$ 0.02} & {\bf 1.00}\footnotesize{$\pm$ 0.00} & 0.53\footnotesize{$\pm$0.02} & {\bf 1.00}\footnotesize{$\pm$ 0.00} & {\bf 1.00}\footnotesize{$\pm$ 0.00} &{\bf 1.00}\footnotesize{$\pm$ 0.00} & {\bf 1.00}\footnotesize{$\pm$ 0.00} & 0.99\footnotesize{$\pm$ 0.01} & 0.827\footnotesize{$\pm$ 0.210} \\
        SVM (J) & 0.546 \footnotesize{$\pm$ 0.021} & 0.99 \footnotesize{$\pm$ 0.00} & 0.53\footnotesize{$\pm$ 0.02} & {\bf 1.00}\footnotesize{$\pm$ 0.00} & 0.57\footnotesize{$\pm$0.02} & 0.98\footnotesize{$\pm$ 0.04} & {\bf 1.00}\footnotesize{$\pm$ 0.00} & {\bf 1.00}\footnotesize{$\pm$ 0.00} & 0.90\footnotesize{$\pm$ 0.04} & 0.77\footnotesize{$\pm$ 0.07} & 0.550\footnotesize{$\pm$ 0.080} \\
        KNN (MM) & 0.679 \footnotesize{$\pm$ 0.076} & 0.74 \footnotesize{$\pm$ 0.24} & 0.49\footnotesize{$\pm$ 0.03} & 0.67\footnotesize{$\pm$ 0.18} & 0.50\footnotesize{$\pm$0.02} & 0.70\footnotesize{$\pm$ 0.27} & 0.72\footnotesize{$\pm$ 0.26} & 0.73\footnotesize{$\pm$ 0.26} & 0.54\footnotesize{$\pm$ 0.16} & 0.52\footnotesize{$\pm$ 0.15} & 0.634\footnotesize{$\pm$ 0.129} \\
        KNN (J) & 0.534 \footnotesize{$\pm$ 0.039} & 0.65 \footnotesize{$\pm$ 0.16} & 0.48\footnotesize{$\pm$ 0.03} & 0.70\footnotesize{$\pm$ 0.20} & 0.51\footnotesize{$\pm$0.03} & 0.70\footnotesize{$\pm$ 0.29} & 0.61\footnotesize{$\pm$ 0.24} & 0.52\footnotesize{$\pm$ 0.16} & 0.55\footnotesize{$\pm$ 0.19} & 0.54\footnotesize{$\pm$ 0.19} & 0.501\footnotesize{$\pm$ 0.007} \\
        Log. regr. & 0.607 \footnotesize{$\pm$ 0.058} & {\bf 1.00} \footnotesize{$\pm$ 0.00} & 0.54\footnotesize{$\pm$ 0.04} & 0.99\footnotesize{$\pm$ 0.00} & 0.51\footnotesize{$\pm$0.04} & {\bf 1.00}\footnotesize{$\pm$ 0.00} & {\bf 1.00}\footnotesize{$\pm$ 0.00} & 0.93\footnotesize{$\pm$ 0.15} & 0.60\footnotesize{$\pm$ 0.19} & 0.43\footnotesize{$\pm$ 0.16} & 0.826\footnotesize{$\pm$ 0.211} \\
        Burden test & 0.699 \footnotesize{$\pm$ 0.041} & {\bf 1.00} \footnotesize{$\pm$ 0.00} & 0.64\footnotesize{$\pm$ 0.05} & {\bf 1.00}\footnotesize{$\pm$ 0.00} & 0.89\footnotesize{$\pm$0.02} & {\bf 1.00}\footnotesize{$\pm$ 0.00} & {\bf 1.00}\footnotesize{$\pm$ 0.00} & {\bf 1.00}\footnotesize{$\pm$ 0.00} & {\bf 1.00}\footnotesize{$\pm$ 0.00} & 0.79\footnotesize{$\pm$ 0.28} & 0.549\footnotesize{$\pm$ 0.074} \\
        Log. MIL (KMER) & 0.582 \footnotesize{$\pm$ 0.065} & 0.54 \footnotesize{$\pm$ 0.07} & 0.51\footnotesize{$\pm$ 0.03} & 0.99\footnotesize{$\pm$ 0.00} & 0.62\footnotesize{$\pm$0.15} & {\bf 1.00}\footnotesize{$\pm$ 0.00} & 0.72\footnotesize{$\pm$ 0.11} & 0.64\footnotesize{$\pm$ 0.14} & 0.57\footnotesize{$\pm$ 0.15} & 0.53\footnotesize{$\pm$ 0.13} & 0.665\footnotesize{$\pm$ 0.224} \\
        Log. MIL (TCR\textbeta) & 0.515 \footnotesize{$\pm$ 0.073} & 0.50 \footnotesize{$\pm$ 0.03} & 0.50\footnotesize{$\pm$ 0.02} & 0.99\footnotesize{$\pm$ 0.00} & 0.78\footnotesize{$\pm$0.03} & 0.54\footnotesize{$\pm$ 0.09} & 0.57\footnotesize{$\pm$ 0.16} & 0.47\footnotesize{$\pm$ 0.09} & 0.51\footnotesize{$\pm$ 0.07} & 0.50\footnotesize{$\pm$ 0.12} & 0.501\footnotesize{$\pm$ 0.016} \\
        \midrule
        Known motif b. & -- & 1.00 \footnotesize{$\pm$ 0.00} & 0.70\footnotesize{$\pm$ 0.03} & 0.99\footnotesize{$\pm$ 0.00} & 0.62\footnotesize{$\pm$0.04} & 1.00\footnotesize{$\pm$ 0.00} & 1.00\footnotesize{$\pm$ 0.00} & 1.00\footnotesize{$\pm$ 0.00} & 1.00\footnotesize{$\pm$ 0.00} & 1.00\footnotesize{$\pm$ 0.00} & 0.890\footnotesize{$\pm$ 0.168} \\
        Known motif c. & -- & 0.92 \footnotesize{$\pm$ 0.00} & 0.56\footnotesize{$\pm$ 0.03} & 0.65\footnotesize{$\pm$ 0.03} & 0.52\footnotesize{$\pm$0.03} & 1.00\footnotesize{$\pm$ 0.00} & 1.00\footnotesize{$\pm$ 0.00} & 0.99\footnotesize{$\pm$ 0.01} & 0.72\footnotesize{$\pm$ 0.09} & 0.63\footnotesize{$\pm$ 0.09} & 0.738\footnotesize{$\pm$ 0.202} \\     
    \bottomrule
    \end{tabular}
        }%
        \caption[Results of immune repertoire classification across all datasets]{Results immune repertoire classification across all datasets. Results are given in terms of AUC of the competing methods on all datasets. The reported errors are standard deviations across $5$ cross-validation (CV) folds (except for the column ``Simulated'').
        \textbf{Real-world CMV:} Average performance over 
        $5$ CV folds on the \emph{cytomegalovirus (CMV) dataset}~\cite{Emerson:17}.
        \textbf{Real-world data with implanted signals:} Average performance over
        $5$ CV folds for each 
        of the four datasets. A signal 
        was implanted with a frequency (=wittness rate) of 
        $1\%$ or $0.1\%$. Either a single motif (``OM'') 
        or multiple motifs (``MM'') were implanted.
        \textbf{LSTM-generated data:} Average performance over
        $5$ CV folds for each 
        of the $5$ datasets. In each dataset, a signal 
        was implanted with a frequency of $10\%$,
        $1\%$, $0.5\%$, $0.1\%$, and $0.05\%$, respectively.
        \textbf{Simulated:} Here we report the mean over 18 simulated datasets with implanted signals and varying difficulties. The error reported is the standard deviation of the AUC values across the 18 datasets.
        }%
        \label{tab:results_full}%
    \end{center}%
\end{table}
\endgroup

\paragraph{Multiple Instance Learning Benchmark Datasets.}
Classical benchmarking datasets comprise
UCSB breast cancer classification \citep{Kandemir:14}, and 
the Elephant, Fox, Tiger datasets \citep{Andrews:03}.

Elephant, Fox and Tiger are MIL datasets for image annotation which comprise
color images from the Corel dataset that have been preprocessed and
segmented. An image consists
of a set of segments (or blobs), 
each characterized by color, texture and shape
descriptors. 
The datasets have 100 positive and 100 negative
example images. 
The latter have been randomly drawn from a pool of photos of
other animals. 
Elephant has 1391 instances and 230 features. 
Fox has 1320 instances and 230 features.
Tiger has 1220 instances and 230 features.
Furthermore, we use the 
UCSB breast cancer classification \citep{Kandemir:14}
dataset, which consists of 2,002 instances across 58 input objects. An instance 
represents a patch of a histopathological image of cancerous or normal tissue.
The layer \texttt{HopfieldPooling} is used, which allows for 
computing a per-input-object representation by extracting an 
average of instances that are indicative for one of the two classes. 
The input to the \texttt{HopfieldPooling} layer
is a set of embedded instances $\BY$ and
a trainable but fixed state (query) pattern $\BQ$ used for averaging of class-indicative instances. 
This averaging enables a compression of variable-sized bags to a 
fixed-sized representation to discriminate the bags.
We performed a manual hyperparameter search on a validation set.
In detail, we used the following architecture to perform the given task
on the Elephant, Fox, Tiger and UCSCB breast cancer datasets:
(I) we apply fully connected linear embedding layers
with ReLU activation. (II) The output of this embedding serves
as the input to our \texttt{HopfieldPooling} layer where the above described
pooling operation is performed. (III) Thereafter we use 'ReLU - Linear blocks'
as the final linear output layers that perform the classification.
Among other hyperparameters, different hidden layer widths
(for the fully connected pre- and post-\texttt{HopfieldPooling} layers),
learning rates and batch sizes were tried.
Additionally our focus resided on the hyperparameters of the \texttt{HopfieldPooling} layer. Among those were the number of heads,
the head dimension and the scaling factor \textbeta{}.
\begin{table*}[t]
\centering
\begin{tabular}{lc}
\hline 
\textbf{parameter} & \textbf{values} \\
\hline 
learning rates & $\{10^{-3}$, $10^{-5}\}$ \\
learning rate decay (\textgamma{}) & $\{0.98,0.96,0.94\}$ \\
embedding layers & $\{1, 2, 3\}$ \\
layer widths & $\{32, 64, 256, 1024, 2048\}$ \\
number of heads & $\{8, 12, 16, 32 \}$ \\
head dimensions & $\{16, 32, 64 \}$ \\
scaling factors & $\{0.1, 1.0, 10.0 \}$ \\
hidden dimensions & $\{ 32, 64, 128 \}$ \\
bag dropout & $\{0.0, 0.75\}$ \\
\hline
\end{tabular}%
\caption[Hyperparameter selection for MIL datasets]{Hyperparameter search\--{}space of a manual hyperparameter selection on the respective validation sets of the Elephant, Fox, Tiger and UCSB breast cancer datasets.}
\label{tab:hyperparameters_mil}
\end{table*}
All models were trained for 160 epochs using the AdamW~optimizer~\citep{loshchilov:17} with exponential
learning rate decay (see~\tablename{}~\ref{tab:hyperparameters_mil}), and validated by 10-fold nested cross validation repeated five times with different splits on the data sets. The reported ROC AUC scores are the average of these repetitions. As overfitting imposed quite a problem, bag dropout was applied as the regularization technique of choice.


\clearpage
\subsubsection{Experiment 3: Classification on Small UCI Benchmark Datasets}
\label{appsec:uci}

\paragraph{Motivation.}
Datasets with a small number of samples, like the UCI benchmark datasets, are particularly difficult for neural networks to generalize on. 
In contrast to their performance on larger datasets, they are consistently outperformed by methods like e.g.\ gradient boosting, random forests (RF) and support vector machines (SVMs).      
Finding samples or even learning prototypes that are highly indicative for the class of a sample (query) suggest the use of Hopfield networks.
We applied a modern Hopfield network via the layer {\tt Hopfield}.
The input vector is mapped to $\BR$
using a self-normalizing net (SNN) and $\BW_K$ is learned,
where the dimension of $\BW_K$ (the number of stored fixed pattern)
is a hyperparameter.
The output $\BZ$ of {\tt Hopfield} enters
the output layer.

\paragraph{Methods compared.}
Modern Hopfield networks via the layer Hopfield
are compared to 
17 groups of methods \citep{Fernandez:14,Klambauer:17}: 
\begin{enumerate}
    \item Support Vector Machines
    \item Random Forest
    \item Multivariate adaptive regression splines (MARS)
    \item Boosting
    \item Rule-based Methods
    \item Logistic and Multinomial Regression (LMR)
    \item Discriminant Analysis (DA)
    \item Bagging
    \item Nearest Neighbor
    \item Decision Trees
    \item Other Ensembles
    \item Neural Networks (standard NN, BatchNorm, WeighNorm, MSRAinit, LayerNorm, ResNet, Self-Normalizing Nets)
    \item Bayesian Methods 
    \item Other Methods 
    \item Generalized linear models (GLM) 
    \item Partial Least Squares and Principal Component Regression (PLSR)
    \item Stacking (Wolpert)
\end{enumerate}

\paragraph{Experimental design and implementation details.}
As specified in the main paper,
we consider $75$ datasets of the
{\it UC Irvine Machine Learning Repository},
which contain less than $1,000$ samples per dataset,
following the dataset separation into large and small dataset in \citet{Klambauer:17}.
On each dataset,
we performed a grid-search to determine the best hyperparameter setting and model per dataset.
The hyperparameter search-space of the grid-search is listed in Table~\ref{tab:uci_search_space}.
All models were trained for $100$ epochs with a mini-batch size of $4$ samples using the cross entropy loss and
the PyTorch SGD module for stochastic gradient descent without momentum and without weight decay or dropout.
After each epoch, the model accuracy was computed on a separated validation set.
Using early stopping,
the model with the best validation set accuracy averaged over $16$ consecutive epochs was selected as final model.
This final model was then evaluated against a separated test set to determine the accuracy,
as reported in Tables~\ref{tab:uci_results} and Table~\url{uci_detailed_results.csv} in the supplemental materials.

As network architecture,
we use $\{0, 1, 7\}$ fully connected embedding layers with SELU
\citet{Klambauer:17} activation functions and
$\{32, 128, 1024\}$ hidden units per embedding layer.
These embedding layers are followed by the layer {\tt Hopfield}.
The number of hidden units is also used as number of dimensions 
for the Hopfield association space with a number of $\{1, 32\}$ heads.
The layer {\tt Hopfield} is followed by a mapping to the output vector,
which has as dimension the number of classes.
Finally, the softmax function is applied to obtain the predicted probability for a class.

\begin{table*}[t]
\centering
\begin{tabular}{lc}
\hline 
\textbf{parameter} & \textbf{values} \\
\hline 
learning rates & $\{0.05\}$ \\
embedding layers & $\{0, 1, 7\}$ \\
hidden units & $\{32, 128, 1024\}$ \\
heads & $\{1, 32\}$ \\
$\beta$ & $\{1.0, 0.1, 0.001\}$ \\
\# stored patterns & $\{1, 8\}\cdot n\_classes$ \\
\hline
\end{tabular}%
\caption[Hyperparameter selection for small UCI benchmark datasets]{Hyperparameter search-space for grid-search on small UCI benchmark datasets.
All models were trained for $100$ epochs using stochastic gradient descent 
with early stopping based on the validation set accuracy and 
a minibatch size of $4$ samples.
The number of stored patterns is depending on the number of target classes of the individual tasks. \label{tab:uci_search_space}}
\end{table*}

\paragraph{Results.}
We compared the performance of 25 methods based on their method rank.
For this we computed the rank per method per dataset based on the accuracy on the test set,
which was then averaged over all 75 datasets for each method to obtain the method rank.
For the baseline methods we used the scores summarized by~\citep{Klambauer:17}.


\subsubsection{Experiment 4: Drug Design Benchmark Datasets}
\label{sec:drug-design}
\paragraph{Experimental design and implementation details.}
We test Hopfield layers on 4 classification datasets from MoleculeNet \citep{Wu:17},
which are challenging for deep learning methods.
The first dataset is HIV, which 
was introduced by the Drug Therapeutics Program (DTP) AIDS Antiviral Screen.
The second dataset is BACE, which 
has IC50 measurements for 
binding affinities of inhibitors (molecules) to the human $\beta$-secretase 1 (BACE-1). 
The third dataset is BBBP (blood-brain barrier permeability), 
which  stems from modeling and predicting 
the blood-brain barrier permeability \citep{Martins:12}.
The fourth dataset is SIDER (Side Effect Resource) \cite{Kuhn:16} and contains 1427 approved drugs. 
These datasets represent four areas of modeling tasks in drug discovery, 
concretely to develop accurate models for predicting  
a) new anti-virals (HIV), 
b) new protein inhibitors (BACE),
c) metabolic effects (BBBP), and
d) side effects of a chemical compound (SIDER).

We implemented a Hopfield layer {\tt HopfieldLayer},  
in which we used the training-input as stored-pattern $\BY$ or key, 
the training-label as pattern-projection $\BY \BW_V$ or value and 
the input as state-pattern $\BR$ or query. As described in section \ref{chap:Hopfield_layer} by concatenation of input $\Bz_i$ and target $\Bt_i$ the matrices $\BW_K$ and  $\BW_V$
can be designed such that inside the softmax the input $\Bz_i$ is used and outside the softmax the target $\Bt_i$. 

All hyperparameters were selected on separate validation sets and 
we selected the model with the highest validation AUC 
on five different random splits.

\begin{table*}[ht]
\centering
\begin{tabular}{lc}
\hline 
\textbf{parameter} & \textbf{values} \\
\hline 
beta & $\{0.0001, 0.001, 0.01, 0.1, 0.2, 0.3\}$ \\
learning rates & $\{0.0002\}$ \\
heads & $\{1, 32, 128, 512\}$ \\
dropout & $\{0.0, 0.1, 0.2\}$ \\
state-pattern bias & $\{0.0, -0.1, -0.125, 0.15, -0.2\}$ \\
association-activation & \{None, LeakyReLU \} \\
state- and stored-pattern static & \{False, True\} \\
normalize state- and stored-pattern & \{False, True\} \\
normalize association projection & \{False, True\} \\
learnable stored-pattern & \{False, True\} \\
\hline
\end{tabular}%
\caption[Hyperparameter selection for drug design datasets]{Hyperparameter search-space for grid-search on HIV, BACE, BBBP and SIDER.
All models were trained if applicable for $4$ epochs using 
Adam and a batch size of $1$ sample. \label{tab:drug_search_space}}
\end{table*}

\paragraph{Results.}

We compared the Hopfield layer {\tt Hopfieldlayer}
to Support Vector Machines (SVMs) \citep{Cortes:95,Schoelkopf:02book}, 
Extreme Gradient Boosting (XGBoost) \citep{Chen:16},
Random Forest (RF) \citep{Breiman:01},
Deep Neural Networks (DNNs) \citep{LeCun:15,Schmidhuber:15}, and to 
graph neural networks (GNN) like
Graph Convolutional Networks (GCNs) \citep{Kipf:16},
Graph Attention Networks (GATs) \citep{Velickovic:18},
Message Passing Neural Networks (MPNNs) \citep{Gilmer:17}, and
Attentive FP \citep{Xiong:20}.
Our architecture with {\tt HopfieldLayer} has reached state-of-the-art 
for predicting side 
effects on SIDER $0.672\pm0.019$ 
as well as for predicting $\beta$-secretase BACE $0.902\pm0.023$. 
See Table~\ref{tab:drug_discovery} for all results, where the results
of other methods are taken from \cite{Jiang:20}.

\begin{table}[H]
\centering
\caption[Results on drug design benchmark datasets]{Results on drug design benchmark datasets. Predictive performance (ROCAUC) on test set as reported by \cite{Jiang:20} for 50 random splits}
\label{tab:drug_discovery}
\begin{tabular}{lcccc}
\toprule
\textbf{Model}  &    \textbf{HIV}           &    \textbf{BACE}  &    \textbf{BBBP}  &   \textbf{SIDER}  \\ 
\midrule
SVM		    	&	{$0.822\pm0.020$}		& {$0.893\pm0.020$}	    &	{$0.919\pm0.028$}	&	{$0.630\pm0.021$} \\
XGBoost			&	{$0.816\pm0.020$}			&	{$0.889\pm0.021$}	& {$\mathbf{0.926\pm0.026}$}	&	{$0.642\pm0.020$} \\
RF				&	{$0.820\pm0.016$}			&	{$0.890\pm0.022$}& {$\mathbf{0.927 \pm0.025}$}	&	{$0.646\pm0.022$} \\
GCN		&	{$\mathbf{0.834\pm0.025}$}&	{$0.898\pm0.019$}	&	{$0.903\pm0.027$}	&	{$0.634\pm0.026$} \\
GAT		  &	{$0.826\pm0.030$}			&	{$0.886\pm0.023$}	    &	{$0.898\pm0.033$}	&	{$0.627\pm0.024$} \\
DNN			    &	{$0.797\pm0.018$}			&	{$0.890\pm0.024$}	    &	{$0.898\pm0.033$}	&	{$0.627\pm0.024$} \\
MPNN			&	{$0.811\pm0.031$}			&	{$0.838\pm0.027$}	    &	{$0.879\pm0.037$}	&	{$0.598\pm0.031$} \\
Attentive FP 	&	{$0.822\pm0.026$}			&	{$0.876\pm0.023$}	    &	{$0.887\pm0.032$}	&	{$0.623\pm0.026$} \\
Hopfield (ours)	&	{$0.815\pm0.023$}	&	$\mathbf{0.902\pm0.023}$    &	{$0.910\pm0.026$}	&	$\mathbf{0.672\pm0.019}$ \\
\bottomrule
\end{tabular}
\end{table}


\subsection{PyTorch Implementation of Hopfield Layers}
The implementation is available at: \url{https://github.com/ml-jku/hopfield-layers}
\label{chap:Hopfield_layer}
    
\subsubsection{Introduction}
\label{sec:C_Introduction}

In this section, we describe the implementation
of Hopfield layers in PyTorch \citep{Paszke:17, Paszke:19}
and, additionally, provide a brief usage manual.
Possible applications for a Hopfield layer
in a deep network architecture comprise:
\begin{itemize}
\item multiple instance learning (MIL) \citep{Dietterich:97},
\item processing of and learning with point sets \citep{Qi:17ieee,Qi:17,Xu:18spider},
\item set-based and permutation invariant learning \citep{Guttenberg:16,Ravanbakhsh:16,Zaheer:17,Korshunova:18,Ilse:18,Zhai:20},
\item attention-based learning \citep{Vaswani:17},
\item associative learning,
\item natural language processing,
\item sequence analysis and time series prediction, and
\item storing and retrieving reference or experienced data, e.g.\ 
to store training data and retrieve it by the model 
or to store experiences for reinforcement learning.
\end{itemize}
The Hopfield layer in a deep neural network architecture can implement:
\begin{itemize}
\item a memory (storage) with associative retrieval \citep{Danihelka:16,Ba:16},
\item conditional pooling and averaging operations \citep{Wang:18,Ilse:20},
\item combining data by associations \citep{Agrawal:93}, 
\item associative credit assignment (e.g.\ Rescorla-Wagner model or value estimation) \citep{Sutton:18book},
and 
\item attention mechanisms \citep{Vaswani:17, Bahdanau:14}.
\end{itemize}
In particular, a Hopfield layer can substitute
attention layers in architectures of transformer and BERT models.
The Hopfield layer
is designed to be used as plug-in replacement
for existing layers like 
\begin{itemize}
\item pooling layers (max-pooling or average pooling),
\item permutation equivariant layers \citep{Guttenberg:16,Ravanbakhsh:16}, 
\item GRU \& LSTM layers, and 
\item attention layers. 
\end{itemize}
In contrast to classical Hopfield networks, the Hopfield layer is based 
on the modern Hopfield networks with continuous states that have increased storage capacity, 
as discussed in the main paper. 
Like classical Hopfield networks, 
the dynamics of the single heads of a Hopfield layer 
follow a energy minimization dynamics.
The energy minimization empowers 
our Hopfield layer with several advantages
over other architectural designs like memory cells, associative memory, or
attention mechanisms.
For example, the Hopfield layer has more functionality than
a transformer self-attention layer \citep{Vaswani:17} as
described in Sec.~\ref{sec:functionality}.
Possible use cases are given in Sec.~\ref{sec:hopfield_layer_usage}.
Source code will be provided under {\tt github}.

\subsubsection{Functionality}\label{sec:functionality}

\begin{figure}[htp]
    \begin{center}
    \includegraphics[width=0.8\textwidth]{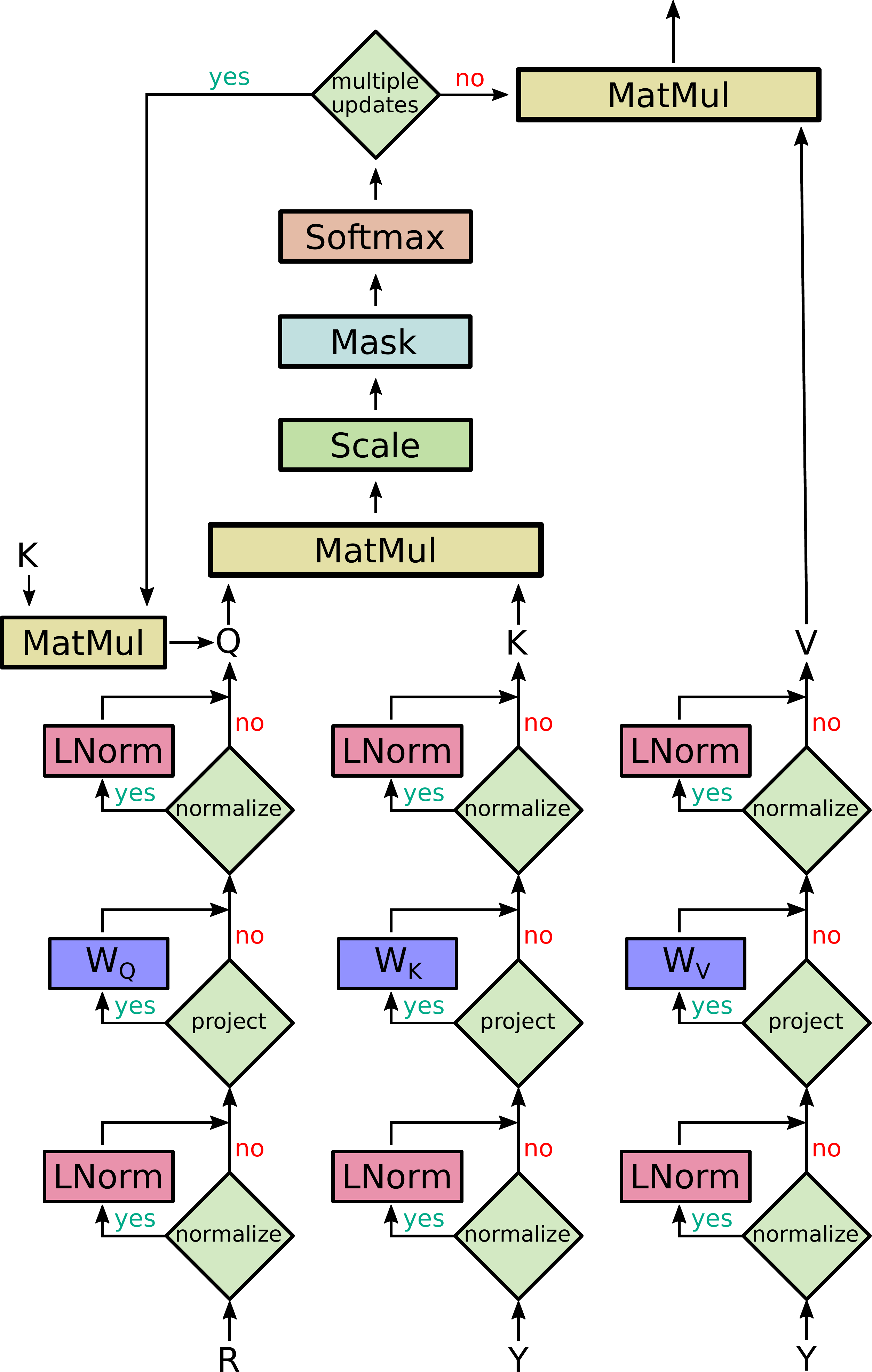}\\
   \caption[A flowchart of the Hopfield layer]{
   A flowchart of the Hopfield layer.
   First, the raw state (query) patterns
   $\BR$ and the raw stored (key) patterns
   $\BY$ are optionally normalized (with layer normalization), 
   projected and optionally normalized (with layer normalization) again.
   The default setting is a layer normalization of the input
   patterns, and no layer normalization of the projected patterns.
   The raw stored patterns $\BY$ can in principle be also two different input tensors.
   Optionally, multiple updates take place in the 
   projected space of $\BQ$ and $\BK$. 
   This update rule is obtained e.g.\ from the full update
   Eq.~\eqref{eq:FullUpdate2set} or the simplified update
   Eq.~\eqref{eq:SimpleUpdate2set}
   in the appendix.
   }
   \label{fig:sketch_hopfield_layer}
   \end{center}
\end{figure}

Non-standard functionalities that are added by a Hopfield layer are 
\begin{itemize}
\item \textit{Association of two sets},
\item \textit{Multiple Updates} for precise fixed points, 
\item \textit{Variable Beta} that determines the kind of fixed points,  
\item \textit{Dimension of the associative space} for controlling the storage capacity, 
\item \textit{Static Patterns} for fixed pattern search,
and 
\item \textit{Pattern Normalization} to control the fixed point dynamics by norm of
the patterns and shift of the patterns.
\end{itemize}
A functional sketch of our Hopfield layer is shown in Fig.~\ref{fig:sketch_hopfield_layer}.

{\textbullet \em Association of two sets.}
The Hopfield layer makes it possible
to associate two sets of vectors.
This general functionality allows
\begin{itemize}
\item for transformer-like self-attention,
\item for decoder-encoder attention, 
\item for time series prediction (maybe with positional encoding),
\item for sequence analysis, 
\item for multiple instance learning, 
\item for learning with point sets,
\item for combining data sources by associations,
\item for constructing a memory, 
\item for averaging and pooling operations, and 
\item for many more.
\end{itemize}
The first set of vectors consists of $S$ {\em raw state patterns} 
$\BR = (\Br_1, \ldots , \Br_S)^T$ with $\Br_s \in \dR^{d_r}$
and the second set of vectors consists of $N$ {\em raw stored patterns} 
$\BY = (\By_1,\ldots, \By_N)^T$ with $\By_i \in \dR^{d_y}$.
Both the $S$ raw state patterns and $N$ raw stored patterns 
are mapped to 
an associative space in $\dR^{d_k}$ via the matrices $\BW_Q \in \dR^{d_r \times d_k}$
and $\BW_K \in \dR^{d_y \times d_k}$, respectively.
We define a matrix $\BQ$ ($\BXi^T$) of {\em state patterns} $\Bxi_n= \BW_Q \Br_n$ in 
an associative space $\dR^{d_k}$ and a matrix $\BK$ ($\BX^T$) of 
{\em stored patterns} $\Bx_i = \BW_K \By_s$ in the associative space $\dR^{d_k}$:
\begin{align}
    \label{eq:Q_mapping}
    \BQ \ &= \ \BXi^T  \ = \ \BR \ \BW_Q \ , \\
    \label{eq:K_mapping}
    \BK \ &= \ \BX^T  \ = \ \BY \ \BW_K \ .
\end{align}

In the main paper, Eq.~\eqref{eq:update}
defines the novel update rule:
\begin{align}
  \Bxi^{\nn} \ &= \ f(\Bxi) \ = \   
  \BX \ \soft ( \beta \ \BX^T \Bxi) \ ,
\end{align} 
For multiple patterns, Eq.~\eqref{eq:update} becomes:
\begin{align}
\label{eq:update_matrix}
 \BXi^{\nn} \ &= \ f(\BXi) \  = \   
    \BX \ \soft ( \beta \ \BX^T \BXi) \ ,
\end{align}
where $\BXi = (\Bxi_1, \ldots, \Bxi_N)$ is the matrix
of $N$ state (query) patterns, $\BX$
is the matrix of stored (key) patterns,
and $\BXi^{\nn}$ 
is the matrix of new state patterns,
which are averages over stored patterns.
A new state pattern can also be very similar to a single stored pattern, in which case 
we call the stored pattern to be retrieved.

These matrices allow to rewrite Eq.~\eqref{eq:update_matrix} as:
\begin{align}
\label{eq:update_matrix_mapped}
  \left( \BQ^{\nn}  \right)^T \ &= \  \BK^T \soft (\beta \ \BK \ \BQ^T)  \ .
\end{align}
For $\beta = 1/\sqrt{d_k}$ and changing in Eq.~\eqref{eq:update_matrix_mapped}
$\soft \in \dR^N$ to a row vector (and evaluating a row vector), we obtain:
\begin{align}
  \BQ^{\nn} \ &= \ \soft (1/\sqrt{d_k} \ \BQ \ \BK^T) \ \BK \ , 
\end{align}
where $\BQ^{\nn}$ is again the matrix of new state patterns.
The new state patterns $\BXi^{\nn}$ are projected 
via $\BW_V$ to the result patterns $\BZ=\BXi^{\nn}\BW_V$,
where $\BW_V \in \dR^{d_k \times d_v}$.
With the pattern projection $\BV = \BK \BW_V$, we obtain
the update rule Eq.~\eqref{eq:transformer_attention} from the main paper:
\begin{align}
\label{eq:transformer_attention_appendix}
  \BZ \ &= \ \soft(1/ \sqrt{d_k} \ \BQ \ \BK^T) \ \BV \ . 
\end{align}

{\textbullet \em Multiple Updates.}
The update Eq.~\eqref{eq:update_matrix_mapped} 
can be iteratively applied to the initial state $\Bxi$
of every Hopfield layer head. 
After the last update, the new states $\BXi^{\nn}$ are projected
via $\BW_V$ to the result patterns $\BZ=\BXi^{\nn}\BW_V$.
Therefore, the Hopfield layer allows multiple update steps 
in the forward pass without changing the number of parameters.
The number of update steps can be given for every Hopfield 
head individually. 
Furthermore, it is possible to set a threshold
for the number of updates of every Hopfield 
head based on $\NRM{\Bxi - \Bxi^{\nn}}_{2}$. 
In the general case of multiple initial states $\BXi$, the maximum over the individual norms is taken.

{\textbullet \em Variable $\beta$.}
In the main paper, 
we have identified $\beta$ as a crucial
parameter for the fixed point dynamics of the Hopfield network,
which governs the operating mode of the attention heads.
In appendix, e.g.\ in Lemma~\ref{th:banach} or 
in Eq.~\eqref{eq:bound_max_softmax_center} and Eq.~\eqref{eq:bound_max_softmax_fixed}, 
we showed that the characteristics of the fixed points of the new
modern Hopfield network are
determined by: $\beta$, $M$ (maximal pattern norm), 
$m_{\max}$ (spread of the similar patterns), and $\NRM{\Bm_{\Bx}}$
(center of the similar patterns).
Low values of $\beta$ induce global averaging and higher values of $\beta$
metastable states.
In the transformer attention, the $\beta$ parameter
is set to $\beta = 1/\sqrt{d_k}$ as in Eq.~\eqref{eq:transformer_attention_appendix}.
The Hopfield layer, however, allows to freely choose $\beta > 0$, since the fixed
point dynamics does not only depend on the dimension of the associative space $d_k$. 
Additionally, $\beta$ heavily influences the gradient flow to the matrices $\BW_Q$
and $\BW_K$. Thus, finding the right $\beta$ for the respective application 
can be crucial. 

{\textbullet \em Variable dimension of the associative space.}
Theorem~\ref{th:mainStorage} says that the storage capacity of
the modern Hopfield network grows exponentially with the 
dimension of the associative space.
However higher dimension of the associative space
also means less averaging and smaller metastable states. 
The dimension of the associative space trades off storage capacity against 
the size of metastable states, e.g.\ over how many pattern is averaged.
In Eq.~\eqref{eq:K_mapping} and in Eq.~\eqref{eq:Q_mapping},
we assumed $N$ raw state patterns $\BR = (\Br_1, \ldots, \Br_N)^T$  and
$S$ raw stored patterns $\BY = (\By_1, \ldots, \By_S)^T$ 
that are mapped to 
a $d_k$-dimensional associative space 
via the matrices $\BW_Q \in \dR^{d_r \times d_k}$ and 
$\BW_K \in \dR^{d_y \times d_k}$, respectively. 
In the associative space $\dR^{d_k}$, we obtain the state patterns
$\BQ = \BXi^T = \BR  \BW_Q$ and 
the stored patterns $\BK = \BX^T   =  \BY \ \BW_K$.
The Hopfield view relates 
the dimension $d_k$ to the number of input patterns $N$ that have to be processed.
The storage capacity depends exponentially on
the dimension $d_k$ (the dimension of the associative space) and the 
size to metastable states is governed by this dimension, too.
Consequently, $d_k$ should be chosen with respect 
to the number $N$ of patterns
one wants to store and the desired size of metastable states,
which is the number of patterns one wants to average over.
For example, if the input consists of many low dimensional input patterns,
it makes sense to project the patterns into a higher dimensional space to allow a proper fixed point dynamics.
Intuitively, this coincides with the construction of a 
richer feature space for the patterns.

{\textbullet \em Static Patterns.}
In Eq.~\eqref{eq:K_mapping} and Eq.~\eqref{eq:Q_mapping}, 
the $N$ raw state patterns $\BR = (\Br_1, \ldots, \Br_N)^T$
and $S$ raw stored patterns $\BY = (\By_1, \ldots, \By_S)^T$
are mapped to 
an associative space via the matrices $\BW_Q \in \dR^{d_r \times d_k}$ 
and $\BW_K \in \dR^{d_y \times d_k}$, which gives
the state patterns $\BQ = \BXi^T = \BR  \BW_Q$ and 
the stored patterns $\BK = \BX^T   =  \BY \ \BW_K$.
We allow for static state and static stored patterns.
Static pattern means that the pattern does not depend on the
network input, i.e.\ it is determined by the bias weights and
remains constant across different network inputs.
Static state patterns allow to determine whether particular fixed patterns
are among the stored patterns and vice versa. 
The static pattern functionality is typically needed if particular patterns
must be identified in the data, e.g.\ as described 
for immune repertoire classification in the main paper,
where a fixed $d_k$-dimensional state vector $\Bxi$ is used.

{\textbullet \em Pattern Normalization.}
In the appendix, e.g.\ in Lemma~\ref{th:banach} or 
in Eq.~\eqref{eq:bound_max_softmax_center} and Eq.~\eqref{eq:bound_max_softmax_fixed}, 
we showed that the characteristics of the fixed points of the new
modern Hopfield network are
determined by: $\beta$, $M$ (maximal pattern norm), 
$m_{\max}$ (spread of the similar patterns), and $\NRM{\Bm_{\Bx}}$
(center of the similar patterns).
We already discussed the parameter $\beta$ while 
the spread of the similar patterns $m_{\max}$ is given by the data.
The remaining variables $M$ and $\Bm_{\Bx}$ 
that both control the fixed point dynamics are
adjusted pattern normalization.
$M$ is the maximal pattern norm and 
$\Bm_{\Bx}$ the center of the similar patterns.
Theorem~\ref{th:mainStorage} says 
that larger $M$ allows for more patterns to
be stored. However, the size of metastable
states will decrease with increasing $M$.
The vector $\Bm_{\Bx}$ says
how well the (similar) patterns are centered. 
If the norm $\NRM{\Bm_{\Bx}}$ is large,
then this leads to smaller metastable states.
The two parameters $M$ and $\Bm_{\Bx}$ 
are controlled by pattern normalization and determine
the size and convergence properties of metastable states.
These two parameters are important for creating large gradients 
if heads start with global averaging which has small gradient.
These two parameters can shift a head towards 
small metastable states which have 
largest gradient as shown in Fig.~\ref{fig:fine_grained}(b).
We allow for three different pattern normalizations, where the first is the default setting:
\begin{itemize}
    \item pattern normalization of the input patterns,
    \item pattern normalization after mapping into the associative space, 
    \item no pattern normalization.
\end{itemize}

\subsubsection{Usage}
\label{sec:hopfield_layer_usage}
As outlined in Sec.~\ref{sec:C_Introduction}, there
are a variety of possible use cases for the Hopfield layer, 
e.g.\ to build memory networks or transformer models.
The goal of the implementation is therefore
to provide an easy to use Hopfield module 
that can be used in a wide range of applications, 
be it as part of a larger architecture or as a standalone module.
Consequently, the focus of the Hopfield layer interface is set
on its core parameters: the association of two sets,
the scaling parameter $\beta$, 
the maximum number of updates, 
the dimension of the associative space,
the possible usage of static patterns,
and the pattern normalization.
The integration into the PyTorch framework is built such
that with all the above functionalities disabled,
the ``HopfieldEncoderLayer'' and
the ``HopfieldDecoderLayer'', both extensions of the
Hopfield module,
can be used as a one-to-one plug-in replacement
for the \textit{TransformerEncoderLayer} 
and the \textit{TransformerDecoderLayer}, respectively,
of the PyTorch transformer module.

The Hopfield layer can be used to implement or to substitute different layers:
\begin{itemize}
    \item \textbf{Pooling layers:} We consider the Hopfield layer as a pooling layer if only
    one static state (query) pattern exists. Then, it is de facto
    a pooling over the sequence, which results from the softmax values
    applied on the stored patterns. 
    Therefore, our Hopfield layer can act as a pooling layer.
    \item \textbf{Permutation equivariant layers:} Our Hopfield layer
    can be used as a plug-in replacement for permutation equivariant layers. Since the Hopfield layer is an associative memory it 
    assumes no dependency between the input patterns.
    \item \textbf{GRU \& LSTM layers:} Our Hopfield layer can be used as a plug-in replacement for GRU \& LSTM layers. Optionally, for substituting 
    GRU \& LSTM layers,
    positional encoding might be considered.
    \item \textbf{Attention layers:} Our Hopfield layer can act as an attention layer, where state (query) and stored (key) patterns are different,
    and need to be associated.
    \item Finally, the extensions of the Hopfield layer
    are able to operate as a self-attention layer 
    (HopfieldEncoderLayer) and as cross-attention layer
    (HopfieldDecoderLayer), 
    as described in  \citep{Vaswani:17}.
    As such, it can be used as building block of 
    transformer-based or general architectures.
\end{itemize}

\clearpage
\bibliography{joint_bib}

\begin{thebibliography}{119}
\providecommand{\natexlab}[1]{#1}
\providecommand{\url}[1]{\texttt{#1}}
\expandafter\ifx\csname urlstyle\endcsname\relax
  \providecommand{\doi}[1]{doi: #1}\else
  \providecommand{\doi}{doi: \begingroup \urlstyle{rm}\Url}\fi

\bibitem[Abu-Mostafa \& StJacques(1985)Abu-Mostafa and
  StJacques]{Abu-Mostafa:85}
Y.~Abu-Mostafa and J.-M. StJacques.
\newblock Information capacity of the {Hopfield} model.
\newblock \emph{IEEE Transactions on Information Theory}, 31, 1985.
\newblock \doi{10.1109/tit.1985.1057069}.

\bibitem[Agrawal et~al.(1993)Agrawal, Imieliundefinedski, and
  Swami]{Agrawal:93}
R.~Agrawal, T.~Imieliundefinedski, and A.~Swami.
\newblock Mining association rules between sets of items in large databases.
\newblock \emph{SIGMOD Rec.}, 22\penalty0 (2):\penalty0 207–216, 1993.
\newblock \doi{10.1145/170036.170072}.

\bibitem[Akbar et~al.(2019)Akbar, Robert, Pavlovi{\'c}, Jeliazkov, Snapkov,
  Slabodkin, Weber, Scheffer, Miho, Haff, et~al.]{Akbar:19}
R.~Akbar, P.~A. Robert, M.~Pavlovi{\'c}, J.~R. Jeliazkov, I.~Snapkov,
  A.~Slabodkin, C.~R. Weber, L.~Scheffer, E.~Miho, I.~H. Haff, et~al.
\newblock A compact vocabulary of paratope-epitope interactions enables
  predictability of antibody-antigen binding.
\newblock \emph{bioRxiv}, 2019.

\bibitem[Alzahrani \& Salem(2018)Alzahrani and Salem]{Alzahrani:18}
F.~Alzahrani and A.~Salem.
\newblock Sharp bounds for the lambert $w$ function.
\newblock \emph{Integral Transforms and Special Functions}, 29\penalty0
  (12):\penalty0 971--978, 2018.

\bibitem[Andrews et~al.(2003)Andrews, Tsochantaridis, and Hofmann]{Andrews:03}
S.~Andrews, I.~Tsochantaridis, and T.~Hofmann.
\newblock Support vector machines for multiple-instance learning.
\newblock In S.~Becker, S.~Thrun, and K.~Obermayer (eds.), \emph{Advances in
  Neural Information Processing Systems 15}, pp.\  577--584. MIT Press, 2003.

\bibitem[Ba et~al.(2016{\natexlab{a}})Ba, Hinton, Mnih, Leibo, and
  Ionescu]{Ba:16}
J.~Ba, G.~E. Hinton, V.~Mnih, J.~Z. Leibo, and C.~Ionescu.
\newblock Using fast weights to attend to the recent past.
\newblock In D.~D. Lee, M.~Sugiyama, U.~V. Luxburg, I.~Guyon, and R.~Garnett
  (eds.), \emph{Advances in Neural Information Processing Systems 29}, pp.\
  4331--4339. Curran Associates, Inc., 2016{\natexlab{a}}.

\bibitem[Ba et~al.(2016{\natexlab{b}})Ba, Hinton, Mnih, Leibo, and
  Ionescu]{Ba:16arxiv}
J.~Ba, G.~E. Hinton, V.~Mnih, J.~Z. Leibo, and C.~Ionescu.
\newblock Using fast weights to attend to the recent past.
\newblock \emph{ArXiv}, 1610.06258, 2016{\natexlab{b}}.

\bibitem[Bahdanau et~al.(2014)Bahdanau, Cho, and Bengio]{Bahdanau:14}
D.~Bahdanau, K.~Cho, and Y.~Bengio.
\newblock Neural machine translation by jointly learning to align and
  translate.
\newblock \emph{ArXiv}, 1409.0473, 2014.
\newblock appeared in ICRL 2015.

\bibitem[Banino et~al.(2020)Banino, Badia, K{\"{o}}ster, Chadwick, Zambaldi,
  Hassabis, Barry, Botvinick, Kumaran, and Blundell]{Banino:20}
A.~Banino, A.~P. Badia, R.~K{\"{o}}ster, M.~J. Chadwick, V.~Zambaldi,
  D.~Hassabis, C.~Barry, M.~Botvinick, D.~Kumaran, and C.~Blundell.
\newblock {MEMO:} a deep network for flexible combination of episodic memories.
\newblock \emph{ArXiv}, 2001.10913, 2020.

\bibitem[Barra et~al.(2018)Barra, Beccaria, and Fachechi]{Barra:18}
A.~Barra, M.~Beccaria, and A.~Fachechi.
\newblock A new mechanical approach to handle generalized {Hopfield} neural
  networks.
\newblock \emph{Neural Networks}, 106:\penalty0 205--222, 2018.
\newblock \doi{10.1016/j.neunet.2018.07.010}.

\bibitem[Bauschke \& Combettes(2017)Bauschke and Combettes]{Bauschke:17}
H.~H. Bauschke and P.~L. Combettes.
\newblock \emph{Convex Analysis and Monotone Operator Theory in {Hilbert}
  Spaces}.
\newblock Cham: Springer International Publishing, 2nd edition, 2017.
\newblock ISBN 978-3-319-48310-8.
\newblock \doi{10.1007/978-3-319-48311-5}.

\bibitem[Boyd \& Vandenberghe(2009)Boyd and Vandenberghe]{Boyd:09}
S.~Boyd and L.~Vandenberghe.
\newblock \emph{Convex Optimization}.
\newblock Cambridge University Press, 7th edition, 2009.
\newblock ISBN 978-0-521-83378-3.

\bibitem[Brauchart et~al.(2018)Brauchart, Reznikov, Saff, Sloan, Wang, and
  Womersley]{Brauchart:18}
J.~S. Brauchart, A.~B. Reznikov, E.~B. Saff, I.~H. Sloan, Y.~G. Wang, and R.~S.
  Womersley.
\newblock Random point sets on the sphere - hole radii, covering, and
  separation.
\newblock \emph{Experimental Mathematics}, 27\penalty0 (1):\penalty0 62--81,
  2018.
\newblock \doi{10.1080/10586458.2016.1226209}.

\bibitem[Breiman(2001)]{Breiman:01}
L.~Breiman.
\newblock Random forests.
\newblock \emph{Machine Learning}, 45\penalty0 (1):\penalty0 5--32, 2001.
\newblock \doi{10.1023/A:1010933404324}.

\bibitem[Bruck \& Roychowdhury(1990)Bruck and Roychowdhury]{Bruck:90}
J.~Bruck and V.~P. Roychowdhury.
\newblock On the number of spurious memories in the {Hopfield} model.
\newblock \emph{IEEE Transactions on Information Theory}, 36\penalty0
  (2):\penalty0 393--397, 1990.

\bibitem[Cai et~al.(2013)Cai, Fan, and Jiang]{Cai:13}
T.~Cai, J.~Fan, and T.~Jiang.
\newblock Distributions of angles in random packing on spheres.
\newblock \emph{Journal of Machine Learning Research}, 14\penalty0
  (21):\penalty0 1837--1864, 2013.

\bibitem[Carbonneau et~al.(2018)Carbonneau, Cheplygina, Granger, and
  Gagnon]{Carbonneau:18}
M.-A. Carbonneau, V.~Cheplygina, E.~Granger, and G.~Gagnon.
\newblock Multiple instance learning: a survey of problem characteristics and
  applications.
\newblock \emph{Pattern Recognition}, 77:\penalty0 329--353, 2018.

\bibitem[Carbonneau et~al.(2016)Carbonneau, Granger, Raymond, and
  Gagnon]{Carbonneau:16}
Marc-André Carbonneau, Eric Granger, Alexandre~J. Raymond, and Ghyslain
  Gagnon.
\newblock Robust multiple-instance learning ensembles using random subspace
  instance selection.
\newblock \emph{Pattern Recognition}, 58:\penalty0 83 -- 99, 2016.
\newblock ISSN 0031-3203.
\newblock \doi{https://doi.org/10.1016/j.patcog.2016.03.035}.
\newblock URL
  \url{http://www.sciencedirect.com/science/article/pii/S0031320316300346}.

\bibitem[Carreira-Perpi{\~{n}}{\'a}n \&
  Williams(2003)Carreira-Perpi{\~{n}}{\'a}n and
  Williams]{Carreira-Perpinan:03tr}
M.~Carreira-Perpi{\~{n}}{\'a}n and C.~K.~I. Williams.
\newblock An isotropic {Gaussian} mixture can have more modes than components.
\newblock Technical Report EDI-INF-RR-0185, The University of Edinburgh, School
  of Informatics, 2003.

\bibitem[Carta et~al.(2020)Carta, Sperduti, and Bacciu]{Carta:20}
A.~Carta, A.~Sperduti, and D.~Bacciu.
\newblock Encoding-based memory modules for recurrent neural networks.
\newblock \emph{ArXiv}, 2001.11771, 2020.

\bibitem[Chen \& Guestrin(2016)Chen and Guestrin]{Chen:16}
T.~Chen and C.~Guestrin.
\newblock {XGBoost}: A scalable tree boosting system.
\newblock In \emph{Proceedings of the 22nd ACM SIGKDD International Conference
  on Knowledge Discovery and Data Mining}, pp.\  785--794. ACM, 2016.
\newblock \doi{10.1145/2939672.2939785}.

\bibitem[Chen et~al.(2006)Chen, Bi, and Wang]{Chen:06}
Y.~Chen, J.~Bi, and J.~Z. Wang.
\newblock {MILES}: Multiple-instance learning via embedded instance selection.
\newblock \emph{IEEE Transactions on Pattern Analysis and Machine
  Intelligence}, 28\penalty0 (12):\penalty0 1931--1947, 2006.

\bibitem[Cheplygina et~al.(2016)Cheplygina, Tax, and Loog]{Cheplygina:16}
V~Cheplygina, DM~Tax, and M~Loog.
\newblock Dissimilarity-based ensembles for multiple instance learning.
\newblock \emph{IEEE transactions on neural networks and learning systems},
  27\penalty0 (6):\penalty0 1379, 2016.

\bibitem[Cho et~al.(2014)Cho, vanMerri\"{e}nboer, Gulcehre, Bahdanau, Bougares,
  Schwenk, and Bengio]{Cho:14}
K.~Cho, B.~vanMerri\"{e}nboer, C.~Gulcehre, D.~Bahdanau, F.~Bougares,
  H.~Schwenk, and Y.~Bengio.
\newblock Learning phrase representations using {RNN} encoder{--}decoder for
  statistical machine translation.
\newblock In \emph{Proceedings of the Conference on Empirical Methods in
  Natural Language Processing ({EMNLP})}, pp.\  1724--1734. Association for
  Computational Linguistics, 2014.
\newblock \doi{10.3115/v1/D14-1179}.

\bibitem[Clark et~al.(2020)Clark, Luong, Le, and Manning]{Clark:20}
K.~Clark, M.-T. Luong, Q.~V. Le, and C.~D. Manning.
\newblock {ELECTRA}: Pre-training text encoders as discriminators rather than
  generators.
\newblock \emph{ArXiv}, 2003.10555, 2020.
\newblock appeared in ICLR 2020.

\bibitem[Cortes \& Vapnik(1995)Cortes and Vapnik]{Cortes:95}
C.~Cortes and V.~Vapnik.
\newblock Support-vector networks.
\newblock \emph{Machine learning}, 20\penalty0 (3):\penalty0 273--297, 1995.

\bibitem[Crisanti et~al.(1986)Crisanti, Amit, and Gutfreund]{Crisanti:86}
A.~Crisanti, D.~J. Amit, and H.~Gutfreund.
\newblock Saturation level of the {Hopfield} model for neural network.
\newblock \emph{Europhysics Letters (EPL)}, 2\penalty0 (4):\penalty0 337--341,
  1986.
\newblock \doi{10.1209/0295-5075/2/4/012}.

\bibitem[Danihelka et~al.(2016)Danihelka, Wayne, Uria, Kalchbrenner, and
  Graves]{Danihelka:16}
I.~Danihelka, G.~Wayne, B.~Uria, N.~Kalchbrenner, and A.~Graves.
\newblock Associative long short-term memory.
\newblock In M.~F. Balcan and K.~Q. Weinberger (eds.), \emph{Proceedings of The
  33rd International Conference on Machine Learning}, volume~48 of
  \emph{Proceedings of Machine Learning Research}, pp.\  1986--1994, New York,
  USA, 2016.

\bibitem[Daniluk et~al.(2017)Daniluk, Rockt{\"{a}}schel, Welbl, and
  Riedel]{Daniluk:17}
M.~Daniluk, T.~Rockt{\"{a}}schel, J.~Welbl, and S.~Riedel.
\newblock Frustratingly short attention spans in neural language modeling.
\newblock \emph{ArXiv}, 1702.04521, 2017.
\newblock appeared in ICRL 2017.

\bibitem[Dehghani et~al.(2018)Dehghani, Gouws, Vinyals, Uszkoreit, and
  Kaiser]{Dehghani:18}
M.~Dehghani, S.~Gouws, O.~Vinyals, J.~Uszkoreit, and L.~Kaiser.
\newblock Universal transformers.
\newblock \emph{ArXiv}, 1807.03819, 2018.
\newblock Published at ICLR 2019.

\bibitem[Demircigil et~al.(2017)Demircigil, Heusel, L{\"{o}}we, Upgang, and
  Vermet]{Demircigil:17}
M.~Demircigil, J.~Heusel, M.~L{\"{o}}we, S.~Upgang, and F.~Vermet.
\newblock On a model of associative memory with huge storage capacity.
\newblock \emph{Journal of Statistical Physics}, 168\penalty0 (2):\penalty0
  288--299, 2017.

\bibitem[Devlin et~al.(2018)Devlin, Chang, Lee, and Toutanova]{Devlin:18}
J.~Devlin, M.-W. Chang, K.~Lee, and K.~Toutanova.
\newblock {BERT:} pre-training of deep bidirectional transformers for language
  understanding.
\newblock \emph{ArXiv}, 1810.04805, 2018.

\bibitem[Devlin et~al.(2019)Devlin, Chang, Lee, and Toutanova]{Devlin:19}
J.~Devlin, M.-W. Chang, K.~Lee, and K.~Toutanova.
\newblock {BERT:} pre-training of deep bidirectional transformers for language
  understanding.
\newblock In \emph{Proceedings of the 2019 Conference of the North American
  Chapter of the Association for Computational Linguistics: Human Language
  Technologies, Volume 1 (Long and Short Papers)}, pp.\  4171--4186.
  Association for Computational Linguistics, 2019.

\bibitem[Dietterich et~al.(1997)Dietterich, Lathrop, and
  Lozano-P{\'e}rez]{Dietterich:97}
T.~G. Dietterich, R.~H. Lathrop, and T.~Lozano-P{\'e}rez.
\newblock Solving the multiple instance problem with axis-parallel rectangles.
\newblock \emph{Artificial Intelligence}, 89\penalty0 (1-2):\penalty0 31--71,
  1997.

\bibitem[Emerson et~al.(2017)Emerson, DeWitt, Vignali, Gravley, Hu, Osborne,
  Desmarais, Klinger, Carlson, Hansen, et~al.]{Emerson:17}
R.~O. Emerson, W.~S. DeWitt, M.~Vignali, J.~Gravley, J.~K. Hu, E.~J. Osborne,
  C.~Desmarais, M.~Klinger, C.~S. Carlson, J.~A. Hansen, et~al.
\newblock Immunosequencing identifies signatures of cytomegalovirus exposure
  history and {HLA-mediated} effects on the {T} cell repertoire.
\newblock \emph{Nature Genetics}, 49\penalty0 (5):\penalty0 659, 2017.

\bibitem[Fern{\'a}ndez-Delgado et~al.(2014)Fern{\'a}ndez-Delgado, Cernadas,
  Barro, and Amorim]{Fernandez:14}
M.~Fern{\'a}ndez-Delgado, E.~Cernadas, S.~Barro, and D.~Amorim.
\newblock Do we need hundreds of classifiers to solve real world classification
  problems?
\newblock \emph{The Journal of Machine Learning Research}, 15\penalty0
  (1):\penalty0 3133--3181, 2014.

\bibitem[Folli et~al.(2017)Folli, Leonetti, and Ruocco]{Folli:17}
V.~Folli, M.~Leonetti, and G.~Ruocco.
\newblock On the maximum storage capacity of the {Hopfield} model.
\newblock \emph{Frontiers in Computational Neuroscience}, 10\penalty0 (144),
  2017.
\newblock \doi{10.3389/fncom.2016.00144}.

\bibitem[Gao \& Pavel(2017)Gao and Pavel]{Gao:17}
B.~Gao and L.~Pavel.
\newblock On the properties of the softmax function with application in game
  theory and reinforcement learning.
\newblock \emph{ArXiv}, 1704.00805, 2017.

\bibitem[Garling(2017)]{Garling:17}
D.~J.~H. Garling.
\newblock \emph{Analysis on {Polish} Spaces and an Introduction to Optimal
  Transportation}.
\newblock London Mathematical Society Student Texts. Cambridge University
  Press, 2017.
\newblock ISBN 1108421571.
\newblock \doi{10.1017/9781108377362}.

\bibitem[Gilmer et~al.(2017)Gilmer, Schoenholz, Riley, Vinyals, and
  Dahl]{Gilmer:17}
J.~Gilmer, S.~S. Schoenholz, P.~F. Riley, O.~Vinyals, and G.~E. Dahl.
\newblock Neural message passing for quantum chemistry.
\newblock In \emph{Proceedings of the 34th International Conference on Machine
  Learning (ICML)}, volume~70, pp.\  1263–1272. JMLR.org, 2017.

\bibitem[Graves et~al.(2014)Graves, Wayne, and Danihelka]{Graves:14}
A.~Graves, G.~Wayne, and I.~Danihelka.
\newblock Neural turing machines.
\newblock \emph{ArXiv}, 1410.5401, 2014.

\bibitem[Guttenberg et~al.(2016)Guttenberg, Virgo, Witkowski, Aoki, and
  Kanai]{Guttenberg:16}
N.~Guttenberg, N.~Virgo, O.~Witkowski, H.~Aoki, and R.~Kanai.
\newblock Permutation-equivariant neural networks applied to dynamics
  prediction.
\newblock \emph{arXiv}, 1612.04530, 2016.

\bibitem[Hertz et~al.(1991)Hertz, Krogh, and Palmer]{Hertz:91}
J.~Hertz, A.~Krogh, and R.~G. Palmer.
\newblock \emph{Introduction to the Theory of Neural Computation}.
\newblock Addison-Wesley Longman Publishing Co., Inc., Redwood City, CA, 1991.
\newblock ISBN 0201503956.

\bibitem[Hochreiter(1991)]{Hochreiter:91a}
S.~Hochreiter.
\newblock {Untersuchungen zu dynamischen neuronalen Netzen. Diploma thesis,
  Institut f\"{u}r Informatik, Lehrstuhl Prof. Brauer, Technische
  Universit\"{a}t M\"{u}nchen}, 1991.
\newblock Advisor: J. Schmidhuber.

\bibitem[Hochreiter \& Schmidhuber(1997)Hochreiter and
  Schmidhuber]{Hochreiter:97}
S.~Hochreiter and J.~Schmidhuber.
\newblock Long short-term memory.
\newblock \emph{Neural Comput.}, 9\penalty0 (8):\penalty0 1735--1780, 1997.

\bibitem[Hoorfar \& Hassani(2008)Hoorfar and Hassani]{Hoorfar:08}
A.~Hoorfar and M.~Hassani.
\newblock Inequalities on the {Lambert} $w$ function and hyperpower function.
\newblock \emph{Journal of Inequalities in Pure and Applied Mathematics},
  9\penalty0 (2):\penalty0 1–5, 2008.

\bibitem[Hopfield(1982)]{Hopfield:82}
J.~J. Hopfield.
\newblock Neural networks and physical systems with emergent collective
  computational abilities.
\newblock \emph{Proceedings of the National Academy of Sciences}, 79\penalty0
  (8):\penalty0 2554--2558, 1982.

\bibitem[Hopfield(1984)]{Hopfield:84}
J.~J. Hopfield.
\newblock Neurons with graded response have collective computational properties
  like those of two-state neurons.
\newblock \emph{Proceedings of the National Academy of Sciences}, 81\penalty0
  (10):\penalty0 3088--3092, 1984.
\newblock \doi{10.1073/pnas.81.10.3088}.

\bibitem[Ilse et~al.(2018)Ilse, Tomczak, and Welling]{Ilse:18}
M.~Ilse, J.~M. Tomczak, and M.~Welling.
\newblock Attention-based deep multiple instance learning.
\newblock \emph{International Conference on Machine Learning (ICML)}, pp.\
  3376--3391, 2018.

\bibitem[Ilse et~al.(2020)Ilse, Tomczak, and Welling]{Ilse:20}
M.~Ilse, J.~M. Tomczak, and M.~Welling.
\newblock Deep multiple instance learning for digital histopathology.
\newblock In \emph{Handbook of Medical Image Computing and Computer Assisted
  Intervention}, pp.\  521--546. Elsevier, 2020.

\bibitem[Jiang et~al.(2020)Jiang, Wu, Hsieh, Chen, Liao, Wang, Shen, Cao, Wu,
  and Hou]{Jiang:20}
D.~Jiang, Z.~Wu, C.-Y. Hsieh, G.~Chen, B.~Liao, Z.~Wang, C.~Shen, D.~Cao,
  J.~Wu, and T.~Hou.
\newblock Could graph neural networks learn better molecular representation for
  drug discovery? a comparison study of descriptor-based and graph-based
  models.
\newblock \emph{Journal of Cheminformatics}, 2020.
\newblock \doi{10.21203/rs.3.rs-81439/v1}.

\bibitem[Kandemir et~al.(2014)Kandemir, Zhang, and Hamprecht]{Kandemir:14}
M.~Kandemir, C.~Zhang, and F.~A. Hamprecht.
\newblock Empowering multiple instance histopathology cancer diagnosis by cell
  graphs.
\newblock In \emph{International Conference on Medical Image Computing and
  Computer-Assisted Intervention}, pp.\  228--235. Springer, 2014.

\bibitem[Khan et~al.(2018)Khan, Arif, Siddique, and Oishe]{Khan:18}
M.~M.~R. Khan, R.~B. Arif, M.~A.~B. Siddique, and M.~R. Oishe.
\newblock Study and observation of the variation of accuracies of {KNN}, {SVM},
  {LMNN}, {ENN} algorithms on eleven different datasets from {UCI} machine
  learning repository.
\newblock In \emph{4th International Conference on Electrical Engineering and
  Information \& Communication Technology (iCEEiCT)}, pp.\  124--129. IEEE,
  2018.

\bibitem[Kipf \& Welling(2016)Kipf and Welling]{Kipf:16}
T.~N. Kipf and M.~Welling.
\newblock Semi-supervised classification with graph convolutional networks.
\newblock \emph{ArXiv}, 1609.02907, 2016.
\newblock in International Conference On Learning Representations (ICLR) 2017.

\bibitem[Klambauer et~al.(2017{\natexlab{a}})Klambauer, Unterthiner, Mayr, and
  Hochreiter]{Klambauer:17}
G.~Klambauer, T.~Unterthiner, A.~Mayr, and S.~Hochreiter.
\newblock Self-normalizing neural networks.
\newblock In \emph{Advances in Neural Information Processing Systems}, pp.\
  971--980, 2017{\natexlab{a}}.

\bibitem[Klambauer et~al.(2017{\natexlab{b}})Klambauer, Unterthiner, Mayr, and
  Hochreiter]{Klambauer:17arxiv}
G.~Klambauer, T.~Unterthiner, A.~Mayr, and S.~Hochreiter.
\newblock Self-normalizing neural networks.
\newblock \emph{ArXiv}, 1706.02515, 2017{\natexlab{b}}.

\bibitem[Koiran(1994)]{Koiran:94}
P.~Koiran.
\newblock Dynamics of discrete time, continuous state {Hopfield} networks.
\newblock \emph{Neural Computation}, 6\penalty0 (3):\penalty0 459--468, 1994.
\newblock \doi{10.1162/neco.1994.6.3.459}.

\bibitem[Korshunova et~al.(2018)Korshunova, Degrave, Huszar, Gal, Gretton, and
  Dambre]{Korshunova:18}
I.~Korshunova, J.~Degrave, F.~Huszar, Y.~Gal, A.~Gretton, and J.~Dambre.
\newblock {BRUNO}: A deep recurrent model for exchangeable data.
\newblock In S.~Bengio, H.~Wallach, H.~Larochelle, K.~Grauman, N.~Cesa-Bianchi,
  and R.~Garnett (eds.), \emph{Advances in Neural Information Processing
  Systems 31}, pp.\  7190--7198. Curran Associates, Inc., 2018.

\bibitem[Krotov \& Hopfield(2016)Krotov and Hopfield]{Krotov:16}
D.~Krotov and J.~J. Hopfield.
\newblock Dense associative memory for pattern recognition.
\newblock In D.~D. Lee, M.~Sugiyama, U.~V. Luxburg, I.~Guyon, and R.~Garnett
  (eds.), \emph{Advances in Neural Information Processing Systems}, pp.\
  1172--1180. Curran Associates, Inc., 2016.

\bibitem[Krotov \& Hopfield(2018)Krotov and Hopfield]{Krotov:18}
D.~Krotov and J.~J. Hopfield.
\newblock Dense associative memory is robust to adversarial inputs.
\newblock \emph{Neural Computation}, 30\penalty0 (12):\penalty0 3151--3167,
  2018.

\bibitem[Krotov \& Hopfield(2020)Krotov and Hopfield]{Krotov:20}
D.~Krotov and J.~J. Hopfield.
\newblock Large associative memory problem in neurobiology and machine
  learning.
\newblock \emph{ArXiv}, 2008.06996, 2020.

\bibitem[K{\"u}{\c{c}}{\"u}ka{\c{s}}c{\i} \&
  Baydo{\u{g}}an(2018)K{\"u}{\c{c}}{\"u}ka{\c{s}}c{\i} and
  Baydo{\u{g}}an]{Kuccukacsci:18}
E.~{\c{S}}. K{\"u}{\c{c}}{\"u}ka{\c{s}}c{\i} and M.~G. Baydo{\u{g}}an.
\newblock Bag encoding strategies in multiple instance learning problems.
\newblock \emph{Information Sciences}, 467:\penalty0 559--578, 2018.

\bibitem[Kuhn et~al.(2016)Kuhn, Letunic, Jensen, and Bork]{Kuhn:16}
M.~Kuhn, I.~Letunic, L.~J. Jensen, and P.~Bork.
\newblock The {SIDER} database of drugs and side effects.
\newblock \emph{Nucleic Acids Research}, 44\penalty0 (D1):\penalty0
  D1075--D1079, 2016.
\newblock \doi{10.1093/nar/gkv1075}.

\bibitem[LeCun et~al.(2015)LeCun, Bengio, and Hinton]{LeCun:15}
Y.~LeCun, Y.~Bengio, and G.~Hinton.
\newblock Deep learning.
\newblock \emph{Nature}, 521:\penalty0 436--444, 2015.

\bibitem[Lipp \& Boyd(2016)Lipp and Boyd]{Lipp:16}
T.~Lipp and S.~Boyd.
\newblock Variations and extension of the convex–concave procedure.
\newblock \emph{Optimization and Engineering}, 17\penalty0 (2):\penalty0
  263--287, 2016.
\newblock \doi{10.1007/s11081-015-9294-x}.

\bibitem[Loshchilov \& Hutter(2017)Loshchilov and Hutter]{loshchilov:17}
Ilya Loshchilov and Frank Hutter.
\newblock Decoupled weight decay regularization.
\newblock \emph{arXiv preprint arXiv:1711.05101}, 2017.

\bibitem[Maron \& Lozano-P{\'e}rez(1998)Maron and Lozano-P{\'e}rez]{Maron:98}
O.~Maron and T.~Lozano-P{\'e}rez.
\newblock A framework for multiple-instance learning.
\newblock In M.~I. Jordan, M.~J. Kearns, and S.~A. Solla (eds.), \emph{Advances
  in Neural Information Processing Systems}, pp.\  570--576. MIT Press, 1998.

\bibitem[Martins et~al.(2012)Martins, Teixeira, Pinheiro, and
  Falcao]{Martins:12}
I.~F. Martins, A.~L. Teixeira, L.~Pinheiro, and A.~O. Falcao.
\newblock A {Bayesian} approach to in silico blood-brain barrier penetration
  modeling.
\newblock \emph{Journal of Chemical Information and Modeling}, 52\penalty0
  (6):\penalty0 1686--1697, 2012.
\newblock \doi{10.1021/ci300124c}.

\bibitem[Mazza(1997)]{Mazza:93}
C.~Mazza.
\newblock On the storage capacity of nonlinear neural networks.
\newblock \emph{Neural Networks}, 10\penalty0 (4):\penalty0 593--597, 1997.
\newblock \doi{10.1016/S0893-6080(97)00017-8}.

\bibitem[McEliece et~al.(1987)McEliece, Posner, Rodemich, and
  Venkatesh]{McEliece:87}
R.~J. McEliece, E.~C. Posner, E.~R. Rodemich, and S.~S. Venkatesh.
\newblock The capacity of the {Hopfield} associative memory.
\newblock \emph{IEEE Trans. Inf. Theor.}, 33\penalty0 (4):\penalty0 461--482,
  1987.
\newblock \doi{10.1109/TIT.1987.1057328}.

\bibitem[Meyer(1976)]{Meyer:76}
R.~R. Meyer.
\newblock Sufficient conditions for the convergence of monotonic mathematical
  programming algorithms.
\newblock \emph{Journal of Computer and System Sciences}, 12\penalty0
  (1):\penalty0 108--121, 1976.
\newblock \doi{10.1016/S0022-0000(76)80021-9}.

\bibitem[Olver et~al.(2010)Olver, Lozier, Boisvert, and Clark]{Olver:10}
F.~W.~J. Olver, D.~W. Lozier, R.~F. Boisvert, and C.~W. Clark.
\newblock \emph{{NIST} handbook of mathematical functions}.
\newblock Cambridge University Press, 1 pap/cdr edition, 2010.
\newblock ISBN 9780521192255.

\bibitem[Paszke et~al.(2017)Paszke, Gross, Chintala, Chanan, Yang, DeVito, Lin,
  Desmaison, Antiga, and Lerer]{Paszke:17}
A.~Paszke, S.~Gross, S.~Chintala, G.~Chanan, E.~Yang, Z.~DeVito, Z.~Lin,
  A.~Desmaison, L.~Antiga, and A.~Lerer.
\newblock Automatic differentiation in {PyTorch}.
\newblock In \emph{Workshop in Advances in Neural Information Processing
  Systems (NeurIPS)}, 2017.

\bibitem[Paszke et~al.(2019)Paszke, Gross, Massa, Lerer, Bradbury, Chanan,
  Killeen, Lin, Gimelshein, Antiga, et~al.]{Paszke:19}
A.~Paszke, S.~Gross, F.~Massa, A.~Lerer, J.~Bradbury, G.~Chanan, T.~Killeen,
  Z.~Lin, N.~Gimelshein, L.~Antiga, et~al.
\newblock {PyTorch}: An imperative style, high-performance deep learning
  library.
\newblock In \emph{Advances in Neural Information Processing Systems}, pp.\
  8026--8037, 2019.

\bibitem[Qi et~al.(2017{\natexlab{a}})Qi, Su, Kaichun, and Guibas]{Qi:17ieee}
C.~R. Qi, H.~Su, M.~Kaichun, and L.~J. Guibas.
\newblock {PointNet}: Deep learning on point sets for 3d classification and
  segmentation.
\newblock In \emph{IEEE Conference on Computer Vision and Pattern Recognition
  (CVPR)}, pp.\  77--85, 2017{\natexlab{a}}.
\newblock \doi{10.1109/CVPR.2017.16}.

\bibitem[Qi et~al.(2017{\natexlab{b}})Qi, Yi, Su, and Guibas]{Qi:17}
C.~R. Qi, L.~Yi, H.~Su, and L.~J. Guibas.
\newblock {PointNet++}: Deep hierarchical feature learning on point sets in a
  metric space.
\newblock In \emph{31st International Conference on Neural Information
  Processing Systems}, pp.\  5105–5114. Curran Associates Inc.,
  2017{\natexlab{b}}.

\bibitem[Rangarajan et~al.(1996)Rangarajan, Gold, and Mjolsness]{Rangarajan:96}
A.~Rangarajan, S.~Gold, and E.~Mjolsness.
\newblock A novel optimizing network architecture with applications.
\newblock \emph{Neural Computation}, 8\penalty0 (5):\penalty0 1041--1060, 1996.
\newblock \doi{10.1162/neco.1996.8.5.1041}.

\bibitem[Rangarajan et~al.(1999)Rangarajan, Yuille, and
  E.~Mjolsness]{Rangarajan:99}
A.~Rangarajan, A.~Yuille, and Eric E.~Mjolsness.
\newblock Convergence properties of the softassign quadratic assignment
  algorithm.
\newblock \emph{Neural Computation}, 11\penalty0 (6):\penalty0 1455--1474,
  1999.
\newblock \doi{10.1162/089976699300016313}.

\bibitem[Ravanbakhsh et~al.(2016)Ravanbakhsh, Schneider, and
  Poczos]{Ravanbakhsh:16}
S.~Ravanbakhsh, J.~Schneider, and B.~Poczos.
\newblock Deep learning with sets and point clouds.
\newblock \emph{arXiv}, 1611.04500, 2016.

\bibitem[Schlag \& Schmidhuber(2018)Schlag and Schmidhuber]{Schlag:18}
I.~Schlag and J.~Schmidhuber.
\newblock Learning to reason with third order tensor products.
\newblock In S.~Bengio, H.~Wallach, H.~Larochelle, K.~Grauman, N.~Cesa-Bianchi,
  and R.~Garnett (eds.), \emph{Advances in Neural Information Processing
  Systems 31}, pp.\  9981--9993. Curran Associates, Inc., 2018.

\bibitem[Schlag et~al.(2019)Schlag, Smolensky, Fernandez, Jojic, Schmidhuber,
  and Gao]{Schlag:19}
I.~Schlag, P.~Smolensky, R.~Fernandez, N.~Jojic, J.~Schmidhuber, and J.~Gao.
\newblock Enhancing the transformer with explicit relational encoding for math
  problem solving.
\newblock \emph{arXiv}, 1910.06611, 2019.

\bibitem[Schlag et~al.(2021)Schlag, Irie, and Schmidhuber]{Schlag:21}
I.~Schlag, K.~Irie, and J.~Schmidhuber.
\newblock Linear transformers are secretly fast weight memory systems.
\newblock \emph{arXiv}, 2102.11174, 2021.

\bibitem[Schmidhuber(1992)]{Schmidhuber:92fastmem}
J.~Schmidhuber.
\newblock Learning to control fast-weight memories: An alternative to dynamic
  recurrent networks.
\newblock In \emph{Neural Computations, Volume: 4, Issue: 1}, pp.\  131 -- 139.
  MIT Press, 1992.

\bibitem[Schmidhuber(2015)]{Schmidhuber:15}
J.~Schmidhuber.
\newblock Deep learning in neural networks: An overview.
\newblock \emph{Neural Networks}, 61:\penalty0 85--117, 2015.
\newblock \doi{10.1016/j.neunet.2014.09.003}.

\bibitem[Sch\"{o}lkopf \& Smola(2002)Sch\"{o}lkopf and
  Smola]{Schoelkopf:02book}
B.~Sch\"{o}lkopf and A.~J. Smola.
\newblock \emph{Learning with Kernels -- Support Vector Machines,
  Regularization, Optimization, and Beyond}.
\newblock MIT Press, Cambridge, MA, 2002.

\bibitem[Sriperumbudur \& Lanckriet(2009)Sriperumbudur and
  Lanckriet]{Sriperumbudur:09}
B.~K. Sriperumbudur and G.~R. Lanckriet.
\newblock On the convergence of the concave-convex procedure.
\newblock In Y.~Bengio, D.~Schuurmans, J.~D. Lafferty, C.~K.~I. Williams, and
  A.~Culotta (eds.), \emph{Advances in Neural Information Processing Systems
  22}, pp.\  1759--1767. Curran Associates, Inc., 2009.

\bibitem[Subramanian et~al.(2016)Subramanian, Ramsundar, Pande, and
  Denny]{Subramanian:16}
G.~Subramanian, B.~Ramsundar, V.~Pande, and R.~A. Denny.
\newblock Computational modeling of {$\beta$-Secretase 1 (BACE-1)} inhibitors
  using ligand based approaches.
\newblock \emph{Journal of Chemical Information and Modeling}, 56\penalty0
  (10):\penalty0 1936--1949, 2016.
\newblock \doi{10.1021/acs.jcim.6b00290}.

\bibitem[Sukhbaatar et~al.(2015{\natexlab{a}})Sukhbaatar, Szlam, Weston, and
  Fergus]{Sukhbaatar:15}
S.~Sukhbaatar, A.~Szlam, J.~Weston, and R.~Fergus.
\newblock End-to-end memory networks.
\newblock In C.~Cortes, N.~D. Lawrence, D.~D. Lee, M.~Sugiyama, and R.~Garnett
  (eds.), \emph{Advances in Neural Information Processing Systems 28}, pp.\
  2440--2448. Curran Associates, Inc., 2015{\natexlab{a}}.

\bibitem[Sukhbaatar et~al.(2015{\natexlab{b}})Sukhbaatar, Szlam, Weston, and
  Fergus]{Sukhbaatar:15arxiv}
S.~Sukhbaatar, A.~Szlam, J.~Weston, and R.~Fergus.
\newblock End-to-end memory networks.
\newblock \emph{ArXiv}, 1503.08895, 2015{\natexlab{b}}.

\bibitem[Sutton \& Barto(2018)Sutton and Barto]{Sutton:18book}
R.~S. Sutton and A.~G. Barto.
\newblock \emph{Reinforcement Learning: An Introduction}.
\newblock MIT Press, Cambridge, MA, 2 edition, 2018.

\bibitem[Tanaka \& Edwards(1980)Tanaka and Edwards]{Tanaka:80}
F.~Tanaka and S.~F. Edwards.
\newblock Analytic theory of the ground state properties of a spin glass. {I.}
  {Ising} spin glass.
\newblock \emph{Journal of Physics F: Metal Physics}, 10\penalty0
  (12):\penalty0 2769--2778, 1980.
\newblock \doi{10.1088/0305-4608/10/12/017}.

\bibitem[Tay et~al.(2020)Tay, Bahri, Metzler, Juan, Zhao, and Zheng]{Tay:20}
Y.~Tay, D.~Bahri, D.~Metzler, D.-C. Juan, Z.~Zhao, and C.~Zheng.
\newblock Synthesizer: Rethinking self-attention in transformer models.
\newblock \emph{ArXiv}, 2005.00743, 2020.

\bibitem[Toneva \& Wehbe(2019{\natexlab{a}})Toneva and Wehbe]{Toneva:19}
M.~Toneva and L.~Wehbe.
\newblock Interpreting and improving natural-language processing (in machines)
  with natural language-processing (in the brain).
\newblock In H.~Wallach, H.~Larochelle, A.~Beygelzimer, F.~d\textquotesingle
  Alch\'{e}-Buc, E.~Fox, and R.~Garnett (eds.), \emph{Advances in Neural
  Information Processing Systems 32}, pp.\  14954--14964. Curran Associates,
  Inc., 2019{\natexlab{a}}.

\bibitem[Toneva \& Wehbe(2019{\natexlab{b}})Toneva and Wehbe]{Toneva:19arxiv}
M.~Toneva and L.~Wehbe.
\newblock Interpreting and improving natural-language processing (in machines)
  with natural language-processing (in the brain).
\newblock \emph{arXiv}, 1905.11833, 2019{\natexlab{b}}.

\bibitem[Torres et~al.(2002)Torres, Pantic, and H.~J.~Kappen]{Torres:02}
J.~J. Torres, L.~Pantic, and Hilbert H.~J.~Kappen.
\newblock Storage capacity of attractor neural networks with depressing
  synapses.
\newblock \emph{Phys. Rev. E}, 66:\penalty0 061910, 2002.
\newblock \doi{10.1103/PhysRevE.66.061910}.

\bibitem[Vaswani et~al.(2017{\natexlab{a}})Vaswani, Shazeer, Parmar, Uszkoreit,
  Jones, Gomez, Kaiser, and Polosukhin]{Vaswani:17}
A.~Vaswani, N.~Shazeer, N.~Parmar, J.~Uszkoreit, L.~Jones, A.~N. Gomez,
  L.~Kaiser, and I.~Polosukhin.
\newblock Attention is all you need.
\newblock In I.~Guyon, U.~V. Luxburg, S.~Bengio, H.~Wallach, R.~Fergus,
  S.~Vishwanathan, and R.~Garnett (eds.), \emph{Advances in Neural Information
  Processing Systems 30}, pp.\  5998--6008. Curran Associates, Inc.,
  2017{\natexlab{a}}.

\bibitem[Vaswani et~al.(2017{\natexlab{b}})Vaswani, Shazeer, Parmar, Uszkoreit,
  Jones, Gomez, Kaiser, and Polosukhin]{Vaswani:17arxiv}
A.~Vaswani, N.~Shazeer, N.~Parmar, J.~Uszkoreit, L.~Jones, A.~N. Gomez,
  L.~Kaiser, and I.~Polosukhin.
\newblock Attention is all you need.
\newblock \emph{ArXiv}, 1706.03762, 2017{\natexlab{b}}.

\bibitem[Veli\u{c}kovi\'{c} et~al.(2018)Veli\u{c}kovi\'{c}, Cucurull, Casanova,
  Romero, Li\`{o}, and Bengio]{Velickovic:18}
P.~Veli\u{c}kovi\'{c}, G.~Cucurull, A.~Casanova, A.~Romero, P.~Li\`{o}, and
  Y.~Bengio.
\newblock Graph attention networks.
\newblock \emph{arXiv}, 1710.10903, 2018.
\newblock in International Conference On Learning Representations (ICLR) 2018.

\bibitem[Wainberg et~al.(2016)Wainberg, Alipanahi, and Frey]{Wainberg:16}
M.~Wainberg, B.~Alipanahi, and B.~J. Frey.
\newblock Are random forests truly the best classifiers?
\newblock \emph{The Journal of Machine Learning Research}, 17\penalty0
  (1):\penalty0 3837--3841, 2016.

\bibitem[Wainrib \& Touboul(2013)Wainrib and Touboul]{Wainrib:13}
G.~Wainrib and J.~Touboul.
\newblock Topological and dynamical complexity of random neural networks.
\newblock \emph{Phys. Rev. Lett.}, 110:\penalty0 118101, 2013.
\newblock \doi{10.1103/PhysRevLett.110.118101}.

\bibitem[Wang(2000)]{Wang:00}
J.~Wang.
\newblock Solving the multiple-instance problem: A lazy learning approach.
\newblock In \emph{Proceedings of the 17th International Conference on Machine
  Learning (ICML)}, 2000.

\bibitem[Wang et~al.(2018)Wang, Yan, Tang, Bai, and Liu]{Wang:18}
X.~Wang, Y.~Yan, P.~Tang, X.~Bai, and W.~Liu.
\newblock Revisiting multiple instance neural networks.
\newblock \emph{Pattern Recognition}, 74:\penalty0 15--24, 2018.

\bibitem[Weber et~al.(2020)Weber, Akbar, Yermanos, Pavlovi{\'c}, Snapkov,
  Sandve, Reddy, and Greiff]{Weber:19}
C.~R. Weber, R.~Akbar, A.~Yermanos, M.~Pavlovi{\'c}, I.~Snapkov, G.~K. Sandve,
  S.~T. Reddy, and V.~Greiff.
\newblock {immuneSIM:} tunable multi-feature simulation of {B-} and {T-cell}
  receptor repertoires for immunoinformatics benchmarking.
\newblock \emph{Bioinformatics}, 36\penalty0 (11):\penalty0 3594--3596, 2020.
\newblock \doi{10.1093/bioinformatics/btaa158}.

\bibitem[Weston et~al.(2014)Weston, Chopra, and Bordes]{Weston:14}
J.~Weston, S.~Chopra, and A.~Bordes.
\newblock Memory networks.
\newblock \emph{ArXiv}, 1410.3916, 2014.

\bibitem[Widrich et~al.(2020{\natexlab{a}})Widrich, Sch\"{a}fl, Pavlovi{\'c},
  Ramsauer, Gruber, Holzleitner, Brandstetter, Sandve, Greiff, Hochreiter, and
  Klambauer]{Widrich:20}
M.~Widrich, B.~Sch\"{a}fl, M.~Pavlovi{\'c}, H.~Ramsauer, L.~Gruber,
  M.~Holzleitner, J.~Brandstetter, G.~K. Sandve, V.~Greiff, S.~Hochreiter, and
  G.~Klambauer.
\newblock Modern {Hopfield} networks and attention for immune repertoire
  classification.
\newblock \emph{ArXiv}, 2007.13505, 2020{\natexlab{a}}.

\bibitem[Widrich et~al.(2020{\natexlab{b}})Widrich, Sch\"{a}fl, Pavlovi{\'c},
  Ramsauer, Gruber, Holzleitner, Brandstetter, Sandve, Greiff, Hochreiter, and
  Klambauer]{Widrich:20nips}
M.~Widrich, B.~Sch\"{a}fl, M.~Pavlovi{\'c}, H.~Ramsauer, L.~Gruber,
  M.~Holzleitner, J.~Brandstetter, G.~K. Sandve, V.~Greiff, S.~Hochreiter, and
  G.~Klambauer.
\newblock Modern {Hopfield} networks and attention for immune repertoire
  classification.
\newblock In \emph{Advances in Neural Information Processing Systems}. Curran
  Associates, Inc., 2020{\natexlab{b}}.

\bibitem[Wolf et~al.(2019)Wolf, Debut, Sanh, Chaumond, Delangue, Moi, Cistac,
  Rault, Louf, Funtowicz, and Brew]{Wolf:19}
T.~Wolf, L.~Debut, V.~Sanh, J.~Chaumond, C.~Delangue, A.~Moi, P.~Cistac,
  T.~Rault, R.~Louf, M.~Funtowicz, and J.~Brew.
\newblock {HuggingFace's} transformers: State-of-the-art natural language
  processing.
\newblock \emph{ArXiv}, 1910.03771, 2019.

\bibitem[Wu(1983)]{Wu:83}
J.~C.~F. Wu.
\newblock On the convergence properties of the em algorithm.
\newblock \emph{Ann. Statist.}, 11\penalty0 (1):\penalty0 95--103, 1983.
\newblock \doi{10.1214/aos/1176346060}.

\bibitem[Wu et~al.(2018)Wu, Liu, Li, and Wu]{Wu:18}
X.~Wu, X.~Liu, W.~Li, and Q.~Wu.
\newblock Improved expressivity through dendritic neural networks.
\newblock In S.~Bengio, H.~Wallach, H.~Larochelle, K.~Grauman, N.~Cesa-Bianchi,
  and R.~Garnett (eds.), \emph{Advances in Neural Information Processing
  Systems 31}, pp.\  8057--8068. Curran Associates, Inc., 2018.

\bibitem[Wu et~al.(2017)Wu, Ramsundar, Feinberg, Gomes, Geniesse, Pappu,
  Leswing, and Pande]{Wu:17}
Z.~Wu, B.~Ramsundar, E.~N. Feinberg, J.~Gomes, C.~Geniesse, A.~S. Pappu,
  K.~Leswing, and V.~Pande.
\newblock {MoleculeNet}: A benchmark for molecular machine learning.
\newblock \emph{arXiv}, 1703.00564, 2017.

\bibitem[Xiong et~al.(2020)Xiong, Wang, Liu, Zhong, Wan, Li, Li, Luo, Chen,
  Jiang, and Zheng]{Xiong:20}
Z.~Xiong, D.~Wang, X.~Liu, F.~Zhong, X.~Wan, X.~Li, Z.~Li, X.~Luo, K.~Chen,
  H.~Jiang, and M.~Zheng.
\newblock Pushing the boundaries of molecular representation for drug discovery
  with the graph attention mechanism.
\newblock \emph{Journal of Medicinal Chemistry}, 63\penalty0 (16):\penalty0
  8749--8760, 2020.
\newblock \doi{10.1021/acs.jmedchem.9b00959}.

\bibitem[Xu et~al.(2018)Xu, Fan, Xu, Zeng, and Qiao]{Xu:18spider}
Y.~Xu, T.~Fan, M.~Xu, L.~Zeng, and Y.~Qiao.
\newblock {SpiderCNN}: Deep learning on point sets with parameterized
  convolutional filters.
\newblock In V.~Ferrari, M.~Hebert, C.~Sminchisescu, and Y.~Weiss (eds.),
  \emph{European Conference on Computer Vision (ECCV)}, pp.\  90--105. Springer
  International Publishing, 2018.

\bibitem[Yuille \& Rangarajan(2002)Yuille and Rangarajan]{Yuille:02}
A.~L. Yuille and A.~Rangarajan.
\newblock The concave-convex procedure {(CCCP)}.
\newblock In T.~G. Dietterich, S.~Becker, and Z.~Ghahramani (eds.),
  \emph{Advances in Neural Information Processing Systems 14}, pp.\
  1033--1040. MIT Press, 2002.

\bibitem[Yuille \& Rangarajan(2003)Yuille and Rangarajan]{Yuille:03}
A.~L. Yuille and A.~Rangarajan.
\newblock The concave-convex procedure.
\newblock \emph{Neural Computation}, 15\penalty0 (4):\penalty0 915--936, 2003.
\newblock \doi{10.1162/08997660360581958}.

\bibitem[Zaheer et~al.(2017)Zaheer, Kottur, Ravanbakhsh, Poczos, Salakhutdinov,
  and Smola]{Zaheer:17}
M.~Zaheer, S.~Kottur, S.~Ravanbakhsh, B.~Poczos, R.~R. Salakhutdinov, and A.~J.
  Smola.
\newblock Deep sets.
\newblock In I.~Guyon, U.~V. Luxburg, S.~Bengio, H.~Wallach, R.~Fergus,
  S.~Vishwanathan, and R.~Garnett (eds.), \emph{Advances in Neural Information
  Processing Systems 30}, pp.\  3391--3401. Curran Associates, Inc., 2017.

\bibitem[Zangwill(1969)]{Zangwill:69}
W.~I. Zangwill.
\newblock \emph{Nonlinear programming: a unified approach}.
\newblock Prentice-Hall international series in management. Englewood Cliffs,
  N.J., 1969.
\newblock ISBN 9780136235798.

\bibitem[Zhai et~al.(2020)Zhai, Talbott, Bautista, Guestrin, and
  Susskind]{Zhai:20}
S.~Zhai, W.~Talbott, M.~A. Bautista, C.~Guestrin, and J.~M. Susskind.
\newblock Set distribution networks: a generative model for sets of images.
\newblock \emph{arXiv}, 2006.10705, 2020.

\bibitem[Zhang \& Zhou(2017)Zhang and Zhou]{Zhang:17}
W.~Zhang and B.~Zhou.
\newblock Learning to update auto-associative memory in recurrent neural
  networks for improving sequence memorization.
\newblock \emph{ArXiv}, 1709.06493, 2017.

\bibitem[Zhu et~al.(2015)Zhu, Kiros, Zemel, Salakhutdinov, Urtasun, Torralba,
  and Fidler]{Zhu:15}
Y.~Zhu, R.~Kiros, R.~S. Zemel, R.~Salakhutdinov, R.~Urtasun, A.~Torralba, and
  S.~Fidler.
\newblock Aligning books and movies: Towards story-like visual explanations by
  watching movies and reading books.
\newblock \emph{Proceedings of the IEEE international conference on computer
  vision}, pp.\  19--27, 2015.
\newblock arXiv 1506.06724.

\end{thebibliography}
\bibliographystyle{iclr2021_conference}

\end{document}




\maketitle

\newlength{\auxparskip}
\setlength{\auxparskip}{\parskip} 
\setlength{\parskip}{0pt}
\tableofcontents

\newpage 

\section*{Introduction}
This document is a supplement to the
paper "Hopfield Networks is All You Need".

\section{Details on experiments with transformer architectures}

\subsection{Experimental setup: learning dynamics, BERT-small} 
Transformer architectures are known for their high computational demands. 
To investigate the learning dynamics of such a model and at the same time keeping training time manageable, we adopted the BERT-small setting from ELECTRA \cite{clark2020electra}. It has $12$ layers, $4$ heads and 
a reduced hidden size, the sequence length is shortened from $512$ to $128$ tokens and the batch size is reduced from
$256$ to $128$. Additionally, the hidden dimension is reduced from $768$ to $256$ and the embedding dimension is
reduced from $768$ to $128$ \cite{clark2020electra}. The training of such a BERT-small model for $1.45$ million update steps takes roughly four days on a single NVIDIA V100 GPU.

As the code base we use the \textit{transformers} repository from Hugging Face, Inc \cite{Wolf:19}.
We aim to reproduce the dataset of \cite{Devlin:19} as close as possible, which consists of
the English Wikipedia dataset and the Toronto BookCorpus dataset \cite{Zhu:15}. Due to recent copyright claims the later is not publicly available anymore. Therefore, the pre-training experiments use an uncased snapshot of the original BookCorpus dataset. 

\paragraph{Learning dynamics.}

\begin{figure}[htp]
    \centering
    \includegraphics[width=0.9\textwidth]{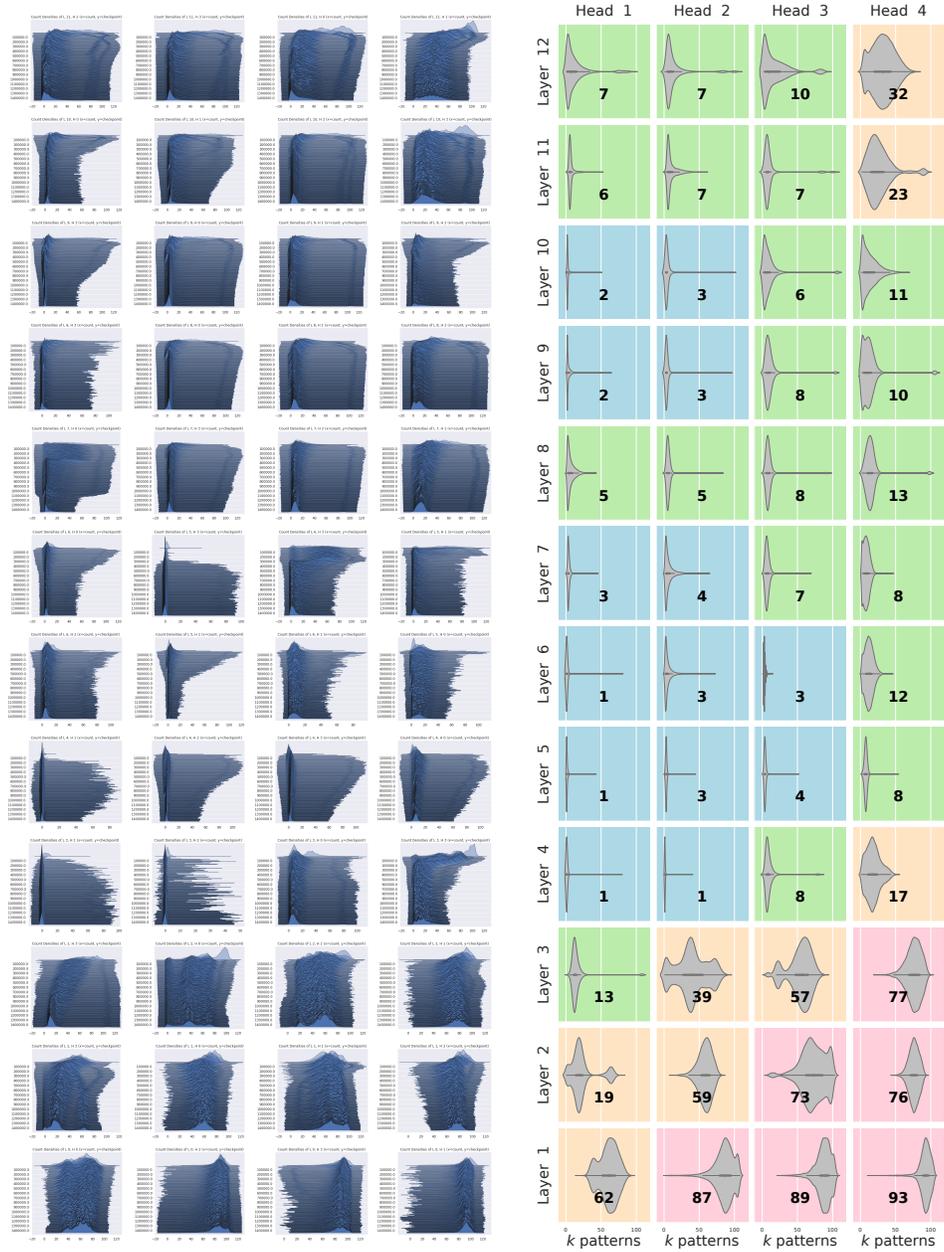}
    \caption{\textbf{Left}: Ridge plots of the distribution of counts $k$ over time for BERT-small \textbf{Right}: Violin plot of counts $k$ after $1,450000$ steps, divided into the four classes from the main paper. The thresholds were adapted to the shorter sequence length.}
    \label{fig:count_small}
\end{figure}

To better understand how operation modes in attention heads develop, we tracked the distribution of counts $k$ (see main paper) over time in a BERT-small model.
At the end of training we visualized the count distribution, grouped into four classes (see Figure~\ref{fig:count_small}). 
The thresholds for the classes were chosen according to the thresholds of Figure~2 in the main paper.
However, they are divided by a factor of $4$ to adapt to the shorter sequence length of $128$ compared to $512$.
From this plot it is clear, that the attention in heads of \textbf{Class IV} commit very early to the operation mode of small metastable stats. 

\begin{figure}
    \centering
    \begin{minipage}[t]{0.49\linewidth}
    \centering
    \includegraphics[width=.9\textwidth]{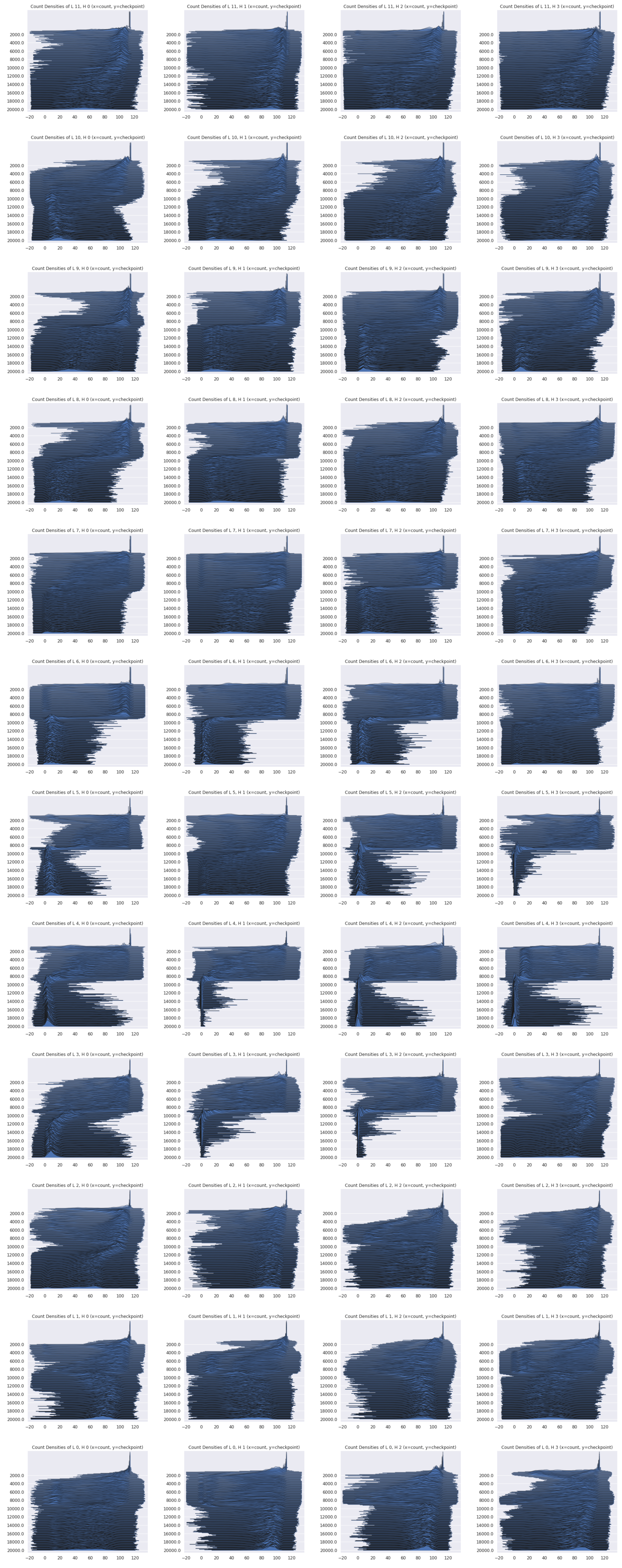} \\
    (a) Densities
    \end{minipage}
    \begin{minipage}[t]{0.49\linewidth}
    \centering
    \includegraphics[width=.9\textwidth]{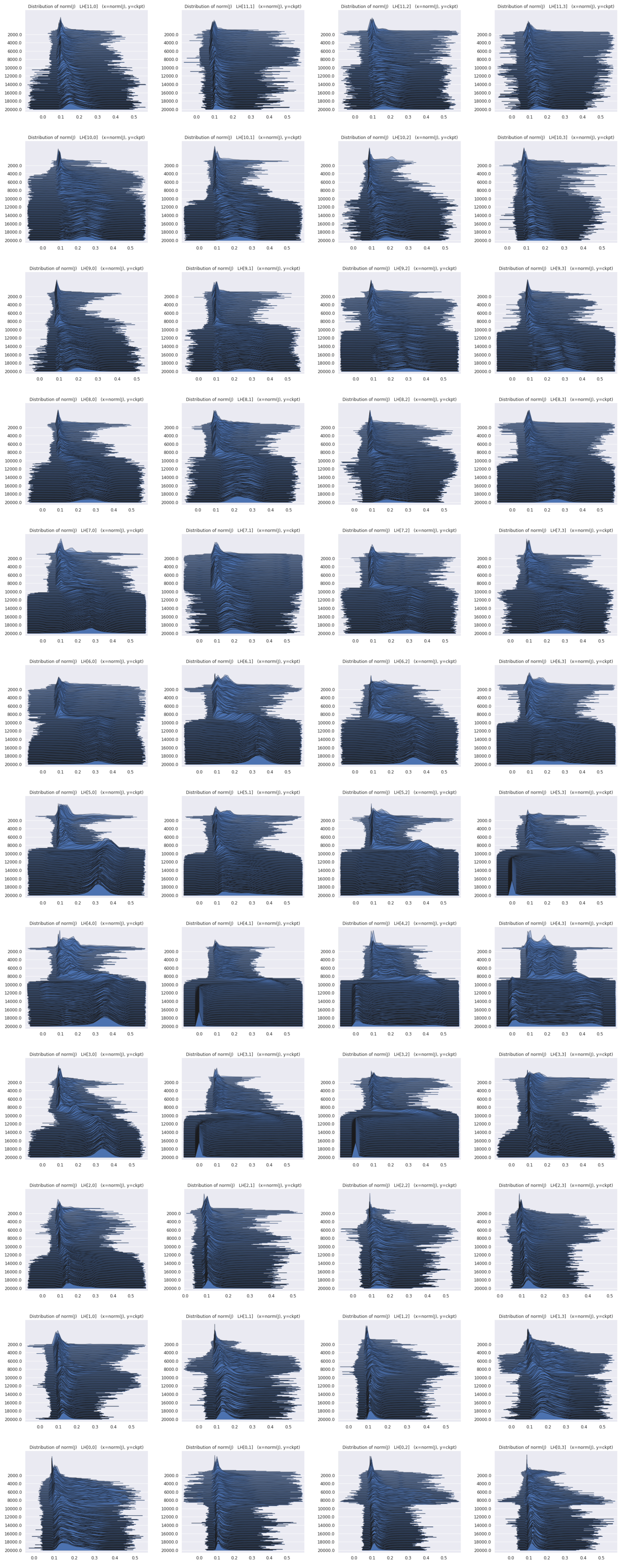} \\ 
    (b) Norm of Jacobian
    \end{minipage}
    \caption{\textbf{(a)}: change of count density during training is depicted for the first $20,000$ steps.
    \textbf{(b)}: the corresponding distribution of the Frobenius norm of the Jacobian of the $\soft$ function
    is depicted.}
    \label{fig:fine_grained}
\end{figure}

To observe this behavior in the early phase of training, we created a ridge plot of the distributions of counts $k$ for the 
first $20,000$ steps (see Figure~\ref{fig:fine_grained} (a)). This plot shows that the attention in heads of middle layers often 
change the operation mode to \textbf{Class IV} around $9,000$ to $10,000$ steps. At the same time the second big drop in the loss occurs.
The question arises whether this is functionally important or whether it is an artefact which could be even 
harmful. To check if the attention mechanism is still able to learn after the change in the operation mode 
we analyzed the gradient flow through the $\soft$ function.
For every token we calculate the Frobenius norm of the Jacobian of the $\soft$ over multiple samples.
Then, for every head we plot the distribution of the norm (see Figure~\ref{fig:fine_grained} (b)). We can see that 
the attention 
in heads of \textbf{Class IV} remain almost unchanged during the rest of the training. 


\subsection{Gaussian layers}
\begin{figure}[htp]
    \centering
    \includegraphics[width=0.8\textwidth]{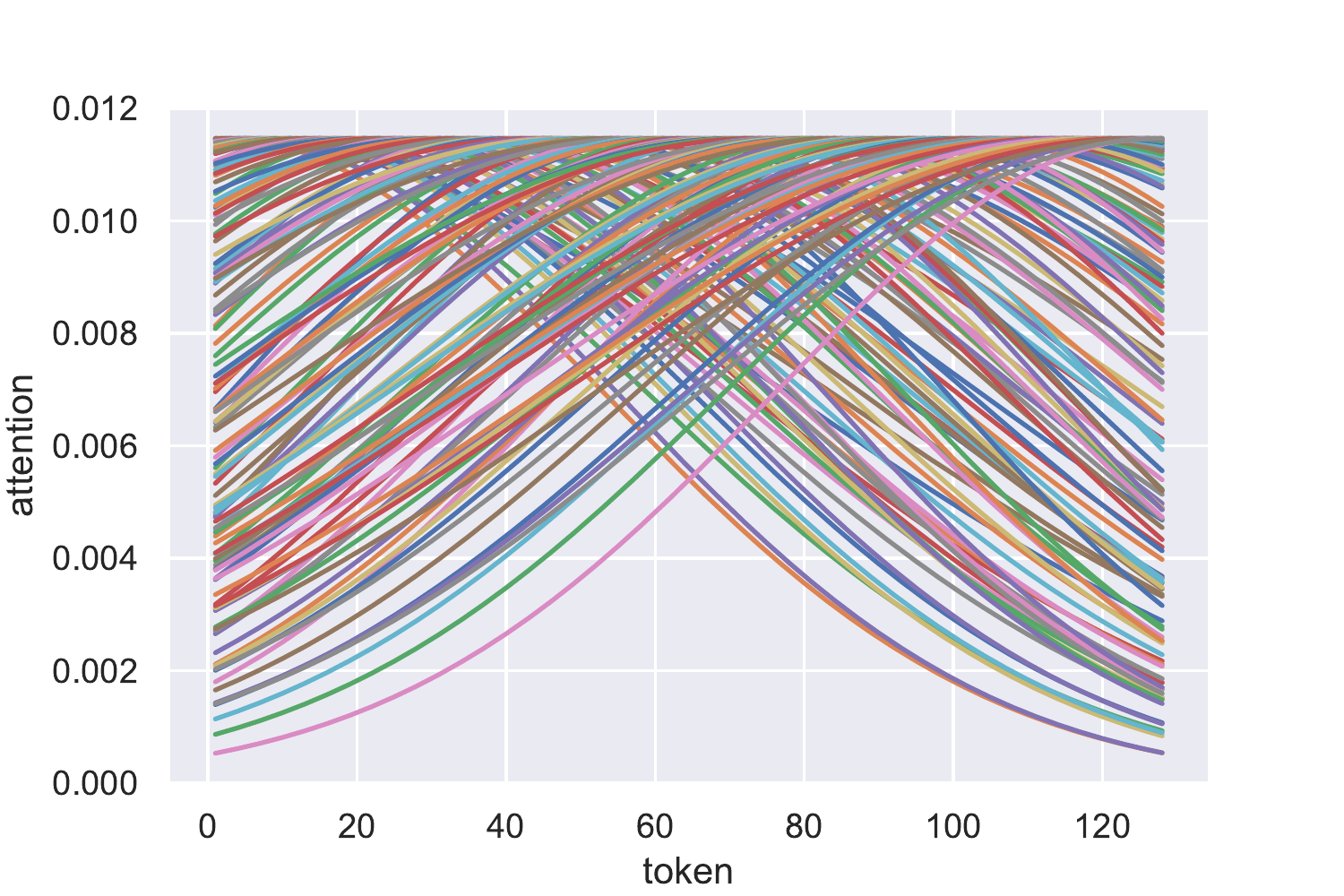}
    \caption{Attentions of a Gaussian averaging head at initialization for sequence length $N = 128$. Every line depicts one Gaussian kernel. Here, the location parameters are initialized with the value of the supporting points $\mu_i = s_i$.
   \label{fig:avg_init}}
\end{figure}

The self-attention mechanism proposed in \cite{vaswani2017attention} utilizes the $\soft$ function to compute
the coefficients of a convex combination over the embedded tokens, where the $\soft$ is conditioned on the input. 
However, our analysis showed that especially in lower layers many heads perform averaging over a very 
large number of patterns. This suggests that at this level neither the dependency on the input 
nor a fine grained attention to individual positions is necessary. As an alternative to the original 
mechanism we propose Gaussian averaging heads which 
are computationally more efficient. Here, the $\soft$ function is replaced by a discrete Gaussian kernel,
where the location $\mu$ and the scale $\sigma$ are learned. In detail, for a sequence length of 
$N$ tokens we are given a vector of location parameters $\boldsymbol{\mu} = (\mu_1, \ldots, \mu_N)^T$ and a vector of
corresponding scale parameters $\boldsymbol{\sigma} = (\sigma_1, \ldots, \sigma_N)^T$.
We subdivide the interval $[-1,1]$ into $N$ equidistant supporting points
$\{s_j\}_{j=1}^N$, where 
\begin{align*}
    s_j = \frac{(j-1) - 0.5~(N-1)}{0.5~(N-1)}.
\end{align*}
The attention $[A]_{i,j}$ from the $i$-th token to the $j$-th position is calculated as
\begin{align*}
    [A]_{i,j} = \frac{1}{z_i} \exp \left\{ -\frac{1}{2} \big(\frac{ s_j - \mu_i}{\sigma_i}\big)^2  \right\},
\end{align*}
where $z_i$ normalizes the $i$-th row of the attention matrix $A$ to sum up to one:
\begin{align*}
    z_i = \sum_{j=1}^N \exp \left\{ -\frac{1}{2} \big(\frac{ s_j - \mu_i}{\sigma_i}\big)^2 \right\}.
\end{align*}
For initialization we uniformly sample a location vector $\boldsymbol{\mu} \in [-1,1]^N$ and 
a scale vector $\boldsymbol{\sigma} \in [0.75, 1.25]^N$ per head. A simple way to consider the individual 
position of each token at initialization is to use the supporting points $\mu_i = s_i$
(see Figure~\ref{fig:avg_init}). In practice no difference to the random initialization was observed.

\paragraph{Number of parameters.}
Gaussian averaging heads can reduce the number of parameters significantly.
For an input size of $N$ tokens, there are $2 \cdot N$ parameters per head. 
In contrast, a standard self-attention head 
with word embedding dimension $d_y$ and projection dimension $d_k$ has two weight matrices
$W_Q, W_K \in \mathbb{R}^{d_k \times d_y}$, which together amount to $2 \cdot d_k\cdot d_y$ parameters.
As a concrete example, the BERT-base model from \cite{Devlin:19} has an embedding dimension $d_y = 768$, 
 a projection dimension $d_k = 64$ and a sequence length of $N = 512$. 
Compared to the Gaussian head, in this case $ (2 \cdot 768 \cdot 64) / (2 \cdot 512) ~ = 95.5$ times more 
parameters are trained for the attention mechanism itself. 
Only for very long sequences (and given
that the word embedding dimension stays the same) the dependence on $N$ may become a disadvantage.
But of course, due to the independence from the input the Gaussian averaging head is less expressive 
in comparison to the original attention mechanism.
A recently proposed input independent replacement for self-attention is the so called Random Synthesizer 
\cite{Tay:20}. Here the $\soft$-attention is directly parametrized with an $N \times N$ 
matrix. This amounts to $0.5 \cdot N$ more parameters than Gaussian averaging. 

\clearpage
\section{Immune repertoire classification}
\label{sec:deeprc_details}
\label{sec:deeprc_implementation_details}

\subsection{The DeepRC approach.} 
We propose
a Deep Learning architecture, called \textbf{Deep} \textbf{R}epertoire \textbf{C}lassification (DeepRC), with a transformer-like 
attention mechanism for immune 
repertoire classification and compare it against 
other machine learning approaches. 
For DeepRC, we consider \emph{immune repertoires as input objects},
which are represented as bags of instances. In a bag, \emph{each instance is
an immune receptor sequence} and each bag can contain a large number of sequences. 
Note that we will use $\Bz_i$ to denote the \emph{sequence-representation} of 
the $i$-th sequence and $\Bz$ to denote the \emph{repertoire-representation}. 
At the core, DeepRC consists of a transformer-like 
attention-mechanism that
extracts the most important information from each repertoire.
We first give an overview of the attention mechanism and then 
provide details on each of the sub-networks 
$h_1$, $h_2$, and $o$ of DeepRC.
(see Fig.~\ref{fig:deeprc_architecture})

\paragraph{Attention mechanism in DeepRC.}
This mechanism is based on the three
matrices $\BK$ (the keys), $\BQ$ (the queries), and $\BV$
(the values) together with a parameter
$\beta$.

\textit{Values.} 
DeepRC uses a 1D convolutional network $h_1$ \citep{lecun1998gradient,hu2014convolutional,kelley2016basset}
that supplies a sequence-representation $\Bz_i=h_1(s_i)$,
which acts as the values $\BV=\BZ= (\Bz_1, \ldots, \Bz_{N})$ 
in the attention mechanism (see Figure~\ref{fig:deeprc_architecture}). 

\textit{Keys.}
A second neural network $h_2$,
which shares its first layers with $h_1$,
is used to obtain keys 
$\BK \in \dR^{N \times d_k}$ for each 
sequence in the repertoire.
This network uses $2$ self-normalizing layers \citep{klambauer2017self} 
with $32$ units per layer (see Figure~\ref{fig:deeprc_architecture}).

\textit{Query.} 
We use a fixed $d_k$-dimensional query vector $\Bxi$ which 
is learned via backpropagation.
For more attention heads, each head has a fixed query vector.
With the quantities introduced above, 
the transformer attention mechanism
of DeepRC is implemented 
as follows:
\begin{align}
    \Bz= \att(\Bxi^T,\BK,\BZ; \frac{1}{\sqrt{d_k}}) = \soft\left(\frac{\Bxi^T \BK^T}{\sqrt{d_k}}\right) \BZ,
\end{align}
%
%
where $\BZ \in \dR^{N \times d_v}$ are the sequence-representations
stacked row-wise, $\BK$ are the keys, and
$\Bz$ is the repertoire-representation
and at the same time a weighted mean of sequence-representations $\Bz_i$. 
The attention mechanism can readily be extended to multiple queries and heads, however,
computational demand could constrain this. 

\paragraph{Attention-pooling.}
Each input object, i.e. repertoire, consists of a large 
number $N$ of sequences,
which are reduced to a single fixed-size feature vector of length $d_v$ 
representing the whole input object by an attention-pooling function. 
To this end, a transformer-like attention 
mechanism adapted to sets is realized in DeepRC which supplies $a_i$ -- the importance of 
the sequence $s_i$. This importance value for 
each sequence, is an interpretable quantity, which is highly 
desired for the immunological problem at hand. 

\paragraph{Classification layer and network parameters.} 
The repertoire-representation $\Bz$ is then 
used as input for a fully-connected output network 
$\hat y= o(\Bz)$ that predicts the immune status, 
where we found it sufficient to train single-layer networks.
In the simplest case, DeepRC predicts a scalar target $y$.
That is, it predicts the class label,
e.g. the immune status of an immune repertoire,
using one output value.
However, since DeepRC is an end-to-end deep learning model, multiple 
targets may be predicted simultaneously in classification or 
regression settings or a mix of both.
This allows for the introduction of additional information into the 
system via auxiliary targets such as age, sex, or other metadata.

\paragraph{Network parameters, training, and inference.}
DeepRC is trained using standard gradient descent methods 
to minimize a regularized cross-entropy loss. 
The network parameters are $\bm \theta_1, \bm \theta_2, \bm \theta_o$ for the 
sub-networks $h_1, h_2$, and $o$, respectively, 
and additionally $\Bxi$. In more detail, we train DeepRC using 
Adam \citep{kingma2014adam} with a
batch size of $4$ and a learning rate of $10^{-4}$.

\subsection{The DeepRC architecture}

\begin{figure}[htp]
    \centering
    \includegraphics[width=0.5\textwidth]{figures/architecture_paper_v6.pdf}
    \caption{Schematic representation of the DeepRC architecture. 
  $d_l$ indicates the sequence length.
   \label{fig:deeprc_architecture}}
\end{figure}

\paragraph{Input layer.} For the input layer of the convolutional neural network (CNN), 
the characters in the input sequence, i.e. the amino acids (AAs),
are encoded in a one-hot vector of length $20$.
To also provide information about the position of an AA in the sequence, we add 
$3$ additional input features with values in range $[0, 1]$ to encode the position 
of an AA relative to the sequence.
These $3$ positional features encode whether the AA is located at the beginning, 
the center, or the end of the sequence, respectively, as shown in 
Figure~\ref{fig:positional_encoding}.
We concatenate these $3$ positional features with the one-hot vector of AAs,
which results in a feature vector of length $23$ 
per sequence position.
Each repertoire, now represented as a bag of feature vectors, 
is then normalized to 
unit variance.
Since the cytomegalovirus dataset (\emph{CMV dataset}) provides sequences with an associated 
abundance value per sequence,
which is the number of occurrences of a sequence in a repertoire,
we incorporate this information into the input of DeepRC.
To this end, the one-hot AA features of a sequence are 
multiplied by a scaling factor $\log(c_a)$
before normalization,
where $c_a$ is the abundance of a sequence.
Sequences of different lengths were zero-padded to the maximum sequence length
per batch at the sequence ends.
We feed the sequences with $23$ features per position into the CNN, as 
shown in Figure~\ref{fig:deeprc_architecture}.

\begin{figure}[htp]
    \begin{center}
    \includegraphics[width=0.8\textwidth]{figures/position_encoding.pdf}\\
   \caption{We use 3 input features with values in range $[0,1]$ to encode the relative position of each AA in a sequence with respect to the sequence.
   ``feature 1'' encodes if an AA is close to the sequence start,
   ``feature 2'' to the sequence center, and 
   ``feature 3'' to the sequence end.
   For every position in the sequence, the values of all three features sum up to $1$.
   \label{fig:positional_encoding}}
   \end{center}
\end{figure}

\paragraph{1D CNN for motif recognition.}
In the following,
we describe how DeepRC identifies patterns in the individual sequences and 
reduces each sequence in the input object to a fixed-size feature vector.
DeepRC employs 1D convolution layers to extract
patterns, where trainable weight kernels are convolved over the sequence positions.
In principle, also recurrent neural networks (RNNs) or transformer networks could be used
instead of 1D CNNs, however, (a) the computational complexity of the network 
must be low to be able to process millions of sequences for a single update. Additionally, 
(b) the learned network should be able to provide insights in the recognized patterns in form
of motifs. Both properties (a) and (b) are fulfilled by 1D convolution operations that are used
by DeepRC. 

We use one 1D CNN layer \citep{hu2014convolutional}
with SELU activation functions \citep{klambauer2017self}
to identify
the relevant patterns in the input sequences with a computationally light-weight operation.
The larger the kernel size,
the more surrounding sequence positions are taken into account,
which influences the length of the motifs that can be extracted.
We therefore adjust the kernel size 
during hyperparameter search.
In prior works \citep{ostmeyer2019biophysicochemical},  
a k-mer size of $4$ yielded good predictive performance,
which could indicate that a kernel size in the range of $4$ may 
be a proficient choice.
For $d_v$ trainable kernels, this produces a feature vector 
of length $d_v$ at each sequence position. 
Subsequently, global max-pooling over all sequence positions of a sequence reduces the sequence-representations $\Bz_i$ to vectors of the fixed length $d_v$. 
Given the challenging size of the input data per repertoire,
the computation of the CNN activations and weight updates is 
performed using 16-bit floating point values.
A list of hyperparameters evaluated for DeepRC is given in 
Table~\ref{tab:deeprc_settings}.

\paragraph{Regularization.}
We apply random and attention-based subsampling of repertoire sequences to
reduce over-fitting and decrease computational effort.
During training, each repertoire is subsampled to $10,000$ input sequences,
which are randomly drawn from the respective repertoire.
This can also be interpreted as random drop-out \citep{hinton2012improving} on 
the input sequences or attention weights.
During training and evaluation,
the attention weights computed by the attention network are furthermore used to 
rank the input sequences.
Based on this ranking, the repertoire is reduced to the $10\%$ of sequences 
with the highest attention weights.
These top $10\%$ of sequences are then used to compute the weight updates and
the prediction for the repertoire.
Additionally, one might employ further regularization techniques,
which we did not investigate further due to high computational demands.
Such regularization techniques include
$l1$ and $l2$ weight decay,
noise in the form of random AA permutations in the input sequences,
noise on the attention weights,
or random shuffling of sequences between repertoires that belong to the negative class.
The last regularization technique assumes that the sequences in positive-class repertoires 
carry a signal, such as an AA motif corresponding to an immune response,
whereas the sequences in negative-class repertoires do not.
Hence, the sequences can be shuffled randomly between negative class repertoires without obscuring the 
signal in the positive class repertoires.

\paragraph{Hyperparameters.} 
For the hyperparameter search of DeepRC for the category ``simulated 
immunosequencing data'',
we only conducted a full hyperparameter 
search on the more difficult datasets with motif implantation probabilities below $1\%$,
as described in Table~\ref{tab:deeprc_settings}.
This process was repeated for all $5$ folds of the 5-fold cross-validation (CV) and 
the average score on the $5$ test sets constitutes the final score of a method.
Table~\ref{tab:deeprc_settings} provides an overview of the hyperparameter 
search,
which was conducted as a grid-search
for each of the datasets in a nested 5-fold CV procedure,
as described in section~\ref{subsec:methods_compared}.

\paragraph{Computation time and optimization.}
We took measures on the implementation level to address the high computational demands, especially GPU memory consumption, in order to make the large number of experiments feasible.
We train the DeepRC model with a small batch size of $4$ samples
and perform computation of inference and updates of the 1D CNN using 16-bit floating point values.
The rest of the network is trained using 32-bit floating point values.
The Adam parameter for numerical stability was therefore
increased from the default value of $\epsilon=10^{-8}$ to $\epsilon=10^{-4}$.
Training was performed on various GPU types,
mainly \texttt{NVIDIA RTX 2080 Ti}.
Computation times were highly dependent on the number of sequences in the repertoires
and the number and sizes of CNN kernels.
A single update on an \texttt{NVIDIA RTX 2080 Ti} GPU took approximately $0.0129$ to $0.0135$ seconds,
while requiring approximately $8$ to $11$ GB GPU memory.
Taking these optimizations and 
GPUs with larger memory ($\geq16$ GB) into account,
it would already be feasible to train DeepRC,
possibly with multi-head attention and a larger network architecture,
on larger datasets.
Our network implementation is based on \texttt{PyTorch 1.3.1} \citep{paszke2019pytorch}.

\subsection{Details on experiments}
\label{sec:experiments}
\subsubsection{Datasets}\label{sec:datasets}
We aimed at constructing immune repertoire classification 
scenarios with varying degree of realism and 
difficulties in order to compare and analyze
the suggested machine learning methods. To this end, we either use
simulated or experimentally-observed immune receptor sequences and we implant signals,
which are sequence motifs \citep{akbar2019compact, weber2019immunesim}, into sequences of repertoires of the positive class. It has been shown previously that interaction of immune receptors with antigens occur via short sequence stretches \citep{akbar2019compact}. Thus, implantation of short motif sequences simulating an immune signal is biologically meaningful.
Our benchmarking study comprises four different categories of datasets:
(a) Simulated immunosequencing data with implanted signals (where the signal is defined as sets of motifs), 
(b) LSTM-generated immunosequencing data with implanted signals,
(c) real-world immunosequencing data with implanted signals, and 
(d) real-world immunosequencing data.
Each of the first three categories consists of multiple datasets with varying 
difficulty depending on the type of the implanted signal and the ratio of sequences with the implanted signal.
The ratio of sequences with the implanted signal,
where each sequence carries at most 1 implanted signal,
corresponds to the \emph{witness rate} (WR).
We consider binary classification tasks to simulate the immune status
of healthy and diseased individuals. We randomly generate immune repertoires
with varying numbers of sequences, where we implant sequence motifs in the 
repertoires of the diseased individuals, i.e. the positive class. The sequences of a repertoire are 
also randomly generated by different procedures (detailed below).
Each sequence is composed of $20$ different characters, corresponding to amino acids, 
and has an average length of $14.5$ AAs.

\paragraph{Simulated immunosequencing data.}
In the first category,
we aim at investigating the impact of the signal frequency, i.e. the WR, and the signal 
complexity on the performance of the different methods.
To this end, we created $18$ datasets,
whereas each dataset contains a large number of repertoires
with a large number of random AA sequences per repertoire.
We then implanted signals in the AA sequences of the positive class repertoires,
where the $18$ datasets differ in frequency and complexity of the implanted signals.
In detail, the AAs were sampled randomly independent of their respective position in the sequence,
while the frequencies of AAs, 
distribution of sequence lengths, 
and distribution of the number of sequences per repertoire,
following the respective distributions observed in the real-world \emph{CMV dataset}~\citep{emerson2017immunosequencing}.
For this, we first sampled the number of sequences of a repertoire from 
a Gaussian $\mathcal N(\mu=316k, \sigma=132k)$ distribution and 
rounded to the nearest positive integer.
We re-sampled if the size was below $5k$. 
We then generated random sequences of AAs with a length of $\mathcal N(\mu=14.5, \sigma=1.8)$,
again rounded the nearest positive integers.
Each simulated repertoire was then randomly assigned to
either the positive or negative class,
with $2,500$ repertoires per class.
In the repertoires assigned to the positive class,
we implanted motifs with an average length of $4$ AAs,
following the results of the experimental analysis of antigen-binding motifs in antibodies and T-cell receptor sequences by \cite{akbar2019compact}.
We varied the characteristics of the implanted motifs for each of the $18$ datasets with respect to the following parameters:
(a) $\rho$, the probability of a motif being implanted in a sequence of a positive repertoire, 
i.e. the average ratio of sequences containing the motif, which is the witness rate.  
(b) The number of wildcard positions in the motif.
A wildcard position contains a random AA, which is randomly sampled for each sequence.
Wildcard positions are located in the center of the implanted motif.
(c) The number of deletion positions in the implanted motif.
A deletion position has a probability of $0.5$ of being removed from the motif. 
Deletion positions are located in the center of the implanted motifs. 

In this way, we generated $18$ different datasets of variable 
difficulty containing in total roughly $28.7$ billion sequences. 
See Table~\ref{tab:lstm_simulations} for an overview of the properties 
of the implanted motifs in the $18$ datasets.

\paragraph{LSTM-generated data.}
In the second dataset category,
we investigate the impact of the signal frequency and complexity
in combination with more plausible immune receptor sequences
by taking into account the positional AA distributions
and other sequence properties.
To this end, we trained an LSTM \citep{hochreiter1997long} in a standard 
next character prediction \citep{graves2013generating}
setting to create AA sequences with properties similar to experimentally observed immune receptor sequences.

In the first step, the LSTM model  
was trained on all immuno-sequences in the \emph{CMV dataset} 
\citep{emerson2017immunosequencing} that contain valid information 
about sequence abundance and have a known CMV label.
Such an LSTM model is able to capture various properties of these sequence,
including position-dependent probability distributions and
combinations, relationships, and order of AAs.
We then used the trained LSTM model to generate $1,000$ repertoires
in an autoregressive fashion,
starting with a start sequence that was randomly 
sampled from the trained-on dataset.
Based on a visual inspection of the frequencies of 4-mers,
the similarity of LSTM generated sequences and real sequences
was deemed sufficient for the purpose of generating the AA sequences
for the datasets in this category.

After generation,
each repertoire was assigned to either the positive or negative class,
with $500$ repertoires per class.
We implanted motifs of length $4$ with varying properties
in the center of the sequences of the positive class
to obtain $5$ different datasets.
Each sequence in the positive repertoires has a probability 
$\rho$ to carry the motif,
which was varied throughout $5$ datasets
and corresponds to the WR (see Table~\ref{tab:lstm_simulations}).
Each position in the motif has a probability of $0.9$ to be implanted
and consequently a probability of $0.1$ that the original AA in the sequence remains,
which can be seen as noise on the motif.

\begin{table}[htp]%
    \begin{center}%
        \begin{tabular}{lccc}%
            \toprule
                                   & Simulated                   & LSTM gen. & Real-world\\
            \midrule
            seq. per bag & $N(316k, 132k)$ & $N (285k, 156k)$ & $10k$\\
            repertoires              & $5,000$                             & $1,000$ & $1,500$\\
            motif noise          & $0\%$                                 & $10\%$ & $*$  \\
            wildcards            & $\{0;1;2\}$                         & $0$ & $0$ \\
            deletions            & $\{0;1\}$                           & $0$ & $0$ \\
            mot. freq. $\rho$     & $\{1; 0.1;$                      & $\{10;1;0.5;$ & $\{1;0.1\}$ \\
            (in $\%$)            & $0.01\}$                           & $0.1;0.05\}$ &  \\
            \bottomrule
        \end{tabular}%
        \caption{Properties of simulated repertoires,
        variations of motifs, 
        and motif frequencies for 
        the datasets in categories 
        ``simulated immunosequencing data'',
        ``LSTM-generated data'',
        and ``real-world data with implanted signals''.
        Noise types for $*$ are explained in paragraph ``real-world data with implanted signals''.}%
        \label{tab:lstm_simulations}%
    \end{center}%
\end{table}

\paragraph{Real-world data with implanted signals.}
In the third category, we implanted signals into experimentally obtained 
immuno-sequences, where we considered $4$ dataset variations.
Each dataset consists of $750$ repertoires for each of the two classes, 
where each repertoire consists of $10k$ sequences. In this way, we 
aim to simulate datasets with a \emph{low sequencing coverage}, which
means that only relatively few sequences per repertoire are available.
The sequences were randomly sampled from healthy (CMV negative) individuals from the \emph{CMV dataset} (see below paragraph for explanation).
Two signal types were considered:
(a) \textbf{One signal with one motif.}
The AA motif \texttt{LDR} was implanted in a certain fraction 
of sequences. The pattern is randomly altered at one of the 
three positions with probabilities $0.2$, $0.6$, and $0.2$, 
respectively. 
(b) \textbf{One signal with multiple motifs.}
One of the three possible motifs \texttt{LDR}, \texttt{CAS}, and
\texttt{GL-N} was implanted with equal probability. Again, the motifs 
were randomly altered before implantation. 
The AA motif \texttt{LDR} changed as described above. 
The AA motif \texttt{CAS} was altered at the second position with 
probability $0.6$ and with probability $0.3$ at the first position. 
The pattern \texttt{GL-N}, where \texttt{-} denotes a gap location,
is randomly altered at the first position with probability $0.6$ and 
the gap has a length of $0$, $1$, or $2$ AAs with equal probability.

Additionally, the datasets differ in the values for $\rho$, the 
average ratio of sequences carrying a signal, 
which were chosen as $1\%$ or $0.1\%$.
The motifs were implanted at positions $107$, $109$, and $114$ according to the IMGT numbering scheme for immune receptor sequences \citep{lefranc2003imgt} 
with probabilities $0.3$, $0.35$ and $0.2$, respectively.
With the remaining $0.15$ chance, 
the motif is implanted at any other sequence position.
This means that the motif occurrence in the simulated sequences is biased towards the middle of the sequence. 

\textbf{Real-world data: CMV dataset.}
We used a real-world dataset of $785$ repertoires,
each of which containing between $4,371$ to $973,081$ (avg. $299,319$) TCR sequences
with a length of $1$ to $27$ (avg. $14.5$) AAs, 
originally collected and provided by \citep{emerson2017immunosequencing}.
$340$ out of $785$ repertoires 
were labelled as positive for cytomegalovirus (CMV) serostatus, which we consider
as the positive class, $420$ repertoires with negative CMV serostatus, considered
as negative class, and $25$ repertoires with unknown status.
We changed the number of 
sequence counts per repertoire from $-1$ to $1$ 
for $3$ sequences.
Furthermore, 
we exclude a total of $99$ repertoires with unknown CMV status or unknown 
information about the sequence abundance within a repertoire, 
reducing the dataset for our analysis to $686$ repertoires, 
$312$ of which with positive and $374$ with negative CMV status.


\subsubsection{Methods compared}\label{subsec:methods_compared}
We evaluate and compare the performance of 
DeepRC against a set of machine learning methods that
serve as baseline, were suggested, or 
can readily be adapted to immune repertoire 
classification. In this section, we describe these compared 
methods.

\paragraph{Known motif.}
This method serves as an estimate for the achievable classification performance 
using prior knowledge about which motif was implanted. 
Note that this does not necessarily lead to perfect predictive performance
since motifs are implanted with a certain amount of noise
and could also be present in the negative class by chance.
The \emph{known motif} method counts how often the known implanted motif occurs 
per sequence for each repertoire and uses this count to rank the repertoires.
From this ranking, 
the Area Under the receiver operator Curve (AUC) is computed as performance measure.
Probabilistic AA changes in the known motif are not considered for this count,
with the exception of gap positions.
We consider two versions of this method:
(a) \textbf{Known motif binary:} counts the occurrence of the known motif in a sequence
and (b) \textbf{Known motif continuous:} counts the maximum number of overlapping AAs between the 
known motif and all sequence positions,
which corresponds to a convolution operation with a binary kernel followed by max-pooling.
Since the implanted signal is not known in the experimentally obtained \emph{CMV dataset},
this method cannot be applied to this dataset.

\paragraph{SVM.}
The Support Vector Machine (SVM) approach uses a fixed mapping from a bag 
of sequences to the corresponding k-mer counts. 
The function $h_{\text{kmer}}$ maps each sequence $s_i$ to a
vector representing the occurrence of k-mers in the sequence. To avoid confusion 
with the sequence-representation obtained from the CNN layers of 
DeepRC, we denote $\Bu_i=h_{\text{kmer}}(s_i)$, which is analogous to $\Bz_i$. 
Specifically, $u_{im}= \left(h_{\text{kmer}}(s_i)\right)_m = \#\{p_m \in s_i\}$, where $\# \{p_m \in s_i\}$
denotes how often the k-mer pattern $p_m$ occurs in sequence $s_i$. Afterwards, 
average-pooling is applied to obtain $\Bu=1/N \sum_{i=1}^N \Bu_i$, 
the \emph{k-mer representation} of the input object $X$. For 
two input objects $X^{(n)}$ and $X^{(l)}$ with representations
$\Bu^{(n)}$ and $\Bu^{(l)}$, respectively, we implement the \emph{MinMax kernel}~\citep{ralaivola2005graph} as follows:
\begin{equation}
    \begin{split}
        k(X^{(n)}, X^{(l)})&= k_{\mathrm{MinMax}} (\Bu^{(n)},\Bu^{(l)}) \\
        &= \frac{\sum_{m=1}^{d_u} \min(u^{(n)}_{m},u^{(l)}_{m})}{\sum_{m=1}^{d_u}  \max(u^{(n)}_{m},u^{(l)}_{m})},
    \end{split}
\end{equation}
where $u^{(n)}_{m}$ is the $m$-th element of the vector $u^{(n)}$. 
The \emph{Jaccard kernel}~\citep{levandowsky1971distance} is identical to the MinMax kernel
except that it operates on binary $\Bu^{(n)}$.
We used a standard C-SVM, as introduced by \citep{cortes1995support}. 
The corresponding hyperparameter $C$ is optimized by random search. The settings of the 
full hyperparameter search as well as the respective value ranges are 
given in \tablename~\ref{tab:svm_settings}.

\paragraph{KNN.}
The same \emph{k-mer representation} of a repertoire, as introduced above for the SVM baseline,
is used for the K-Nearest Neighbor (KNN) approach. As this method clusters samples 
according to distances between them, the previous kernel definitions cannot be applied 
directly. It is therefore necessary to transform the MinMax as well as the Jaccard 
kernel from similarities to distances by constructing 
the following \citep{levandowsky1971distance}:
\begin{equation}
    \begin{split}
        d_{\mathrm{MinMax}} (\Bu^{(n)},\Bu^{(l)}) &= 1 - k_{\mathrm{MinMax}} (\Bu^{(n)},\Bu^{(l)}), \\
        d_{\mathrm{Jaccard}} (\Bu^{(n)},\Bu^{(l)}) &= 1 - k_{\mathrm{Jaccard}} (\Bu^{(n)},\Bu^{(l)}).
    \end{split}
\end{equation}
The amount of neighbors is treated as the hyperparameter and optimized by an 
exhaustive grid search. The settings of the full hyperparameter search as well 
as the respective value ranges are given in \tablename~\ref{tab:knn_settings}.

\paragraph{Logistic regression.}
We implemented logistic regression on the \emph{k-mer 
representation} $\Bu$ of an immune repertoire. The model
is trained by gradient descent using the Adam optimizer~\citep{kingma2014adam}.
The learning rate is treated as the hyperparameter and optimized by grid search.
Furthermore, we explored two regularization settings 
using combinations of $l1$ and $l2$ weight decay.
The settings of the full hyperparameter search as well as the respective value ranges
are given in \tablename~\ref{tab:logistic_regression_settings}.

\paragraph{Burden test.} 
We implemented a burden test \citep{emerson2017immunosequencing,li2008methods,wu2011rare} in a machine learning
setting. The burden test first identifies sequences or k-mers that are associated
with the individual's class, i.e., immune status, and then calculates a burden 
score per individual. Concretely, for each k-mer or sequence, the phi coefficient
of the contingency table for absence or presence and positive or negative immune status
is calculated. Then, $J$ k-mers or sequences with the highest phi coefficients 
are selected as the set of associated k-mers or sequences. $J$ is a hyperparameter that 
is selected on a validation set. Additionally, we consider the type of input 
features, sequences or k-mers, as a hyperparameter. For inference, a burden score
per individual is calculated as the sum of associated k-mers or sequences it carries. 
This score is used as raw prediction and to rank the individuals.
Hence, we have extended the burden test by \citep{emerson2017immunosequencing} 
to k-mers and to adaptive thresholds that are adjusted on a validation set.

\paragraph{Logistic MIL (Ostmeyer et al).}
The logistic multiple instance learning (MIL) approach for immune 
repertoire classification \citep{ostmeyer2019biophysicochemical} applies a logistic 
regression model to each k-mer representation in a bag.
The resulting scores are then 
summarized by max-pooling to obtain a prediction for the bag.
Each amino acid of each k-mer is represented by $5$ features, the so-called 
Atchley factors~\citep{atchley2005solving}. As k-mers of length $4$ are used, 
this gives a total of $4\times5=20$ features.
One additional feature per 4-mer is added, which represents the relative 
frequency of this 4-mer with respect to its containing bag, resulting in 
$21$ features per 4-mer.
Two options for the relative frequency feature exist, which are
(a) whether the frequency of the 4-mer (``4MER'') or
(b) the frequency of the sequence in which the 4-mer appeared (``TCR\textbeta'') is used.
We optimized the learning rate, batch size, and early stopping parameter 
on the validation set. The settings of the full hyperparameter search as
well as the respective value ranges are given in
\tablename~\ref{tab:mil_settings}.

\subsection{Details on hyperparameter selection}
\label{sec:hyperparams}
For all competing methods a hyperparameter search was performed,
for which we split each 
of the $5$ training sets into an inner training set and 
inner validation set.
The models were trained on the inner training set and 
evaluated on the inner validation set.
The model with the highest AUC score on the inner validation 
set is then used to 
calculate the score on the respective test set.
Here
we report the hyperparameter sets and search strategy that is
used for all methods. 

\paragraph{DeepRC.}
The set of hyperparameters of DeepRC is shown in 
Table~\ref{tab:deeprc_settings}. These hyperparameter
combinations are adjusted by grid-search.

\begin{table}[htp]%
    \begin{center}%
        \begin{tabular}{lc}%
            \toprule
            learning rate & $10^{-4}$ \\
            number of kernels ($d_v$) & $\{8;16;32;64^*;128^*;256^*\}$ \\
            number of CNN layers & $\{1\}$ \\
            number of layers in key-NN & $\{2\}$ \\
            number of units in key-NN & $\{32\}$ \\
            kernel size & $\{5;7;9\}$ \\
            subsampled seqences & $10,000$ \\
            batch size & $4$ \\
            \bottomrule
        \end{tabular}%
        \caption{\textbf{DeepRC hyperparameter search space.}
        Every $5\cdot10^{3}$ updates, the current model was evaluated against the validation fold.
        The early stopping hyperparameter was determined by selecting the model with the best loss
        on the validation fold after $10^{5}$ updates.
        $^*$: Experiments for $\{64;128;256\}$ kernels were omitted for datasets with  motif implantation probabilities $\geq 1\%$ in the category ``simulated immunosequencing data''.
        }%
        \label{tab:deeprc_settings}%
    \end{center}%
\end{table}

\paragraph{Known motif.}
This method does not have hyperparameters and has been applied 
to all datasets except for the \emph{CMV dataset}. 

\paragraph{SVM.}
The corresponding hyperparameter $C$ of the SVM is optimized by randomly 
drawing $10^{3}$ values in the range of $[-6; 6]$ 
according to a uniform distribution. These values act 
as the exponents of a power of $10$ and are applied 
for each of the two kernel types (see Table~\ref{tab:svm_settings}).

\begin{table}[htp]%
    \begin{center}%
        \begin{tabular}{lc}
            \toprule
            $C$ & $10^{[-6;6]}]$ \\
            type of kernel & $\{\mathrm{MinMax}; \mathrm{Jaccard}\}$ \\
            number of trials & $10^{3}$ \\
            \bottomrule
        \end{tabular}%
        \caption{Settings used in the hyperparameter search of the SVM baseline approach. 
        The number of trials defines the quantity of random values of the 
        $C$ penalty term (per type of kernel).}%
        \label{tab:svm_settings}%
    \end{center}%
\end{table}

\paragraph{KNN.}
The amount of neighbors is treated as the hyperparameter and optimized by grid search operating in the discrete range of $[1; \max\{N, 10^{3}\}]$ with a step size of $1$. The corresponding tight upper bound is automatically defined by the total amount of samples $N\in\mathbb{N}_{> 0}$ in the training set, capped at $10^{3}$ (see Table~\ref{tab:knn_settings}).

\begin{table}[htp]%
    \begin{center}%
        \begin{tabular}{lc}
            \toprule
            number of neighbors & $\left[1; \max\{N, 10^{3}\}\right]$ \\
            type of kernel & $\{\mathrm{MinMax}; \mathrm{Jaccard}\}$ \\
            \bottomrule
        \end{tabular}%
        \caption{Settings used in the hyperparameter search of the KNN baseline approach. 
        The number of trials (per type of kernel) is automatically defined by the total 
        amount of samples $N\in\mathbb{N}_{> 0}$ in the training set, capped at $10^{3}$.}%
        \label{tab:knn_settings}%
    \end{center}%
\end{table}

\paragraph{Logistic regression.}
The hyperparameter optimization strategy that was employed 
for Logistic Regression was grid search across the hyperparameters
given in Table~\ref{tab:logistic_regression_settings}.

\begin{table}[htp]%
    \begin{center}%
        \begin{tabular}{lc}
            \toprule
            learning rate & $10^{-\{2; 3; 4\}}$ \\
            batch size & $4$ \\
            max. updates & $10^{5}$ \\
            coefficient $\beta_{1}$ (Adam) & $0.9$ \\
            coefficient $\beta_{2}$ (Adam) & $0.999$ \\
            weight decay weightings & $\{(l1=10^{-7}, l2=10^{-3});(l1=10^{-7}, l2=10^{-5})\}$ \\
            \bottomrule
        \end{tabular}%
        \caption{Settings used in the hyperparameter search of the logistic regression baseline approach.}%
        \label{tab:logistic_regression_settings}%
    \end{center}%
\end{table}

\paragraph{Burden test.}
The burden test selects two hyperparameters: the number of features
in the burden set and the type of features, see Table~\ref{tab:burden_settings}.

\begin{table}[htp]%
    \begin{center}%
        \begin{tabular}{lc}
            \toprule
            number of features in burden set & $\left[50,100,150,250\right]$ \\
            type of features & $\{\mathrm{4MER}; \mathrm{sequence}\}$ \\
            \bottomrule
        \end{tabular}%
        \caption{Settings used in the hyperparameter search of the burden test approach.}%
        \label{tab:burden_settings}%
    \end{center}%
\end{table}

\paragraph{Logistic MIL.}
For this method, we adjusted the learning rate as well as the batch size as the
hyperparameters by randomly drawing $25$ different hyperparameter 
combinations from a uniform distribution. The corresponding 
range of the learning rate is $[-4.5; -1.5]$, which
acts as the exponent of a power of $10$. The batch size lies 
within the range of $[1; 32]$. For each hyperparameter 
combination, a model is optimized by gradient descent using
Adam, whereas the early stopping parameter is adjusted 
according to the corresponding validation set (see Table~\ref{tab:mil_settings}).

\begin{table}[htp]%
    \begin{center}%
        \begin{tabular}{lc}
            \toprule
            learning rate & $10^{[-4.5;-1.5]}$ \\
            batch size & $\left[1;32\right]$ \\
            relative abundance term & $\{\mathrm{4MER}; \mathrm{TCR}\text{\textbeta}\}$ \\
            number of trials & $25$ \\
            max. epochs & $10^{2}$ \\
            coefficient $\beta_{1}$ (Adam) & $0.9$ \\
            coefficient $\beta_{2}$ (Adam) & $0.999$ \\
            \bottomrule
        \end{tabular}%
        \caption{Settings used in the hyperparameter search of the logistic MIL baseline approach. 
        The number of trials (per type of relative abundance) defines the quantity of combinations of random values of the learning rate 
        as well as the batch size.}%
        \label{tab:mil_settings}%
    \end{center}%
\end{table}

\subsection{Summarized results}
An overview of results on all datasets is given in Table~\ref{tab:results_full}. 
In each of the four categories, 
``real-world data'', 
``real-world data with implanted signals'',
``LSTM-generated data'',
and ``simulated immunosequencing data'', 
DeepRC outperforms all competing methods 
with respect to average AUC.
Across categories, the runner-up methods are either the SVM for 
MIL problems with MinMax kernel or the burden test. 

\begingroup
\setlength{\tabcolsep}{2pt} 
\renewcommand{\arraystretch}{1.5} 

\begin{table}[htp]%
    \begin{center}%
        \resizebox{\textwidth}{!}{%
        \begin{tabular}{lcacacacacac}
         \toprule
        & Real-world & \multicolumn{4}{c}{ Real-world data with  implanted signals} & \multicolumn{5}{c}{ LSTM-generated data} &  Simulated\\
    & CMV & OM 1\% & OM 0.1\% & MM 1\% & MM 0.1\% & 10\% & 1\% & 0.5\% & 0.1\% & 0.05\%  & avg.\\
        \cmidrule(r){2-2}
        \cmidrule(r){3-6}
        \cmidrule(r){7-11}
        \cmidrule(r){12-12}
        DeepRC & {\bf 0.832} \footnotesize{$\pm$ 0.022} & {\bf 1.00} \footnotesize{$\pm$ 0.00} & {\bf 0.98}\footnotesize{$\pm$ 0.01} & {\bf 1.00}\footnotesize{$\pm$ 0.00} & {\bf 0.94}\footnotesize{$\pm$0.01} & {\bf 1.00}\footnotesize{$\pm$ 0.00} & {\bf 1.00}\footnotesize{$\pm$ 0.00} & {\bf 1.00}\footnotesize{$\pm$ 0.00} & {\bf 1.00}\footnotesize{$\pm$ 0.00} & {\bf 1.00}\footnotesize{$\pm$ 0.00} & {\bf 0.846}\footnotesize{$\pm$ 0.223} \\
        SVM (MM) & 0.825 \footnotesize{$\pm$ 0.022} & {\bf 1.00} \footnotesize{$\pm$ 0.00} & 0.58\footnotesize{$\pm$ 0.02} & {\bf 1.00}\footnotesize{$\pm$ 0.00} & 0.53\footnotesize{$\pm$0.02} & {\bf 1.00}\footnotesize{$\pm$ 0.00} & {\bf 1.00}\footnotesize{$\pm$ 0.00} &{\bf 1.00}\footnotesize{$\pm$ 0.00} & {\bf 1.00}\footnotesize{$\pm$ 0.00} & 0.99\footnotesize{$\pm$ 0.01} & 0.827\footnotesize{$\pm$ 0.210} \\
        SVM (J) & 0.546 \footnotesize{$\pm$ 0.021} & 0.99 \footnotesize{$\pm$ 0.00} & 0.53\footnotesize{$\pm$ 0.02} & {\bf 1.00}\footnotesize{$\pm$ 0.00} & 0.57\footnotesize{$\pm$0.02} & 0.98\footnotesize{$\pm$ 0.04} & {\bf 1.00}\footnotesize{$\pm$ 0.00} & {\bf 1.00}\footnotesize{$\pm$ 0.00} & 0.90\footnotesize{$\pm$ 0.04} & 0.77\footnotesize{$\pm$ 0.07} & 0.550\footnotesize{$\pm$ 0.080} \\
        KNN (MM) & 0.679 \footnotesize{$\pm$ 0.076} & 0.74 \footnotesize{$\pm$ 0.24} & 0.49\footnotesize{$\pm$ 0.03} & 0.67\footnotesize{$\pm$ 0.18} & 0.50\footnotesize{$\pm$0.02} & 0.70\footnotesize{$\pm$ 0.27} & 0.72\footnotesize{$\pm$ 0.26} & 0.73\footnotesize{$\pm$ 0.26} & 0.54\footnotesize{$\pm$ 0.16} & 0.52\footnotesize{$\pm$ 0.15} & 0.634\footnotesize{$\pm$ 0.129} \\
        KNN (J) & 0.534 \footnotesize{$\pm$ 0.039} & 0.65 \footnotesize{$\pm$ 0.16} & 0.48\footnotesize{$\pm$ 0.03} & 0.70\footnotesize{$\pm$ 0.20} & 0.51\footnotesize{$\pm$0.03} & 0.70\footnotesize{$\pm$ 0.29} & 0.61\footnotesize{$\pm$ 0.24} & 0.52\footnotesize{$\pm$ 0.16} & 0.55\footnotesize{$\pm$ 0.19} & 0.54\footnotesize{$\pm$ 0.19} & 0.501\footnotesize{$\pm$ 0.007} \\
        Log. regr. & 0.607 \footnotesize{$\pm$ 0.058} & {\bf 1.00} \footnotesize{$\pm$ 0.00} & 0.54\footnotesize{$\pm$ 0.04} & 0.99\footnotesize{$\pm$ 0.00} & 0.51\footnotesize{$\pm$0.04} & {\bf 1.00}\footnotesize{$\pm$ 0.00} & {\bf 1.00}\footnotesize{$\pm$ 0.00} & 0.93\footnotesize{$\pm$ 0.15} & 0.60\footnotesize{$\pm$ 0.19} & 0.43\footnotesize{$\pm$ 0.16} & 0.826\footnotesize{$\pm$ 0.211} \\
        Burden test & 0.699 \footnotesize{$\pm$ 0.041} & {\bf 1.00} \footnotesize{$\pm$ 0.00} & 0.64\footnotesize{$\pm$ 0.05} & {\bf 1.00}\footnotesize{$\pm$ 0.00} & 0.89\footnotesize{$\pm$0.02} & {\bf 1.00}\footnotesize{$\pm$ 0.00} & {\bf 1.00}\footnotesize{$\pm$ 0.00} & {\bf 1.00}\footnotesize{$\pm$ 0.00} & {\bf 1.00}\footnotesize{$\pm$ 0.00} & 0.79\footnotesize{$\pm$ 0.28} & 0.549\footnotesize{$\pm$ 0.074} \\
        Log. MIL (KMER) & 0.582 \footnotesize{$\pm$ 0.065} & 0.54 \footnotesize{$\pm$ 0.07} & 0.51\footnotesize{$\pm$ 0.03} & 0.99\footnotesize{$\pm$ 0.00} & 0.62\footnotesize{$\pm$0.15} & {\bf 1.00}\footnotesize{$\pm$ 0.00} & 0.72\footnotesize{$\pm$ 0.11} & 0.64\footnotesize{$\pm$ 0.14} & 0.57\footnotesize{$\pm$ 0.15} & 0.53\footnotesize{$\pm$ 0.13} & 0.665\footnotesize{$\pm$ 0.224} \\
        Log. MIL (TCR\textbeta) & 0.515 \footnotesize{$\pm$ 0.073} & 0.50 \footnotesize{$\pm$ 0.03} & 0.50\footnotesize{$\pm$ 0.02} & 0.99\footnotesize{$\pm$ 0.00} & 0.78\footnotesize{$\pm$0.03} & 0.54\footnotesize{$\pm$ 0.09} & 0.57\footnotesize{$\pm$ 0.16} & 0.47\footnotesize{$\pm$ 0.09} & 0.51\footnotesize{$\pm$ 0.07} & 0.50\footnotesize{$\pm$ 0.12} & 0.501\footnotesize{$\pm$ 0.016} \\
        \midrule
        Known motif b. & -- & 1.00 \footnotesize{$\pm$ 0.00} & 0.70\footnotesize{$\pm$ 0.03} & 0.99\footnotesize{$\pm$ 0.00} & 0.62\footnotesize{$\pm$0.04} & 1.00\footnotesize{$\pm$ 0.00} & 1.00\footnotesize{$\pm$ 0.00} & 1.00\footnotesize{$\pm$ 0.00} & 1.00\footnotesize{$\pm$ 0.00} & 1.00\footnotesize{$\pm$ 0.00} & 0.890\footnotesize{$\pm$ 0.168} \\
        Known motif c. & -- & 0.92 \footnotesize{$\pm$ 0.00} & 0.56\footnotesize{$\pm$ 0.03} & 0.65\footnotesize{$\pm$ 0.03} & 0.52\footnotesize{$\pm$0.03} & 1.00\footnotesize{$\pm$ 0.00} & 1.00\footnotesize{$\pm$ 0.00} & 0.99\footnotesize{$\pm$ 0.01} & 0.72\footnotesize{$\pm$ 0.09} & 0.63\footnotesize{$\pm$ 0.09} & 0.738\footnotesize{$\pm$ 0.202} \\     
    \bottomrule
    \end{tabular}
        }%
        \caption{Results in terms of AUC of the competing methods on all datasets. The reported errors are standard deviations across $5$ cross-validation (CV) folds (except for the column ``Simulated'').
        \textbf{Real-world CMV:} Average performance over 
        $5$ CV folds on the \emph{cytomegalovirus (CMV) dataset}~\cite{emerson2017immunosequencing}.
        \textbf{Real-world data with implanted signals:} Average performance over
        $5$ CV folds for each 
        of the four datasets. A signal 
        was implanted with a frequency (=wittness rate) of 
        $1\%$ or $0.1\%$. Either a single motif (``OM'') 
        or multiple motifs (``MM'') were implanted.
        \textbf{LSTM-generated data:} Average performance over
        $5$ CV folds for each 
        of the $5$ datasets. In each dataset, a signal 
        was implanted with a frequency of $10\%$,
        $1\%$, $0.5\%$, $0.1\%$, and $0.05\%$, respectively.
        \textbf{Simulated:} Here we report the mean over 18 simulated datasets with implanted signals and varying difficulties (see 
        Table~\ref{tab:results_naive} for details). The error reported is the standard deviation of the AUC values across the 18 datasets.
        }%
        \label{tab:results_full}%
    \end{center}%
\end{table}
\endgroup

\clearpage
\subsection{Detailed results}
\label{sec:detailed_results}
In this section, we report the detailed results on all four categories 
of datasets (a) simulated immunosequencing data (Table~\ref{tab:results_naive})
(b) LSTM-generated data (Table~\ref{tab:results_lstm}),
(c) real-world data with implanted signals (Table~\ref{tab:results_milena}), 
and (d) real-world data on the \emph{CMV dataset} (Table~\ref{tab:results_cmv}).

\paragraph{Simulated immunosequencing data.}
In this setting the complexity of the implanted signal is in focus and varies 
throughout 18 simulated datasets.
Some datasets are challenging for the methods because
the implanted motif is hidden by noise and others because
only a small fraction of sequences carries the motif, 
and hence have a low witness rate.
These difficulties
become evident by the method called ``known motif binary'', which assumes
the implanted motif is known.
The performance of this method ranges from a
perfect AUC of $1.000$ in several datasets to an 
AUC of $0.532$ in dataset '17'.
DeepRC outperforms all other methods with an average AUC of $0.846\pm0.223$, 
followed by the SVM with MinMax kernel with an average AUC of $0.827\pm0.210$ (see Tab.~\ref{tab:results_naive}).
The predictive
performance of all methods suffers if the signal occurs only in an extremely
small fraction of sequences. 
In datasets, in which only $0.01\%$ of the sequences
carry the motif, all AUC values are below~$0.550$.

\begingroup
\setlength{\tabcolsep}{2pt} 
\renewcommand{\arraystretch}{1.5} 
\begingroup
\setlength{\tabcolsep}{2pt} 
\renewcommand{\arraystretch}{1.5} 
\begin{table*}[h]%
    \begin{center}%
        \resizebox{0.98\textwidth}{!}{%
        \begin{tabular}{l acacacacacacacacaca}%
            \toprule
            ID &0 &1 & 2 & 3 & 4 & 5 & 6 & 7 & 8 & 9 &10 &11 &12 &13 &14 &15 &16 &17 & avg.\\
motif freq. $\rho$ & 1\% & 0.1\% & 0.01\% & 1\% & 0.1\% & 0.01\% & 1\% & 0.1\% & 0.01\% & 1\% & 0.1\% & 0.01\% & 1\% & 0.1\% & 0.01\% & 1\% & 0.1\% & 0.01\% & -- \\
implanted motif & \texttt{SFEN} & \texttt{SFEN} & \texttt{SFEN} & \texttt{SF\textsuperscript{d}EN} & \texttt{SF\textsuperscript{d}EN} & \texttt{SF\textsuperscript{d}EN} & \texttt{SFZN} & \texttt{SFZN} & \texttt{SFZN} & \texttt{SF\textsuperscript{d}ZN} & \texttt{SF\textsuperscript{d}ZN} & \texttt{SF\textsuperscript{d}ZN} & \texttt{SZZN} & \texttt{SZZN} & \texttt{SZZN} & \texttt{SZ\textsuperscript{d}ZN} & \texttt{SZ\textsuperscript{d}ZN} & \texttt{SZ\textsuperscript{d}ZN} & --\\
\midrule
DeepRC & {\bf 1.000} & {\bf 1.000} & 0.703 & {\bf 1.000} & {\bf 1.000} & 0.600 & {\bf 1.000} & {\bf 1.000} & 0.509 & {\bf 1.000} & {\bf 1.000} & 0.492 & {\bf 1.000} & {\bf 0.997} & 0.487 & 0.999 & {\bf 0.942} & 0.492 & {\bf 0.864} \\
& \footnotesize{$\pm$ 0.000} & \footnotesize{$\pm$ 0.000} & \footnotesize{$\pm$ 0.271} & \footnotesize{$\pm$ 0.000} & \footnotesize{$\pm$ 0.000} & \footnotesize{$\pm$ 0.218} & \footnotesize{$\pm$ 0.000} & \footnotesize{$\pm$ 0.000} & \footnotesize{$\pm$ 0.029} & \footnotesize{$\pm$ 0.000} & \footnotesize{$\pm$ 0.001} & \footnotesize{$\pm$ 0.017} & \footnotesize{$\pm$ 0.001} & \footnotesize{$\pm$ 0.002} & \footnotesize{$\pm$ 0.023} & \footnotesize{$\pm$ 0.001} & \footnotesize{$\pm$ 0.048} & \footnotesize{$\pm$ 0.013} & \footnotesize{$\pm$ 0.223} \\
SVM (MinMax) & {\bf 1.000} & {\bf 1.000} & 0.764 & {\bf 1.000} & {\bf 1.000} & 0.603 & {\bf 1.000} & 0.998 & {\bf 0.539} & {\bf 1.000} & 0.994 & {\bf 0.529} & {\bf 1.000} & 0.741 & {\bf 0.513} & {\bf 1.000} & 0.706 & 0.503 & 0.827 \\
& \footnotesize{$\pm$ 0.000} & \footnotesize{$\pm$ 0.000} & \footnotesize{$\pm$ 0.016} & \footnotesize{$\pm$ 0.000} & \footnotesize{$\pm$ 0.000} & \footnotesize{$\pm$ 0.021} & \footnotesize{$\pm$ 0.000} & \footnotesize{$\pm$ 0.002} & \footnotesize{$\pm$ 0.024} & \footnotesize{$\pm$ 0.000} & \footnotesize{$\pm$ 0.004} & \footnotesize{$\pm$ 0.016} & \footnotesize{$\pm$ 0.000} & \footnotesize{$\pm$ 0.024} & \footnotesize{$\pm$ 0.006} & \footnotesize{$\pm$ 0.000} & \footnotesize{$\pm$ 0.013} & \footnotesize{$\pm$ 0.013} & \footnotesize{$\pm$ 0.210} \\
SVM (Jaccard) & 0.783 & 0.505 & 0.500 & 0.656 & 0.504 & 0.492 & 0.629 & 0.499 & 0.505 & 0.594 & 0.508 & 0.497 & 0.620 & 0.496 & 0.506 & 0.595 & 0.507 & 0.505 & 0.550 \\
& \footnotesize{$\pm$ 0.010} & \footnotesize{$\pm$ 0.009} & \footnotesize{$\pm$ 0.010} & \footnotesize{$\pm$ 0.009} & \footnotesize{$\pm$ 0.018} & \footnotesize{$\pm$ 0.018} & \footnotesize{$\pm$ 0.011} & \footnotesize{$\pm$ 0.010} & \footnotesize{$\pm$ 0.009} & \footnotesize{$\pm$ 0.007} & \footnotesize{$\pm$ 0.017} & \footnotesize{$\pm$ 0.013} & \footnotesize{$\pm$ 0.007} & \footnotesize{$\pm$ 0.006} & \footnotesize{$\pm$ 0.019} & \footnotesize{$\pm$ 0.013} & \footnotesize{$\pm$ 0.012} & \footnotesize{$\pm$ 0.017} & \footnotesize{$\pm$ 0.080} \\
KNN (MinMax) & 0.669 & 0.802 & 0.503 & 0.722 & 0.757 & 0.493 & 0.766 & 0.678 & 0.496 & 0.762 & 0.652 & 0.489 & 0.797 & 0.512 & 0.498 & 0.796 & 0.511 & 0.503 & 0.634 \\
& \footnotesize{$\pm$ 0.204} & \footnotesize{$\pm$ 0.265} & \footnotesize{$\pm$ 0.038} & \footnotesize{$\pm$ 0.214} & \footnotesize{$\pm$ 0.255} & \footnotesize{$\pm$ 0.017} & \footnotesize{$\pm$ 0.241} & \footnotesize{$\pm$ 0.165} & \footnotesize{$\pm$ 0.014} & \footnotesize{$\pm$ 0.237} & \footnotesize{$\pm$ 0.139} & \footnotesize{$\pm$ 0.015} & \footnotesize{$\pm$ 0.271} & \footnotesize{$\pm$ 0.023} & \footnotesize{$\pm$ 0.014} & \footnotesize{$\pm$ 0.270} & \footnotesize{$\pm$ 0.037} & \footnotesize{$\pm$ 0.006} & \footnotesize{$\pm$ 0.129} \\
KNN (Jaccard) & 0.516 & 0.493 & 0.497 & 0.506 & 0.500 & 0.492 & 0.509 & 0.493 & 0.497 & 0.495 & 0.504 & 0.500 & 0.502 & 0.497 & 0.500 & 0.502 & 0.503 & {\bf 0.513} & 0.501 \\
& \footnotesize{$\pm$ 0.035} & \footnotesize{$\pm$ 0.020} & \footnotesize{$\pm$ 0.013} & \footnotesize{$\pm$ 0.015} & \footnotesize{$\pm$ 0.019} & \footnotesize{$\pm$ 0.014} & \footnotesize{$\pm$ 0.017} & \footnotesize{$\pm$ 0.011} & \footnotesize{$\pm$ 0.018} & \footnotesize{$\pm$ 0.013} & \footnotesize{$\pm$ 0.004} & \footnotesize{$\pm$ 0.017} & \footnotesize{$\pm$ 0.011} & \footnotesize{$\pm$ 0.017} & \footnotesize{$\pm$ 0.022} & \footnotesize{$\pm$ 0.015} & \footnotesize{$\pm$ 0.020} & \footnotesize{$\pm$ 0.012} & \footnotesize{$\pm$ 0.007} \\
Logistic regression & {\bf 1.000} & {\bf 1.000} & {\bf 0.786} & {\bf 1.000} & {\bf 1.000} & {\bf 0.607} & {\bf 1.000} & 0.997 & 0.527 & {\bf 1.000} & 0.992 & 0.526 & {\bf 1.000} & 0.719 & 0.505 & {\bf 1.000} & 0.694 & 0.510 & 0.826 \\
& \footnotesize{$\pm$ 0.000} & \footnotesize{$\pm$ 0.000} & \footnotesize{$\pm$ 0.009} & \footnotesize{$\pm$ 0.000} & \footnotesize{$\pm$ 0.000} & \footnotesize{$\pm$ 0.025} & \footnotesize{$\pm$ 0.000} & \footnotesize{$\pm$ 0.002} & \footnotesize{$\pm$ 0.018} & \footnotesize{$\pm$ 0.000} & \footnotesize{$\pm$ 0.004} & \footnotesize{$\pm$ 0.019} & \footnotesize{$\pm$ 0.000} & \footnotesize{$\pm$ 0.019} & \footnotesize{$\pm$ 0.015} & \footnotesize{$\pm$ 0.001} & \footnotesize{$\pm$ 0.021} & \footnotesize{$\pm$ 0.017} & \footnotesize{$\pm$ 0.211} \\
Logistic MIL (KMER) & {\bf 1.000} & {\bf 1.000} & 0.509 & {\bf 1.000} & 0.783 & 0.489 & {\bf 1.000} & 0.544 & 0.517 & {\bf 1.000} & 0.529 & 0.483 & 0.579 & 0.498 & 0.502 & 0.550 & 0.488 & 0.498 & 0.665 \\
& \footnotesize{$\pm$ 0.000} & \footnotesize{$\pm$ 0.000} & \footnotesize{$\pm$ 0.039} & \footnotesize{$\pm$ 0.000} & \footnotesize{$\pm$ 0.216} & \footnotesize{$\pm$ 0.023} & \footnotesize{$\pm$ 0.000} & \footnotesize{$\pm$ 0.038} & \footnotesize{$\pm$ 0.018} & \footnotesize{$\pm$ 0.000} & \footnotesize{$\pm$ 0.043} & \footnotesize{$\pm$ 0.007} & \footnotesize{$\pm$ 0.042} & \footnotesize{$\pm$ 0.017} & \footnotesize{$\pm$ 0.018} & \footnotesize{$\pm$ 0.051} & \footnotesize{$\pm$ 0.009} & \footnotesize{$\pm$ 0.005} & \footnotesize{$\pm$ 0.224} \\
Logistic MIL (TCR\textbeta) & 0.544 & 0.505 & 0.493 & 0.487 & 0.476 & 0.500 & 0.520 & 0.495 & 0.510 & 0.492 & 0.506 & 0.503 & 0.509 & 0.505 & 0.500 & 0.475 & 0.489 & 0.500 & 0.501 \\
& \footnotesize{$\pm$ 0.078} & \footnotesize{$\pm$ 0.014} & \footnotesize{$\pm$ 0.018} & \footnotesize{$\pm$ 0.021} & \footnotesize{$\pm$ 0.019} & \footnotesize{$\pm$ 0.022} & \footnotesize{$\pm$ 0.053} & \footnotesize{$\pm$ 0.009} & \footnotesize{$\pm$ 0.022} & \footnotesize{$\pm$ 0.014} & \footnotesize{$\pm$ 0.019} & \footnotesize{$\pm$ 0.010} & \footnotesize{$\pm$ 0.034} & \footnotesize{$\pm$ 0.009} & \footnotesize{$\pm$ 0.011} & \footnotesize{$\pm$ 0.013} & \footnotesize{$\pm$ 0.024} & \footnotesize{$\pm$ 0.019} & \footnotesize{$\pm$ 0.016} \\
\midrule
Known motif b. & 1.000 & 1.000 & 0.973 & 1.000 & 1.000 & 0.865 & 1.000 & 1.000 & 0.700 & 1.000 & 0.989 & 0.609 & 1.000 & 0.946 & 0.570 & 1.000 & 0.834 & 0.532 & 0.890 \\
& \footnotesize{$\pm$ 0.000} & \footnotesize{$\pm$ 0.000} & \footnotesize{$\pm$ 0.004} & \footnotesize{$\pm$ 0.000} & \footnotesize{$\pm$ 0.000} & \footnotesize{$\pm$ 0.004} & \footnotesize{$\pm$ 0.000} & \footnotesize{$\pm$ 0.000} & \footnotesize{$\pm$ 0.020} & \footnotesize{$\pm$ 0.000} & \footnotesize{$\pm$ 0.002} & \footnotesize{$\pm$ 0.017} & \footnotesize{$\pm$ 0.000} & \footnotesize{$\pm$ 0.010} & \footnotesize{$\pm$ 0.024} & \footnotesize{$\pm$ 0.000} & \footnotesize{$\pm$ 0.016} & \footnotesize{$\pm$ 0.020} & \footnotesize{$\pm$ 0.168} \\
Known motif c. & 0.999 & 0.720 & 0.529 & 0.999 & 0.698 & 0.534 & 0.999 & 0.694 & 0.532 & 1.000 & 0.696 & 0.527 & 0.997 & 0.666 & 0.520 & 0.998 & 0.668 & 0.509 & 0.738 \\
& \footnotesize{$\pm$ 0.001} & \footnotesize{$\pm$ 0.014} & \footnotesize{$\pm$ 0.020} & \footnotesize{$\pm$ 0.001} & \footnotesize{$\pm$ 0.013} & \footnotesize{$\pm$ 0.017} & \footnotesize{$\pm$ 0.001} & \footnotesize{$\pm$ 0.012} & \footnotesize{$\pm$ 0.012} & \footnotesize{$\pm$ 0.001} & \footnotesize{$\pm$ 0.018} & \footnotesize{$\pm$ 0.018} & \footnotesize{$\pm$ 0.002} & \footnotesize{$\pm$ 0.010} & \footnotesize{$\pm$ 0.009} & \footnotesize{$\pm$ 0.002} & \footnotesize{$\pm$ 0.012} & \footnotesize{$\pm$ 0.013} & \footnotesize{$\pm$ 0.202} \\

      \bottomrule
        \end{tabular}%
        }%
        \caption{AUC estimates based on 5-fold CV for all $18$ datasets in category ``simulated immunosequencing data''.
        The reported errors are standard deviations across the $5$ cross-validation folds except 
        for the last column ``avg.'', in which they show standard deviations across datasets.
        Wildcard characters in motifs are indicated by \texttt{Z},
        characters with $50\%$ probability of being removed by \texttt{\textsuperscript{d}.}
        }%
        \label{tab:results_naive}%
    \end{center}%
\end{table*}
\endgroup

\paragraph{LSTM-generated data.}
On the LSTM-generated data, in which we implanted noisy motifs 
with frequencies of $10\%$, $1\%$, $0.5\%$, $0.1\%$, and $0.05\%$, DeepRC yields almost
perfect predictive performance with an average AUC of $1.000\pm0.001$. The second best method, 
SVM with MinMax kernel, has a similar predictive performance 
to DeepRC on all datasets but the other 
competing methods have a lower predictive performance 
on datasets with low frequency of the signal ($0.05\%$) (see Tab.~\ref{tab:results_lstm}).

\begin{table}[htp]
    \centering
    \resizebox{0.8\columnwidth}{!}{
    \begin{tabular}{lacacac}
    \toprule
ID & 0 & 1 & 2 & 3 & 4 & avg.  \\
motif freq. $\rho$ & 10\% & 1\% & 0.5\% & 0.1\% & 0.05\% & -- \\
implanted motif & \texttt{G\textsuperscript{r}S\textsuperscript{r}A\textsuperscript{r}F\textsuperscript{r}} & \texttt{G\textsuperscript{r}S\textsuperscript{r}A\textsuperscript{r}F\textsuperscript{r}} &\texttt{G\textsuperscript{r}S\textsuperscript{r}A\textsuperscript{r}F\textsuperscript{r}} &\texttt{G\textsuperscript{r}S\textsuperscript{r}A\textsuperscript{r}F\textsuperscript{r}} &\texttt{G\textsuperscript{r}S\textsuperscript{r}A\textsuperscript{r}F\textsuperscript{r}} & --\\
\midrule
DeepRC & {\bf 1.000} \footnotesize{$\pm$ 0.000} & {\bf 1.000} \footnotesize{$\pm$ 0.000} & {\bf 1.000} \footnotesize{$\pm$ 0.000} & {\bf 1.000} \footnotesize{$\pm$ 0.000} & {\bf 0.998} \footnotesize{$\pm$ 0.002} & {\bf 1.000} \footnotesize{$\pm$ 0.001} \\
SVM (MinMax) & {\bf 1.000} \footnotesize{$\pm$ 0.000} & {\bf 1.000} \footnotesize{$\pm$ 0.000} & 0.999 \footnotesize{$\pm$ 0.001} & 0.999 \footnotesize{$\pm$ 0.002} & 0.985 \footnotesize{$\pm$ 0.014} & 0.997 \footnotesize{$\pm$ 0.007} \\
SVM (Jaccard) & 0.981 \footnotesize{$\pm$ 0.041} & {\bf 1.000} \footnotesize{$\pm$ 0.000} & {\bf 1.000} \footnotesize{$\pm$ 0.000} & 0.904 \footnotesize{$\pm$ 0.036} & 0.768 \footnotesize{$\pm$ 0.068} & 0.931 \footnotesize{$\pm$ 0.099} \\
KNN (MinMax) & 0.699 \footnotesize{$\pm$ 0.272} & 0.717 \footnotesize{$\pm$ 0.263} & 0.732 \footnotesize{$\pm$ 0.263} & 0.536 \footnotesize{$\pm$ 0.156} & 0.516 \footnotesize{$\pm$ 0.153} & 0.640 \footnotesize{$\pm$ 0.105} \\
KNN (Jaccard) & 0.698 \footnotesize{$\pm$ 0.285} & 0.606 \footnotesize{$\pm$ 0.237} & 0.523 \footnotesize{$\pm$ 0.164} & 0.550 \footnotesize{$\pm$ 0.186} & 0.539 \footnotesize{$\pm$ 0.194} & 0.583 \footnotesize{$\pm$ 0.071} \\
Logistic regression & {\bf 1.000} \footnotesize{$\pm$ 0.000} & {\bf 1.000} \footnotesize{$\pm$ 0.000} & 0.934 \footnotesize{$\pm$ 0.147} & 0.604 \footnotesize{$\pm$ 0.193} & 0.427 \footnotesize{$\pm$ 0.156} & 0.793 \footnotesize{$\pm$ 0.262} \\
Logistic MIL (KMER) & 0.997 \footnotesize{$\pm$ 0.004} & 0.718 \footnotesize{$\pm$ 0.112} & 0.637 \footnotesize{$\pm$ 0.144} & 0.571 \footnotesize{$\pm$ 0.146} & 0.528 \footnotesize{$\pm$ 0.129} & 0.690 \footnotesize{$\pm$ 0.186} \\
Logistic MIL (TCR\textbeta) & 0.541 \footnotesize{$\pm$ 0.086} & 0.566 \footnotesize{$\pm$ 0.162} & 0.468 \footnotesize{$\pm$ 0.086} & 0.505 \footnotesize{$\pm$ 0.067} & 0.500 \footnotesize{$\pm$ 0.121} & 0.516 \footnotesize{$\pm$ 0.038} \\
\midrule
Known motif b. & 1.000 \footnotesize{$\pm$ 0.000} & 1.000 \footnotesize{$\pm$ 0.000} & 1.000 \footnotesize{$\pm$ 0.000} & 0.999 \footnotesize{$\pm$ 0.003} & 0.999 \footnotesize{$\pm$ 0.003} & 1.000 \footnotesize{$\pm$ 0.001} \\
Known motif c. & 1.000 \footnotesize{$\pm$ 0.000} & 1.000 \footnotesize{$\pm$ 0.000} & 0.989 \footnotesize{$\pm$ 0.011} & 0.722 \footnotesize{$\pm$ 0.085} & 0.626 \footnotesize{$\pm$ 0.094} & 0.867 \footnotesize{$\pm$ 0.180} \\ 
\bottomrule
        \end{tabular} 
        }
    \caption{
            AUC estimates based on 5-fold CV for all $5$ datasets in category ``LSTM-generated data''.
            The reported errors are standard deviations across the $5$ cross-validation folds except 
            for the last column ``avg.'', in which they show standard deviations across datasets.
            Characters affected by noise, as described in \ref{sec:datasets}, paragraph ``LSTM-generated data'', are indicated by \texttt{\textsuperscript{r}.}
            }
    \label{tab:results_lstm}
\end{table}

\paragraph{Real-world data with implanted motifs.}
In this dataset category, we used real immunosequences and
implanted single or multiple noisy motifs. Again, 
DeepRC outperforms all other methods with an average AUC of $0.980\pm0.029$,
with the second best method being the burden test with 
an average AUC of $0.883\pm0.170$.
Notably, all methods except for DeepRC have difficulties with noisy
motifs at a frequency of $0.1\%$ 
(see Tab.~\ref{tab:results_milena}). 

\begin{table*}[htp]
    \centering
    \begin{tabular}{lacaca}
    \toprule
  & OM 1\% & OM 0.1\%  & MM 1\% & MM 0.1\% & Avg.  \\
  \midrule
DeepRC & {\bf 1.000} \footnotesize{$\pm$ 0.000} & {\bf 0.984} \footnotesize{$\pm$ 0.008} & 0.999 \footnotesize{$\pm$ 0.001} & {\bf 0.938} \footnotesize{$\pm$ 0.009} & {\bf 0.980} \footnotesize{$\pm$ 0.029} \\
SVM (MinMax) & {\bf 1.000} \footnotesize{$\pm$ 0.000} & 0.578 \footnotesize{$\pm$ 0.020} & {\bf 1.000} \footnotesize{$\pm$ 0.000} & 0.531 \footnotesize{$\pm$ 0.019} & 0.777 \footnotesize{$\pm$ 0.258} \\
SVM (Jaccard) & 0.988 \footnotesize{$\pm$ 0.003} & 0.527 \footnotesize{$\pm$ 0.016} & {\bf 1.000} \footnotesize{$\pm$ 0.000} & 0.574 \footnotesize{$\pm$ 0.019} & 0.772 \footnotesize{$\pm$ 0.257} \\
KNN (MinMax) & 0.744 \footnotesize{$\pm$ 0.237} & 0.486 \footnotesize{$\pm$ 0.031} & 0.674 \footnotesize{$\pm$ 0.182} & 0.500 \footnotesize{$\pm$ 0.022} & 0.601 \footnotesize{$\pm$ 0.128} \\
KNN (Jaccard) & 0.652 \footnotesize{$\pm$ 0.155} & 0.484 \footnotesize{$\pm$ 0.025} & 0.695 \footnotesize{$\pm$ 0.200} & 0.508 \footnotesize{$\pm$ 0.025} & 0.585 \footnotesize{$\pm$ 0.104} \\
Logistic regression & {\bf 1.000} \footnotesize{$\pm$ 0.000} & 0.544 \footnotesize{$\pm$ 0.035} & 0.991 \footnotesize{$\pm$ 0.003} & 0.512 \footnotesize{$\pm$ 0.035} & 0.762 \footnotesize{$\pm$ 0.270} \\
Logistic MIL (KMER) & 0.541 \footnotesize{$\pm$ 0.074} & 0.506 \footnotesize{$\pm$ 0.034} & 0.994 \footnotesize{$\pm$ 0.004} & 0.620 \footnotesize{$\pm$ 0.153} & 0.665 \footnotesize{$\pm$ 0.224} \\
Logistic MIL (TCR\textbeta) & 0.503 \footnotesize{$\pm$ 0.032} & 0.501 \footnotesize{$\pm$ 0.016} & 0.992 \footnotesize{$\pm$ 0.003} & 0.782 \footnotesize{$\pm$ 0.030} & 0.695 \footnotesize{$\pm$ 0.238} \\
\midrule
Known motif b. & 1.000 \footnotesize{$\pm$ 0.000} & 0.704 \footnotesize{$\pm$ 0.028} & 0.994 \footnotesize{$\pm$ 0.003} & 0.620 \footnotesize{$\pm$ 0.038} & 0.830 \footnotesize{$\pm$ 0.196} \\
Known motif c. & 0.920 \footnotesize{$\pm$ 0.004} & 0.562 \footnotesize{$\pm$ 0.028} & 0.647 \footnotesize{$\pm$ 0.030} & 0.515 \footnotesize{$\pm$ 0.031} & 0.661 \footnotesize{$\pm$ 0.181} \\ 
   \bottomrule
    \end{tabular}
    \caption{
            AUC estimates based on 5-fold CV for all $4$ datasets in category ``real-world data with implanted signals''.
            The reported errors are standard deviations across the $5$ cross-validation folds except 
            for the last column ``avg.'', in which they show standard deviations across datasets.
            \textbf{One Motif 1\%:} In this dataset, a single motif with a frequency of 1\% was implanted. 
            \textbf{One 0.1\%:} In this dataset, a single motif with a frequency of 0.1\% was implanted.
            \textbf{Multi 1\%:} In this dataset, multiple motifs with a frequency of 1\% were implanted.
            \textbf{Multi 0.1\%:} In this dataset, multiple motifs with a frequency of 0.1\% were implanted.
            A detailed description of the motifs is provided in section~\ref{sec:datasets}, paragraph ``Real-world data with implanted signals.''.}
    \label{tab:results_milena}
\end{table*}

\paragraph{Real-world data.}
On the real-world dataset, in which the immune status of persons affected
by the cytomegalovirus has to be predicted, the competing methods yield 
predictive AUCs between $0.515$ and $0.825$.
This dataset differs from the dataset that was used in  
\cite{emerson2017immunosequencing} (see Sec.~\ref{sec:datasets}).
The best performing 
method is DeepRC with an AUC of $0.832\pm0.002$, followed by
the SVM with MinMax kernel (AUC $0.825\pm0.022$) and the burden
test with an AUC of $0.699\pm0.041$ (see Tab.~\ref{tab:results_cmv}).

\begin{table}[htp]
    \centering
    \begin{tabular}{lacac}
    \toprule
    & AUC & F1 score & balanced accuracy & accuracy \\
    \midrule
    DeepRC & {\bf0.832} \footnotesize{$\pm$ 0.002} & {\bf 0.721} \footnotesize{$\pm$ 0.030} & {\bf 0.734} \footnotesize{$\pm$ 0.032} & 0.735 \footnotesize{$\pm$ 0.037} \\
    SVM (MinMax) & 0.825 \footnotesize{$\pm$ 0.022} & 0.680 \footnotesize{$\pm$ 0.056} & {\bf 0.734} \footnotesize{$\pm$ 0.037} & {\bf 0.742} \footnotesize{$\pm$ 0.031} \\
    SVM (Jaccard) & 0.546 \footnotesize{$\pm$ 0.021} & 0.272 \footnotesize{$\pm$ 0.184} & 0.523 \footnotesize{$\pm$ 0.026} & 0.542 \footnotesize{$\pm$ 0.032} \\
    KNN (MinMax) & 0.679 \footnotesize{$\pm$ 0.076} & 0.000 \footnotesize{$\pm$ 0.000} & 0.500 \footnotesize{$\pm$ 0.000} & 0.545 \footnotesize{$\pm$ 0.044} \\
    KNN (Jaccard) & 0.534 \footnotesize{$\pm$ 0.039} & 0.073 \footnotesize{$\pm$ 0.101} & 0.508 \footnotesize{$\pm$ 0.012} & 0.551 \footnotesize{$\pm$ 0.042} \\
    Logistic regression & 0.607 \footnotesize{$\pm$ 0.058} & 0.244 \footnotesize{$\pm$ 0.206} & 0.552 \footnotesize{$\pm$ 0.049} & 0.590 \footnotesize{$\pm$ 0.019} \\
    Logistic MIL (KMER) & 0.582 \footnotesize{$\pm$ 0.065} & 0.118 \footnotesize{$\pm$ 0.264} & 0.503 \footnotesize{$\pm$ 0.007} & 0.515 \footnotesize{$\pm$ 0.058} \\
    Logistic MIL (TCR\textbeta) & 0.515 \footnotesize{$\pm$ 0.073} & 0.000 \footnotesize{$\pm$ 0.000} & 0.496 \footnotesize{$\pm$ 0.008} & 0.541 \footnotesize{$\pm$ 0.039} \\
    \bottomrule
    \end{tabular} 
    \caption{Results on the \emph{CMV dataset} (real-world data) in terms of AUC, F1 score, balanced accuracy, and accuracy. For F1 score, balanced accuracy, and accuracy, all methods use their default thresholds. Each entry shows mean and standard deviation across $5$ 
    cross-validation folds.}
    \label{tab:results_cmv}
\end{table}

\clearpage 

\bibliography{references_supplement_B}
\bibliographystyle{plain}